\documentclass[]{article}
\usepackage{lmodern}
\usepackage{amssymb,amsmath}
\usepackage{ifxetex,ifluatex}
\usepackage{fixltx2e} 
\ifnum 0\ifxetex 1\fi\ifluatex 1\fi=0 
  \usepackage[T1]{fontenc}
  \usepackage[utf8]{inputenc}
\else 
  \ifxetex
    \usepackage{mathspec}
  \else
    \usepackage{fontspec}
  \fi
  \defaultfontfeatures{Ligatures=TeX,Scale=MatchLowercase}
\fi
\IfFileExists{upquote.sty}{\usepackage{upquote}}{}
\IfFileExists{microtype.sty}{%
\usepackage{microtype}
\UseMicrotypeSet[protrusion]{basicmath} 
}{}
\usepackage[unicode=true]{hyperref}
\PassOptionsToPackage{usenames,dvipsnames}{color} 
\hypersetup{
            pdftitle={mlr Tutorial},
            pdfauthor={Julia Schiffner; Bernd Bischl; Michel Lang; Jakob Richter; Zachary M. Jones; Philipp Probst; Florian Pfisterer; Mason Gallo; Dominik Kirchhoff; Tobias Kühn; Janek Thomas; Lars Kotthoff},
            colorlinks=true,
            linkcolor=Maroon,
            citecolor=Blue,
            urlcolor=Blue,
            breaklinks=true}
\usepackage{listings}
\usepackage{longtable,booktabs}
\usepackage{graphicx,grffile}
\makeatletter
\def\maxwidth{\ifdim\Gin@nat@width>\linewidth\linewidth\else\Gin@nat@width\fi}
\def\maxheight{\ifdim\Gin@nat@height>\textheight\textheight\else\Gin@nat@height\fi}
\makeatother
\setkeys{Gin}{width=\maxwidth,height=\maxheight,keepaspectratio}
\IfFileExists{parskip.sty}{%
\usepackage{parskip}
}{
\setlength{\parindent}{0pt}
\setlength{\parskip}{6pt plus 2pt minus 1pt}
}
\providecommand{\tightlist}{%
  \setlength{\itemsep}{0pt}\setlength{\parskip}{0pt}}
\ifx\paragraph\undefined\else
\let\oldparagraph\paragraph
\renewcommand{\paragraph}[1]{\oldparagraph{#1}\mbox{}}
\fi
\ifx\subparagraph\undefined\else
\let\oldsubparagraph\subparagraph
\renewcommand{\subparagraph}[1]{\oldsubparagraph{#1}\mbox{}}
\fi
\usepackage{fancyhdr}
\title{mlr Tutorial}
\author{Julia Schiffner \and Bernd Bischl \and Michel Lang \and Jakob Richter \and Zachary M. Jones \and Philipp Probst \and Florian Pfisterer \and Mason Gallo \and Dominik Kirchhoff \and Tobias Kühn \and Janek Thomas \and Lars Kotthoff}
\date{}

\begin{document}
\maketitle

{
\hypersetup{linkcolor=black}
\setcounter{tocdepth}{3}
\tableofcontents
}
\section{mlr Tutorial}\label{mlr-tutorial}

This web page provides an in-depth introduction on how to use the
\href{http://www.rdocumentation.org/packages/mlr/}{mlr} framework for
machine learning experiments in \textbf{R}.

We focus on the comprehension of the basic functions and applications.
More detailed technical information can be found in the
\href{http://www.rdocumentation.org/packages/mlr/}{manual pages} which
are regularly updated and reflect the documentation of the
\href{http://cran.r-project.org/web/packages/mlr/index.html}{current
package version on CRAN}.

An offline version of this tutorial is available for download

\begin{itemize}
\tightlist
\item
  \href{https://mlr-org.github.io/mlr-tutorial/release/mlr_tutorial.zip}{here}
  for the current mlr release on CRAN
\item
  and
  \href{https://mlr-org.github.io/mlr-tutorial/devel/mlr_tutorial.zip}{here}
  for the mlr devel version on Github.
\end{itemize}

The tutorial explains the basic analysis of a data set step by step.
Please refer to sections of the menu above: Basics, Advanced, Extend and
Appendix.

During the tutorial we present various simple examples from
classification, regression, cluster and survival analysis to illustrate
the main features of the package.

Enjoy reading!

\subsection{Quick start}\label{quick-start}

A simple stratified cross-validation of
\href{http://www.rdocumentation.org/packages/MASS/functions/lda.html}{linear
discriminant analysis} with
\href{http://www.rdocumentation.org/packages/mlr/}{mlr}.

\begin{lstlisting}[language=R]
library(mlr)
data(iris)

## Define the task
task = makeClassifTask(id = "tutorial", data = iris, target = "Species")

## Define the learner
lrn = makeLearner("classif.lda")

## Define the resampling strategy
rdesc = makeResampleDesc(method = "CV", stratify = TRUE)

## Do the resampling
r = resample(learner = lrn, task = task, resampling = rdesc, show.info = FALSE)

## Get the mean misclassification error
r$aggr
#> mmce.test.mean 
#>           0.02
\end{lstlisting}

\section{Basics}\label{basics}

\hypertarget{learning-tasks}{\subsection{Learning
Tasks}\label{learning-tasks}}

Learning tasks encapsulate the data set and further relevant information
about a machine learning problem, for example the name of the target
variable for supervised problems.

\subsubsection{Task types and creation}\label{task-types-and-creation}

The tasks are organized in a hierarchy, with the generic
\href{http://www.rdocumentation.org/packages/mlr/functions/Task.html}{Task}
at the top. The following tasks can be instantiated and all inherit from
the virtual superclass
\href{http://www.rdocumentation.org/packages/mlr/functions/Task.html}{Task}:

\begin{itemize}
\tightlist
\item
  \href{http://www.rdocumentation.org/packages/mlr/functions/Task.html}{RegrTask}
  for regression problems,
\item
  \href{http://www.rdocumentation.org/packages/mlr/functions/Task.html}{ClassifTask}
  for binary and multi-class classification problems
  (\protect\hyperlink{cost-sensitive-classification}{cost-sensitive
  classification} with class-dependent costs can be handled as well),
\item
  \href{http://www.rdocumentation.org/packages/mlr/functions/Task.html}{SurvTask}
  for survival analysis,
\item
  \href{http://www.rdocumentation.org/packages/mlr/functions/Task.html}{ClusterTask}
  for cluster analysis,
\item
  \href{http://www.rdocumentation.org/packages/mlr/functions/Task.html}{MultilabelTask}
  for multilabel classification problems,
\item
  \href{http://www.rdocumentation.org/packages/mlr/functions/Task.html}{CostSensTask}
  for general
  \protect\hyperlink{cost-sensitive-classification}{cost-sensitive
  classification} (with example-specific costs).
\end{itemize}

To create a task, just call \lstinline!make<TaskType>!, e.g.,
\href{http://www.rdocumentation.org/packages/mlr/functions/Task.html}{makeClassifTask}.
All tasks require an identifier (argument \lstinline!id!) and a
\href{http://www.rdocumentation.org/packages/base/functions/data.frame.html}{data.frame}
(argument \lstinline!data!). If no ID is provided it is automatically
generated using the variable name of the data. The ID will be later used
to name results, for example of
\protect\hyperlink{benchmark-experiments}{benchmark experiments}, and to
annotate plots. Depending on the nature of the learning problem,
additional arguments may be required and are discussed in the following
sections.

\paragraph{Regression}\label{regression}

For supervised learning like regression (as well as classification and
survival analysis) we, in addition to \lstinline!data!, have to specify
the name of the \lstinline!target! variable.

\begin{lstlisting}[language=R]
data(BostonHousing, package = "mlbench")
regr.task = makeRegrTask(id = "bh", data = BostonHousing, target = "medv")
regr.task
#> Supervised task: bh
#> Type: regr
#> Target: medv
#> Observations: 506
#> Features:
#> numerics  factors  ordered 
#>       12        1        0 
#> Missings: FALSE
#> Has weights: FALSE
#> Has blocking: FALSE
\end{lstlisting}

As you can see, the
\href{http://www.rdocumentation.org/packages/mlr/functions/Task.html}{Task}
records the type of the learning problem and basic information about the
data set, e.g., the types of the features
(\href{http://www.rdocumentation.org/packages/base/functions/numeric.html}{numeric}
vectors,
\href{http://www.rdocumentation.org/packages/base/functions/factor.html}{factors}
or
\href{http://www.rdocumentation.org/packages/base/functions/factor.html}{ordered
factors}), the number of observations, or whether missing values are
present.

Creating tasks for classification and survival analysis follows the same
scheme, the data type of the target variables included in
\lstinline!data! is simply different. For each of these learning
problems some specifics are described below.

\paragraph{Classification}\label{classification}

For classification the target column has to be a
\href{http://www.rdocumentation.org/packages/base/functions/factor.html}{factor}.

In the following example we define a classification task for the
\href{http://www.rdocumentation.org/packages/mlbench/functions/BreastCancer.html}{BreastCancer}
data set and exclude the variable \lstinline!Id! from all further model
fitting and evaluation.

\begin{lstlisting}[language=R]
data(BreastCancer, package = "mlbench")
df = BreastCancer
df$Id = NULL
classif.task = makeClassifTask(id = "BreastCancer", data = df, target = "Class")
classif.task
#> Supervised task: BreastCancer
#> Type: classif
#> Target: Class
#> Observations: 699
#> Features:
#> numerics  factors  ordered 
#>        0        4        5 
#> Missings: TRUE
#> Has weights: FALSE
#> Has blocking: FALSE
#> Classes: 2
#>    benign malignant 
#>       458       241 
#> Positive class: benign
\end{lstlisting}

In binary classification the two classes are usually referred to as
\emph{positive} and \emph{negative} class with the positive class being
the category of greater interest. This is relevant for many
\protect\hyperlink{evaluating-learner-performance}{performance measures}
like the \emph{true positive rate} or
\protect\hyperlink{roc-analysis-and-performance-curves}{ROC curves}.
Moreover, \href{http://www.rdocumentation.org/packages/mlr/}{mlr}, where
possible, permits to set options (like the
\href{http://www.rdocumentation.org/packages/mlr/functions/setThreshold.html}{decision
threshold} or
\href{http://www.rdocumentation.org/packages/mlr/functions/makeWeightedClassesWrapper.html}{class
weights}) and returns and plots results (like class posterior
probabilities) for the positive class only.

\href{http://www.rdocumentation.org/packages/mlr/functions/Task.html}{makeClassifTask}
by default selects the first factor level of the target variable as the
positive class, in the above example \lstinline!benign!. Class
\lstinline!malignant! can be manually selected as follows:

\begin{lstlisting}[language=R]
classif.task = makeClassifTask(id = "BreastCancer", data = df, target = "Class", positive = "malignant")
\end{lstlisting}

\paragraph{Survival analysis}\label{survival-analysis}

Survival tasks use two target columns. For left and right censored
problems these consist of the survival time and a binary event
indicator. For interval censored data the two target columns must be
specified in the \lstinline!"interval2"! format (see
\href{http://www.rdocumentation.org/packages/survival/functions/Surv.html}{Surv}).

\begin{lstlisting}[language=R]
data(lung, package = "survival")
lung$status = (lung$status == 2) # convert to logical
surv.task = makeSurvTask(data = lung, target = c("time", "status"))
surv.task
#> Supervised task: lung
#> Type: surv
#> Target: time,status
#> Observations: 228
#> Features:
#> numerics  factors  ordered 
#>        8        0        0 
#> Missings: TRUE
#> Has weights: FALSE
#> Has blocking: FALSE
\end{lstlisting}

The type of censoring can be specified via the argument
\lstinline!censoring!, which defaults to \lstinline!"rcens"! for right
censored data.

\hypertarget{multilabel-classification}{\paragraph{Multilabel
classification}\label{multilabel-classification}}

In multilabel classification each object can belong to more than one
category at the same time.

The \lstinline!data! are expected to contain as many target columns as
there are class labels. The target columns should be logical vectors
that indicate which class labels are present. The names of the target
columns are taken as class labels and need to be passed to the
\lstinline!target! argument of
\href{http://www.rdocumentation.org/packages/mlr/functions/Task.html}{makeMultilabelTask}.

In the following example we get the data of the yeast data set, extract
the label names, and pass them to the \lstinline!target! argument in
\href{http://www.rdocumentation.org/packages/mlr/functions/Task.html}{makeMultilabelTask}.

\begin{lstlisting}[language=R]
yeast = getTaskData(yeast.task)

labels = colnames(yeast)[1:14]
yeast.task = makeMultilabelTask(id = "multi", data = yeast, target = labels)
yeast.task
#> Supervised task: multi
#> Type: multilabel
#> Target: label1,label2,label3,label4,label5,label6,label7,label8,label9,label10,label11,label12,label13,label14
#> Observations: 2417
#> Features:
#> numerics  factors  ordered 
#>      103        0        0 
#> Missings: FALSE
#> Has weights: FALSE
#> Has blocking: FALSE
#> Classes: 14
#>  label1  label2  label3  label4  label5  label6  label7  label8  label9 
#>     762    1038     983     862     722     597     428     480     178 
#> label10 label11 label12 label13 label14 
#>     253     289    1816    1799      34
\end{lstlisting}

See also the tutorial page on
\protect\hyperlink{multilabel-classification}{multilabel
classification}.

\paragraph{Cluster analysis}\label{cluster-analysis}

As cluster analysis is unsupervised, the only mandatory argument to
construct a cluster analysis task is the \lstinline!data!. Below we
create a learning task from the data set
\href{http://www.rdocumentation.org/packages/datasets/functions/mtcars.html}{mtcars}.

\begin{lstlisting}[language=R]
data(mtcars, package = "datasets")
cluster.task = makeClusterTask(data = mtcars)
cluster.task
#> Unsupervised task: mtcars
#> Type: cluster
#> Observations: 32
#> Features:
#> numerics  factors  ordered 
#>       11        0        0 
#> Missings: FALSE
#> Has weights: FALSE
#> Has blocking: FALSE
\end{lstlisting}

\hypertarget{cost-sensitive-classification}{\paragraph{Cost-sensitive
classification}\label{cost-sensitive-classification}}

The standard objective in classification is to obtain a high prediction
accuracy, i.e., to minimize the number of errors. All types of
misclassification errors are thereby deemed equally severe. However, in
many applications different kinds of errors cause different costs.

In case of \emph{class-dependent costs}, that solely depend on the
actual and predicted class labels, it is sufficient to create an
ordinary
\href{http://www.rdocumentation.org/packages/mlr/functions/Task.html}{ClassifTask}.

In order to handle \emph{example-specific costs} it is necessary to
generate a
\href{http://www.rdocumentation.org/packages/mlr/functions/Task.html}{CostSensTask}.
In this scenario, each example \((x, y)\) is associated with an
individual cost vector of length \(K\) with \(K\) denoting the number of
classes. The \(k\)-th component indicates the cost of assigning \(x\) to
class \(k\). Naturally, it is assumed that the cost of the intended
class label \(y\) is minimal.

As the cost vector contains all relevant information about the intended
class \(y\), only the feature values \(x\) and a \lstinline!cost!
matrix, which contains the cost vectors for all examples in the data
set, are required to create the
\href{http://www.rdocumentation.org/packages/mlr/functions/Task.html}{CostSensTask}.

In the following example we use the
\href{http://www.rdocumentation.org/packages/datasets/functions/iris.html}{iris}
data and an artificial cost matrix (which is generated as proposed by
\href{http://dx.doi.org/10.1145/1102351.1102358}{Beygelzimer et al.,
2005}):

\begin{lstlisting}[language=R]
df = iris
cost = matrix(runif(150 * 3, 0, 2000), 150) * (1 - diag(3))[df$Species,]
df$Species = NULL

costsens.task = makeCostSensTask(data = df, cost = cost)
costsens.task
#> Supervised task: df
#> Type: costsens
#> Observations: 150
#> Features:
#> numerics  factors  ordered 
#>        4        0        0 
#> Missings: FALSE
#> Has blocking: FALSE
#> Classes: 3
#> y1, y2, y3
\end{lstlisting}

For more details see the page about
\protect\hyperlink{cost-sensitive-classification}{cost-sensitive
classification}.

\subsubsection{Further settings}\label{further-settings}

The
\href{http://www.rdocumentation.org/packages/mlr/functions/Task.html}{Task}
help page also lists several other arguments to describe further details
of the learning problem.

For example, we could include a \lstinline!blocking! factor in the task.
This would indicate that some observations ``belong together'' and
should not be separated when splitting the data into training and test
sets for \protect\hyperlink{resampling}{resampling}.

Another option is to assign \lstinline!weights! to observations. These
can simply indicate observation frequencies or result from the sampling
scheme used to collect the data.\\
Note that you should use this option only if the weights really belong
to the task. If you plan to train some learning algorithms with
different weights on the same
\href{http://www.rdocumentation.org/packages/mlr/functions/Task.html}{Task},
\href{http://www.rdocumentation.org/packages/mlr/}{mlr} offers several
other ways to set observation or class weights (for supervised
classification). See for example the tutorial page about
\protect\hyperlink{training-a-learner}{training} or function
\href{http://www.rdocumentation.org/packages/mlr/functions/makeWeightedClassesWrapper.html}{makeWeightedClassesWrapper}.

\subsubsection{Accessing a learning
task}\label{accessing-a-learning-task}

We provide many operators to access the elements stored in a
\href{http://www.rdocumentation.org/packages/mlr/functions/Task.html}{Task}.
The most important ones are listed in the documentation of
\href{http://www.rdocumentation.org/packages/mlr/functions/Task.html}{Task}
and
\href{http://www.rdocumentation.org/packages/mlr/functions/getTaskData.html}{getTaskData}.

To access the
\href{http://www.rdocumentation.org/packages/mlr/functions/TaskDesc.html}{task
description} that contains basic information about the task you can use:

\begin{lstlisting}[language=R]
getTaskDescription(classif.task)
#> $id
#> [1] "BreastCancer"
#> 
#> $type
#> [1] "classif"
#> 
#> $target
#> [1] "Class"
#> 
#> $size
#> [1] 699
#> 
#> $n.feat
#> numerics  factors  ordered 
#>        0        4        5 
#> 
#> $has.missings
#> [1] TRUE
#> 
#> $has.weights
#> [1] FALSE
#> 
#> $has.blocking
#> [1] FALSE
#> 
#> $class.levels
#> [1] "benign"    "malignant"
#> 
#> $positive
#> [1] "malignant"
#> 
#> $negative
#> [1] "benign"
#> 
#> attr(,"class")
#> [1] "TaskDescClassif"    "TaskDescSupervised" "TaskDesc"
\end{lstlisting}

Note that
\href{http://www.rdocumentation.org/packages/mlr/functions/TaskDesc.html}{task
descriptions} have slightly different elements for different types of
\href{http://www.rdocumentation.org/packages/mlr/functions/Task.html}{Task}s.
Frequently required elements can also be accessed directly.

\begin{lstlisting}[language=R]
### Get the ID
getTaskId(classif.task)
#> [1] "BreastCancer"

### Get the type of task
getTaskType(classif.task)
#> [1] "classif"

### Get the names of the target columns
getTaskTargetNames(classif.task)
#> [1] "Class"

### Get the number of observations
getTaskSize(classif.task)
#> [1] 699

### Get the number of input variables
getTaskNFeats(classif.task)
#> [1] 9

### Get the class levels in classif.task
getTaskClassLevels(classif.task)
#> [1] "benign"    "malignant"
\end{lstlisting}

Moreover, \href{http://www.rdocumentation.org/packages/mlr/}{mlr}
provides several functions to extract data from a
\href{http://www.rdocumentation.org/packages/mlr/functions/Task.html}{Task}.

\begin{lstlisting}[language=R]
### Accessing the data set in classif.task
str(getTaskData(classif.task))
#> 'data.frame':    699 obs. of  10 variables:
#>  $ Cl.thickness   : Ord.factor w/ 10 levels "1"<"2"<"3"<"4"<..: 5 5 3 6 4 8 1 2 2 4 ...
#>  $ Cell.size      : Ord.factor w/ 10 levels "1"<"2"<"3"<"4"<..: 1 4 1 8 1 10 1 1 1 2 ...
#>  $ Cell.shape     : Ord.factor w/ 10 levels "1"<"2"<"3"<"4"<..: 1 4 1 8 1 10 1 2 1 1 ...
#>  $ Marg.adhesion  : Ord.factor w/ 10 levels "1"<"2"<"3"<"4"<..: 1 5 1 1 3 8 1 1 1 1 ...
#>  $ Epith.c.size   : Ord.factor w/ 10 levels "1"<"2"<"3"<"4"<..: 2 7 2 3 2 7 2 2 2 2 ...
#>  $ Bare.nuclei    : Factor w/ 10 levels "1","2","3","4",..: 1 10 2 4 1 10 10 1 1 1 ...
#>  $ Bl.cromatin    : Factor w/ 10 levels "1","2","3","4",..: 3 3 3 3 3 9 3 3 1 2 ...
#>  $ Normal.nucleoli: Factor w/ 10 levels "1","2","3","4",..: 1 2 1 7 1 7 1 1 1 1 ...
#>  $ Mitoses        : Factor w/ 9 levels "1","2","3","4",..: 1 1 1 1 1 1 1 1 5 1 ...
#>  $ Class          : Factor w/ 2 levels "benign","malignant": 1 1 1 1 1 2 1 1 1 1 ...

### Get the names of the input variables in cluster.task
getTaskFeatureNames(cluster.task)
#>  [1] "mpg"  "cyl"  "disp" "hp"   "drat" "wt"   "qsec" "vs"   "am"   "gear"
#> [11] "carb"

### Get the values of the target variables in surv.task
head(getTaskTargets(surv.task))
#>   time status
#> 1  306   TRUE
#> 2  455   TRUE
#> 3 1010  FALSE
#> 4  210   TRUE
#> 5  883   TRUE
#> 6 1022  FALSE

### Get the cost matrix in costsens.task
head(getTaskCosts(costsens.task))
#>      y1        y2         y3
#> [1,]  0 1589.5664  674.44434
#> [2,]  0 1173.4364  828.40682
#> [3,]  0  942.7611 1095.33713
#> [4,]  0 1049.5562  477.82496
#> [5,]  0 1121.8899   90.85237
#> [6,]  0 1819.9830  841.06686
\end{lstlisting}

Note that
\href{http://www.rdocumentation.org/packages/mlr/functions/getTaskData.html}{getTaskData}
offers many options for converting the data set into a convenient
format. This especially comes in handy when you
\protect\hyperlink{integrating-another-learner}{integrate a new learner}
from another \textbf{R} package into
\href{http://www.rdocumentation.org/packages/mlr/}{mlr}. In this regard
function
\href{http://www.rdocumentation.org/packages/mlr/functions/getTaskFormula.html}{getTaskFormula}
is also useful.

\subsubsection{Modifying a learning
task}\label{modifying-a-learning-task}

\href{http://www.rdocumentation.org/packages/mlr/}{mlr} provides several
functions to alter an existing
\href{http://www.rdocumentation.org/packages/mlr/functions/Task.html}{Task},
which is often more convenient than creating a new
\href{http://www.rdocumentation.org/packages/mlr/functions/Task.html}{Task}
from scratch. Here are some examples.

\begin{lstlisting}[language=R]
### Select observations and/or features
cluster.task = subsetTask(cluster.task, subset = 4:17)

### It may happen, especially after selecting observations, that features are constant.
### These should be removed.
removeConstantFeatures(cluster.task)
#> Removing 1 columns: am
#> Unsupervised task: mtcars
#> Type: cluster
#> Observations: 14
#> Features:
#> numerics  factors  ordered 
#>       10        0        0 
#> Missings: FALSE
#> Has weights: FALSE
#> Has blocking: FALSE

### Remove selected features
dropFeatures(surv.task, c("meal.cal", "wt.loss"))
#> Supervised task: lung
#> Type: surv
#> Target: time,status
#> Observations: 228
#> Features:
#> numerics  factors  ordered 
#>        6        0        0 
#> Missings: TRUE
#> Has weights: FALSE
#> Has blocking: FALSE

### Standardize numerical features
task = normalizeFeatures(cluster.task, method = "range")
summary(getTaskData(task))
#>       mpg              cyl              disp              hp        
#>  Min.   :0.0000   Min.   :0.0000   Min.   :0.0000   Min.   :0.0000  
#>  1st Qu.:0.3161   1st Qu.:0.5000   1st Qu.:0.1242   1st Qu.:0.2801  
#>  Median :0.5107   Median :1.0000   Median :0.4076   Median :0.6311  
#>  Mean   :0.4872   Mean   :0.7143   Mean   :0.4430   Mean   :0.5308  
#>  3rd Qu.:0.6196   3rd Qu.:1.0000   3rd Qu.:0.6618   3rd Qu.:0.7473  
#>  Max.   :1.0000   Max.   :1.0000   Max.   :1.0000   Max.   :1.0000  
#>       drat              wt              qsec              vs        
#>  Min.   :0.0000   Min.   :0.0000   Min.   :0.0000   Min.   :0.0000  
#>  1st Qu.:0.2672   1st Qu.:0.1275   1st Qu.:0.2302   1st Qu.:0.0000  
#>  Median :0.3060   Median :0.1605   Median :0.3045   Median :0.0000  
#>  Mean   :0.4544   Mean   :0.3268   Mean   :0.3752   Mean   :0.4286  
#>  3rd Qu.:0.7026   3rd Qu.:0.3727   3rd Qu.:0.4908   3rd Qu.:1.0000  
#>  Max.   :1.0000   Max.   :1.0000   Max.   :1.0000   Max.   :1.0000  
#>        am           gear             carb       
#>  Min.   :0.5   Min.   :0.0000   Min.   :0.0000  
#>  1st Qu.:0.5   1st Qu.:0.0000   1st Qu.:0.3333  
#>  Median :0.5   Median :0.0000   Median :0.6667  
#>  Mean   :0.5   Mean   :0.2857   Mean   :0.6429  
#>  3rd Qu.:0.5   3rd Qu.:0.7500   3rd Qu.:1.0000  
#>  Max.   :0.5   Max.   :1.0000   Max.   :1.0000
\end{lstlisting}

For more functions and more detailed explanations have a look at the
\protect\hyperlink{data-preprocessing}{data preprocessing} page.

\subsubsection{Example tasks and convenience
functions}\label{example-tasks-and-convenience-functions}

For your convenience
\href{http://www.rdocumentation.org/packages/mlr/}{mlr} provides
pre-defined
\href{http://www.rdocumentation.org/packages/mlr/functions/Task.html}{Task}s
for each type of learning problem. These are also used throughout this
tutorial in order to get shorter and more readable code. A list of all
\href{http://www.rdocumentation.org/packages/mlr/functions/Task.html}{Task}s
can be found in the \protect\hyperlink{example-tasks}{Appendix}.

Moreover, \href{http://www.rdocumentation.org/packages/mlr/}{mlr}'s
function
\href{http://www.rdocumentation.org/packages/mlr/functions/convertMLBenchObjToTask.html}{convertMLBenchObjToTask}
can generate
\href{http://www.rdocumentation.org/packages/mlr/functions/Task.html}{Task}s
from the data sets and data generating functions in package
\href{http://www.rdocumentation.org/packages/mlbench/}{mlbench}.

\hypertarget{learners}{\subsection{Learners}\label{learners}}

The following classes provide a unified interface to all popular machine
learning methods in \textbf{R}: (cost-sensitive) classification,
regression, survival analysis, and clustering. Many are already
integrated in \href{http://www.rdocumentation.org/packages/mlr/}{mlr},
others are not, but the package is specifically designed to make
extensions simple.

Section \protect\hyperlink{integrated-learners}{integrated learners}
shows the already implemented machine learning methods and their
properties. If your favorite method is missing, either
\href{https://github.com/mlr-org/mlr/issues}{open an issue} or take a
look at \protect\hyperlink{integrating-another-learner}{how to integrate
a learning method yourself}. This basic introduction demonstrates how to
use already implemented learners.

\subsubsection{Constructing a learner}\label{constructing-a-learner}

A learner in \href{http://www.rdocumentation.org/packages/mlr/}{mlr} is
generated by calling
\href{http://www.rdocumentation.org/packages/mlr/functions/makeLearner.html}{makeLearner}.
In the constructor you need to specify which learning method you want to
use. Moreover, you can:

\begin{itemize}
\tightlist
\item
  Set hyperparameters.
\item
  Control the output for later prediction, e.g., for classification
  whether you want a factor of predicted class labels or probabilities.
\item
  Set an ID to name the object (some methods will later use this ID to
  name results or annotate plots).
\end{itemize}

\begin{lstlisting}[language=R]
### Classification tree, set it up for predicting probabilities
classif.lrn = makeLearner("classif.randomForest", predict.type = "prob", fix.factors.prediction = TRUE)

### Regression gradient boosting machine, specify hyperparameters via a list
regr.lrn = makeLearner("regr.gbm", par.vals = list(n.trees = 500, interaction.depth = 3))

### Cox proportional hazards model with custom name
surv.lrn = makeLearner("surv.coxph", id = "cph")

### K-means with 5 clusters
cluster.lrn = makeLearner("cluster.kmeans", centers = 5)

### Multilabel Random Ferns classification algorithm
multilabel.lrn = makeLearner("multilabel.rFerns")
\end{lstlisting}

The first argument specifies which algorithm to use. The naming
convention is \lstinline!classif.<R_method_name>! for classification
methods, \lstinline!regr.<R_method_name>! for regression methods,
\lstinline!surv.<R_method_name>! for survival analysis,
\lstinline!cluster.<R_method_name>! for clustering methods, and
\lstinline!multilabel.<R_method_name>! for multilabel classification.

Hyperparameter values can be specified either via the \lstinline!...!
argument or as a
\href{http://www.rdocumentation.org/packages/base/functions/list.html}{list}
via \lstinline!par.vals!.

Occasionally,
\href{http://www.rdocumentation.org/packages/base/functions/factor.html}{factor}
features may cause problems when fewer levels are present in the test
data set than in the training data. By setting
\lstinline!fix.factors.prediction = TRUE! these are avoided by adding a
factor level for missing data in the test data set.

Let's have a look at two of the learners created above.

\begin{lstlisting}[language=R]
classif.lrn
#> Learner classif.randomForest from package randomForest
#> Type: classif
#> Name: Random Forest; Short name: rf
#> Class: classif.randomForest
#> Properties: twoclass,multiclass,numerics,factors,ordered,prob,class.weights,featimp
#> Predict-Type: prob
#> Hyperparameters:

surv.lrn
#> Learner cph from package survival
#> Type: surv
#> Name: Cox Proportional Hazard Model; Short name: coxph
#> Class: surv.coxph
#> Properties: numerics,factors,weights,rcens
#> Predict-Type: response
#> Hyperparameters:
\end{lstlisting}

All generated learners are objects of class
\href{http://www.rdocumentation.org/packages/mlr/functions/makeLearner.html}{Learner}.
This class contains the properties of the method, e.g., which types of
features it can handle, what kind of output is possible during
prediction, and whether multi-class problems, observations weights or
missing values are supported.

As you might have noticed, there is currently no special learner class
for cost-sensitive classification. For ordinary misclassification costs
you can use standard classification methods. For example-dependent costs
there are several ways to generate cost-sensitive learners from ordinary
regression and classification learners. This is explained in greater
detail in the section about
\protect\hyperlink{cost-sensitive-classification}{cost-sensitive
classification}.

\subsubsection{Accessing a learner}\label{accessing-a-learner}

The
\href{http://www.rdocumentation.org/packages/mlr/functions/makeLearner.html}{Learner}
object is a
\href{http://www.rdocumentation.org/packages/base/functions/list.html}{list}
and the following elements contain information regarding the
hyperparameters and the type of prediction.

\begin{lstlisting}[language=R]
### Get the configured hyperparameter settings that deviate from the defaults
cluster.lrn$par.vals
#> $centers
#> [1] 5

### Get the set of hyperparameters
classif.lrn$par.set
#>                      Type  len   Def   Constr Req Tunable Trafo
#> ntree             integer    -   500 1 to Inf   -    TRUE     -
#> mtry              integer    -     - 1 to Inf   -    TRUE     -
#> replace           logical    -  TRUE        -   -    TRUE     -
#> classwt     numericvector <NA>     - 0 to Inf   -    TRUE     -
#> cutoff      numericvector <NA>     -   0 to 1   -    TRUE     -
#> strata            untyped    -     -        -   -    TRUE     -
#> sampsize    integervector <NA>     - 1 to Inf   -    TRUE     -
#> nodesize          integer    -     1 1 to Inf   -    TRUE     -
#> maxnodes          integer    -     - 1 to Inf   -    TRUE     -
#> importance        logical    - FALSE        -   -    TRUE     -
#> localImp          logical    - FALSE        -   -    TRUE     -
#> proximity         logical    - FALSE        -   -   FALSE     -
#> oob.prox          logical    -     -        -   Y   FALSE     -
#> norm.votes        logical    -  TRUE        -   -   FALSE     -
#> do.trace          logical    - FALSE        -   -   FALSE     -
#> keep.forest       logical    -  TRUE        -   -   FALSE     -
#> keep.inbag        logical    - FALSE        -   -   FALSE     -

### Get the type of prediction
regr.lrn$predict.type
#> [1] "response"
\end{lstlisting}

Slot \lstinline!$par.set! is an object of class
\href{http://www.rdocumentation.org/packages/ParamHelpers/functions/makeParamSet.html}{ParamSet}.
It contains, among others, the type of hyperparameters (e.g., numeric,
logical), potential default values and the range of allowed values.

Moreover, \href{http://www.rdocumentation.org/packages/mlr/}{mlr}
provides function
\href{http://www.rdocumentation.org/packages/mlr/functions/getHyperPars.html}{getHyperPars}
to access the current hyperparameter setting of a
\href{http://www.rdocumentation.org/packages/mlr/functions/makeLearner.html}{Learner}
and
\href{http://www.rdocumentation.org/packages/mlr/functions/getParamSet.html}{getParamSet}
to get a description of all possible settings. These are particularly
useful in case of wrapped
\href{http://www.rdocumentation.org/packages/mlr/functions/makeLearner.html}{Learner}s,
for example if a learner is fused with a feature selection strategy, and
both, the learner as well the feature selection method, have
hyperparameters. For details see the section on
\protect\hyperlink{wrapper}{wrapped learners}.

\begin{lstlisting}[language=R]
### Get current hyperparameter settings
getHyperPars(cluster.lrn)
#> $centers
#> [1] 5

### Get a description of all possible hyperparameter settings
getParamSet(classif.lrn)
#>                      Type  len   Def   Constr Req Tunable Trafo
#> ntree             integer    -   500 1 to Inf   -    TRUE     -
#> mtry              integer    -     - 1 to Inf   -    TRUE     -
#> replace           logical    -  TRUE        -   -    TRUE     -
#> classwt     numericvector <NA>     - 0 to Inf   -    TRUE     -
#> cutoff      numericvector <NA>     -   0 to 1   -    TRUE     -
#> strata            untyped    -     -        -   -    TRUE     -
#> sampsize    integervector <NA>     - 1 to Inf   -    TRUE     -
#> nodesize          integer    -     1 1 to Inf   -    TRUE     -
#> maxnodes          integer    -     - 1 to Inf   -    TRUE     -
#> importance        logical    - FALSE        -   -    TRUE     -
#> localImp          logical    - FALSE        -   -    TRUE     -
#> proximity         logical    - FALSE        -   -   FALSE     -
#> oob.prox          logical    -     -        -   Y   FALSE     -
#> norm.votes        logical    -  TRUE        -   -   FALSE     -
#> do.trace          logical    - FALSE        -   -   FALSE     -
#> keep.forest       logical    -  TRUE        -   -   FALSE     -
#> keep.inbag        logical    - FALSE        -   -   FALSE     -
\end{lstlisting}

We can also use
\href{http://www.rdocumentation.org/packages/mlr/functions/getParamSet.html}{getParamSet}
to get a quick overview about the available hyperparameters and defaults
of a learning method without explicitly constructing it (by calling
\href{http://www.rdocumentation.org/packages/mlr/functions/makeLearner.html}{makeLearner}).

\begin{lstlisting}[language=R]
getParamSet("classif.randomForest")
#>                      Type  len   Def   Constr Req Tunable Trafo
#> ntree             integer    -   500 1 to Inf   -    TRUE     -
#> mtry              integer    -     - 1 to Inf   -    TRUE     -
#> replace           logical    -  TRUE        -   -    TRUE     -
#> classwt     numericvector <NA>     - 0 to Inf   -    TRUE     -
#> cutoff      numericvector <NA>     -   0 to 1   -    TRUE     -
#> strata            untyped    -     -        -   -    TRUE     -
#> sampsize    integervector <NA>     - 1 to Inf   -    TRUE     -
#> nodesize          integer    -     1 1 to Inf   -    TRUE     -
#> maxnodes          integer    -     - 1 to Inf   -    TRUE     -
#> importance        logical    - FALSE        -   -    TRUE     -
#> localImp          logical    - FALSE        -   -    TRUE     -
#> proximity         logical    - FALSE        -   -   FALSE     -
#> oob.prox          logical    -     -        -   Y   FALSE     -
#> norm.votes        logical    -  TRUE        -   -   FALSE     -
#> do.trace          logical    - FALSE        -   -   FALSE     -
#> keep.forest       logical    -  TRUE        -   -   FALSE     -
#> keep.inbag        logical    - FALSE        -   -   FALSE     -
\end{lstlisting}

\subsubsection{Modifying a learner}\label{modifying-a-learner}

There are also some functions that enable you to change certain aspects
of a
\href{http://www.rdocumentation.org/packages/mlr/functions/makeLearner.html}{Learner}
without needing to create a new
\href{http://www.rdocumentation.org/packages/mlr/functions/makeLearner.html}{Learner}
from scratch. Here are some examples.

\begin{lstlisting}[language=R]
### Change the ID
surv.lrn = setLearnerId(surv.lrn, "CoxModel")
surv.lrn
#> Learner CoxModel from package survival
#> Type: surv
#> Name: Cox Proportional Hazard Model; Short name: coxph
#> Class: surv.coxph
#> Properties: numerics,factors,weights,rcens
#> Predict-Type: response
#> Hyperparameters:

### Change the prediction type, predict a factor with class labels instead of probabilities
classif.lrn = setPredictType(classif.lrn, "response")

### Change hyperparameter values
cluster.lrn = setHyperPars(cluster.lrn, centers = 4)

### Go back to default hyperparameter values
regr.lrn = removeHyperPars(regr.lrn, c("n.trees", "interaction.depth"))
\end{lstlisting}

\subsubsection{Listing learners}\label{listing-learners}

A list of all learners integrated in
\href{http://www.rdocumentation.org/packages/mlr/}{mlr} and their
respective properties is shown in the
\protect\hyperlink{integrated-learners}{Appendix}.

If you would like a list of available learners, maybe only with certain
properties or suitable for a certain learning
\href{http://www.rdocumentation.org/packages/mlr/functions/Task.html}{Task}
use function
\href{http://www.rdocumentation.org/packages/mlr/functions/listLearners.html}{listLearners}.

\begin{lstlisting}[language=R]
### List everything in mlr
lrns = listLearners()
head(lrns[c("class", "package")])
#>                 class      package
#> 1         classif.ada          ada
#> 2      classif.avNNet         nnet
#> 3 classif.bartMachine  bartMachine
#> 4         classif.bdk      kohonen
#> 5    classif.binomial        stats
#> 6  classif.blackboost mboost,party

### List classifiers that can output probabilities
lrns = listLearners("classif", properties = "prob")
head(lrns[c("class", "package")])
#>                 class      package
#> 1         classif.ada          ada
#> 2      classif.avNNet         nnet
#> 3 classif.bartMachine  bartMachine
#> 4         classif.bdk      kohonen
#> 5    classif.binomial        stats
#> 6  classif.blackboost mboost,party

### List classifiers that can be applied to iris (i.e., multiclass) and output probabilities
lrns = listLearners(iris.task, properties = "prob")
head(lrns[c("class", "package")])
#>              class      package
#> 1   classif.avNNet         nnet
#> 2      classif.bdk      kohonen
#> 3 classif.boosting adabag,rpart
#> 4      classif.C50          C50
#> 5  classif.cforest        party
#> 6    classif.ctree        party

### The calls above return character vectors, but you can also create learner objects
head(listLearners("cluster", create = TRUE), 2)
#> [[1]]
#> Learner cluster.cmeans from package e1071,clue
#> Type: cluster
#> Name: Fuzzy C-Means Clustering; Short name: cmeans
#> Class: cluster.cmeans
#> Properties: numerics,prob
#> Predict-Type: response
#> Hyperparameters: centers=2
#> 
#> 
#> [[2]]
#> Learner cluster.Cobweb from package RWeka
#> Type: cluster
#> Name: Cobweb Clustering Algorithm; Short name: cobweb
#> Class: cluster.Cobweb
#> Properties: numerics
#> Predict-Type: response
#> Hyperparameters:
\end{lstlisting}

\hypertarget{training-a-learner}{\subsection{Training a
Learner}\label{training-a-learner}}

Training a learner means fitting a model to a given data set. In
\href{http://www.rdocumentation.org/packages/mlr/}{mlr} this can be done
by calling function
\href{http://www.rdocumentation.org/packages/mlr/functions/train.html}{train}
on a
\href{http://www.rdocumentation.org/packages/mlr/functions/makeLearner.html}{Learner}
and a suitable
\href{http://www.rdocumentation.org/packages/mlr/functions/Task.html}{Task}.

We start with a classification example and perform a
\href{http://www.rdocumentation.org/packages/MASS/functions/lda.html}{linear
discriminant analysis} on the
\href{http://www.rdocumentation.org/packages/datasets/functions/iris.html}{iris}
data set.

\begin{lstlisting}[language=R]
### Generate the task
task = makeClassifTask(data = iris, target = "Species")

### Generate the learner
lrn = makeLearner("classif.lda")

### Train the learner
mod = train(lrn, task)
mod
#> Model for learner.id=classif.lda; learner.class=classif.lda
#> Trained on: task.id = iris; obs = 150; features = 4
#> Hyperparameters:
\end{lstlisting}

In the above example creating the
\href{http://www.rdocumentation.org/packages/mlr/functions/makeLearner.html}{Learner}
explicitly is not absolutely necessary. As a general rule, you have to
generate the
\href{http://www.rdocumentation.org/packages/mlr/functions/makeLearner.html}{Learner}
yourself if you want to change any defaults, e.g., setting
hyperparameter values or altering the predict type. Otherwise,
\href{http://www.rdocumentation.org/packages/mlr/functions/train.html}{train}
and many other functions also accept the class name of the learner and
call
\href{http://www.rdocumentation.org/packages/mlr/functions/makeLearner.html}{makeLearner}
internally with default settings.

\begin{lstlisting}[language=R]
mod = train("classif.lda", task)
mod
#> Model for learner.id=classif.lda; learner.class=classif.lda
#> Trained on: task.id = iris; obs = 150; features = 4
#> Hyperparameters:
\end{lstlisting}

Training a learner works the same way for every type of learning
problem. Below is a survival analysis example where a
\href{http://www.rdocumentation.org/packages/survival/functions/coxph.html}{Cox
proportional hazards model} is fitted to the
\href{http://www.rdocumentation.org/packages/survival/functions/lung.html}{lung}
data set. Note that we use the corresponding
\href{http://www.rdocumentation.org/packages/mlr/functions/lung.task.html}{lung.task}
provided by \href{http://www.rdocumentation.org/packages/mlr/}{mlr}. All
available
\href{http://www.rdocumentation.org/packages/mlr/functions/Task.html}{Task}s
are listed in the \protect\hyperlink{example-tasks}{Appendix}.

\begin{lstlisting}[language=R]
mod = train("surv.coxph", lung.task)
mod
#> Model for learner.id=surv.coxph; learner.class=surv.coxph
#> Trained on: task.id = lung-example; obs = 167; features = 8
#> Hyperparameters:
\end{lstlisting}

\subsubsection{Accessing learner models}\label{accessing-learner-models}

Function
\href{http://www.rdocumentation.org/packages/mlr/functions/train.html}{train}
returns an object of class
\href{http://www.rdocumentation.org/packages/mlr/functions/makeWrappedModel.html}{WrappedModel},
which encapsulates the fitted model, i.e., the output of the underlying
\textbf{R} learning method. Additionally, it contains some information
about the
\href{http://www.rdocumentation.org/packages/mlr/functions/makeLearner.html}{Learner},
the
\href{http://www.rdocumentation.org/packages/mlr/functions/Task.html}{Task},
the features and observations used for training, and the training time.
A
\href{http://www.rdocumentation.org/packages/mlr/functions/makeWrappedModel.html}{WrappedModel}
can subsequently be used to make a
\href{http://www.rdocumentation.org/packages/mlr/functions/predict.WrappedModel.html}{prediction}
for new observations.

The fitted model in slot \lstinline!$learner.model! of the
\href{http://www.rdocumentation.org/packages/mlr/functions/makeWrappedModel.html}{WrappedModel}
object can be accessed using function
\href{http://www.rdocumentation.org/packages/mlr/functions/getLearnerModel.html}{getLearnerModel}.

In the following example we cluster the
\href{http://www.rdocumentation.org/packages/cluster/functions/ruspini.html}{Ruspini}
data set (which has four groups and two features) by \(K\)-means with
\(K = 4\) and extract the output of the underlying
\href{http://www.rdocumentation.org/packages/stats/functions/kmeans.html}{kmeans}
function.

\begin{lstlisting}[language=R]
data(ruspini, package = "cluster")
plot(y ~ x, ruspini)
\end{lstlisting}

\includegraphics{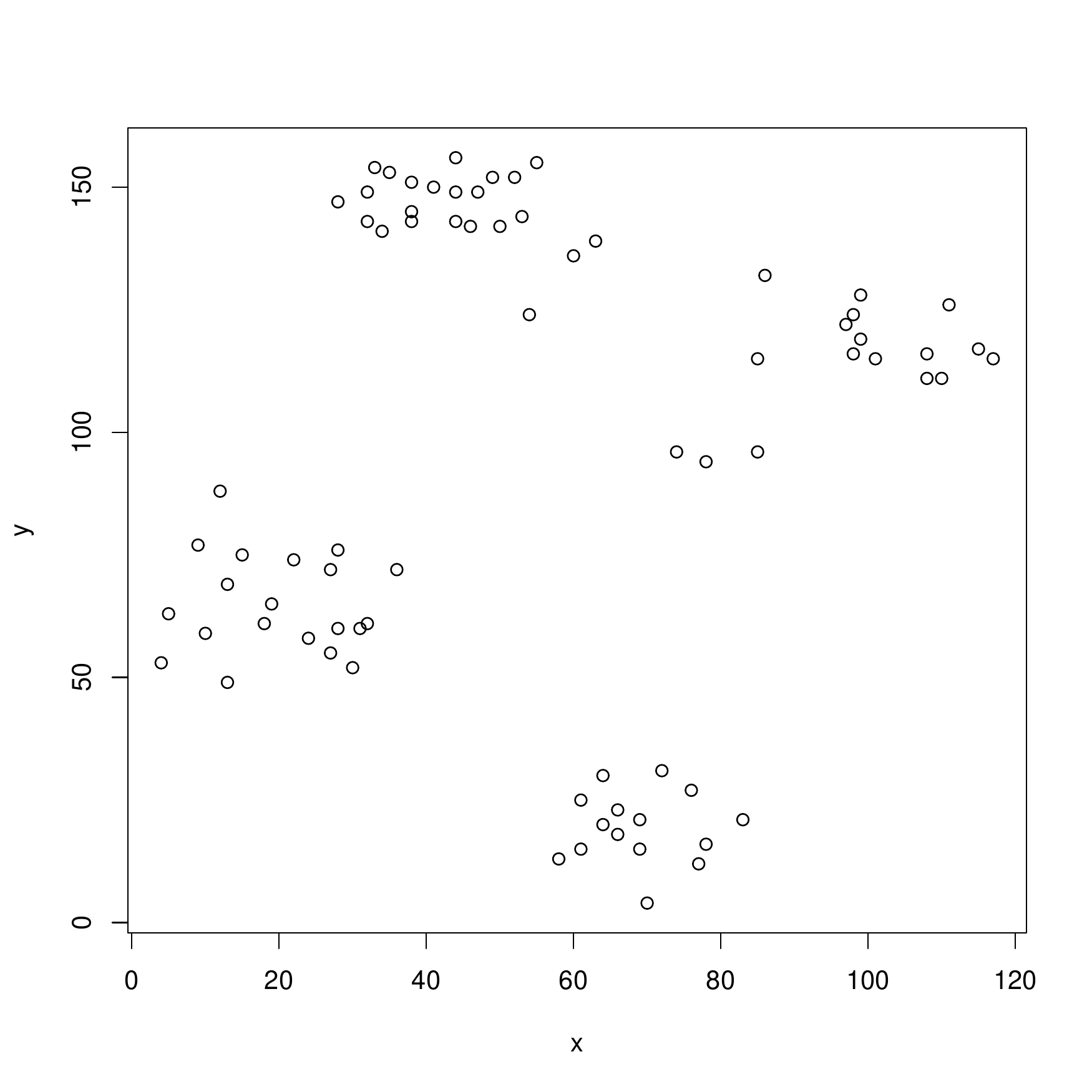}

\begin{lstlisting}[language=R]
### Generate the task
ruspini.task = makeClusterTask(data = ruspini)

### Generate the learner
lrn = makeLearner("cluster.kmeans", centers = 4)

### Train the learner
mod = train(lrn, ruspini.task)
mod
#> Model for learner.id=cluster.kmeans; learner.class=cluster.kmeans
#> Trained on: task.id = ruspini; obs = 75; features = 2
#> Hyperparameters: centers=4

### Peak into mod
names(mod)
#> [1] "learner"       "learner.model" "task.desc"     "subset"       
#> [5] "features"      "factor.levels" "time"

mod$learner
#> Learner cluster.kmeans from package stats,clue
#> Type: cluster
#> Name: K-Means; Short name: kmeans
#> Class: cluster.kmeans
#> Properties: numerics,prob
#> Predict-Type: response
#> Hyperparameters: centers=4

mod$features
#> [1] "x" "y"

mod$time
#> [1] 0.001

### Extract the fitted model
getLearnerModel(mod)
#> K-means clustering with 4 clusters of sizes 23, 17, 15, 20
#> 
#> Cluster means:
#>          x        y
#> 1 43.91304 146.0435
#> 2 98.17647 114.8824
#> 3 68.93333  19.4000
#> 4 20.15000  64.9500
#> 
#> Clustering vector:
#>  1  2  3  4  5  6  7  8  9 10 11 12 13 14 15 16 17 18 19 20 21 22 23 24 25 
#>  4  4  4  4  4  4  4  4  4  4  4  4  4  4  4  4  4  4  4  4  1  1  1  1  1 
#> 26 27 28 29 30 31 32 33 34 35 36 37 38 39 40 41 42 43 44 45 46 47 48 49 50 
#>  1  1  1  1  1  1  1  1  1  1  1  1  1  1  1  1  1  1  2  2  2  2  2  2  2 
#> 51 52 53 54 55 56 57 58 59 60 61 62 63 64 65 66 67 68 69 70 71 72 73 74 75 
#>  2  2  2  2  2  2  2  2  2  2  3  3  3  3  3  3  3  3  3  3  3  3  3  3  3 
#> 
#> Within cluster sum of squares by cluster:
#> [1] 3176.783 4558.235 1456.533 3689.500
#>  (between_SS / total_SS =  94.7 %)
#> 
#> Available components:
#> 
#> [1] "cluster"      "centers"      "totss"        "withinss"    
#> [5] "tot.withinss" "betweenss"    "size"         "iter"        
#> [9] "ifault"
\end{lstlisting}

\subsubsection{Further options and
comments}\label{further-options-and-comments}

By default, the whole data set in the
\href{http://www.rdocumentation.org/packages/mlr/functions/Task.html}{Task}
is used for training. The \lstinline!subset! argument of
\href{http://www.rdocumentation.org/packages/mlr/functions/train.html}{train}
takes a logical or integer vector that indicates which observations to
use, for example if you want to split your data into a training and a
test set or if you want to fit separate models to different subgroups in
the data.

Below we fit a
\href{http://www.rdocumentation.org/packages/stats/functions/lm.html}{linear
regression model} to the
\href{http://www.rdocumentation.org/packages/mlbench/functions/BostonHousing.html}{BostonHousing}
data set
(\href{http://www.rdocumentation.org/packages/mlr/functions/bh.task.html}{bh.task})
and randomly select 1/3 of the data set for training.

\begin{lstlisting}[language=R]
### Get the number of observations
n = getTaskSize(bh.task)

### Use 1/3 of the observations for training
train.set = sample(n, size = n/3)

### Train the learner
mod = train("regr.lm", bh.task, subset = train.set)
mod
#> Model for learner.id=regr.lm; learner.class=regr.lm
#> Trained on: task.id = BostonHousing-example; obs = 168; features = 13
#> Hyperparameters:
\end{lstlisting}

Note, for later, that all standard
\protect\hyperlink{resampling}{resampling strategies} are supported.
Therefore you usually do not have to subset the data yourself.

Moreover, if the learner supports this, you can specify observation
\lstinline!weights! that reflect the relevance of observations in the
training process. Weights can be useful in many regards, for example to
express the reliability of the training observations, reduce the
influence of outliers or, if the data were collected over a longer time
period, increase the influence of recent data. In supervised
classification weights can be used to incorporate misclassification
costs or account for class imbalance.

For example in the
\href{http://www.rdocumentation.org/packages/mlbench/functions/BreastCancer.html}{BreastCancer}
data set class \lstinline!benign! is almost twice as frequent as class
\lstinline!malignant!. In order to grant both classes equal importance
in training the classifier we can weight the examples according to the
inverse class frequencies in the data set as shown in the following
\textbf{R} code.

\begin{lstlisting}[language=R]
### Calculate the observation weights
target = getTaskTargets(bc.task)
tab = as.numeric(table(target))
w = 1/tab[target]

train("classif.rpart", task = bc.task, weights = w)
#> Model for learner.id=classif.rpart; learner.class=classif.rpart
#> Trained on: task.id = BreastCancer-example; obs = 683; features = 9
#> Hyperparameters: xval=0
\end{lstlisting}

Note, for later, that
\href{http://www.rdocumentation.org/packages/mlr/}{mlr} offers much more
functionality to deal with
\protect\hyperlink{imbalanced-classification-problems}{imbalanced
classification problems}.

As another side remark for more advanced readers: By varying the weights
in the calls to
\href{http://www.rdocumentation.org/packages/mlr/functions/train.html}{train},
you could also implement your own variant of a general boosting type
algorithm on arbitrary
\href{http://www.rdocumentation.org/packages/mlr/}{mlr} base learners.

As you may recall, it is also possible to set observation weights when
creating the
\href{http://www.rdocumentation.org/packages/mlr/functions/Task.html}{Task}.
As a general rule, you should specify them in
\href{http://www.rdocumentation.org/packages/mlr/functions/Task.html}{make*Task}
if the weights really ``belong'' to the task and always should be used.
Otherwise, pass them to
\href{http://www.rdocumentation.org/packages/mlr/functions/train.html}{train}.
The weights in
\href{http://www.rdocumentation.org/packages/mlr/functions/train.html}{train}
take precedence over the weights in
\href{http://www.rdocumentation.org/packages/mlr/functions/Task.html}{Task}.

\hypertarget{predicting-outcomes-for-new-data}{\subsection{Predicting
Outcomes for New Data}\label{predicting-outcomes-for-new-data}}

Predicting the target values for new observations is implemented the
same way as most of the other predict methods in \textbf{R}. In general,
all you need to do is call
\href{http://www.rdocumentation.org/packages/mlr/functions/predict.WrappedModel.html}{predict}
on the object returned by
\href{http://www.rdocumentation.org/packages/mlr/functions/train.html}{train}
and pass the data you want predictions for.

There are two ways to pass the data:

\begin{itemize}
\tightlist
\item
  Either pass the
  \href{http://www.rdocumentation.org/packages/mlr/functions/Task.html}{Task}
  via the \lstinline!task! argument or
\item
  pass a
  \href{http://www.rdocumentation.org/packages/base/functions/data.frame.html}{data
  frame} via the \lstinline!newdata! argument.
\end{itemize}

The first way is preferable if you want predictions for data already
included in a
\href{http://www.rdocumentation.org/packages/mlr/functions/Task.html}{Task}.

Just as
\href{http://www.rdocumentation.org/packages/mlr/functions/train.html}{train},
the
\href{http://www.rdocumentation.org/packages/mlr/functions/predict.WrappedModel.html}{predict}
function has a \lstinline!subset! argument, so you can set aside
different portions of the data in
\href{http://www.rdocumentation.org/packages/mlr/functions/Task.html}{Task}
for training and prediction (more advanced methods for splitting the
data in train and test set are described in the
\protect\hyperlink{resampling}{section on resampling}).

In the following example we fit a
\href{http://www.rdocumentation.org/packages/gbm/functions/gbm.html}{gradient
boosting machine} to every second observation of the
\href{http://www.rdocumentation.org/packages/mlbench/functions/BostonHousing.html}{BostonHousing}
data set and make predictions on the remaining data in
\href{http://www.rdocumentation.org/packages/mlr/functions/bh.task.html}{bh.task}.

\begin{lstlisting}[language=R]
n = getTaskSize(bh.task)
train.set = seq(1, n, by = 2)
test.set = seq(2, n, by = 2)
lrn = makeLearner("regr.gbm", n.trees = 100)
mod = train(lrn, bh.task, subset = train.set)

task.pred = predict(mod, task = bh.task, subset = test.set)
task.pred
#> Prediction: 253 observations
#> predict.type: response
#> threshold: 
#> time: 0.00
#>    id truth response
#> 2   2  21.6 22.28539
#> 4   4  33.4 23.33968
#> 6   6  28.7 22.40896
#> 8   8  27.1 22.12750
#> 10 10  18.9 22.12750
#> 12 12  18.9 22.12750
#> ... (253 rows, 3 cols)
\end{lstlisting}

The second way is useful if you want to predict data not included in the
\href{http://www.rdocumentation.org/packages/mlr/functions/Task.html}{Task}.

Here we cluster the \lstinline!iris! data set without the target
variable. All observations with an odd index are included in the
\href{http://www.rdocumentation.org/packages/mlr/functions/Task.html}{Task}
and used for training. Predictions are made for the remaining
observations.

\begin{lstlisting}[language=R]
n = nrow(iris)
iris.train = iris[seq(1, n, by = 2), -5]
iris.test = iris[seq(2, n, by = 2), -5]
task = makeClusterTask(data = iris.train)
mod = train("cluster.kmeans", task)

newdata.pred = predict(mod, newdata = iris.test)
newdata.pred
#> Prediction: 75 observations
#> predict.type: response
#> threshold: 
#> time: 0.00
#>    response
#> 2         2
#> 4         2
#> 6         2
#> 8         2
#> 10        2
#> 12        2
#> ... (75 rows, 1 cols)
\end{lstlisting}

Note that for supervised learning you do not have to remove the target
columns from the data. These columns are automatically removed prior to
calling the underlying \lstinline!predict! method of the learner.

\subsubsection{Accessing the prediction}\label{accessing-the-prediction}

Function
\href{http://www.rdocumentation.org/packages/mlr/functions/predict.html}{predict}
returns a named
\href{http://www.rdocumentation.org/packages/base/functions/list.html}{list}
of class
\href{http://www.rdocumentation.org/packages/mlr/functions/Prediction.html}{Prediction}.
Its most important element is \lstinline!$data! which is a
\href{http://www.rdocumentation.org/packages/base/functions/data.frame.html}{data
frame} that contains columns with the true values of the target variable
(in case of supervised learning problems) and the predictions. Use
\href{http://www.rdocumentation.org/packages/mlr/functions/Prediction.html}{as.data.frame}
for direct access.

In the following the predictions on the
\href{http://www.rdocumentation.org/packages/mlbench/functions/BostonHousing.html}{BostonHousing}
and the
\href{http://www.rdocumentation.org/packages/datasets/functions/iris.html}{iris}
data sets are shown. As you may recall, the predictions in the first
case were made from a
\href{http://www.rdocumentation.org/packages/mlr/functions/Task.html}{Task}
and in the second case from a
\href{http://www.rdocumentation.org/packages/base/functions/data.frame.html}{data
frame}.

\begin{lstlisting}[language=R]
### Result of predict with data passed via task argument
head(as.data.frame(task.pred))
#>    id truth response
#> 2   2  21.6 22.28539
#> 4   4  33.4 23.33968
#> 6   6  28.7 22.40896
#> 8   8  27.1 22.12750
#> 10 10  18.9 22.12750
#> 12 12  18.9 22.12750

### Result of predict with data passed via newdata argument
head(as.data.frame(newdata.pred))
#>    response
#> 2         2
#> 4         2
#> 6         2
#> 8         2
#> 10        2
#> 12        2
\end{lstlisting}

As you can see when predicting from a
\href{http://www.rdocumentation.org/packages/mlr/functions/Task.html}{Task},
the resulting
\href{http://www.rdocumentation.org/packages/base/functions/data.frame.html}{data
frame} contains an additional column, called \lstinline!id!, which tells
us which element in the original data set the prediction corresponds to.

A direct way to access the true and predicted values of the target
variable(s) is provided by functions
\href{http://www.rdocumentation.org/packages/mlr/functions/getPredictionResponse.html}{getPredictionTruth}
and
\href{http://www.rdocumentation.org/packages/mlr/functions/getPredictionResponse.html}{getPredictionResponse}.

\begin{lstlisting}[language=R]
head(getPredictionTruth(task.pred))
#> [1] 21.6 33.4 28.7 27.1 18.9 18.9

head(getPredictionResponse(task.pred))
#> [1] 22.28539 23.33968 22.40896 22.12750 22.12750 22.12750
\end{lstlisting}

\paragraph{Extract Probabilities}\label{extract-probabilities}

The predicted probabilities can be extracted from the
\href{http://www.rdocumentation.org/packages/mlr/functions/Prediction.html}{Prediction}
using the function
\href{http://www.rdocumentation.org/packages/mlr/functions/getPredictionProbabilities.html}{getPredictionProbabilities}.
(Function
\href{http://www.rdocumentation.org/packages/mlr/functions/getProbabilities.html}{getProbabilities}
has been deprecated in favor of
\href{http://www.rdocumentation.org/packages/mlr/functions/getPredictionProbabilities.html}{getPredictionProbabilities}
in \href{http://www.rdocumentation.org/packages/mlr/}{mlr} version 2.5.)
Here is another cluster analysis example. We use
\href{http://www.rdocumentation.org/packages/e1071/functions/cmeans.html}{fuzzy
c-means clustering} on the
\href{http://www.rdocumentation.org/packages/datasets/functions/mtcars.html}{mtcars}
data set.

\begin{lstlisting}[language=R]
lrn = makeLearner("cluster.cmeans", predict.type = "prob")
mod = train(lrn, mtcars.task)

pred = predict(mod, task = mtcars.task)
head(getPredictionProbabilities(pred))
#>                            1           2
#> Mazda RX4         0.97959529 0.020404714
#> Mazda RX4 Wag     0.97963550 0.020364495
#> Datsun 710        0.99265984 0.007340164
#> Hornet 4 Drive    0.54292079 0.457079211
#> Hornet Sportabout 0.01870622 0.981293776
#> Valiant           0.75746556 0.242534444
\end{lstlisting}

For classification problems there are some more things worth mentioning.
By default, class labels are predicted.

\begin{lstlisting}[language=R]
### Linear discriminant analysis on the iris data set
mod = train("classif.lda", task = iris.task)

pred = predict(mod, task = iris.task)
pred
#> Prediction: 150 observations
#> predict.type: response
#> threshold: 
#> time: 0.00
#>   id  truth response
#> 1  1 setosa   setosa
#> 2  2 setosa   setosa
#> 3  3 setosa   setosa
#> 4  4 setosa   setosa
#> 5  5 setosa   setosa
#> 6  6 setosa   setosa
#> ... (150 rows, 3 cols)
\end{lstlisting}

In order to get predicted posterior probabilities we have to create a
\href{http://www.rdocumentation.org/packages/mlr/functions/makeLearner.html}{Learner}
with the appropriate \lstinline!predict.type!.

\begin{lstlisting}[language=R]
lrn = makeLearner("classif.rpart", predict.type = "prob")
mod = train(lrn, iris.task)

pred = predict(mod, newdata = iris)
head(as.data.frame(pred))
#>    truth prob.setosa prob.versicolor prob.virginica response
#> 1 setosa           1               0              0   setosa
#> 2 setosa           1               0              0   setosa
#> 3 setosa           1               0              0   setosa
#> 4 setosa           1               0              0   setosa
#> 5 setosa           1               0              0   setosa
#> 6 setosa           1               0              0   setosa
\end{lstlisting}

In addition to the probabilities, class labels are predicted by choosing
the class with the maximum probability and breaking ties at random.

As mentioned above, the predicted posterior probabilities can be
accessed via the
\href{http://www.rdocumentation.org/packages/mlr/functions/getPredictionProbabilities.html}{getPredictionProbabilities}
function.

\begin{lstlisting}[language=R]
head(getPredictionProbabilities(pred))
#>   setosa versicolor virginica
#> 1      1          0         0
#> 2      1          0         0
#> 3      1          0         0
#> 4      1          0         0
#> 5      1          0         0
#> 6      1          0         0
\end{lstlisting}

\paragraph{Confusion matrix}\label{confusion-matrix}

A confusion matrix can be obtained by calling
\href{http://www.rdocumentation.org/packages/mlr/functions/calculateConfusionMatrix.html}{calculateConfusionMatrix}.
The columns represent predicted and the rows true class labels.

\begin{lstlisting}[language=R]
calculateConfusionMatrix(pred)
#>             predicted
#> true         setosa versicolor virginica -err.-
#>   setosa         50          0         0      0
#>   versicolor      0         49         1      1
#>   virginica       0          5        45      5
#>   -err.-          0          5         1      6
\end{lstlisting}

You can see the number of correctly classified observations on the
diagonal of the matrix. Misclassified observations are on the
off-diagonal. The total number of errors for single (true and predicted)
classes is shown in the \lstinline!-err.-! row and column, respectively.

To get relative frequencies additional to the absolute numbers we can
set \lstinline!relative = TRUE!.

\begin{lstlisting}[language=R]
conf.matrix = calculateConfusionMatrix(pred, relative = TRUE)
conf.matrix
#> Relative confusion matrix (normalized by row/column):
#>             predicted
#> true         setosa    versicolor virginica -err.-   
#>   setosa     1.00/1.00 0.00/0.00  0.00/0.00 0.00     
#>   versicolor 0.00/0.00 0.98/0.91  0.02/0.02 0.02     
#>   virginica  0.00/0.00 0.10/0.09  0.90/0.98 0.10     
#>   -err.-          0.00      0.09       0.02 0.08     
#> 
#> 
#> Absolute confusion matrix:
#>             predicted
#> true         setosa versicolor virginica -err.-
#>   setosa         50          0         0      0
#>   versicolor      0         49         1      1
#>   virginica       0          5        45      5
#>   -err.-          0          5         1      6
\end{lstlisting}

It is possible to normalize by either row or column, therefore every
element of the above relative confusion matrix contains two values. The
first is the relative frequency grouped by row (the true label) and the
second value grouped by column (the predicted label).

If you want to access the relative values directly you can do this
through the \lstinline!$relative.row! and \lstinline!$relative.col!
members of the returned object \lstinline!conf.matrix!. For more details
see the
\href{http://www.rdocumentation.org/packages/mlr/functions/ConfusionMatrix.html}{ConfusionMatrix}
documentation page.

\begin{lstlisting}[language=R]
conf.matrix$relative.row
#>            setosa versicolor virginica -err-
#> setosa          1       0.00      0.00  0.00
#> versicolor      0       0.98      0.02  0.02
#> virginica       0       0.10      0.90  0.10
\end{lstlisting}

Finally, we can also add the absolute number of observations for each
predicted and true class label to the matrix (both absolute and
relative) by setting \lstinline!sums = TRUE!.

\begin{lstlisting}[language=R]
calculateConfusionMatrix(pred, relative = TRUE, sums = TRUE)
#> Relative confusion matrix (normalized by row/column):
#>             predicted
#> true         setosa    versicolor virginica -err.-    -n- 
#>   setosa     1.00/1.00 0.00/0.00  0.00/0.00 0.00      50  
#>   versicolor 0.00/0.00 0.98/0.91  0.02/0.02 0.02      54  
#>   virginica  0.00/0.00 0.10/0.09  0.90/0.98 0.10      46  
#>   -err.-          0.00      0.09       0.02 0.08      <NA>
#>   -n-        50        50         50        <NA>      150 
#> 
#> 
#> Absolute confusion matrix:
#>            setosa versicolor virginica -err.- -n-
#> setosa         50          0         0      0  50
#> versicolor      0         49         1      1  50
#> virginica       0          5        45      5  50
#> -err.-          0          5         1      6  NA
#> -n-            50         54        46     NA 150
\end{lstlisting}

\subsubsection{Adjusting the threshold}\label{adjusting-the-threshold}

We can set the threshold value that is used to map the predicted
posterior probabilities to class labels. Note that for this purpose we
need to create a
\href{http://www.rdocumentation.org/packages/mlr/functions/makeLearner.html}{Learner}
that predicts probabilities. For binary classification, the threshold
determines when the \emph{positive} class is predicted. The default is
0.5. Now, we set the threshold for the positive class to 0.9 (that is,
an example is assigned to the positive class if its posterior
probability exceeds 0.9). Which of the two classes is the positive one
can be seen by accessing the
\href{http://www.rdocumentation.org/packages/mlr/functions/Task.html}{Task}.
To illustrate binary classification, we use the
\href{http://www.rdocumentation.org/packages/mlbench/functions/Sonar.html}{Sonar}
data set from the
\href{http://www.rdocumentation.org/packages/mlbench/}{mlbench} package.

\begin{lstlisting}[language=R]
lrn = makeLearner("classif.rpart", predict.type = "prob")
mod = train(lrn, task = sonar.task)

### Label of the positive class
getTaskDescription(sonar.task)$positive
#> [1] "M"

### Default threshold
pred1 = predict(mod, sonar.task)
pred1$threshold
#>   M   R 
#> 0.5 0.5

### Set the threshold value for the positive class
pred2 = setThreshold(pred1, 0.9)
pred2$threshold
#>   M   R 
#> 0.9 0.1

pred2
#> Prediction: 208 observations
#> predict.type: prob
#> threshold: M=0.90,R=0.10
#> time: 0.00
#>   id truth    prob.M    prob.R response
#> 1  1     R 0.1060606 0.8939394        R
#> 2  2     R 0.7333333 0.2666667        R
#> 3  3     R 0.0000000 1.0000000        R
#> 4  4     R 0.1060606 0.8939394        R
#> 5  5     R 0.9250000 0.0750000        M
#> 6  6     R 0.0000000 1.0000000        R
#> ... (208 rows, 5 cols)

### We can also set the effect in the confusion matrix
calculateConfusionMatrix(pred1)
#>         predicted
#> true      M  R -err.-
#>   M      95 16     16
#>   R      10 87     10
#>   -err.- 10 16     26

calculateConfusionMatrix(pred2)
#>         predicted
#> true      M  R -err.-
#>   M      84 27     27
#>   R       6 91      6
#>   -err.-  6 27     33
\end{lstlisting}

Note that in the binary case
\href{http://www.rdocumentation.org/packages/mlr/functions/getPredictionProbabilities.html}{getPredictionProbabilities}
by default extracts the posterior probabilities of the positive class
only.

\begin{lstlisting}[language=R]
head(getPredictionProbabilities(pred1))
#> [1] 0.1060606 0.7333333 0.0000000 0.1060606 0.9250000 0.0000000

### But we can change that, too
head(getPredictionProbabilities(pred1, cl = c("M", "R")))
#>           M         R
#> 1 0.1060606 0.8939394
#> 2 0.7333333 0.2666667
#> 3 0.0000000 1.0000000
#> 4 0.1060606 0.8939394
#> 5 0.9250000 0.0750000
#> 6 0.0000000 1.0000000
\end{lstlisting}

It works similarly for multiclass classification. The threshold has to
be given by a named vector specifying the values by which each
probability will be divided. The class with the maximum resulting value
is then selected.

\begin{lstlisting}[language=R]
lrn = makeLearner("classif.rpart", predict.type = "prob")
mod = train(lrn, iris.task)
pred = predict(mod, newdata = iris)
pred$threshold
#>     setosa versicolor  virginica 
#>  0.3333333  0.3333333  0.3333333
table(as.data.frame(pred)$response)
#> 
#>     setosa versicolor  virginica 
#>         50         54         46
pred = setThreshold(pred, c(setosa = 0.01, versicolor = 50, virginica = 1))
pred$threshold
#>     setosa versicolor  virginica 
#>       0.01      50.00       1.00
table(as.data.frame(pred)$response)
#> 
#>     setosa versicolor  virginica 
#>         50          0        100
\end{lstlisting}

If you are interested in tuning the threshold (vector) have a look at
the section about
\protect\hyperlink{roc-analysis-and-performance-curves}{performance
curves and threshold tuning}.

\subsubsection{Visualizing the
prediction}\label{visualizing-the-prediction}

The function
\href{http://www.rdocumentation.org/packages/mlr/functions/plotLearnerPrediction.html}{plotLearnerPrediction}
allows to visualize predictions, e.g., for teaching purposes or
exploring models. It trains the chosen learning method for 1 or 2
selected features and then displays the predictions with
\href{http://www.rdocumentation.org/packages/ggplot2/functions/ggplot.html}{ggplot}.

For \emph{classification}, we get a scatter plot of 2 features (by
default the first 2 in the data set). The type of symbol shows the true
class labels of the data points. Symbols with white border indicate
misclassified observations. The posterior probabilities (if the learner
under consideration supports this) are represented by the background
color where higher saturation means larger probabilities.

The plot title displays the ID of the
\href{http://www.rdocumentation.org/packages/mlr/functions/makeLearner.html}{Learner}
(in the following example CART), its parameters, its training
performance and its cross-validation performance.
\protect\hyperlink{implemented-performance-measures}{mmce} stands for
\emph{mean misclassification error}, i.e., the error rate. See the
sections on
\protect\hyperlink{evaluating-learner-performance}{performance} and
\protect\hyperlink{resampling}{resampling} for further explanations.

\begin{lstlisting}[language=R]
lrn = makeLearner("classif.rpart", id = "CART")
plotLearnerPrediction(lrn, task = iris.task)
\end{lstlisting}

\includegraphics{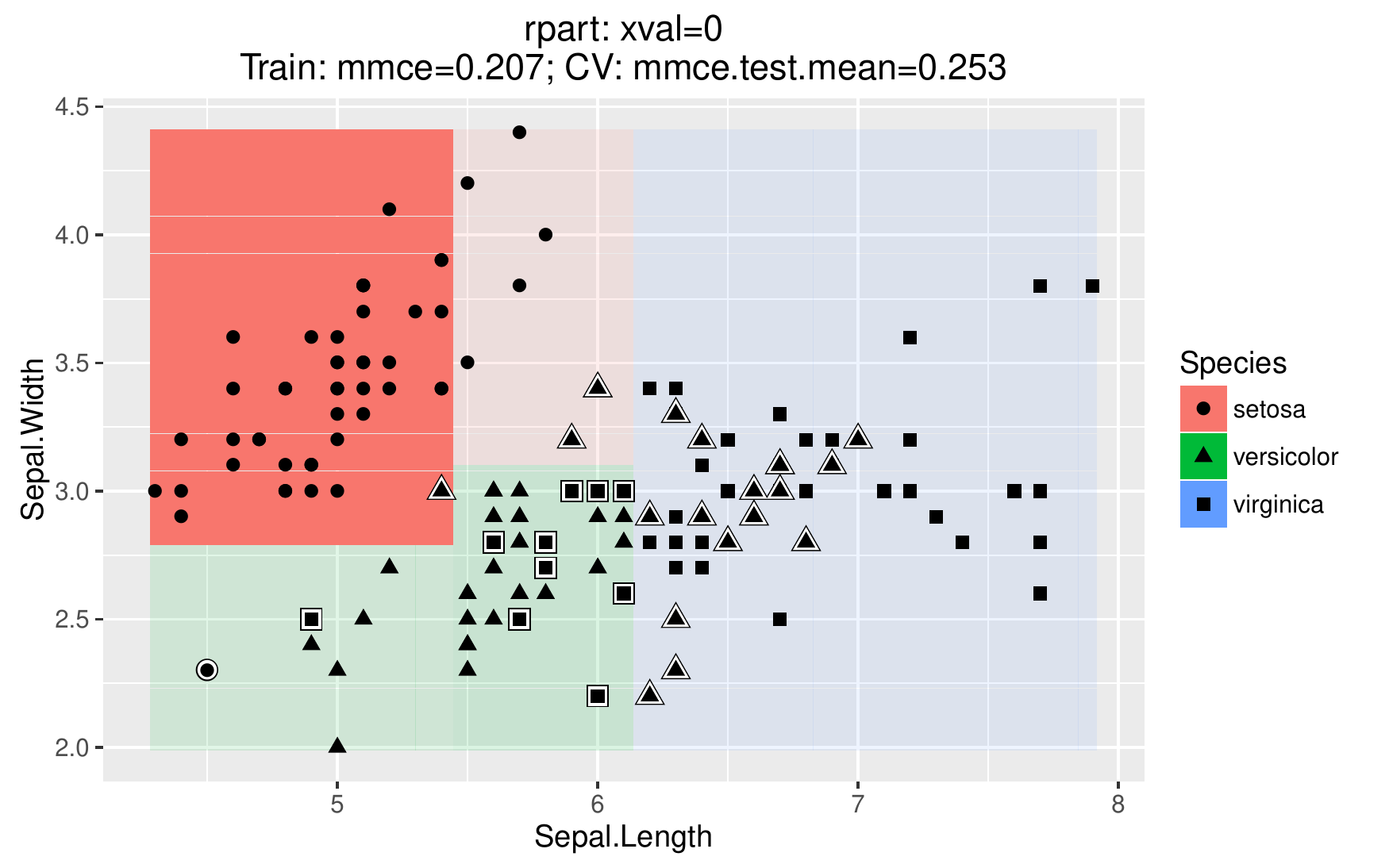}

For \emph{clustering} we also get a scatter plot of two selected
features. The color of the points indicates the predicted cluster.

\begin{lstlisting}[language=R]
lrn = makeLearner("cluster.kmeans")
plotLearnerPrediction(lrn, task = mtcars.task, features = c("disp", "drat"), cv = 0)
\end{lstlisting}

\includegraphics{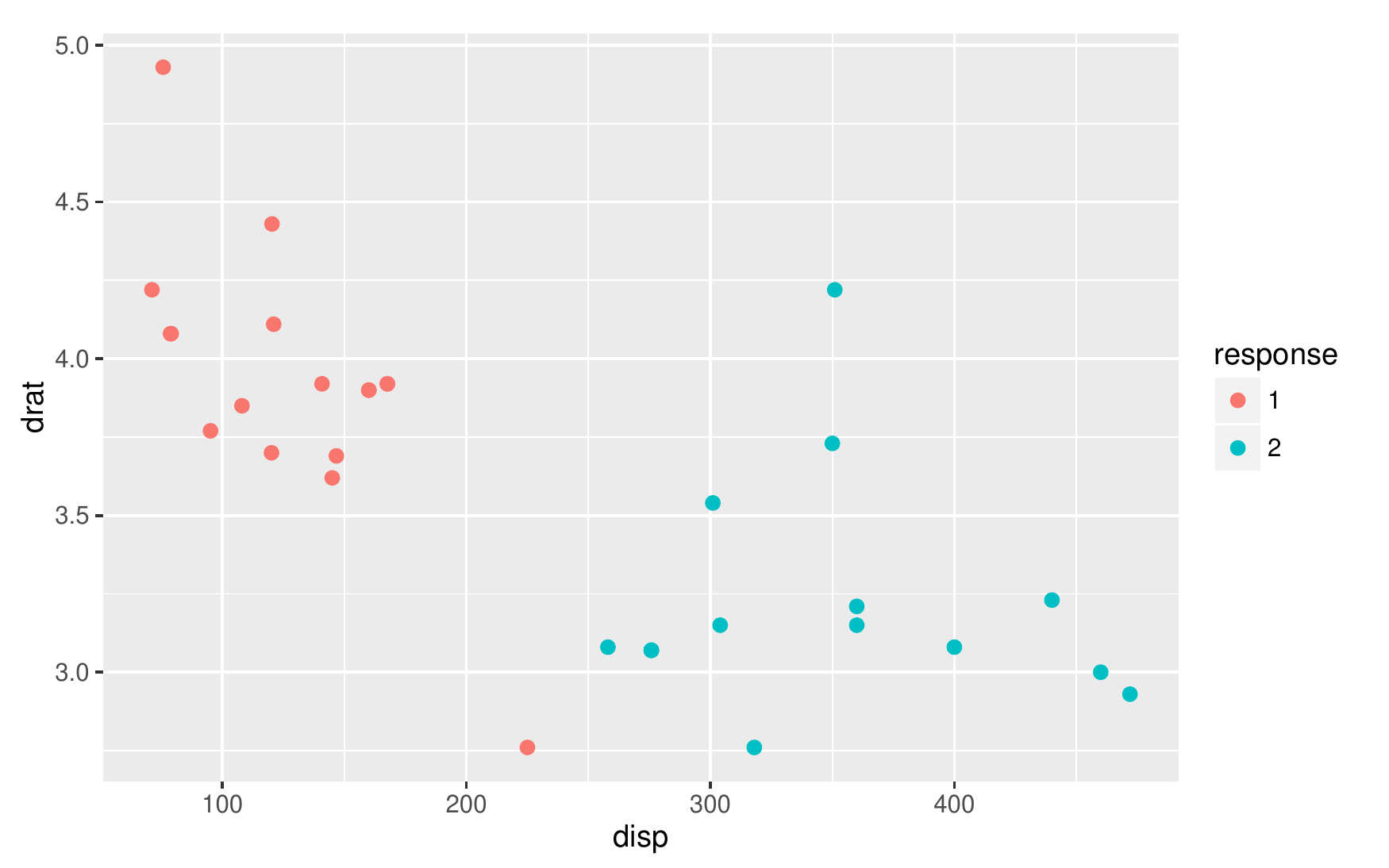}

For \emph{regression}, there are two types of plots. The 1D plot shows
the target values in relation to a single feature, the regression curve
and, if the chosen learner supports this, the estimated standard error.

\begin{lstlisting}[language=R]
plotLearnerPrediction("regr.lm", features = "lstat", task = bh.task)
\end{lstlisting}

\includegraphics{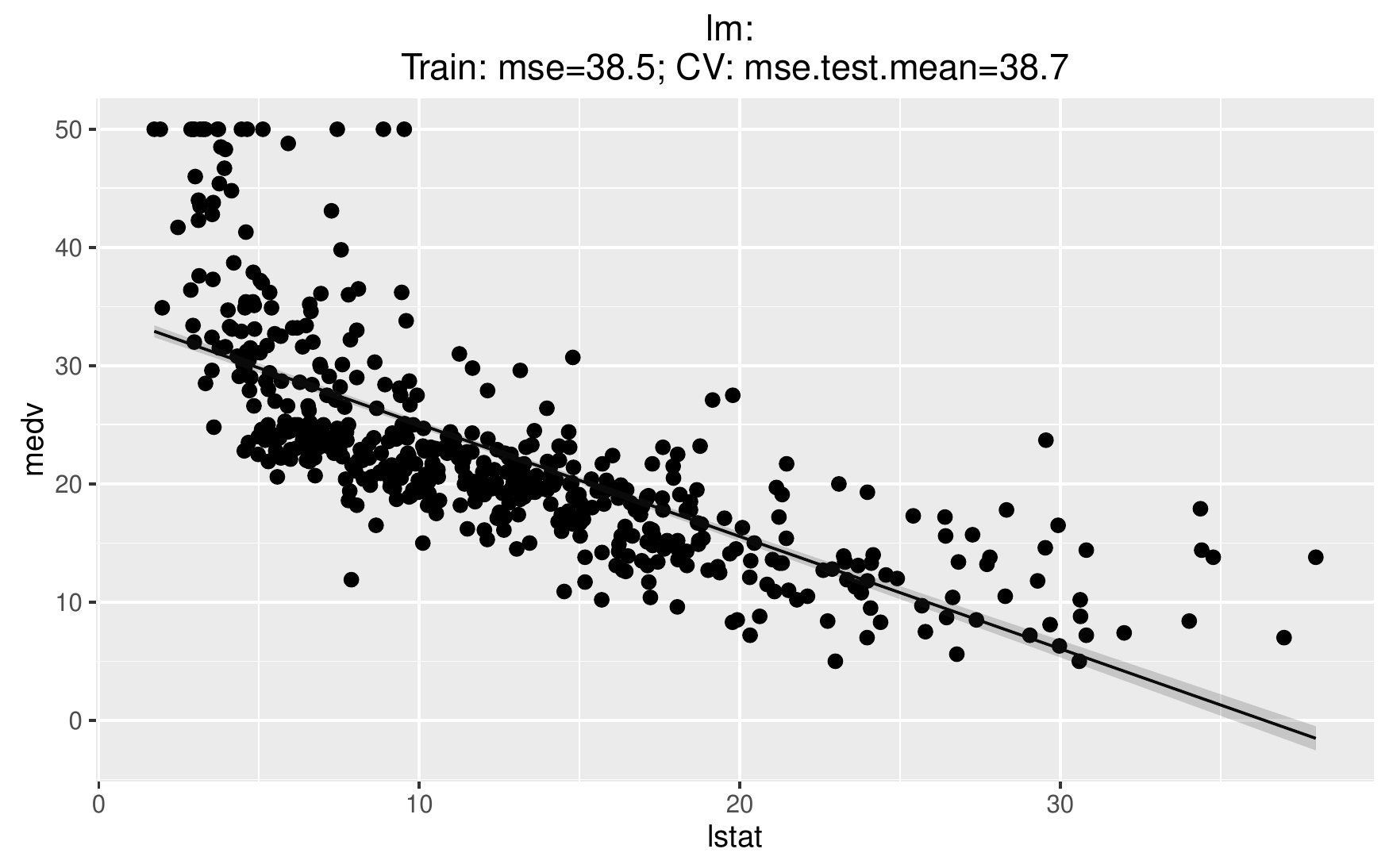}

The 2D variant, as in the classification case, generates a scatter plot
of 2 features. The fill color of the dots illustrates the value of the
target variable \lstinline!"medv"!, the background colors show the
estimated mean. The plot does not represent the estimated standard
error.

\begin{lstlisting}[language=R]
plotLearnerPrediction("regr.lm", features = c("lstat", "rm"), task = bh.task)
\end{lstlisting}

\includegraphics{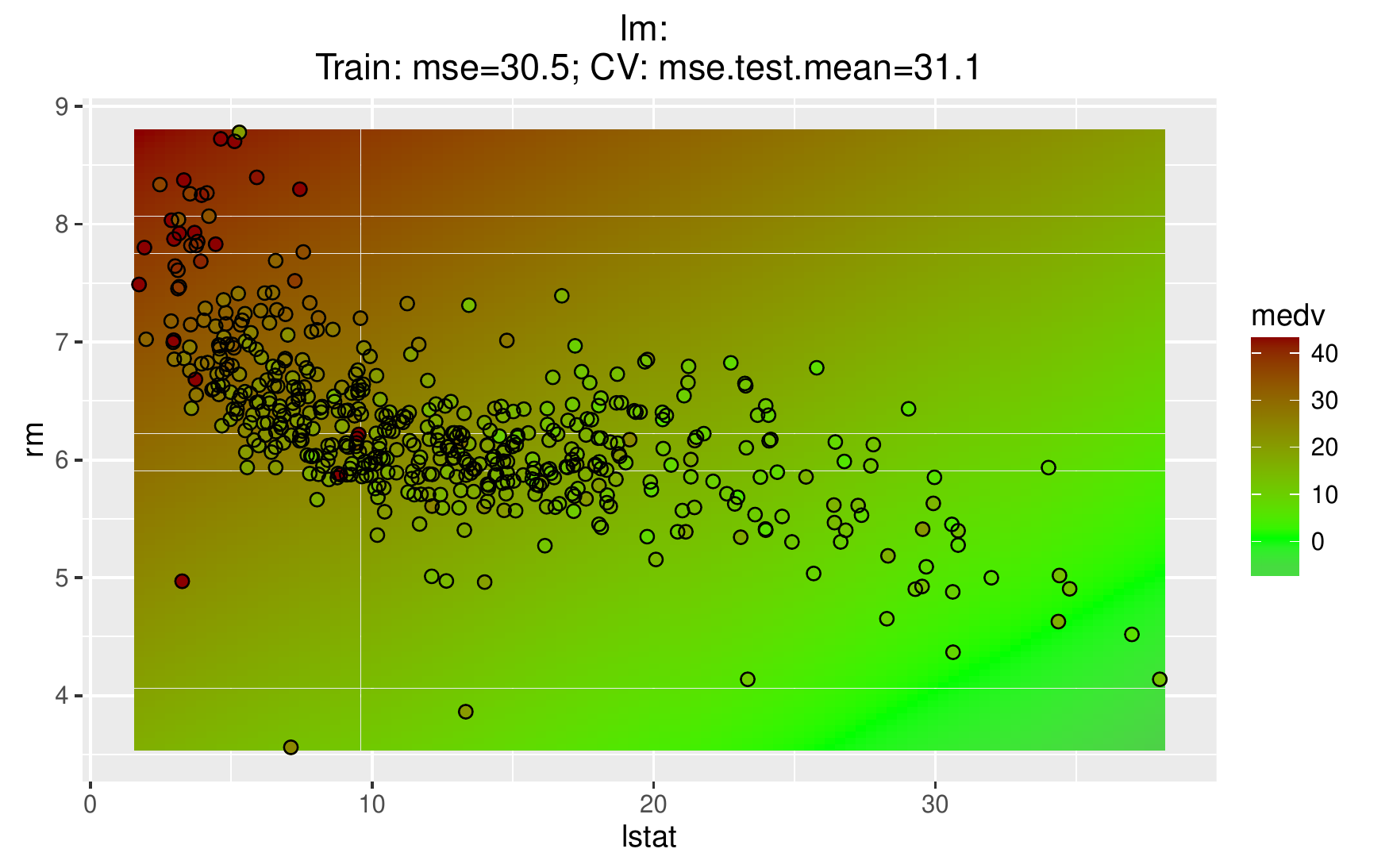}

\hypertarget{evaluating-learner-performance}{\subsection{Evaluating
Learner Performance}\label{evaluating-learner-performance}}

The quality of the predictions of a model in
\href{http://www.rdocumentation.org/packages/mlr/}{mlr} can be assessed
with respect to a number of different performance measures. In order to
calculate the performance measures, call
\href{http://www.rdocumentation.org/packages/mlr/functions/performance.html}{performance}
on the object returned by
\href{http://www.rdocumentation.org/packages/mlr/functions/predict.WrappedModel.html}{predict}
and specify the desired performance measures.

\subsubsection{Available performance
measures}\label{available-performance-measures}

\href{http://www.rdocumentation.org/packages/mlr/}{mlr} provides a large
number of performance measures for all types of learning problems.
Typical performance measures for \emph{classification} are the mean
misclassification error
(\protect\hyperlink{implemented-performance-measures}{mmce}), accuracy
(\protect\hyperlink{implemented-performance-measures}{acc}) or measures
based on \protect\hyperlink{roc-analysis-and-performance-curves}{ROC
analysis}. For \emph{regression} the mean of squared errors
(\protect\hyperlink{implemented-performance-measures}{mse}) or mean of
absolute errors
(\protect\hyperlink{implemented-performance-measures}{mae}) are usually
considered. For \emph{clustering} tasks, measures such as the Dunn index
(\protect\hyperlink{implemented-performance-measures}{dunn}) are
provided, while for \emph{survival} predictions, the Concordance Index
(\protect\hyperlink{implemented-performance-measures}{cindex}) is
supported, and for \emph{cost-sensitive} predictions the
misclassification penalty
(\protect\hyperlink{implemented-performance-measures}{mcp}) and others.
It is also possible to access the time to train the learner
(\protect\hyperlink{implemented-performance-measures}{timetrain}), the
time to compute the prediction
(\protect\hyperlink{implemented-performance-measures}{timepredict}) and
their sum
(\protect\hyperlink{implemented-performance-measures}{timeboth}) as
performance measures.

To see which performance measures are implemented, have a look at the
\protect\hyperlink{implemented-performance-measures}{table of
performance measures} and the
\href{http://www.rdocumentation.org/packages/mlr/functions/measures.html}{measures}
documentation page.

If you want to implement an additional measure or include a measure with
non-standard misclassification costs, see the section on
\protect\hyperlink{integrating-another-measure}{creating custom
measures}.

\subsubsection{Listing measures}\label{listing-measures}

The properties and requirements of the individual measures are shown in
the \protect\hyperlink{implemented-performance-measures}{table of
performance measures}.

If you would like a list of available measures with certain properties
or suitable for a certain learning
\href{http://www.rdocumentation.org/packages/mlr/functions/Task.html}{Task}
use the function
\href{http://www.rdocumentation.org/packages/mlr/functions/listMeasures.html}{listMeasures}.

\begin{lstlisting}[language=R]
### Performance measures for classification with multiple classes
listMeasures("classif", properties = "classif.multi")
#>  [1] "multiclass.brier" "multiclass.aunp"  "multiclass.aunu" 
#>  [4] "qsr"              "ber"              "logloss"         
#>  [7] "timeboth"         "timepredict"      "acc"             
#> [10] "lsr"              "featperc"         "multiclass.au1p" 
#> [13] "multiclass.au1u"  "ssr"              "timetrain"       
#> [16] "mmce"
### Performance measure suitable for the iris classification task
listMeasures(iris.task)
#>  [1] "multiclass.brier" "multiclass.aunp"  "multiclass.aunu" 
#>  [4] "qsr"              "ber"              "logloss"         
#>  [7] "timeboth"         "timepredict"      "acc"             
#> [10] "lsr"              "featperc"         "multiclass.au1p" 
#> [13] "multiclass.au1u"  "ssr"              "timetrain"       
#> [16] "mmce"
\end{lstlisting}

For convenience there exists a default measure for each type of learning
problem, which is calculated if nothing else is specified. As defaults
we chose the most commonly used measures for the respective types, e.g.,
the mean squared error
(\protect\hyperlink{implemented-performance-measures}{mse}) for
regression and the misclassification rate
(\protect\hyperlink{implemented-performance-measures}{mmce}) for
classification. The help page of function
\href{http://www.rdocumentation.org/packages/mlr/functions/getDefaultMeasure.html}{getDefaultMeasure}
lists all defaults for all types of learning problems. The function
itself returns the default measure for a given task type,
\href{http://www.rdocumentation.org/packages/mlr/functions/Task.html}{Task}
or
\href{http://www.rdocumentation.org/packages/mlr/functions/Learner.html}{Learner}.

\begin{lstlisting}[language=R]
### Get default measure for iris.task
getDefaultMeasure(iris.task)
#> Name: Mean misclassification error
#> Performance measure: mmce
#> Properties: classif,classif.multi,req.pred,req.truth
#> Minimize: TRUE
#> Best: 0; Worst: 1
#> Aggregated by: test.mean
#> Note:

### Get the default measure for linear regression
getDefaultMeasure(makeLearner("regr.lm"))
#> Name: Mean of squared errors
#> Performance measure: mse
#> Properties: regr,req.pred,req.truth
#> Minimize: TRUE
#> Best: 0; Worst: Inf
#> Aggregated by: test.mean
#> Note:
\end{lstlisting}

\subsubsection{Calculate performance
measures}\label{calculate-performance-measures}

In the following example we fit a
\href{http://www.rdocumentation.org/packages/gbm/functions/gbm.html}{gradient
boosting machine} on a subset of the
\href{http://www.rdocumentation.org/packages/mlbench/functions/BostonHousing.html}{BostonHousing}
data set and calculate the default measure mean squared error
(\protect\hyperlink{implemented-performance-measures}{mse}) on the
remaining observations.

\begin{lstlisting}[language=R]
n = getTaskSize(bh.task)
lrn = makeLearner("regr.gbm", n.trees = 1000)
mod = train(lrn, task = bh.task, subset = seq(1, n, 2))
pred = predict(mod, task = bh.task, subset = seq(2, n, 2))

performance(pred)
#>      mse 
#> 42.68414
\end{lstlisting}

The following code computes the median of squared errors
(\protect\hyperlink{implemented-performance-measures}{medse}) instead.

\begin{lstlisting}[language=R]
performance(pred, measures = medse)
#>    medse 
#> 9.134965
\end{lstlisting}

Of course, we can also calculate multiple performance measures at once
by simply passing a list of measures which can also include
\protect\hyperlink{integrating-another-measure}{your own measure}.

Calculate the mean squared error, median squared error and mean absolute
error (\protect\hyperlink{implemented-performance-measures}{mae}).

\begin{lstlisting}[language=R]
performance(pred, measures = list(mse, medse, mae))
#>       mse     medse       mae 
#> 42.684141  9.134965  4.536750
\end{lstlisting}

For the other types of learning problems and measures, calculating the
performance basically works in the same way.

\paragraph{Requirements of performance
measures}\label{requirements-of-performance-measures}

Note that in order to calculate some performance measures it is required
that you pass the
\href{http://www.rdocumentation.org/packages/mlr/functions/Task.html}{Task}
or the
\href{http://www.rdocumentation.org/packages/mlr/functions/makeWrappedModel.html}{fitted
model} in addition to the
\href{http://www.rdocumentation.org/packages/mlr/functions/Prediction.html}{Prediction}.

For example in order to assess the time needed for training
(\protect\hyperlink{implemented-performance-measures}{timetrain}), the
fitted model has to be passed.

\begin{lstlisting}[language=R]
performance(pred, measures = timetrain, model = mod)
#> timetrain 
#>     0.061
\end{lstlisting}

For many performance measures in cluster analysis the
\href{http://www.rdocumentation.org/packages/mlr/functions/Task.html}{Task}
is required.

\begin{lstlisting}[language=R]
lrn = makeLearner("cluster.kmeans", centers = 3)
mod = train(lrn, mtcars.task)
pred = predict(mod, task = mtcars.task)

### Calculate the Dunn index
performance(pred, measures = dunn, task = mtcars.task)
#>      dunn 
#> 0.1462919
\end{lstlisting}

Moreover, some measures require a certain type of prediction. For
example in binary classification in order to calculate the AUC
(\protect\hyperlink{implemented-performance-measures}{auc}) -- the area
under the ROC (receiver operating characteristic) curve -- we have to
make sure that posterior probabilities are predicted. For more
information on ROC analysis, see the section on
\protect\hyperlink{roc-analysis-and-performance-curves}{ROC analysis}.

\begin{lstlisting}[language=R]
lrn = makeLearner("classif.rpart", predict.type = "prob")
mod = train(lrn, task = sonar.task)
pred = predict(mod, task = sonar.task)

performance(pred, measures = auc)
#>       auc 
#> 0.9224018
\end{lstlisting}

Also bear in mind that many of the performance measures that are
available for classification, e.g., the false positive rate
(\protect\hyperlink{implemented-performance-measures}{fpr}), are only
suitable for binary problems.

\subsubsection{Access a performance
measure}\label{access-a-performance-measure}

Performance measures in
\href{http://www.rdocumentation.org/packages/mlr/}{mlr} are objects of
class
\href{http://www.rdocumentation.org/packages/mlr/functions/makeMeasure.html}{Measure}.
If you are interested in the properties or requirements of a single
measure you can access it directly. See the help page of
\href{http://www.rdocumentation.org/packages/mlr/functions/makeMeasure.html}{Measure}
for information on the individual slots.

\begin{lstlisting}[language=R]
### Mean misclassification error
str(mmce)
#> List of 10
#>  $ id        : chr "mmce"
#>  $ minimize  : logi TRUE
#>  $ properties: chr [1:4] "classif" "classif.multi" "req.pred" "req.truth"
#>  $ fun       :function (task, model, pred, feats, extra.args)  
#>  $ extra.args: list()
#>  $ best      : num 0
#>  $ worst     : num 1
#>  $ name      : chr "Mean misclassification error"
#>  $ note      : chr ""
#>  $ aggr      :List of 4
#>   ..$ id        : chr "test.mean"
#>   ..$ name      : chr "Test mean"
#>   ..$ fun       :function (task, perf.test, perf.train, measure, group, pred)  
#>   ..$ properties: chr "req.test"
#>   ..- attr(*, "class")= chr "Aggregation"
#>  - attr(*, "class")= chr "Measure"
\end{lstlisting}

\subsubsection{Binary classification}\label{binary-classification}

For binary classification specialized techniques exist to analyze the
performance.

\paragraph{Plot performance versus
threshold}\label{plot-performance-versus-threshold}

As you may recall (see the previous section on
\protect\hyperlink{predicting-outcomes-for-new-data}{making
predictions}) in binary classification we can adjust the threshold used
to map probabilities to class labels. Helpful in this regard is are the
functions
\href{http://www.rdocumentation.org/packages/mlr/functions/generateThreshVsPerfData.html}{generateThreshVsPerfData}
and
\href{http://www.rdocumentation.org/packages/mlr/functions/plotThreshVsPerf.html}{plotThreshVsPerf},
which generate and plot, respectively, the learner performance versus
the threshold.

For more performance plots and automatic threshold tuning see
\protect\hyperlink{roc-analysis-and-performance-curves}{here}.

In the following example we consider the
\href{http://www.rdocumentation.org/packages/mlbench/functions/Sonar.html}{Sonar}
data set and plot the false positive rate
(\protect\hyperlink{implemented-performance-measures}{fpr}), the false
negative rate
(\protect\hyperlink{implemented-performance-measures}{fnr}) as well as
the misclassification rate
(\protect\hyperlink{implemented-performance-measures}{mmce}) for all
possible threshold values.

\begin{lstlisting}[language=R]
lrn = makeLearner("classif.lda", predict.type = "prob")
n = getTaskSize(sonar.task)
mod = train(lrn, task = sonar.task, subset = seq(1, n, by = 2))
pred = predict(mod, task = sonar.task, subset = seq(2, n, by = 2))

### Performance for the default threshold 0.5
performance(pred, measures = list(fpr, fnr, mmce))
#>       fpr       fnr      mmce 
#> 0.2500000 0.3035714 0.2788462
### Plot false negative and positive rates as well as the error rate versus the threshold
d = generateThreshVsPerfData(pred, measures = list(fpr, fnr, mmce))
plotThreshVsPerf(d)
\end{lstlisting}

\includegraphics{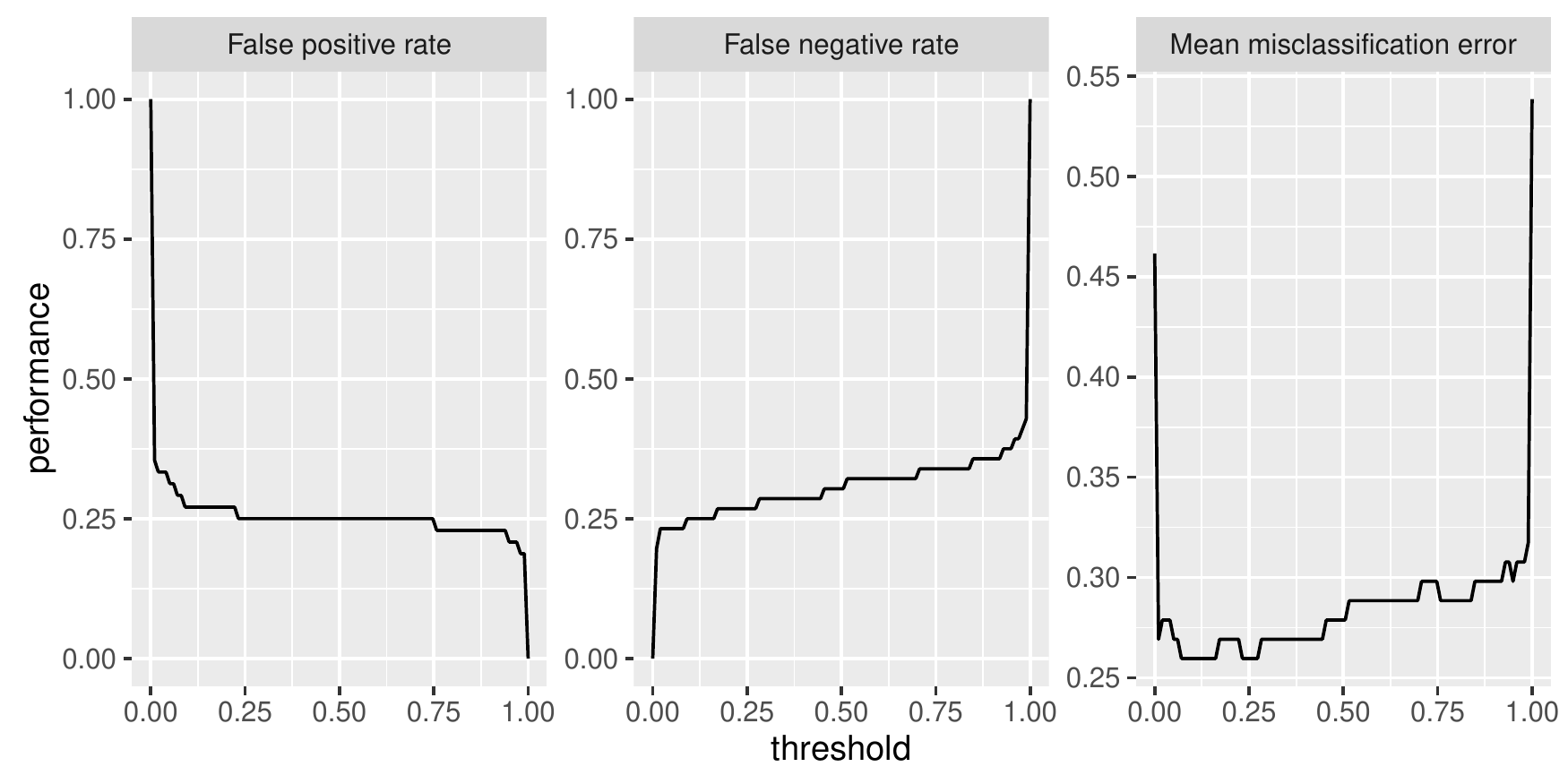}

There is an experimental
\href{http://www.rdocumentation.org/packages/ggvis/}{ggvis} plotting
function
\href{http://www.rdocumentation.org/packages/mlr/functions/plotThreshVsPerfGGVIS.html}{plotThreshVsPerfGGVIS}
which performs similarly to
\href{http://www.rdocumentation.org/packages/mlr/functions/plotThreshVsPerf.html}{plotThreshVsPerf}
but instead of creating facetted subplots to visualize multiple learners
and/or multiple measures, one of them is mapped to an interactive
sidebar which selects what to display.

\begin{lstlisting}[language=R]
plotThreshVsPerfGGVIS(d)
\end{lstlisting}

\paragraph{ROC measures}\label{roc-measures}

For binary classification a large number of specialized measures exist,
which can be nicely formatted into one matrix, see for example the
\href{https://en.wikipedia.org/wiki/Receiver_operating_characteristic}{receiver
operating characteristic page on wikipedia}.

We can generate a similiar table with the
\href{http://www.rdocumentation.org/packages/mlr/functions/calculateROCMeasures.html}{calculateROCMeasures}
function.

\begin{lstlisting}[language=R]
r = calculateROCMeasures(pred)
r
#>     predicted
#> true M         R                            
#>    M 0.7       0.3       tpr: 0.7  fnr: 0.3 
#>    R 0.25      0.75      fpr: 0.25 tnr: 0.75
#>      ppv: 0.76 for: 0.32 lrp: 2.79 acc: 0.72
#>      fdr: 0.24 npv: 0.68 lrm: 0.4  dor: 6.88
#> 
#> 
#> Abbreviations:
#> tpr - True positive rate (Sensitivity, Recall)
#> fpr - False positive rate (Fall-out)
#> fnr - False negative rate (Miss rate)
#> tnr - True negative rate (Specificity)
#> ppv - Positive predictive value (Precision)
#> for - False omission rate
#> lrp - Positive likelihood ratio (LR+)
#> fdr - False discovery rate
#> npv - Negative predictive value
#> acc - Accuracy
#> lrm - Negative likelihood ratio (LR-)
#> dor - Diagnostic odds ratio
\end{lstlisting}

The top left \(2 \times 2\) matrix is the
\href{predict.md\#confusion-matrix}{confusion matrix}, which shows the
relative frequency of correctly and incorrectly classified observations.
Below and to the right a large number of performance measures that can
be inferred from the confusion matrix are added. By default some
additional info about the measures is printed. You can turn this off
using the \lstinline!abbreviations! argument of the
\href{http://www.rdocumentation.org/packages/mlr/functions/calculateROCMeasures.html}{print}
method: \lstinline!print(r, abbreviations = FALSE)!.

\hypertarget{resampling}{\subsection{Resampling}\label{resampling}}

In order to assess the performance of a learning algorithm, resampling
strategies are usually used. The entire data set is split into
(multiple) training and test sets. You train a learner on each training
set, predict on the corresponding test set (sometimes on the training
set as well) and calculate some performance measure. Then the individual
performance values are aggregated, typically by calculating the mean.
There exist various different resampling strategies, for example
cross-validation and bootstrap, to mention just two popular approaches.

\includegraphics{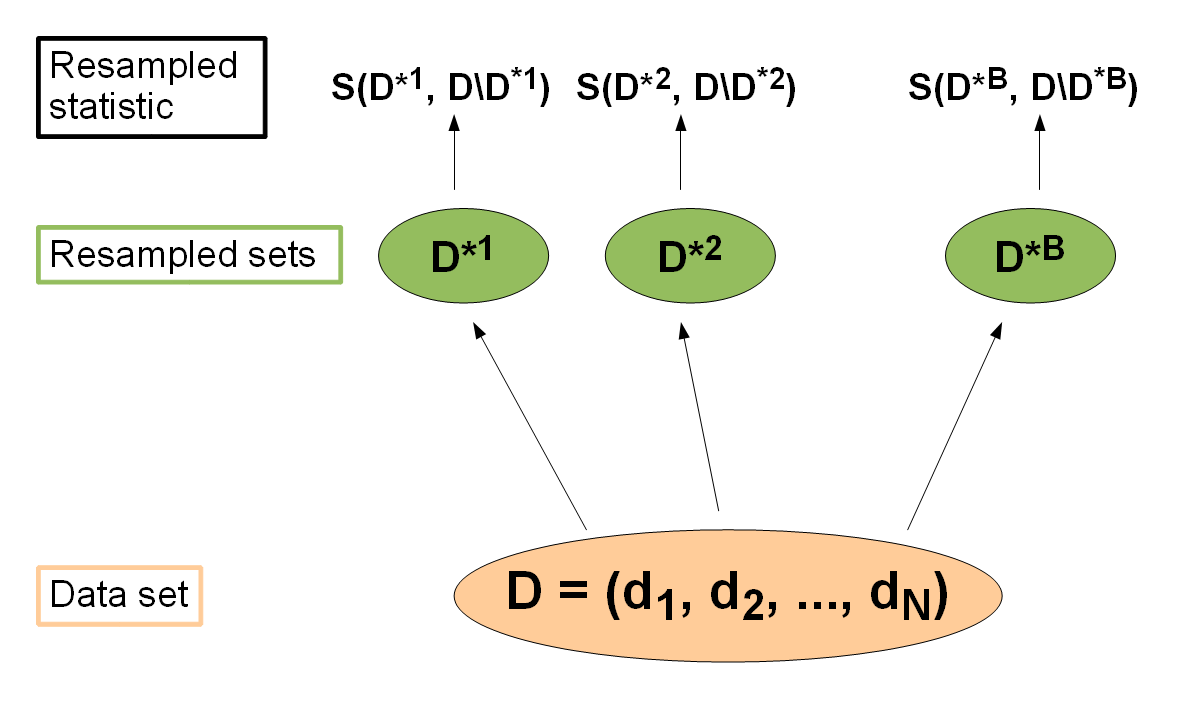}

If you want to read up further details, the paper
\href{http://link.springer.com/chapter/10.1007\%2F978-0-387-47509-7_8}{Resampling
Strategies for Model Assessment and Selection} by Simon is proabably not
a bad choice. Bernd has also published a paper
\href{http://www.mitpressjournals.org/doi/pdf/10.1162/EVCO_a_00069}{Resampling
methods for meta-model validation with recommendations for evolutionary
computation} which contains detailed descriptions and lots of
statistical background information on resampling methods.

In \href{http://www.rdocumentation.org/packages/mlr/}{mlr} the
resampling strategy can be chosen via the function
\href{http://www.rdocumentation.org/packages/mlr/functions/makeResampleDesc.html}{makeResampleDesc}.
The supported resampling strategies are:

\begin{itemize}
\tightlist
\item
  Cross-validation (\lstinline!"CV"!),
\item
  Leave-one-out cross-validation (\lstinline!"LOO""!),
\item
  Repeated cross-validation (\lstinline!"RepCV"!),
\item
  Out-of-bag bootstrap and other variants (\lstinline!"Bootstrap"!),
\item
  Subsampling, also called Monte-Carlo cross-validaton
  (\lstinline!"Subsample"!),
\item
  Holdout (training/test) (\lstinline!"Holdout"!).
\end{itemize}

The
\href{http://www.rdocumentation.org/packages/mlr/functions/resample.html}{resample}
function evaluates the performance of a
\href{http://www.rdocumentation.org/packages/mlr/functions/makeLearner.html}{Learner}
using the specified resampling strategy for a given machine learning
\href{http://www.rdocumentation.org/packages/mlr/functions/Task.html}{Task}.

In the following example the performance of the
\href{http://www.rdocumentation.org/packages/survival/functions/coxph.html}{Cox
proportional hazards model} on the
\href{http://www.rdocumentation.org/packages/survival/functions/lung.html}{lung}
data set is calculated using \emph{3-fold cross-validation}. Generally,
in \emph{\(K\)-fold cross-validation} the data set \(D\) is partitioned
into \(K\) subsets of (approximately) equal size. In the \(i\)-th step
of the \(K\) iterations, the \(i\)-th subset is used for testing, while
the union of the remaining parts forms the training set. The default
performance measure in survival analysis is the concordance index
(\protect\hyperlink{implemented-performance-measures}{cindex}).

\begin{lstlisting}[language=R]
### Specify the resampling strategy (3-fold cross-validation)
rdesc = makeResampleDesc("CV", iters = 3)

### Calculate the performance
r = resample("surv.coxph", lung.task, rdesc)
#> [Resample] cross-validation iter: 1
#> [Resample] cross-validation iter: 2
#> [Resample] cross-validation iter: 3
#> [Resample] Result: cindex.test.mean=0.627
r
#> Resample Result
#> Task: lung-example
#> Learner: surv.coxph
#> Aggr perf: cindex.test.mean=0.627
#> Runtime: 0.0267198
### peak a little bit into r
names(r)
#>  [1] "learner.id"     "task.id"        "measures.train" "measures.test" 
#>  [5] "aggr"           "pred"           "models"         "err.msgs"      
#>  [9] "extract"        "runtime"
r$aggr
#> cindex.test.mean 
#>        0.6271182
r$measures.test
#>   iter    cindex
#> 1    1 0.5783027
#> 2    2 0.6324074
#> 3    3 0.6706444
r$measures.train
#>   iter cindex
#> 1    1     NA
#> 2    2     NA
#> 3    3     NA
\end{lstlisting}

\lstinline!r$measures.test! gives the value of the performance measure
on the 3 individual test data sets. \lstinline!r$aggr! shows the
aggregated performance value. Its name, \lstinline!"cindex.test.mean"!,
indicates the performance measure,
\protect\hyperlink{implemented-performance-measures}{cindex}, and the
method used to aggregate the 3 individual performances.
\href{http://www.rdocumentation.org/packages/mlr/functions/aggregations.html}{test.mean}
is the default method and, as the name implies, takes the mean over the
performances on the 3 test data sets. No predictions on the training
data sets were made and thus \lstinline!r$measures.train! contains
missing values.

If predictions for the training set are required, too, set
\lstinline!predict = "train"!or \lstinline!predict = "both"! in
\href{http://www.rdocumentation.org/packages/mlr/functions/makeResampleDesc.html}{makeResampleDesc}.
This is necessary for some bootstrap methods (\emph{b632} and
\emph{b632+}) and we will see some examples later on.

\lstinline!r$pred! is an object of class
\href{http://www.rdocumentation.org/packages/mlr/functions/ResamplePrediction.html}{ResamplePrediction}.
Just as a
\href{http://www.rdocumentation.org/packages/mlr/functions/Prediction.html}{Prediction}
object (see the section on
\protect\hyperlink{predicting-outcomes-for-new-data}{making
predictions}) \lstinline!r$pred! has an element called
\lstinline!"data"! which is a \lstinline!data.frame! that contains the
predictions and in case of a supervised learning problem the true values
of the target variable.

\begin{lstlisting}[language=R]
head(r$pred$data)
#>   id truth.time truth.event   response iter  set
#> 1  1        455        TRUE -0.4951788    1 test
#> 2  2        210        TRUE  0.9573824    1 test
#> 3  4        310        TRUE  0.8069059    1 test
#> 4 10        613        TRUE  0.1918188    1 test
#> 5 12         61        TRUE  0.6638736    1 test
#> 6 14         81        TRUE -0.1873917    1 test
\end{lstlisting}

The columns \lstinline!iter! and \lstinline!set!indicate the resampling
iteration and if an individual prediction was made on the test or the
training data set.

In the above example the performance measure is the concordance index
(\protect\hyperlink{implemented-performance-measures}{cindex}). Of
course, it is possible to compute multiple performance measures at once
by passing a list of measures (see also the previous section on
\protect\hyperlink{evaluating-learner-performance}{evaluating learner
performance}).

In the following we estimate the Dunn index
(\protect\hyperlink{implemented-performance-measures}{dunn}), the
Davies-Bouldin cluster separation measure
(\protect\hyperlink{implemented-performance-measures}{db}), and the time
for training the learner
(\protect\hyperlink{implemented-performance-measures}{timetrain}) by
\emph{subsampling} with 5 iterations. In each iteration the data set
\(D\) is randomly partitioned into a training and a test set according
to a given percentage, e.g., 2/3 training and 1/3 test set. If there is
just one iteration, the strategy is commonly called \emph{holdout} or
\emph{test sample estimation}.

\begin{lstlisting}[language=R]
### cluster iris feature data
task = makeClusterTask(data = iris[,-5])
### Subsampling with 5 iterations and default split 2/3
rdesc = makeResampleDesc("Subsample", iters = 5)
### Subsampling with 5 iterations and 4/5 training data
rdesc = makeResampleDesc("Subsample", iters = 5, split = 4/5)

### Calculate the three performance measures
r = resample("cluster.kmeans", task, rdesc, measures = list(dunn, db, timetrain))
#> [Resample] subsampling iter: 1
#> [Resample] subsampling iter: 2
#> [Resample] subsampling iter: 3
#> [Resample] subsampling iter: 4
#> [Resample] subsampling iter: 5
#> [Resample] Result: dunn.test.mean=0.274,db.test.mean=0.51,timetrain.test.mean=0.0006
r$aggr
#>      dunn.test.mean        db.test.mean timetrain.test.mean 
#>           0.2738893           0.5103655           0.0006000
\end{lstlisting}

\subsubsection{Stratified resampling}\label{stratified-resampling}

For classification, it is usually desirable to have the same proportion
of the classes in all of the partitions of the original data set.
Stratified resampling ensures this. This is particularly useful in case
of imbalanced classes and small data sets. Otherwise it may happen, for
example, that observations of less frequent classes are missing in some
of the training sets which can decrease the performance of the learner,
or lead to model crashes In order to conduct stratified resampling, set
\lstinline!stratify = TRUE! when calling
\href{http://www.rdocumentation.org/packages/mlr/functions/makeResampleDesc.html}{makeResampleDesc}.

\begin{lstlisting}[language=R]
### 3-fold cross-validation
rdesc = makeResampleDesc("CV", iters = 3, stratify = TRUE)

r = resample("classif.lda", iris.task, rdesc)
#> [Resample] cross-validation iter: 1
#> [Resample] cross-validation iter: 2
#> [Resample] cross-validation iter: 3
#> [Resample] Result: mmce.test.mean=0.02
\end{lstlisting}

Stratification is also available for survival tasks. Here the
stratification balances the censoring rate.

Sometimes it is required to also stratify on the input data, e.g.~to
ensure that all subgroups are represented in all training and test sets.
To stratify on the input columns, specify factor columns of your task
data via \lstinline!stratify.cols!

\begin{lstlisting}[language=R]
rdesc = makeResampleDesc("CV", iters = 3, stratify.cols = "chas")
r = resample("regr.rpart", bh.task, rdesc)
#> [Resample] cross-validation iter: 1
#> [Resample] cross-validation iter: 2
#> [Resample] cross-validation iter: 3
#> [Resample] Result: mse.test.mean=23.2
\end{lstlisting}

\subsubsection{Accessing individual learner
models}\label{accessing-individual-learner-models}

In each resampling iteration a
\href{http://www.rdocumentation.org/packages/mlr/functions/makeLearner.html}{Learner}
is fitted on the respective training set. By default, the resulting
\href{http://www.rdocumentation.org/packages/mlr/functions/makeWrappedModel.html}{WrappedModel}s
are not returned by
\href{http://www.rdocumentation.org/packages/mlr/functions/resample.html}{resample}.
If you want to keep them, set \lstinline!models = TRUE! when calling
\href{http://www.rdocumentation.org/packages/mlr/functions/resample.html}{resample}.

\begin{lstlisting}[language=R]
### 3-fold cross-validation
rdesc = makeResampleDesc("CV", iters = 3)

r = resample("classif.lda", iris.task, rdesc, models = TRUE)
#> [Resample] cross-validation iter: 1
#> [Resample] cross-validation iter: 2
#> [Resample] cross-validation iter: 3
#> [Resample] Result: mmce.test.mean=0.02
r$models
#> [[1]]
#> Model for learner.id=classif.lda; learner.class=classif.lda
#> Trained on: task.id = iris-example; obs = 100; features = 4
#> Hyperparameters: 
#> 
#> [[2]]
#> Model for learner.id=classif.lda; learner.class=classif.lda
#> Trained on: task.id = iris-example; obs = 100; features = 4
#> Hyperparameters: 
#> 
#> [[3]]
#> Model for learner.id=classif.lda; learner.class=classif.lda
#> Trained on: task.id = iris-example; obs = 100; features = 4
#> Hyperparameters:
\end{lstlisting}

Keeping only certain information instead of entire
\href{http://www.rdocumentation.org/packages/mlr/functions/makeWrappedModel.html}{models},
for example the variable importance in a regression tree, can be
achieved using the \lstinline!extract! argument. The function passed to
\lstinline!extract! is applied to each
\href{http://www.rdocumentation.org/packages/mlr/functions/makeWrappedModel.html}{model}
fitted on one of the 3 training sets.

\begin{lstlisting}[language=R]
### 3-fold cross-validation
rdesc = makeResampleDesc("CV", iters = 3)

### Extract the variable importance in a regression tree
r = resample("regr.rpart", bh.task, rdesc,
    extract = function(x) x$learner.model$variable.importance)
#> [Resample] cross-validation iter: 1
#> [Resample] cross-validation iter: 2
#> [Resample] cross-validation iter: 3
#> [Resample] Result: mse.test.mean=30.3
r$extract
#> [[1]]
#>         rm      lstat       crim      indus        age    ptratio 
#> 15228.2872 10742.2277  3893.2744  3651.6232  2601.5262  2551.8492 
#>        dis        nox        rad        tax         zn 
#>  2498.2748  2419.5269  1014.2609   743.3742   308.8209 
#> 
#> [[2]]
#>       lstat         nox         age       indus        crim          rm 
#> 15725.19021  9323.20270  8474.23077  8358.67000  8251.74446  7332.59637 
#>          zn         dis         tax         rad     ptratio           b 
#>  6151.29577  2741.12074  2055.67537  1216.01398   634.78381    71.00088 
#> 
#> [[3]]
#>         rm      lstat        age    ptratio        nox        dis 
#> 15890.9279 13262.3672  4296.4175  3678.6651  3668.4944  3512.2753 
#>       crim        tax      indus         zn          b        rad 
#>  3474.5883  2844.9918  1437.7900  1284.4714   578.6932   496.2382
\end{lstlisting}

\subsubsection{Resample descriptions and resample
instances}\label{resample-descriptions-and-resample-instances}

As shown above, the function
\href{http://www.rdocumentation.org/packages/mlr/functions/makeResampleDesc.html}{makeResampleDesc}
is used to specify the resampling strategy.

\begin{lstlisting}[language=R]
rdesc = makeResampleDesc("CV", iters = 3)
str(rdesc)
#> List of 4
#>  $ id      : chr "cross-validation"
#>  $ iters   : int 3
#>  $ predict : chr "test"
#>  $ stratify: logi FALSE
#>  - attr(*, "class")= chr [1:2] "CVDesc" "ResampleDesc"
\end{lstlisting}

The result \lstinline!rdesc!is an object of class
\href{http://www.rdocumentation.org/packages/mlr/functions/makeResampleDesc.html}{ResampleDesc}
and contains, as the name implies, a description of the resampling
strategy. In principle, this is an instruction for drawing training and
test sets including the necessary parameters like the number of
iterations, the sizes of the training and test sets etc.

Based on this description, the data set is randomly partitioned into
multiple training and test sets. For each iteration, we get a set of
index vectors indicating the training and test examples. These are
stored in a
\href{http://www.rdocumentation.org/packages/mlr/functions/makeResampleInstance.html}{ResampleInstance}.

If a
\href{http://www.rdocumentation.org/packages/mlr/functions/makeResampleDesc.html}{ResampleDesc}
is passed to
\href{http://www.rdocumentation.org/packages/mlr/functions/resample.html}{resample},
it is instantiated internally. Naturally, it is also possible to pass a
\href{http://www.rdocumentation.org/packages/mlr/functions/makeResampleInstance.html}{ResampleInstance}
directly.

A
\href{http://www.rdocumentation.org/packages/mlr/functions/makeResampleInstance.html}{ResampleInstance}
can be created through the function
\href{http://www.rdocumentation.org/packages/mlr/functions/makeResampleInstance.html}{makeResampleInstance}
given a
\href{http://www.rdocumentation.org/packages/mlr/functions/makeResampleDesc.html}{ResampleDesc}
and either the size of the data set at hand or the
\href{http://www.rdocumentation.org/packages/mlr/functions/Task.html}{Task}.
It basically performs the random drawing of indices to separate the data
into training and test sets according to the description.

\begin{lstlisting}[language=R]
### Create a resample instance based an a task
rin = makeResampleInstance(rdesc, task = iris.task)
rin
#> Resample instance for 150 cases.
#> Resample description: cross-validation with 3 iterations.
#> Predict: test
#> Stratification: FALSE

### Create a resample instance given the size of the data set
rin = makeResampleInstance(rdesc, size = nrow(iris))
str(rin)
#> List of 5
#>  $ desc      :List of 4
#>   ..$ id      : chr "cross-validation"
#>   ..$ iters   : int 3
#>   ..$ predict : chr "test"
#>   ..$ stratify: logi FALSE
#>   ..- attr(*, "class")= chr [1:2] "CVDesc" "ResampleDesc"
#>  $ size      : int 150
#>  $ train.inds:List of 3
#>   ..$ : int [1:100] 36 81 6 82 120 110 118 132 105 61 ...
#>   ..$ : int [1:100] 6 119 120 110 121 118 99 100 29 127 ...
#>   ..$ : int [1:100] 36 81 82 119 121 99 132 105 61 115 ...
#>  $ test.inds :List of 3
#>   ..$ : int [1:50] 2 3 4 5 7 9 11 16 22 24 ...
#>   ..$ : int [1:50] 8 12 17 19 20 23 25 27 32 33 ...
#>   ..$ : int [1:50] 1 6 10 13 14 15 18 21 29 31 ...
#>  $ group     : Factor w/ 0 levels: 
#>  - attr(*, "class")= chr "ResampleInstance"

### Access the indices of the training observations in iteration 3
rin$train.inds[[3]]
#>   [1]  36  81  82 119 121  99 132 105  61 115  17  42   4  71   5  79  30
#>  [18] 113 138  19 150  77  58  92 114 133   8 109  33 145  22 111  97  24
#>  [35]   7  44   3  20 134  96  16  43 149   9  46  32 139  87   2  11  52
#>  [52]  86  40 141 142  72  54  48  83  64  90 112 148 129 137 116 143  69
#>  [69]  84  25  80  37  38  75 130 126 135 107 146  26  12  98  55 124  60
#>  [86]  63 117  23  67  73  28 106  76  50 144  59  47 102  56  27
\end{lstlisting}

While having two separate objects, resample descriptions and instances
as well as the
\href{http://www.rdocumentation.org/packages/mlr/functions/resample.html}{resample}
function seems overly complicated, it has several advantages:

\begin{itemize}
\tightlist
\item
  Resample instances allow for paired experiments, that is comparing the
  performance of several learners on exactly the same training and test
  sets. This is particularly useful if you want to add another method to
  a comparison experiment you already did.
\end{itemize}

\begin{lstlisting}[language=R]
rdesc = makeResampleDesc("CV", iters = 3)
rin = makeResampleInstance(rdesc, task = iris.task)

### Calculate the performance of two learners based on the same resample instance
r.lda = resample("classif.lda", iris.task, rin, show.info = FALSE)
r.rpart = resample("classif.rpart", iris.task, rin, show.info = FALSE)
r.lda$aggr
#> mmce.test.mean 
#>     0.02666667
r.rpart$aggr
#> mmce.test.mean 
#>           0.06
\end{lstlisting}

\begin{itemize}
\tightlist
\item
  It is easy to add other resampling methods later on. You can simply
  derive from the
  \href{http://www.rdocumentation.org/packages/mlr/functions/makeResampleInstance.html}{ResampleInstance}
  class, but you do not have to touch any methods that use the
  resampling strategy.
\end{itemize}

As mentioned above, when calling
\href{http://www.rdocumentation.org/packages/mlr/functions/makeResampleInstance.html}{makeResampleInstance}
the index sets are drawn randomly. Mainly for \emph{holdout} (\emph{test
sample}) \emph{estimation} you might want full control about the
training and tests set and specify them manually. This can be done using
the function
\href{http://www.rdocumentation.org/packages/mlr/functions/makeFixedHoldoutInstance.html}{makeFixedHoldoutInstance}.

\begin{lstlisting}[language=R]
rin = makeFixedHoldoutInstance(train.inds = 1:100, test.inds = 101:150, size = 150)
rin
#> Resample instance for 150 cases.
#> Resample description: holdout with 0.67 split rate.
#> Predict: test
#> Stratification: FALSE
\end{lstlisting}

\subsubsection{Aggregating performance
values}\label{aggregating-performance-values}

In resampling we get (for each measure we wish to calculate) one
performance value (on the test set, training set, or both) for each
iteration. Subsequently, these are aggregated. As mentioned above,
mainly the mean over the performance values on the test data sets
(\href{http://www.rdocumentation.org/packages/mlr/functions/aggregations.html}{test.mean})
is calculated.

For example, a 10-fold cross validation computes 10 values for the
chosen performance measure. The aggregated value is the mean of these 10
numbers. \href{http://www.rdocumentation.org/packages/mlr/}{mlr} knows
how to handle it because each
\href{http://www.rdocumentation.org/packages/mlr/functions/makeMeasure.html}{Measure}
knows how it is aggregated:

\begin{lstlisting}[language=R]
### Mean misclassification error
mmce$aggr
#> Aggregation function: test.mean

### Root mean square error
rmse$aggr
#> Aggregation function: test.rmse
\end{lstlisting}

The aggregation method of a
\href{http://www.rdocumentation.org/packages/mlr/functions/makeMeasure.html}{Measure}
can be changed via the function
\href{http://www.rdocumentation.org/packages/mlr/functions/setAggregation.html}{setAggregation}.
See the documentation of
\href{http://www.rdocumentation.org/packages/mlr/functions/aggregations.html}{aggregations}
for available methods.

\paragraph{Example: Different measures and
aggregations}\label{example-different-measures-and-aggregations}

\href{http://www.rdocumentation.org/packages/mlr/functions/aggregations.html}{test.median}
computes the median of the performance values on the test sets.

\begin{lstlisting}[language=R]
### We use the mean error rate and the median of the true positive rates
m1 = mmce
m2 = setAggregation(tpr, test.median)
rdesc = makeResampleDesc("CV", iters = 3)
r = resample("classif.rpart", sonar.task, rdesc, measures = list(m1, m2))
#> [Resample] cross-validation iter: 1
#> [Resample] cross-validation iter: 2
#> [Resample] cross-validation iter: 3
#> [Resample] Result: mmce.test.mean=0.293,tpr.test.median=0.735
r$aggr
#>  mmce.test.mean tpr.test.median 
#>       0.2930987       0.7352941
\end{lstlisting}

\paragraph{Example: Calculating the training
error}\label{example-calculating-the-training-error}

Here we calculate the mean misclassification error
(\protect\hyperlink{implemented-performance-measures}{mmce}) on the
training and the test data sets. Note that we have to set
\lstinline!predict = "both"!when calling
\href{http://www.rdocumentation.org/packages/mlr/functions/makeResampleDesc.html}{makeResampleDesc}
in order to get predictions on both data sets, training and test.

\begin{lstlisting}[language=R]
mmce.train.mean = setAggregation(mmce, train.mean)
rdesc = makeResampleDesc("CV", iters = 3, predict = "both")
r = resample("classif.rpart", iris.task, rdesc, measures = list(mmce, mmce.train.mean))
#> [Resample] cross-validation iter: 1
#> [Resample] cross-validation iter: 2
#> [Resample] cross-validation iter: 3
#> [Resample] Result: mmce.test.mean=0.0467,mmce.train.mean=0.0367
r$measures.train
#>   iter mmce mmce
#> 1    1 0.04 0.04
#> 2    2 0.03 0.03
#> 3    3 0.04 0.04
r$aggr
#>  mmce.test.mean mmce.train.mean 
#>      0.04666667      0.03666667
\end{lstlisting}

\paragraph{Example: Bootstrap}\label{example-bootstrap}

In \emph{out-of-bag bootstrap estimation} \(B\) new data sets \(D_1\) to
\(D_B\) are drawn from the data set \(D\) with replacement, each of the
same size as \(D\). In the \(i\)-th iteration, \(D_i\) forms the
training set, while the remaining elements from \(D\), i.e., elements
not in the training set, form the test set.

The variants \emph{b632} and \emph{b632+} calculate a convex combination
of the training performance and the out-of-bag bootstrap performance and
thus require predictions on the training sets and an appropriate
aggregation strategy.

\begin{lstlisting}[language=R]
rdesc = makeResampleDesc("Bootstrap", predict = "both", iters = 10)
b632.mmce = setAggregation(mmce, b632)
b632plus.mmce = setAggregation(mmce, b632plus)
b632.mmce
#> Name: Mean misclassification error
#> Performance measure: mmce
#> Properties: classif,classif.multi,req.pred,req.truth
#> Minimize: TRUE
#> Best: 0; Worst: 1
#> Aggregated by: b632
#> Note:

r = resample("classif.rpart", iris.task, rdesc,
    measures = list(mmce, b632.mmce, b632plus.mmce), show.info = FALSE)
head(r$measures.train)
#>   iter        mmce        mmce        mmce
#> 1    1 0.026666667 0.026666667 0.026666667
#> 2    2 0.026666667 0.026666667 0.026666667
#> 3    3 0.006666667 0.006666667 0.006666667
#> 4    4 0.026666667 0.026666667 0.026666667
#> 5    5 0.033333333 0.033333333 0.033333333
#> 6    6 0.013333333 0.013333333 0.013333333
r$aggr
#> mmce.test.mean      mmce.b632  mmce.b632plus 
#>     0.07051905     0.05389071     0.05496489
\end{lstlisting}

\subsubsection{Convenience functions}\label{convenience-functions}

When quickly trying out some learners, it can get tedious to write the
\textbf{R} code for generating a resample instance, setting the
aggregation strategy and so on. For this reason
\href{http://www.rdocumentation.org/packages/mlr/}{mlr} provides some
convenience functions for the frequently used resampling strategies, for
example
\href{http://www.rdocumentation.org/packages/mlr/functions/resample.html}{holdout},
\href{http://www.rdocumentation.org/packages/mlr/functions/resample.html}{crossval}
or
\href{http://www.rdocumentation.org/packages/mlr/functions/resample.html}{bootstrapB632}.
But note that you do not have as much control and flexibility as when
using
\href{http://www.rdocumentation.org/packages/mlr/functions/resample.html}{resample}
with a resample description or instance.

\begin{lstlisting}[language=R]
holdout("regr.lm", bh.task, measures = list(mse, mae))
crossval("classif.lda", iris.task, iters = 3, measures = list(mmce, ber))
\end{lstlisting}

\hypertarget{tuning-hyperparameters}{\subsection{Tuning
Hyperparameters}\label{tuning-hyperparameters}}

Many machine learning algorithms have hyperparameters that need to be
set. If selected by the user they can be specified as explained on the
tutorial page on \protect\hyperlink{learners}{Learners} -- simply pass
them to
\href{http://www.rdocumentation.org/packages/mlr/functions/makeLearner.html}{makeLearner}.
Often suitable parameter values are not obvious and it is preferable to
tune the hyperparameters, that is automatically identify values that
lead to the best performance.

\subsubsection{Basics}\label{basics-1}

In order to tune a machine learning algorithm, you have to specify:

\begin{itemize}
\tightlist
\item
  the search space
\item
  the optimization algorithm (aka tuning method)
\item
  an evaluation method, i.e., a resampling strategy and a performance
  measure
\end{itemize}

An example of the search space could be searching values of the
\lstinline!C! parameter for
\href{http://www.rdocumentation.org/packages/kernlab/functions/ksvm.html}{SVM}:

\begin{lstlisting}[language=R]
### ex: create a search space for the C hyperparameter from 0.01 to 0.1
ps = makeParamSet(
  makeNumericParam("C", lower = 0.01, upper = 0.1)
)
\end{lstlisting}

An example of the optimization algorithm could be performing random
search on the space:

\begin{lstlisting}[language=R]
### ex: random search with 100 iterations
ctrl = makeTuneControlRandom(maxit = 100L)
\end{lstlisting}

An example of an evaluation method could be 3-fold CV using accuracy as
the performance measure:

\begin{lstlisting}[language=R]
rdesc = makeResampleDesc("CV", iters = 3L)
measure = acc
\end{lstlisting}

The evaluation method is already covered in detail in
\protect\hyperlink{evaluating-learner-performance}{evaluation of
learning methods} and \protect\hyperlink{resampling}{resampling}.

In this tutorial, we show how to specify the search space and
optimization algorithm, how to do the tuning and how to access the
tuning result, and how to visualize the hyperparameter tuning effects
through several examples.

Throughout this section we consider classification examples. For the
other types of learning problems, you can follow the same process
analogously.

We use the
\href{http://www.rdocumentation.org/packages/mlr/functions/iris.task.html}{iris
classification task} for illustration and tune the hyperparameters of an
SVM (function
\href{http://www.rdocumentation.org/packages/kernlab/functions/ksvm.html}{ksvm}
from the \href{http://www.rdocumentation.org/packages/kernlab/}{kernlab}
package) with a radial basis kernel. The following examples tune the
cost parameter \lstinline!C! and the RBF kernel parameter
\lstinline!sigma! of the
\href{http://www.rdocumentation.org/packages/kernlab/functions/ksvm.html}{ksvm}
function.

\paragraph{Specifying the search
space}\label{specifying-the-search-space}

We first must define a space to search when tuning our learner. For
example, maybe we want to tune several specific values of a
hyperparameter or perhaps we want to define a space from \(10^{-10}\) to
\(10^{10}\) and let the optimization algorithm decide which points to
choose.

In order to define a search space, we create a
\href{http://www.rdocumentation.org/packages/ParamHelpers/functions/makeParamSet.html}{ParamSet}
object, which describes the parameter space we wish to search. This is
done via the function
\href{http://www.rdocumentation.org/packages/ParamHelpers/functions/makeParamSet.html}{makeParamSet}.

For example, we could define a search space with just the values 0.5,
1.0, 1.5, 2.0 for both \lstinline!C! and \lstinline!gamma!. Notice how
we name each parameter as it's defined in the
\href{http://www.rdocumentation.org/packages/kernlab/}{kernlab} package:

\begin{lstlisting}[language=R]
discrete_ps = makeParamSet(
  makeDiscreteParam("C", values = c(0.5, 1.0, 1.5, 2.0)),
  makeDiscreteParam("sigma", values = c(0.5, 1.0, 1.5, 2.0))
)
print(discrete_ps)
#>           Type len Def      Constr Req Tunable Trafo
#> C     discrete   -   - 0.5,1,1.5,2   -    TRUE     -
#> sigma discrete   -   - 0.5,1,1.5,2   -    TRUE     -
\end{lstlisting}

We could also define a continuous search space (using
\href{http://www.rdocumentation.org/packages/ParamHelpers/functions/makeNumericParam.html}{makeNumericParam}
instead of
\href{http://www.rdocumentation.org/packages/ParamHelpers/functions/makeDiscreteParam.html}{makeDiscreteParam})
from \(10^{-10}\) to \(10^{10}\) for both parameters through the use of
the \lstinline!trafo! argument (trafo is short for transformation).
Transformations work like this: All optimizers basically see the
parameters on their original scale (from \(-10\) to \(10\) in this case)
and produce values on this scale during the search. Right before they
are passed to the learning algorithm, the transformation function is
applied.

Notice this time we use
\href{http://www.rdocumentation.org/packages/ParamHelpers/functions/makeNumericParam.html}{makeNumericParam}:

\begin{lstlisting}[language=R]
num_ps = makeParamSet(
  makeNumericParam("C", lower = -10, upper = 10, trafo = function(x) 10^x),
  makeNumericParam("sigma", lower = -10, upper = 10, trafo = function(x) 10^x)
)
\end{lstlisting}

Many other parameters can be created, check out the examples in
\href{http://www.rdocumentation.org/packages/ParamHelpers/functions/makeParamSet.html}{makeParamSet}.

In order to standardize your workflow across several packages, whenever
parameters in the underlying \textbf{R} functions should be passed in a
\href{http://www.rdocumentation.org/packages/base/functions/list.html}{list}
structure, \href{http://www.rdocumentation.org/packages/mlr/}{mlr} tries
to give you direct access to each parameter and get rid of the list
structure!

This is the case with the \lstinline!kpar! argument of
\href{http://www.rdocumentation.org/packages/kernlab/functions/ksvm.html}{ksvm}
which is a list of kernel parameters like \lstinline!sigma!. This allows
us to interface with learners from different packages in the same way
when defining parameters to tune!

\paragraph{Specifying the optimization
algorithm}\label{specifying-the-optimization-algorithm}

Now that we have specified the search space, we need to choose an
optimization algorithm for our parameters to pass to the
\href{http://www.rdocumentation.org/packages/kernlab/functions/ksvm.html}{ksvm}
learner. Optimization algorithms are considered
\href{http://www.rdocumentation.org/packages/mlr/functions/TuneControl.html}{TuneControl}
objects in \href{http://www.rdocumentation.org/packages/mlr/}{mlr}.

A grid search is one of the standard -- albeit slow -- ways to choose an
appropriate set of parameters from a given search space.

In the case of \lstinline!discrete_ps! above, since we have manually
specified the values, grid search will simply be the cross product. We
create the grid search object using the defaults, noting that we will
have \(4 \times 4 = 16\) combinations in the case of
\lstinline!discrete_ps!:

\begin{lstlisting}[language=R]
ctrl = makeTuneControlGrid()
\end{lstlisting}

In the case of \lstinline!num_ps! above, since we have only specified
the upper and lower bounds for the search space, grid search will create
a grid using equally-sized steps. By default, grid search will span the
space in 10 equal-sized steps. The number of steps can be changed with
the \lstinline!resolution! argument. Here we change to 15 equal-sized
steps in the space defined within the
\href{http://www.rdocumentation.org/packages/ParamHelpers/functions/makeParamSet.html}{ParamSet}
object. For \lstinline!num_ps!, this means 15 steps in the form of
\lstinline!10 ^ seq(-10, 10, length.out = 15)!:

\begin{lstlisting}[language=R]
ctrl = makeTuneControlGrid(resolution = 15L)
\end{lstlisting}

Many other types of optimization algorithms are available. Check out
\href{http://www.rdocumentation.org/packages/mlr/functions/TuneControl.html}{TuneControl}
for some examples.

Since grid search is normally too slow in practice, we'll also examine
random search. In the case of \lstinline!discrete_ps!, random search
will randomly choose from the specified values. The \lstinline!maxit!
argument controls the amount of iterations.

\begin{lstlisting}[language=R]
ctrl = makeTuneControlRandom(maxit = 10L)
\end{lstlisting}

In the case of \lstinline!num_ps!, random search will randomly choose
points within the space according to the specified bounds. Perhaps in
this case we would want to increase the amount of iterations to ensure
we adequately cover the space:

\begin{lstlisting}[language=R]
ctrl = makeTuneControlRandom(maxit = 200L)
\end{lstlisting}

\paragraph{Performing the tuning}\label{performing-the-tuning}

Now that we have specified a search space and the optimization
algorithm, it's time to perform the tuning. We will need to define a
resampling strategy and make note of our performance measure.

We will use 3-fold cross-validation to assess the quality of a specific
parameter setting. For this we need to create a resampling description
just like in the \protect\hyperlink{resampling}{resampling} part of the
tutorial.

\begin{lstlisting}[language=R]
rdesc = makeResampleDesc("CV", iters = 3L)
\end{lstlisting}

Finally, by combining all the previous pieces, we can tune the SVM
parameters by calling
\href{http://www.rdocumentation.org/packages/mlr/functions/tuneParams.html}{tuneParams}.
We will use \lstinline!discrete_ps! with grid search:

\begin{lstlisting}[language=R]
discrete_ps = makeParamSet(
  makeDiscreteParam("C", values = c(0.5, 1.0, 1.5, 2.0)),
  makeDiscreteParam("sigma", values = c(0.5, 1.0, 1.5, 2.0))
)
ctrl = makeTuneControlGrid()
rdesc = makeResampleDesc("CV", iters = 3L)
res = tuneParams("classif.ksvm", task = iris.task, resampling = rdesc,
  par.set = discrete_ps, control = ctrl)
#> [Tune] Started tuning learner classif.ksvm for parameter set:
#>           Type len Def      Constr Req Tunable Trafo
#> C     discrete   -   - 0.5,1,1.5,2   -    TRUE     -
#> sigma discrete   -   - 0.5,1,1.5,2   -    TRUE     -
#> With control class: TuneControlGrid
#> Imputation value: 1
#> [Tune-x] 1: C=0.5; sigma=0.5
#> [Tune-y] 1: mmce.test.mean=0.04; time: 0.0 min; memory: 176Mb use, 711Mb max
#> [Tune-x] 2: C=1; sigma=0.5
#> [Tune-y] 2: mmce.test.mean=0.04; time: 0.0 min; memory: 176Mb use, 711Mb max
#> [Tune-x] 3: C=1.5; sigma=0.5
#> [Tune-y] 3: mmce.test.mean=0.0467; time: 0.0 min; memory: 176Mb use, 711Mb max
#> [Tune-x] 4: C=2; sigma=0.5
#> [Tune-y] 4: mmce.test.mean=0.0467; time: 0.0 min; memory: 176Mb use, 711Mb max
#> [Tune-x] 5: C=0.5; sigma=1
#> [Tune-y] 5: mmce.test.mean=0.04; time: 0.0 min; memory: 176Mb use, 711Mb max
#> [Tune-x] 6: C=1; sigma=1
#> [Tune-y] 6: mmce.test.mean=0.0467; time: 0.0 min; memory: 176Mb use, 711Mb max
#> [Tune-x] 7: C=1.5; sigma=1
#> [Tune-y] 7: mmce.test.mean=0.0467; time: 0.0 min; memory: 176Mb use, 711Mb max
#> [Tune-x] 8: C=2; sigma=1
#> [Tune-y] 8: mmce.test.mean=0.0467; time: 0.0 min; memory: 176Mb use, 711Mb max
#> [Tune-x] 9: C=0.5; sigma=1.5
#> [Tune-y] 9: mmce.test.mean=0.0333; time: 0.0 min; memory: 176Mb use, 711Mb max
#> [Tune-x] 10: C=1; sigma=1.5
#> [Tune-y] 10: mmce.test.mean=0.04; time: 0.0 min; memory: 176Mb use, 711Mb max
#> [Tune-x] 11: C=1.5; sigma=1.5
#> [Tune-y] 11: mmce.test.mean=0.04; time: 0.0 min; memory: 176Mb use, 711Mb max
#> [Tune-x] 12: C=2; sigma=1.5
#> [Tune-y] 12: mmce.test.mean=0.0467; time: 0.0 min; memory: 176Mb use, 711Mb max
#> [Tune-x] 13: C=0.5; sigma=2
#> [Tune-y] 13: mmce.test.mean=0.04; time: 0.0 min; memory: 176Mb use, 711Mb max
#> [Tune-x] 14: C=1; sigma=2
#> [Tune-y] 14: mmce.test.mean=0.0333; time: 0.0 min; memory: 176Mb use, 711Mb max
#> [Tune-x] 15: C=1.5; sigma=2
#> [Tune-y] 15: mmce.test.mean=0.04; time: 0.0 min; memory: 176Mb use, 711Mb max
#> [Tune-x] 16: C=2; sigma=2
#> [Tune-y] 16: mmce.test.mean=0.04; time: 0.0 min; memory: 176Mb use, 711Mb max
#> [Tune] Result: C=0.5; sigma=1.5 : mmce.test.mean=0.0333

res
#> Tune result:
#> Op. pars: C=0.5; sigma=1.5
#> mmce.test.mean=0.0333
\end{lstlisting}

\href{http://www.rdocumentation.org/packages/mlr/functions/tuneParams.html}{tuneParams}
simply performs the cross-validation for every element of the
cross-product and selects the parameter setting with the best mean
performance. As no performance measure was specified, by default the
error rate (\protect\hyperlink{implemented-performance-measures}{mmce})
is used.

Note that each
\href{http://www.rdocumentation.org/packages/mlr/functions/makeMeasure.html}{measure}
``knows'' if it is minimized or maximized during tuning.

\begin{lstlisting}[language=R]
### error rate
mmce$minimize
#> [1] TRUE

### accuracy
acc$minimize
#> [1] FALSE
\end{lstlisting}

Of course, you can pass other measures and also a
\href{http://www.rdocumentation.org/packages/base/functions/list.html}{list}
of measures to
\href{http://www.rdocumentation.org/packages/mlr/functions/tuneParams.html}{tuneParams}.
In the latter case the first measure is optimized during tuning, the
others are simply evaluated. If you are interested in optimizing several
measures simultaneously have a look at
\protect\hyperlink{advanced-tuning}{Advanced Tuning}.

In the example below we calculate the accuracy
(\protect\hyperlink{implemented-performance-measures}{acc}) instead of
the error rate. We use function
\href{http://www.rdocumentation.org/packages/mlr/functions/setAggregation.html}{setAggregation},
as described on the \protect\hyperlink{resampling}{resampling} page, to
additionally obtain the standard deviation of the accuracy. We also use
random search with 100 iterations on the \lstinline!num_set! we defined
above and set \lstinline!show.info! to \lstinline!FALSE! to hide the
output for all 100 iterations:

\begin{lstlisting}[language=R]
num_ps = makeParamSet(
  makeNumericParam("C", lower = -10, upper = 10, trafo = function(x) 10^x),
  makeNumericParam("sigma", lower = -10, upper = 10, trafo = function(x) 10^x)
)
ctrl = makeTuneControlRandom(maxit = 100L)
res = tuneParams("classif.ksvm", task = iris.task, resampling = rdesc, par.set = num_ps,
  control = ctrl, measures = list(acc, setAggregation(acc, test.sd)), show.info = FALSE)
res
#> Tune result:
#> Op. pars: C=95.2; sigma=0.0067
#> acc.test.mean=0.987,acc.test.sd=0.0231
\end{lstlisting}

\paragraph{Accessing the tuning
result}\label{accessing-the-tuning-result}

The result object
\href{http://www.rdocumentation.org/packages/mlr/functions/TuneResult.html}{TuneResult}
allows you to access the best found settings \lstinline!$x! and their
estimated performance \lstinline!$y!.

\begin{lstlisting}[language=R]
res$x
#> $C
#> [1] 95.22422
#> 
#> $sigma
#> [1] 0.006695534

res$y
#> acc.test.mean   acc.test.sd 
#>    0.98666667    0.02309401
\end{lstlisting}

We can generate a
\href{http://www.rdocumentation.org/packages/mlr/functions/makeLearner.html}{Learner}
with optimal hyperparameter settings as follows:

\begin{lstlisting}[language=R]
lrn = setHyperPars(makeLearner("classif.ksvm"), par.vals = res$x)
lrn
#> Learner classif.ksvm from package kernlab
#> Type: classif
#> Name: Support Vector Machines; Short name: ksvm
#> Class: classif.ksvm
#> Properties: twoclass,multiclass,numerics,factors,prob,class.weights
#> Predict-Type: response
#> Hyperparameters: fit=FALSE,C=95.2,sigma=0.0067
\end{lstlisting}

Then you can proceed as usual. Here we refit and predict the learner on
the complete
\href{http://www.rdocumentation.org/packages/datasets/functions/iris.html}{iris}
data set:

\begin{lstlisting}[language=R]
m = train(lrn, iris.task)
predict(m, task = iris.task)
#> Prediction: 150 observations
#> predict.type: response
#> threshold: 
#> time: 0.00
#>   id  truth response
#> 1  1 setosa   setosa
#> 2  2 setosa   setosa
#> 3  3 setosa   setosa
#> 4  4 setosa   setosa
#> 5  5 setosa   setosa
#> 6  6 setosa   setosa
#> ... (150 rows, 3 cols)
\end{lstlisting}

But what if you wanted to inspect the other points on the search path,
not just the optimal?

\paragraph{Investigating hyperparameter tuning
effects}\label{investigating-hyperparameter-tuning-effects}

We can inspect all points evaluated during the search by using
\href{http://www.rdocumentation.org/packages/mlr/functions/generateHyperParsEffectData.html}{generateHyperParsEffectData}:

\begin{lstlisting}[language=R]
generateHyperParsEffectData(res)
#> HyperParsEffectData:
#> Hyperparameters: C,sigma
#> Measures: acc.test.mean,acc.test.sd
#> Optimizer: TuneControlRandom
#> Nested CV Used: FALSE
#> Snapshot of data:
#>            C      sigma acc.test.mean acc.test.sd iteration exec.time
#> 1 -9.9783231  1.0531818     0.2733333  0.02309401         1     0.051
#> 2 -0.5292817  3.2214785     0.2733333  0.02309401         2     0.053
#> 3 -0.3544567  4.1644832     0.2733333  0.02309401         3     0.052
#> 4  0.6341910  7.8640461     0.2866667  0.03055050         4     0.052
#> 5  5.7640748 -3.3159251     0.9533333  0.03055050         5     0.051
#> 6 -6.5880397  0.4600323     0.2733333  0.02309401         6     0.052
\end{lstlisting}

Note that the result of
\href{http://www.rdocumentation.org/packages/mlr/functions/generateHyperParsEffectData.html}{generateHyperParsEffectData}
contains the parameter values \emph{on the original scale}. In order to
get the \emph{transformed} parameter values instead, use the
\lstinline!trafo! argument:

\begin{lstlisting}[language=R]
generateHyperParsEffectData(res, trafo = TRUE)
#> HyperParsEffectData:
#> Hyperparameters: C,sigma
#> Measures: acc.test.mean,acc.test.sd
#> Optimizer: TuneControlRandom
#> Nested CV Used: FALSE
#> Snapshot of data:
#>              C        sigma acc.test.mean acc.test.sd iteration exec.time
#> 1 1.051180e-10 1.130269e+01     0.2733333  0.02309401         1     0.051
#> 2 2.956095e-01 1.665246e+03     0.2733333  0.02309401         2     0.053
#> 3 4.421232e-01 1.460438e+04     0.2733333  0.02309401         3     0.052
#> 4 4.307159e+00 7.312168e+07     0.2866667  0.03055050         4     0.052
#> 5 5.808644e+05 4.831421e-04     0.9533333  0.03055050         5     0.051
#> 6 2.582024e-07 2.884246e+00     0.2733333  0.02309401         6     0.052
\end{lstlisting}

Note that we can also generate performance on the train data along with
the validation/test data, as discussed on the
\href{resample.md\#aggregating-performance-values}{resampling} tutorial
page:

\begin{lstlisting}[language=R]
rdesc2 = makeResampleDesc("Holdout", predict = "both")
res2 = tuneParams("classif.ksvm", task = iris.task, resampling = rdesc2, par.set = num_ps,
  control = ctrl, measures = list(acc, setAggregation(acc, train.mean)), show.info = FALSE)
generateHyperParsEffectData(res2)
#> HyperParsEffectData:
#> Hyperparameters: C,sigma
#> Measures: acc.test.mean,acc.train.mean
#> Optimizer: TuneControlRandom
#> Nested CV Used: FALSE
#> Snapshot of data:
#>           C      sigma acc.test.mean acc.train.mean iteration exec.time
#> 1  9.457202 -4.0536025          0.98           0.97         1     0.040
#> 2  9.900523  1.8815923          0.40           1.00         2     0.030
#> 3  2.363975  5.3202458          0.26           1.00         3     0.029
#> 4 -1.530251  4.7579424          0.26           0.37         4     0.031
#> 5 -7.837476  2.4352698          0.26           0.37         5     0.029
#> 6  8.782931 -0.4143757          0.92           1.00         6     0.029
\end{lstlisting}

We can also easily visualize the points evaluated by using
\href{http://www.rdocumentation.org/packages/mlr/functions/plotHyperParsEffect.html}{plotHyperParsEffect}.
In the example below, we plot the performance over iterations, using the
\lstinline!res! from the previous section but instead with 2 performance
measures:

\begin{lstlisting}[language=R]
res = tuneParams("classif.ksvm", task = iris.task, resampling = rdesc, par.set = num_ps,
  control = ctrl, measures = list(acc, mmce), show.info = FALSE)
data = generateHyperParsEffectData(res)
plotHyperParsEffect(data, x = "iteration", y = "acc.test.mean",
  plot.type = "line")
\end{lstlisting}

\includegraphics{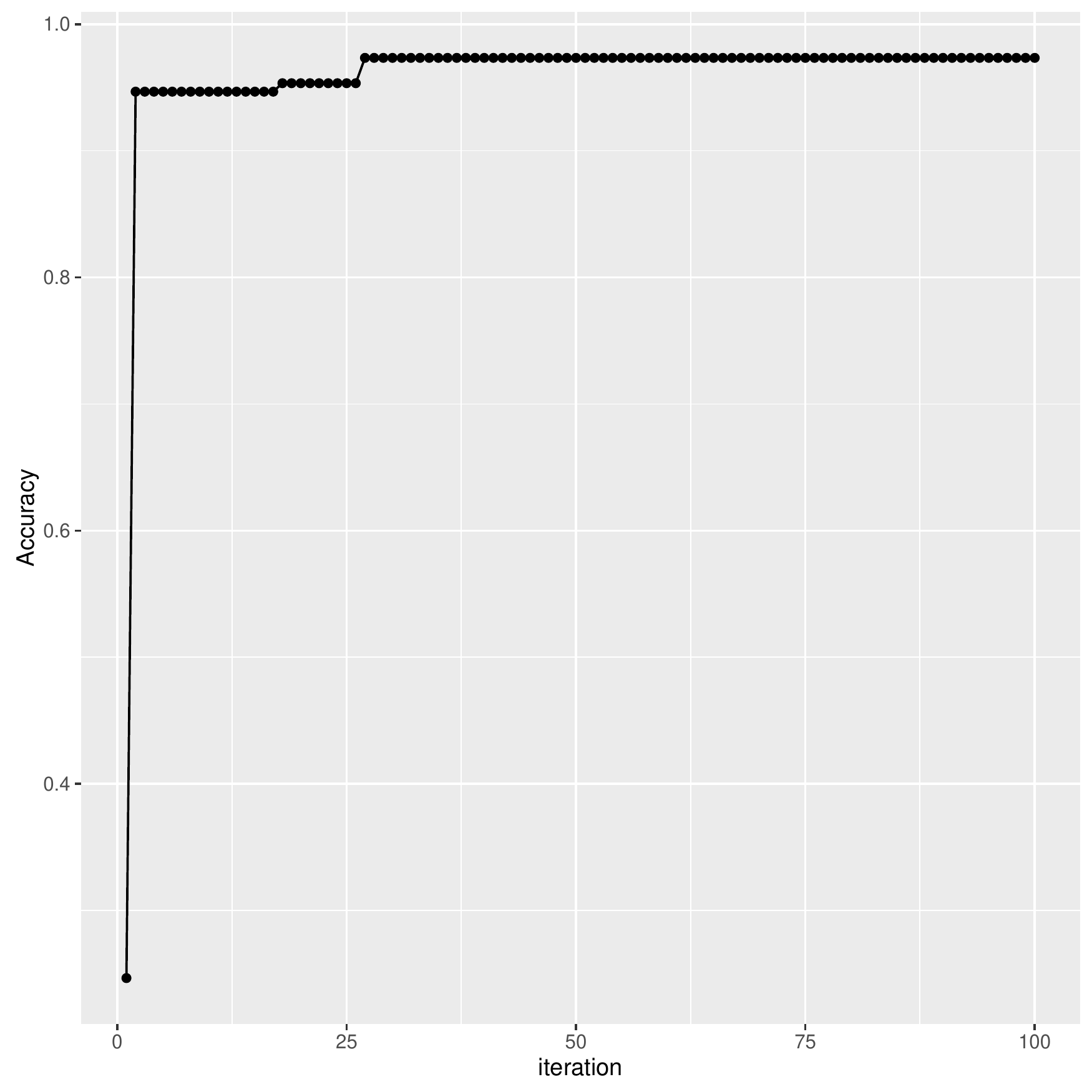}

Note that by default, we only plot the current global optima. This can
be changed with the \lstinline!global.only! argument.

For an in-depth exploration of generating hyperparameter tuning effects
and plotting the data, check out
\protect\hyperlink{evaluating-hyperparameter-tuning}{Hyperparameter
Tuning Effects}.

\subsubsection{Further comments}\label{further-comments}

\begin{itemize}
\item
  Tuning works for all other tasks like regression, survival analysis
  and so on in a completely similar fashion.
\item
  In longer running tuning experiments it is very annoying if the
  computation stops due to numerical or other errors. Have a look at
  \lstinline!on.learner.error! in
  \href{http://www.rdocumentation.org/packages/mlr/functions/configureMlr.html}{configureMlr}
  as well as the examples given in section
  \protect\hyperlink{configuring-mlr}{Configure mlr} of this tutorial.
  You might also want to inform yourself about \lstinline!impute.val! in
  \href{http://www.rdocumentation.org/packages/mlr/functions/TuneControl.html}{TuneControl}.
\item
  As we continually optimize over the same data during tuning, the
  estimated performance value might be optimistically biased. A clean
  approach to ensure unbiased performance estimation is
  \protect\hyperlink{nested-resampling}{nested resampling}, where we
  embed the whole model selection process into an outer resampling loop.
\end{itemize}

\hypertarget{benchmark-experiments}{\subsection{Benchmark
Experiments}\label{benchmark-experiments}}

In a benchmark experiment different learning methods are applied to one
or several data sets with the aim to compare and rank the algorithms
with respect to one or more performance measures.

In \href{http://www.rdocumentation.org/packages/mlr/}{mlr} a benchmark
experiment can be conducted by calling function
\href{http://www.rdocumentation.org/packages/mlr/functions/benchmark.html}{benchmark}
on a
\href{http://www.rdocumentation.org/packages/base/functions/list.html}{list}
of
\href{http://www.rdocumentation.org/packages/mlr/functions/makeLearner.html}{Learner}s
and a
\href{http://www.rdocumentation.org/packages/base/functions/list.html}{list}
of
\href{http://www.rdocumentation.org/packages/mlr/functions/Task.html}{Task}s.
\href{http://www.rdocumentation.org/packages/mlr/functions/benchmark.html}{benchmark}
basically executes
\href{http://www.rdocumentation.org/packages/mlr/functions/resample.html}{resample}
for each combination of
\href{http://www.rdocumentation.org/packages/mlr/functions/makeLearner.html}{Learner}
and
\href{http://www.rdocumentation.org/packages/mlr/functions/Task.html}{Task}.
You can specify an individual resampling strategy for each
\href{http://www.rdocumentation.org/packages/mlr/functions/Task.html}{Task}
and select one or multiple performance measures to be calculated.

\subsubsection{Conducting benchmark
experiments}\label{conducting-benchmark-experiments}

We start with a small example. Two learners,
\href{http://www.rdocumentation.org/packages/MASS/functions/lda.html}{linear
discriminant analysis (lda)} and a
\href{http://www.rdocumentation.org/packages/rpart/functions/rpart.html}{classification
tree (rpart)}, are applied to one classification problem
(\href{http://www.rdocumentation.org/packages/mlr/functions/sonar.task.html}{sonar.task}).
As resampling strategy we choose \lstinline!"Holdout"!. The performance
is thus calculated on a single randomly sampled test data set.

In the example below we create a resample description
(\href{http://www.rdocumentation.org/packages/mlr/functions/makeResampleDesc.html}{ResampleDesc}),
which is automatically instantiated by
\href{http://www.rdocumentation.org/packages/mlr/functions/benchmark.html}{benchmark}.
The instantiation is done only once per
\href{http://www.rdocumentation.org/packages/mlr/functions/Task.html}{Task},
i.e., the same training and test sets are used for all learners. It is
also possible to directly pass a
\href{http://www.rdocumentation.org/packages/mlr/functions/makeResampleInstance.html}{ResampleInstance}.

If you would like to use a \emph{fixed test data set} instead of a
randomly selected one, you can create a suitable
\href{http://www.rdocumentation.org/packages/mlr/functions/makeResampleInstance.html}{ResampleInstance}
through function
\href{http://www.rdocumentation.org/packages/mlr/functions/makeFixedHoldoutInstance.html}{makeFixedHoldoutInstance}.

\begin{lstlisting}[language=R]
### Two learners to be compared
lrns = list(makeLearner("classif.lda"), makeLearner("classif.rpart"))

### Choose the resampling strategy
rdesc = makeResampleDesc("Holdout")

### Conduct the benchmark experiment
bmr = benchmark(lrns, sonar.task, rdesc)
#> Task: Sonar-example, Learner: classif.lda
#> [Resample] holdout iter: 1
#> [Resample] Result: mmce.test.mean= 0.3
#> Task: Sonar-example, Learner: classif.rpart
#> [Resample] holdout iter: 1
#> [Resample] Result: mmce.test.mean=0.286

bmr
#>         task.id    learner.id mmce.test.mean
#> 1 Sonar-example   classif.lda      0.3000000
#> 2 Sonar-example classif.rpart      0.2857143
\end{lstlisting}

In the printed table every row corresponds to one pair of
\href{http://www.rdocumentation.org/packages/mlr/functions/Task.html}{Task}
and
\href{http://www.rdocumentation.org/packages/mlr/functions/makeLearner.html}{Learner}.
The entries show the mean misclassification error
(\protect\hyperlink{implemented-performance-measures}{mmce}), the
default performance measure for classification, on the test data set.

The result \lstinline!bmr! is an object of class
\href{http://www.rdocumentation.org/packages/mlr/functions/BenchmarkResult.html}{BenchmarkResult}.
Basically, it contains a
\href{http://www.rdocumentation.org/packages/base/functions/list.html}{list}
of lists of
\href{http://www.rdocumentation.org/packages/mlr/functions/ResampleResult.html}{ResampleResult}
objects, first ordered by
\href{http://www.rdocumentation.org/packages/mlr/functions/Task.html}{Task}
and then by
\href{http://www.rdocumentation.org/packages/mlr/functions/makeLearner.html}{Learner}.

\subsubsection{Accessing benchmark
results}\label{accessing-benchmark-results}

\href{http://www.rdocumentation.org/packages/mlr/}{mlr} provides several
accessor functions, named \lstinline!getBMR<WhatToExtract>!, that permit
to retrieve information for further analyses. This includes for example
the performances or predictions of the learning algorithms under
consideration.

\paragraph{Learner performances}\label{learner-performances}

Let's have a look at the benchmark result above.
\href{http://www.rdocumentation.org/packages/mlr/functions/getBMRPerformances.html}{getBMRPerformances}
returns individual performances in resampling runs, while
\href{http://www.rdocumentation.org/packages/mlr/functions/getBMRAggrPerformances.html}{getBMRAggrPerformances}
gives the aggregated values.

\begin{lstlisting}[language=R]
getBMRPerformances(bmr)
#> $`Sonar-example`
#> $`Sonar-example`$classif.lda
#>   iter mmce
#> 1    1  0.3
#> 
#> $`Sonar-example`$classif.rpart
#>   iter      mmce
#> 1    1 0.2857143

getBMRAggrPerformances(bmr)
#> $`Sonar-example`
#> $`Sonar-example`$classif.lda
#> mmce.test.mean 
#>            0.3 
#> 
#> $`Sonar-example`$classif.rpart
#> mmce.test.mean 
#>      0.2857143
\end{lstlisting}

Since we used holdout as resampling strategy, individual and aggregated
performance values coincide.

Often it is more convenient to work with
\href{http://www.rdocumentation.org/packages/base/functions/data.frame.html}{data.frame}s.
You can easily convert the result structure by setting
\lstinline!as.df = TRUE!.

\begin{lstlisting}[language=R]
getBMRPerformances(bmr, as.df = TRUE)
#>         task.id    learner.id iter      mmce
#> 1 Sonar-example   classif.lda    1 0.3000000
#> 2 Sonar-example classif.rpart    1 0.2857143

getBMRAggrPerformances(bmr, as.df = TRUE)
#>         task.id    learner.id mmce.test.mean
#> 1 Sonar-example   classif.lda      0.3000000
#> 2 Sonar-example classif.rpart      0.2857143
\end{lstlisting}

\paragraph{Predictions}\label{predictions}

Per default, the
\href{http://www.rdocumentation.org/packages/mlr/functions/BenchmarkResult.html}{BenchmarkResult}
contains the learner predictions. If you do not want to keep them, e.g.,
to conserve memory, set \lstinline!keep.pred = FALSE! when calling
\href{http://www.rdocumentation.org/packages/mlr/functions/benchmark.html}{benchmark}.

You can access the predictions using function
\href{http://www.rdocumentation.org/packages/mlr/functions/getBMRPredictions.html}{getBMRPredictions}.
Per default, you get a
\href{http://www.rdocumentation.org/packages/base/functions/list.html}{list}
of lists of
\href{http://www.rdocumentation.org/packages/mlr/functions/ResamplePrediction.html}{ResamplePrediction}
objects. In most cases you might prefer the
\href{http://www.rdocumentation.org/packages/base/functions/data.frame.html}{data.frame}
version.

\begin{lstlisting}[language=R]
getBMRPredictions(bmr)
#> $`Sonar-example`
#> $`Sonar-example`$classif.lda
#> Resampled Prediction for:
#> Resample description: holdout with 0.67 split rate.
#> Predict: test
#> Stratification: FALSE
#> predict.type: response
#> threshold: 
#> time (mean): 0.00
#>    id truth response iter  set
#> 1 180     M        M    1 test
#> 2 100     M        R    1 test
#> 3  53     R        M    1 test
#> 4  89     R        R    1 test
#> 5  92     R        M    1 test
#> 6  11     R        R    1 test
#> ... (70 rows, 5 cols)
#> 
#> 
#> $`Sonar-example`$classif.rpart
#> Resampled Prediction for:
#> Resample description: holdout with 0.67 split rate.
#> Predict: test
#> Stratification: FALSE
#> predict.type: response
#> threshold: 
#> time (mean): 0.00
#>    id truth response iter  set
#> 1 180     M        M    1 test
#> 2 100     M        M    1 test
#> 3  53     R        R    1 test
#> 4  89     R        M    1 test
#> 5  92     R        M    1 test
#> 6  11     R        R    1 test
#> ... (70 rows, 5 cols)

head(getBMRPredictions(bmr, as.df = TRUE))
#>         task.id  learner.id  id truth response iter  set
#> 1 Sonar-example classif.lda 180     M        M    1 test
#> 2 Sonar-example classif.lda 100     M        R    1 test
#> 3 Sonar-example classif.lda  53     R        M    1 test
#> 4 Sonar-example classif.lda  89     R        R    1 test
#> 5 Sonar-example classif.lda  92     R        M    1 test
#> 6 Sonar-example classif.lda  11     R        R    1 test
\end{lstlisting}

It is also easily possible to access results for certain learners or
tasks via their IDs. For this purpose many ``getter'' functions have a
\lstinline!learner.ids! and a \lstinline!task.ids! argument.

\begin{lstlisting}[language=R]
head(getBMRPredictions(bmr, learner.ids = "classif.rpart", as.df = TRUE))
#>         task.id    learner.id  id truth response iter  set
#> 1 Sonar-example classif.rpart 180     M        M    1 test
#> 2 Sonar-example classif.rpart 100     M        M    1 test
#> 3 Sonar-example classif.rpart  53     R        R    1 test
#> 4 Sonar-example classif.rpart  89     R        M    1 test
#> 5 Sonar-example classif.rpart  92     R        M    1 test
#> 6 Sonar-example classif.rpart  11     R        R    1 test
\end{lstlisting}

If you don't like the default IDs, you can set the IDs of learners and
tasks via the \lstinline!id! option of
\href{http://www.rdocumentation.org/packages/mlr/functions/makeLearner.html}{makeLearner}
and
\href{http://www.rdocumentation.org/packages/mlr/functions/Task.html}{make*Task}.
Moreover, you can conveniently change the ID of a
\href{http://www.rdocumentation.org/packages/mlr/functions/makeLearner.html}{Learner}
via function
\href{http://www.rdocumentation.org/packages/mlr/functions/setLearnerId.html}{setLearnerId}.

\paragraph{IDs}\label{ids}

The IDs of all
\href{http://www.rdocumentation.org/packages/mlr/functions/makeLearner.html}{Learner}s,
\href{http://www.rdocumentation.org/packages/mlr/functions/Task.html}{Task}s
and
\href{http://www.rdocumentation.org/packages/mlr/functions/makeMeasure.html}{Measure}s
in a benchmark experiment can be retrieved as follows:

\begin{lstlisting}[language=R]
getBMRTaskIds(bmr)
#> [1] "Sonar-example"

getBMRLearnerIds(bmr)
#> [1] "classif.lda"   "classif.rpart"

getBMRMeasureIds(bmr)
#> [1] "mmce"
\end{lstlisting}

\paragraph{Learner models}\label{learner-models}

Per default the
\href{http://www.rdocumentation.org/packages/mlr/functions/BenchmarkResult.html}{BenchmarkResult}
also contains the fitted models for all learners on all tasks. If you do
not want to keep them set \lstinline!models = FALSE! when calling
\href{http://www.rdocumentation.org/packages/mlr/functions/benchmark.html}{benchmark}.
The fitted models can be retrieved by function
\href{http://www.rdocumentation.org/packages/mlr/functions/getBMRModels.html}{getBMRModels}.
It returns a
\href{http://www.rdocumentation.org/packages/base/functions/list.html}{list}
of lists of
\href{http://www.rdocumentation.org/packages/mlr/functions/makeWrappedModel.html}{WrappedModel}
objects.

\begin{lstlisting}[language=R]
getBMRModels(bmr)
#> $`Sonar-example`
#> $`Sonar-example`$classif.lda
#> $`Sonar-example`$classif.lda[[1]]
#> Model for learner.id=classif.lda; learner.class=classif.lda
#> Trained on: task.id = Sonar-example; obs = 138; features = 60
#> Hyperparameters: 
#> 
#> 
#> $`Sonar-example`$classif.rpart
#> $`Sonar-example`$classif.rpart[[1]]
#> Model for learner.id=classif.rpart; learner.class=classif.rpart
#> Trained on: task.id = Sonar-example; obs = 138; features = 60
#> Hyperparameters: xval=0

getBMRModels(bmr, learner.ids = "classif.lda")
#> $`Sonar-example`
#> $`Sonar-example`$classif.lda
#> $`Sonar-example`$classif.lda[[1]]
#> Model for learner.id=classif.lda; learner.class=classif.lda
#> Trained on: task.id = Sonar-example; obs = 138; features = 60
#> Hyperparameters:
\end{lstlisting}

\paragraph{Learners and measures}\label{learners-and-measures}

Moreover, you can extract the employed
\href{http://www.rdocumentation.org/packages/mlr/functions/makeLearner.html}{Learner}s
and
\href{http://www.rdocumentation.org/packages/mlr/functions/makeMeasure.html}{Measure}s.

\begin{lstlisting}[language=R]
getBMRLearners(bmr)
#> $classif.lda
#> Learner classif.lda from package MASS
#> Type: classif
#> Name: Linear Discriminant Analysis; Short name: lda
#> Class: classif.lda
#> Properties: twoclass,multiclass,numerics,factors,prob
#> Predict-Type: response
#> Hyperparameters: 
#> 
#> 
#> $classif.rpart
#> Learner classif.rpart from package rpart
#> Type: classif
#> Name: Decision Tree; Short name: rpart
#> Class: classif.rpart
#> Properties: twoclass,multiclass,missings,numerics,factors,ordered,prob,weights,featimp
#> Predict-Type: response
#> Hyperparameters: xval=0

getBMRMeasures(bmr)
#> [[1]]
#> Name: Mean misclassification error
#> Performance measure: mmce
#> Properties: classif,classif.multi,req.pred,req.truth
#> Minimize: TRUE
#> Best: 0; Worst: 1
#> Aggregated by: test.mean
#> Note:
\end{lstlisting}

\subsubsection{Merging benchmark
results}\label{merging-benchmark-results}

Sometimes after completing a benchmark experiment it turns out that you
want to extend it by another
\href{http://www.rdocumentation.org/packages/mlr/functions/makeLearner.html}{Learner}
or another
\href{http://www.rdocumentation.org/packages/mlr/functions/Task.html}{Task}.
In this case you can perform an additional benchmark experiment and then
merge the results to get a single
\href{http://www.rdocumentation.org/packages/mlr/functions/BenchmarkResult.html}{BenchmarkResult}
object that can be accessed and analyzed as usual.

\href{http://www.rdocumentation.org/packages/mlr/}{mlr} provides two
functions to merge results:
\href{http://www.rdocumentation.org/packages/mlr/functions/mergeBenchmarkResultLearner.html}{mergeBenchmarkResultLearner}
combines two or more benchmark results for different sets of learners on
the same
\href{http://www.rdocumentation.org/packages/mlr/functions/Task.html}{Task}s,
while
\href{http://www.rdocumentation.org/packages/mlr/functions/mergeBenchmarkResultTask.html}{mergeBenchmarkResultTask}
fuses results obtained with the same
\href{http://www.rdocumentation.org/packages/mlr/functions/makeLearner.html}{Learner}s
on different sets of
\href{http://www.rdocumentation.org/packages/mlr/functions/Task.html}{Task}s.

For example in the benchmark experiment above we applied
\href{http://www.rdocumentation.org/packages/MASS/functions/lda.html}{lda}
and
\href{http://www.rdocumentation.org/packages/rpart/functions/rpart.html}{rpart}
to the
\href{http://www.rdocumentation.org/packages/mlr/functions/sonar.task.html}{sonar.task}.
We now perform a second experiment using a
\href{http://www.rdocumentation.org/packages/randomForest/functions/randomForest.html}{random
forest} and
\href{http://www.rdocumentation.org/packages/MASS/functions/qda.html}{quadratic
discriminant analysis (qda)} and use
\href{http://www.rdocumentation.org/packages/mlr/functions/mergeBenchmarkResultLearner.html}{mergeBenchmarkResultLearner}
to combine the results.

\begin{lstlisting}[language=R]
### First benchmark result
bmr
#>         task.id    learner.id mmce.test.mean
#> 1 Sonar-example   classif.lda      0.3000000
#> 2 Sonar-example classif.rpart      0.2857143

### Benchmark experiment for the additional learners
lrns2 = list(makeLearner("classif.randomForest"), makeLearner("classif.qda"))
bmr2 = benchmark(lrns2, sonar.task, rdesc, show.info = FALSE)
bmr2
#>         task.id           learner.id mmce.test.mean
#> 1 Sonar-example classif.randomForest      0.2000000
#> 2 Sonar-example          classif.qda      0.5142857

### Merge the results
mergeBenchmarkResultLearner(bmr, bmr2)
#>         task.id           learner.id mmce.test.mean
#> 1 Sonar-example          classif.lda      0.3000000
#> 2 Sonar-example        classif.rpart      0.2857143
#> 3 Sonar-example classif.randomForest      0.2000000
#> 4 Sonar-example          classif.qda      0.5142857
\end{lstlisting}

Note that in the above examples in each case a
\href{http://www.rdocumentation.org/packages/mlr/functions/makeResampleDesc.html}{resample
description} was passed to the
\href{http://www.rdocumentation.org/packages/mlr/functions/benchmark.html}{benchmark}
function. For this reason
\href{http://www.rdocumentation.org/packages/MASS/functions/lda.html}{lda}
and
\href{http://www.rdocumentation.org/packages/rpart/functions/rpart.html}{rpart}
were most likely evaluated on a different training/test set pair than
\href{http://www.rdocumentation.org/packages/randomForest/functions/randomForest.html}{random
forest} and
\href{http://www.rdocumentation.org/packages/MASS/functions/qda.html}{qda}.

Differing training/test set pairs across learners pose an additional
source of variation in the results, which can make it harder to detect
actual performance differences between learners. Therefore, if you
suspect that you will have to extend your benchmark experiment by
another
\href{http://www.rdocumentation.org/packages/mlr/functions/makeLearner.html}{Learner}
later on it's probably easiest to work with
\href{http://www.rdocumentation.org/packages/mlr/functions/makeResampleInstance.html}{ResampleInstance}s
from the start. These can be stored and used for any additional
experiments.

Alternatively, if you used a resample description in the first benchmark
experiment you could also extract the
\href{http://www.rdocumentation.org/packages/mlr/functions/makeResampleInstance.html}{ResampleInstance}s
from the
\href{http://www.rdocumentation.org/packages/mlr/functions/BenchmarkResult.html}{BenchmarkResult}
\lstinline!bmr! and pass these to all further
\href{http://www.rdocumentation.org/packages/mlr/functions/benchmark.html}{benchmark}
calls.

\begin{lstlisting}[language=R]
rin = getBMRPredictions(bmr)[[1]][[1]]$instance
rin
#> Resample instance for 208 cases.
#> Resample description: holdout with 0.67 split rate.
#> Predict: test
#> Stratification: FALSE

### Benchmark experiment for the additional random forest
bmr3 = benchmark(lrns2, sonar.task, rin, show.info = FALSE)
bmr3
#>         task.id           learner.id mmce.test.mean
#> 1 Sonar-example classif.randomForest      0.2714286
#> 2 Sonar-example          classif.qda      0.3857143

### Merge the results
mergeBenchmarkResultLearner(bmr, bmr3)
#>         task.id           learner.id mmce.test.mean
#> 1 Sonar-example          classif.lda      0.3000000
#> 2 Sonar-example        classif.rpart      0.2857143
#> 3 Sonar-example classif.randomForest      0.2714286
#> 4 Sonar-example          classif.qda      0.3857143
\end{lstlisting}

\subsubsection{Benchmark analysis and
visualization}\label{benchmark-analysis-and-visualization}

\href{http://www.rdocumentation.org/packages/mlr/}{mlr} offers several
ways to analyze the results of a benchmark experiment. This includes
visualization, ranking of learning algorithms and hypothesis tests to
assess performance differences between learners.

In order to demonstrate the functionality we conduct a slightly larger
benchmark experiment with three learning algorithms that are applied to
five classification tasks.

\paragraph{Example: Comparing lda, rpart and random
Forest}\label{example-comparing-lda-rpart-and-random-forest}

We consider
\href{http://www.rdocumentation.org/packages/MASS/functions/lda.html}{linear
discriminant analysis (lda)},
\href{http://www.rdocumentation.org/packages/rpart/functions/rpart.html}{classification
trees (rpart)}, and
\href{http://www.rdocumentation.org/packages/randomForest/functions/randomForest.html}{random
forests (randomForest)}. Since the default learner IDs are a little
long, we choose shorter names in the \textbf{R} code below.

We use five classification tasks. Three are already provided by
\href{http://www.rdocumentation.org/packages/mlr/}{mlr}, two more data
sets are taken from package
\href{http://www.rdocumentation.org/packages/mlbench/}{mlbench} and
converted to
\href{http://www.rdocumentation.org/packages/mlr/functions/Task.html}{Task}s
by function
\href{http://www.rdocumentation.org/packages/mlr/functions/convertMLBenchObjToTask.html}{convertMLBenchObjToTask}.

For all tasks 10-fold cross-validation is chosen as resampling strategy.
This is achieved by passing a single
\href{http://www.rdocumentation.org/packages/mlr/functions/makeResampleDesc.html}{resample
description} to
\href{http://www.rdocumentation.org/packages/mlr/functions/benchmark.html}{benchmark},
which is then instantiated automatically once for each
\href{http://www.rdocumentation.org/packages/mlr/functions/Task.html}{Task}.
This way, the same instance is used for all learners applied to a single
task.

It is also possible to choose a different resampling strategy for each
\href{http://www.rdocumentation.org/packages/mlr/functions/Task.html}{Task}
by passing a
\href{http://www.rdocumentation.org/packages/base/functions/list.html}{list}
of the same length as the number of tasks that can contain both
\href{http://www.rdocumentation.org/packages/mlr/functions/makeResampleDesc.html}{resample
descriptions} and
\href{http://www.rdocumentation.org/packages/mlr/functions/makeResampleInstance.html}{resample
instances}.

We use the mean misclassification error
\protect\hyperlink{implemented-performance-measures}{mmce} as primary
performance measure, but also calculate the balanced error rate
(\protect\hyperlink{implemented-performance-measures}{ber}) and the
training time
(\protect\hyperlink{implemented-performance-measures}{timetrain}).

\begin{lstlisting}[language=R]
### Create a list of learners
lrns = list(
  makeLearner("classif.lda", id = "lda"),
  makeLearner("classif.rpart", id = "rpart"),
  makeLearner("classif.randomForest", id = "randomForest")
)

### Get additional Tasks from package mlbench
ring.task = convertMLBenchObjToTask("mlbench.ringnorm", n = 600)
wave.task = convertMLBenchObjToTask("mlbench.waveform", n = 600)

tasks = list(iris.task, sonar.task, pid.task, ring.task, wave.task)
rdesc = makeResampleDesc("CV", iters = 10)
meas = list(mmce, ber, timetrain)
bmr = benchmark(lrns, tasks, rdesc, meas, show.info = FALSE)
bmr
#>                        task.id   learner.id mmce.test.mean ber.test.mean
#> 1                 iris-example          lda     0.02000000    0.02222222
#> 2                 iris-example        rpart     0.08000000    0.07555556
#> 3                 iris-example randomForest     0.05333333    0.05250000
#> 4             mlbench.ringnorm          lda     0.35000000    0.34605671
#> 5             mlbench.ringnorm        rpart     0.17333333    0.17313632
#> 6             mlbench.ringnorm randomForest     0.05833333    0.05806121
#> 7             mlbench.waveform          lda     0.19000000    0.18257244
#> 8             mlbench.waveform        rpart     0.28833333    0.28765247
#> 9             mlbench.waveform randomForest     0.16500000    0.16306057
#> 10 PimaIndiansDiabetes-example          lda     0.22778537    0.27148893
#> 11 PimaIndiansDiabetes-example        rpart     0.25133288    0.28967870
#> 12 PimaIndiansDiabetes-example randomForest     0.23685919    0.27543146
#> 13               Sonar-example          lda     0.24619048    0.23986694
#> 14               Sonar-example        rpart     0.30785714    0.31153361
#> 15               Sonar-example randomForest     0.17785714    0.17442696
#>    timetrain.test.mean
#> 1               0.0022
#> 2               0.0035
#> 3               0.0374
#> 4               0.0062
#> 5               0.0088
#> 6               0.3726
#> 7               0.0066
#> 8               0.0256
#> 9               0.4191
#> 10              0.0035
#> 11              0.0048
#> 12              0.3431
#> 13              0.0127
#> 14              0.0105
#> 15              0.2280
\end{lstlisting}

From the aggregated performance values we can see that for the iris- and
PimaIndiansDiabetes-example
\href{http://www.rdocumentation.org/packages/MASS/functions/lda.html}{linear
discriminant analysis} performs well while for all other tasks the
\href{http://www.rdocumentation.org/packages/randomForest/functions/randomForest.html}{random
forest} seems superior. Training takes longer for the
\href{http://www.rdocumentation.org/packages/randomForest/functions/randomForest.html}{random
forest} than for the other learners.

In order to draw any conclusions from the average performances at least
their variability has to be taken into account or, preferably, the
distribution of performance values across resampling iterations.

The individual performances on the 10 folds for every task, learner, and
measure are retrieved below.

\begin{lstlisting}[language=R]
perf = getBMRPerformances(bmr, as.df = TRUE)
head(perf)
#>        task.id learner.id iter      mmce       ber timetrain
#> 1 iris-example        lda    1 0.0000000 0.0000000     0.002
#> 2 iris-example        lda    2 0.1333333 0.1666667     0.002
#> 3 iris-example        lda    3 0.0000000 0.0000000     0.002
#> 4 iris-example        lda    4 0.0000000 0.0000000     0.003
#> 5 iris-example        lda    5 0.0000000 0.0000000     0.002
#> 6 iris-example        lda    6 0.0000000 0.0000000     0.002
\end{lstlisting}

A closer look at the result reveals that the
\href{http://www.rdocumentation.org/packages/randomForest/functions/randomForest.html}{random
forest} outperforms the
\href{http://www.rdocumentation.org/packages/rpart/functions/rpart.html}{classification
tree} in every instance, while
\href{http://www.rdocumentation.org/packages/MASS/functions/lda.html}{linear
discriminant analysis} performs better than
\href{http://www.rdocumentation.org/packages/rpart/functions/rpart.html}{rpart}
most of the time. Additionally
\href{http://www.rdocumentation.org/packages/MASS/functions/lda.html}{lda}
sometimes even beats the
\href{http://www.rdocumentation.org/packages/randomForest/functions/randomForest.html}{random
forest}. With increasing size of such
\href{http://www.rdocumentation.org/packages/mlr/functions/benchmark.html}{benchmark}
experiments, those tables become almost unreadable and hard to
comprehend.

\href{http://www.rdocumentation.org/packages/mlr/}{mlr} features some
plotting functions to visualize results of benchmark experiments that
you might find useful. Moreover,
\href{http://www.rdocumentation.org/packages/mlr/}{mlr} offers
statistical hypothesis tests to assess performance differences between
learners.

\paragraph{Integrated plots}\label{integrated-plots}

Plots are generated using
\href{http://www.rdocumentation.org/packages/ggplot2/}{ggplot2}. Further
customization, such as renaming plot elements or changing colors, is
easily possible.

\subparagraph{Visualizing performances}\label{visualizing-performances}

\href{http://www.rdocumentation.org/packages/mlr/functions/plotBMRBoxplots.html}{plotBMRBoxplots}
creates box or violin plots which show the distribution of performance
values across resampling iterations for one performance measure and for
all learners and tasks (and thus visualize the output of
\href{http://www.rdocumentation.org/packages/mlr/functions/getBMRPerformances.html}{getBMRPerformances}).

Below are both variants, box and violin plots. The first plot shows the
\protect\hyperlink{implemented-performance-measures}{mmce} and the
second plot the balanced error rate
(\protect\hyperlink{implemented-performance-measures}{ber}). Moreover,
in the second plot we color the boxes according to the learners to make
them better distinguishable.

\begin{lstlisting}[language=R]
plotBMRBoxplots(bmr, measure = mmce)
\end{lstlisting}

\includegraphics{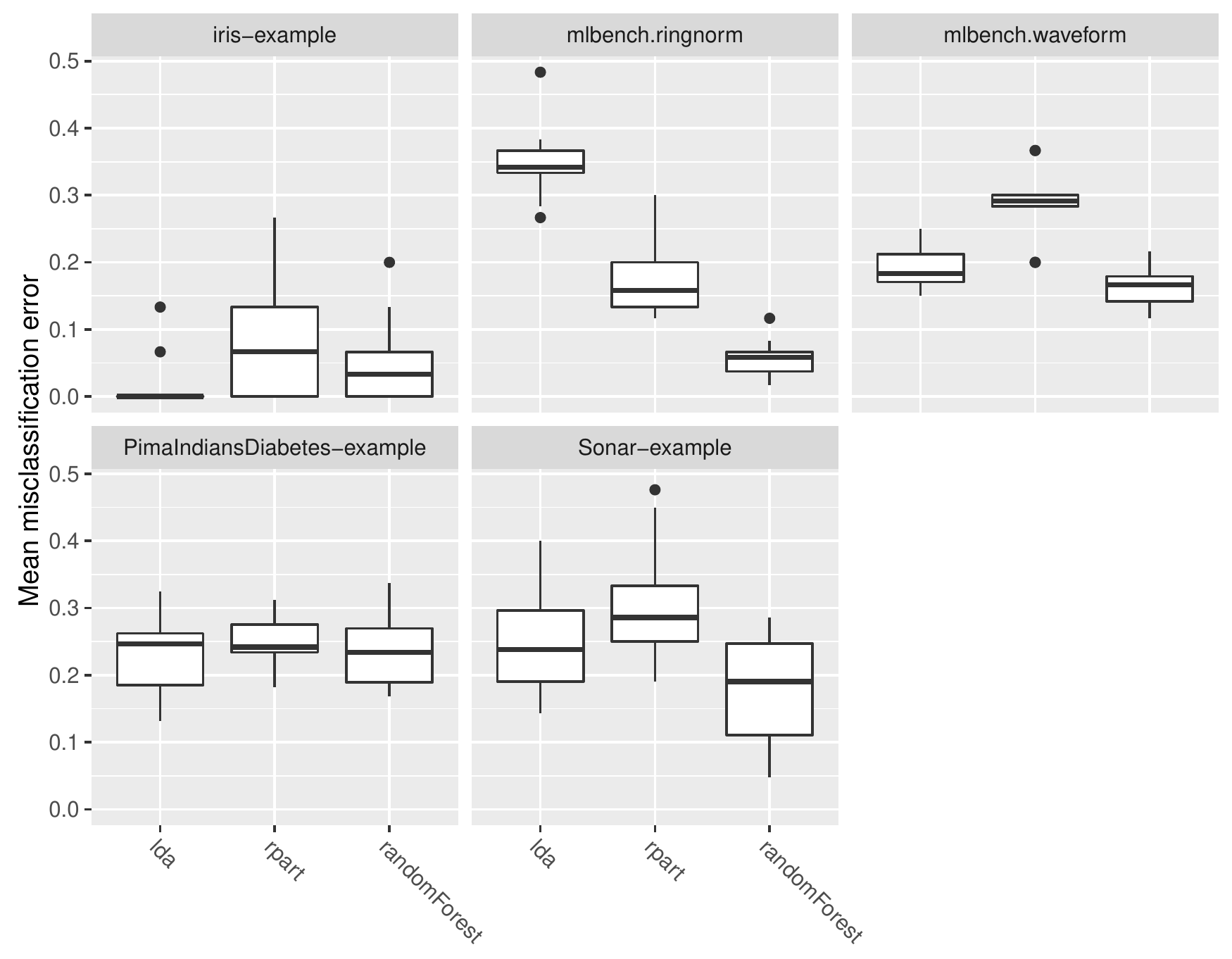}

\begin{lstlisting}[language=R]
plotBMRBoxplots(bmr, measure = ber, style = "violin", pretty.names = FALSE) +
  aes(color = learner.id) +
  theme(strip.text.x = element_text(size = 8))
\end{lstlisting}

\includegraphics{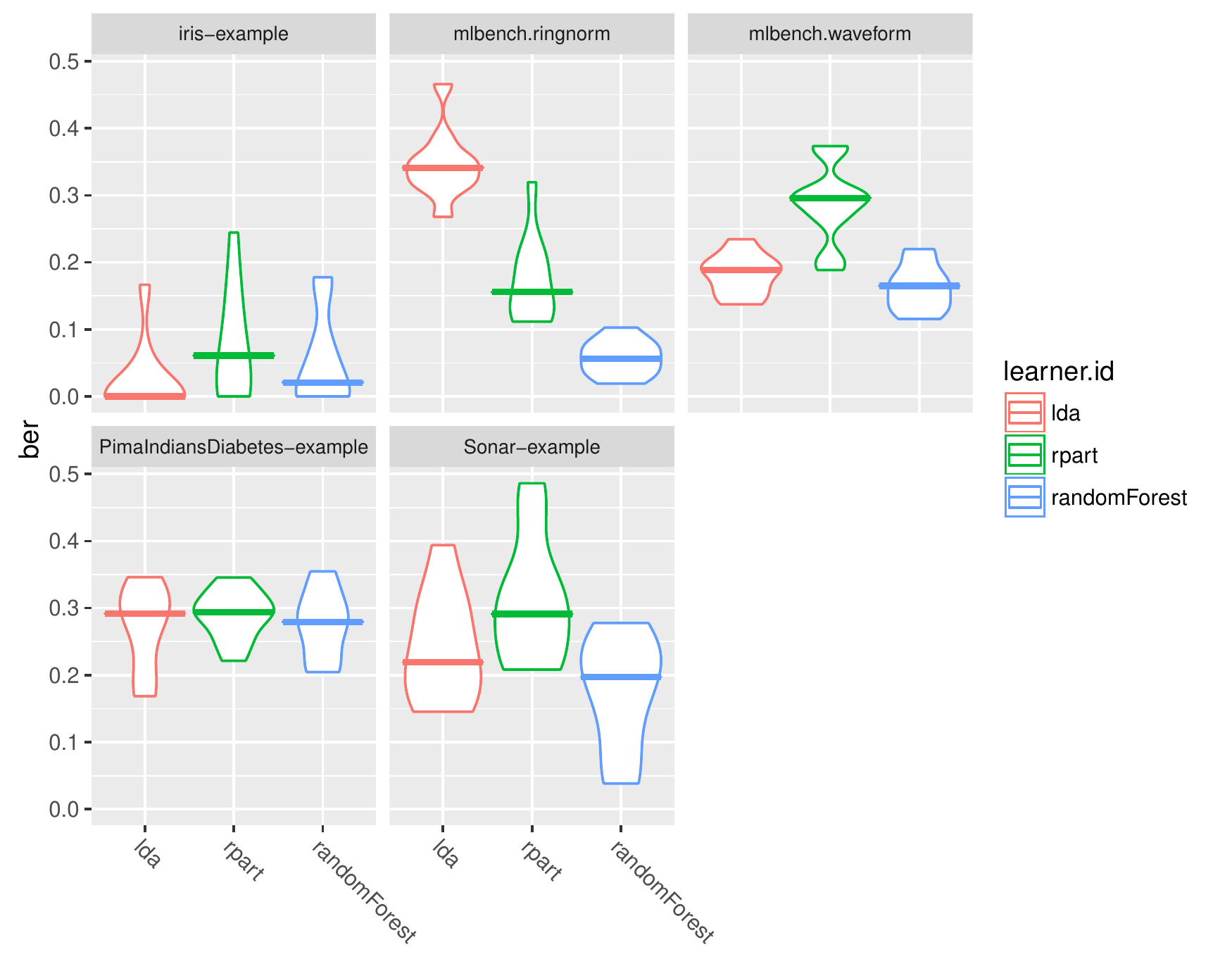}

Note that by default the measure \lstinline!name!s are used as labels
for the y-axis.

\begin{lstlisting}[language=R]
mmce$name
#> [1] "Mean misclassification error"
mmce$id
#> [1] "mmce"
\end{lstlisting}

If you prefer the shorter \lstinline!id!s like mmce and ber set
\lstinline!pretty.names = FALSE! (as done for the second plot). Of
course you can also use the
\href{http://www.rdocumentation.org/packages/ggplot2/functions/labs.html}{ylab}
function to choose a completely different label.

Another thing which probably comes up quite often is changing the panel
headers (which default to the
\href{http://www.rdocumentation.org/packages/mlr/functions/Task.html}{Task}
IDs) and the learner names on the x-axis (which default to the
\href{http://www.rdocumentation.org/packages/mlr/functions/makeLearner.html}{Learner}
IDs). For example looking at the above plots we would like to remove the
``example'' suffixes and the ``mlbench'' prefixes from the panel
headers. Moreover, compared to the other learner names ``randomForest''
seems a little long. Currently, the probably simplest solution is to
change the factor levels of the plotted data as shown below.

\begin{lstlisting}[language=R]
plt = plotBMRBoxplots(bmr, measure = mmce)
head(plt$data)
#>        task.id learner.id iter      mmce       ber timetrain
#> 1 iris-example        lda    1 0.0000000 0.0000000     0.002
#> 2 iris-example        lda    2 0.1333333 0.1666667     0.002
#> 3 iris-example        lda    3 0.0000000 0.0000000     0.002
#> 4 iris-example        lda    4 0.0000000 0.0000000     0.003
#> 5 iris-example        lda    5 0.0000000 0.0000000     0.002
#> 6 iris-example        lda    6 0.0000000 0.0000000     0.002

levels(plt$data$task.id) = c("Iris", "Ringnorm", "Waveform", "Diabetes", "Sonar")
levels(plt$data$learner.id) = c("lda", "rpart", "rF")

plt + ylab("Error rate")
\end{lstlisting}

\includegraphics{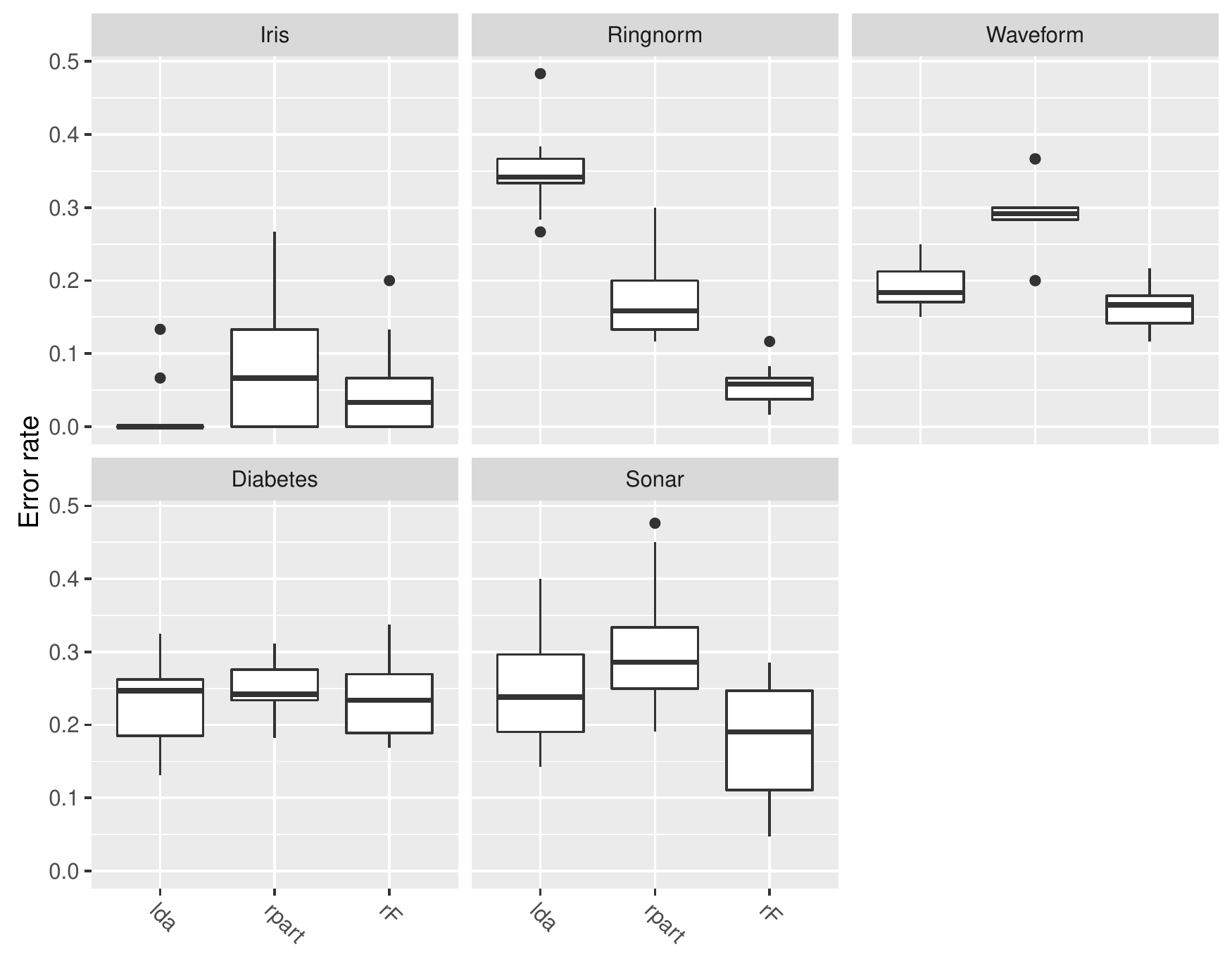}

\subparagraph{Visualizing aggregated
performances}\label{visualizing-aggregated-performances}

The aggregated performance values (resulting from
\href{http://www.rdocumentation.org/packages/mlr/functions/getBMRAggrPerformances.html}{getBMRAggrPerformances})
can be visualized by function
\href{http://www.rdocumentation.org/packages/mlr/functions/plotBMRSummary.html}{plotBMRSummary}.
This plot draws one line for each task on which the aggregated values of
one performance measure for all learners are displayed. By default, the
first measure in the
\href{http://www.rdocumentation.org/packages/base/functions/list.html}{list}
of
\href{http://www.rdocumentation.org/packages/mlr/functions/makeMeasure.html}{Measure}s
passed to
\href{http://www.rdocumentation.org/packages/mlr/functions/benchmark.html}{benchmark}
is used, in our example
\protect\hyperlink{implemented-performance-measures}{mmce}. Moreover, a
small vertical jitter is added to prevent overplotting.

\begin{lstlisting}[language=R]
plotBMRSummary(bmr)
\end{lstlisting}

\includegraphics{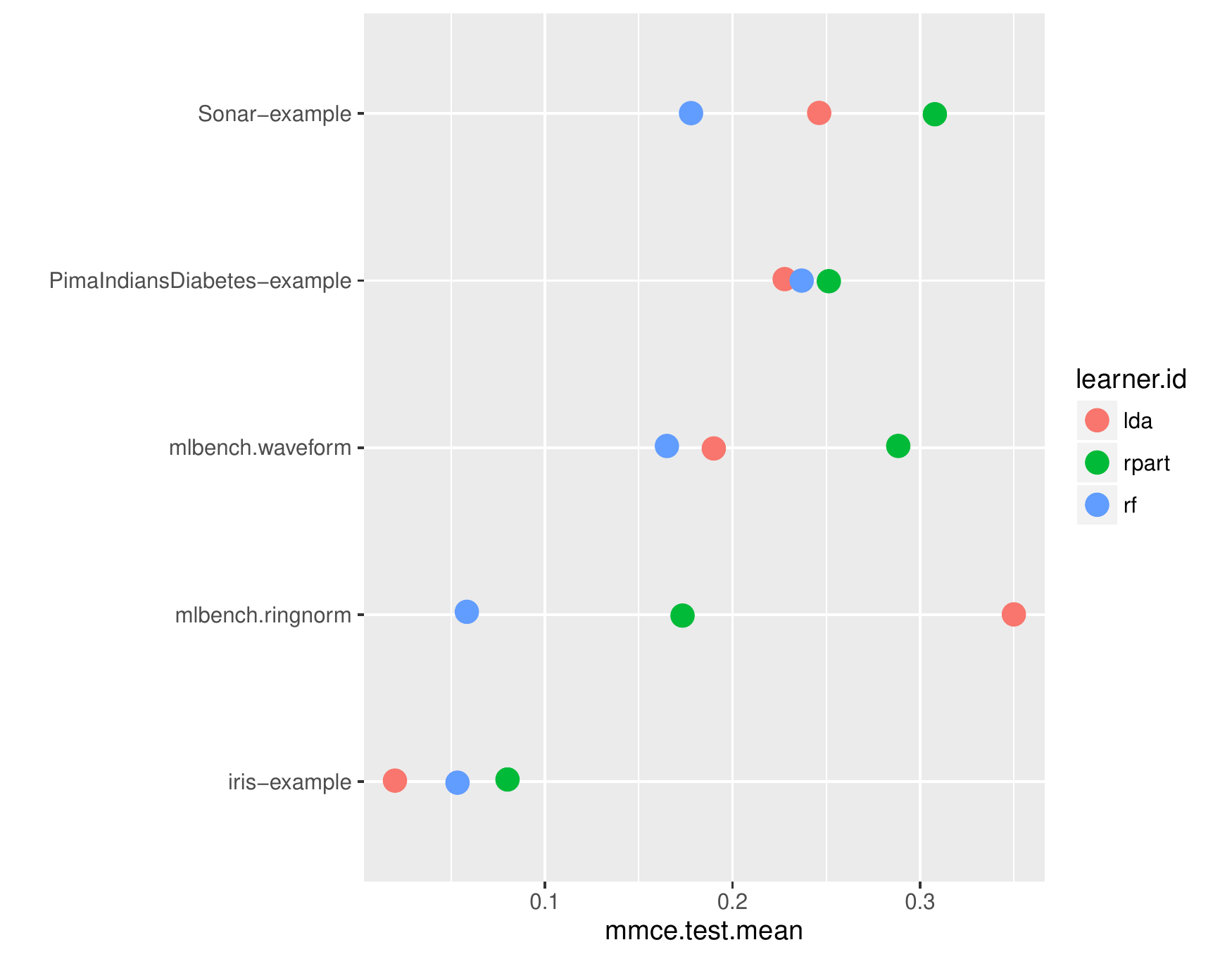}

\subparagraph{Calculating and visualizing
ranks}\label{calculating-and-visualizing-ranks}

Additional to the absolute performance, relative performance, i.e.,
ranking the learners is usually of interest and might provide valuable
additional insight.

Function
\href{http://www.rdocumentation.org/packages/mlr/functions/convertBMRToRankMatrix.html}{convertBMRToRankMatrix}
calculates ranks based on aggregated learner performances of one
measure. We choose the mean misclassification error
(\protect\hyperlink{implemented-performance-measures}{mmce}). The rank
structure can be visualized by
\href{http://www.rdocumentation.org/packages/mlr/functions/plotBMRRanksAsBarChart.html}{plotBMRRanksAsBarChart}.

\begin{lstlisting}[language=R]
m = convertBMRToRankMatrix(bmr, mmce)
m
#>              iris-example mlbench.ringnorm mlbench.waveform
#> lda                     1                3                2
#> rpart                   3                2                3
#> randomForest            2                1                1
#>              PimaIndiansDiabetes-example Sonar-example
#> lda                                    1             2
#> rpart                                  3             3
#> randomForest                           2             1
\end{lstlisting}

Methods with best performance, i.e., with lowest
\protect\hyperlink{implemented-performance-measures}{mmce}, are assigned
the lowest rank.
\href{http://www.rdocumentation.org/packages/MASS/functions/lda.html}{Linear
discriminant analysis} is best for the iris and
PimaIndiansDiabetes-examples while the
\href{http://www.rdocumentation.org/packages/randomForest/functions/randomForest.html}{random
forest} shows best results on the remaining tasks.

\href{http://www.rdocumentation.org/packages/mlr/functions/plotBMRRanksAsBarChart.html}{plotBMRRanksAsBarChart}
with option \lstinline!pos = "tile"! shows a corresponding heat map. The
ranks are displayed on the x-axis and the learners are color-coded.

\begin{lstlisting}[language=R]
plotBMRRanksAsBarChart(bmr, pos = "tile")
\end{lstlisting}

\includegraphics{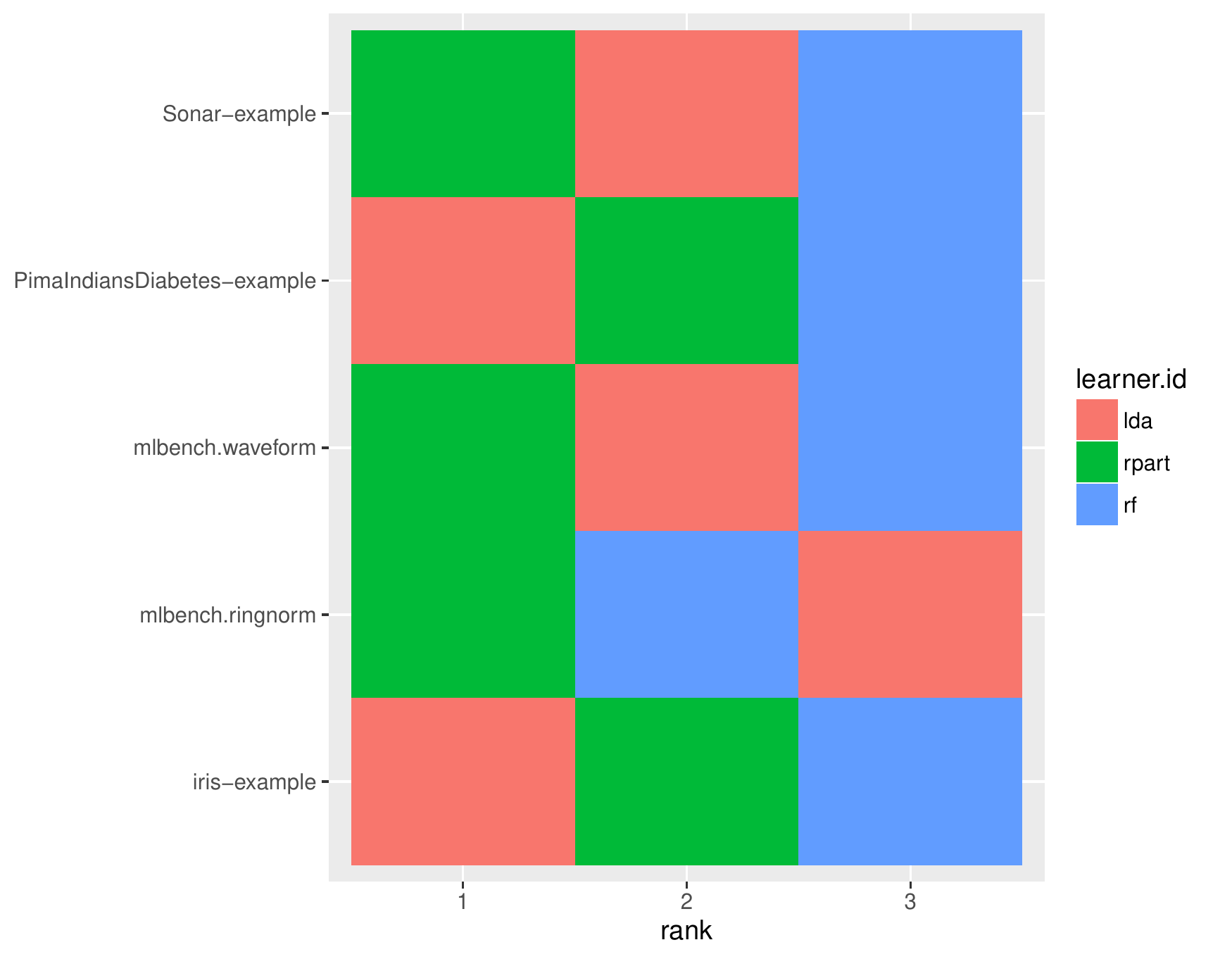}

A similar plot can also be obtained via
\href{http://www.rdocumentation.org/packages/mlr/functions/plotBMRSummary.html}{plotBMRSummary}.
With option \lstinline!trafo = "rank"! the ranks are displayed instead
of the aggregated performances.

\begin{lstlisting}[language=R]
plotBMRSummary(bmr, trafo = "rank", jitter = 0)
\end{lstlisting}

\includegraphics{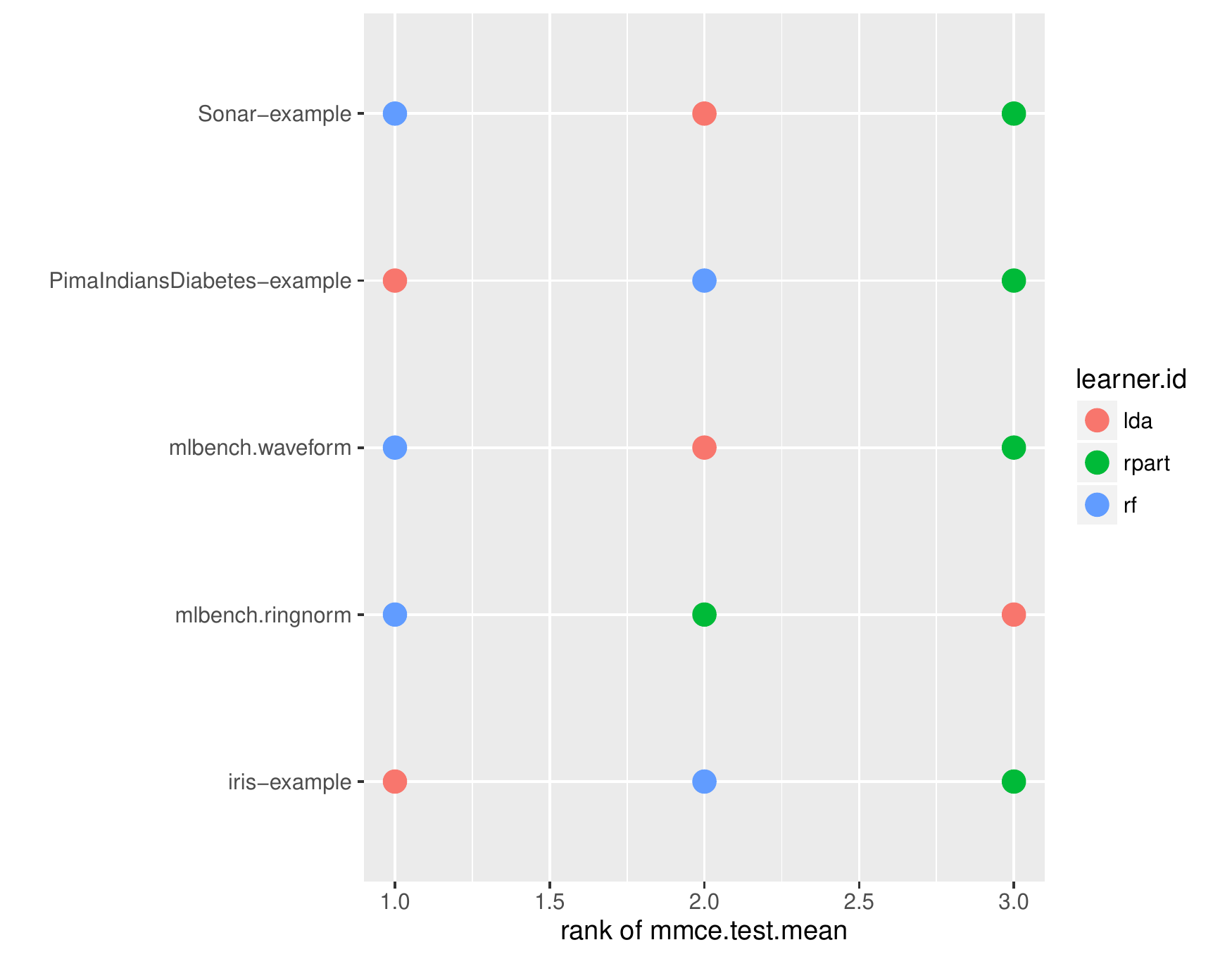}

Alternatively, you can draw stacked bar charts (the default) or bar
charts with juxtaposed bars (\lstinline!pos = "dodge"!) that are better
suited to compare the frequencies of learners within and across ranks.

\begin{lstlisting}[language=R]
plotBMRRanksAsBarChart(bmr)
plotBMRRanksAsBarChart(bmr, pos = "dodge")
\end{lstlisting}

\paragraph{Comparing learners using hypothesis
tests}\label{comparing-learners-using-hypothesis-tests}

Many researchers feel the need to display an algorithm's superiority by
employing some sort of hypothesis testing. As non-parametric tests seem
better suited for such benchmark results the tests provided in
\href{http://www.rdocumentation.org/packages/mlr/}{mlr} are the
\textbf{Overall Friedman test} and the \textbf{Friedman-Nemenyi post hoc
test}.

While the ad hoc
\href{http://www.rdocumentation.org/packages/mlr/functions/friedmanTestBMR.html}{Friedman
test} based on
\href{http://www.rdocumentation.org/packages/stats/functions/friedman.test.html}{friedman.test}
from the \href{http://www.rdocumentation.org/packages/stats/}{stats}
package is testing the hypothesis whether there is a significant
difference between the employed learners, the post hoc
\href{http://www.rdocumentation.org/packages/mlr/functions/friedmanPostHocTestBMR.html}{Friedman-Nemenyi
test} tests for significant differences between all pairs of learners.
\emph{Non parametric} tests often do have less power then their
\emph{parametric} counterparts but less assumptions about underlying
distributions have to be made. This often means many \textbf{data sets}
are needed in order to be able to show significant differences at
reasonable significance levels.

In our example, we want to compare the three learners on the selected
data sets. First we might we want to test the hypothesis whether there
is a difference between the learners.

\begin{lstlisting}[language=R]
friedmanTestBMR(bmr)
#> 
#>  Friedman rank sum test
#> 
#> data:  mmce.test.mean and learner.id and task.id
#> Friedman chi-squared = 5.2, df = 2, p-value = 0.07427
\end{lstlisting}

In order to keep the computation time for this tutorial small, the
\href{http://www.rdocumentation.org/packages/mlr/functions/makeLearner.html}{Learner}s
are only evaluated on five tasks. This also means that we operate on a
relatively low significance level \(\alpha = 0.1\). As we can reject the
null hypothesis of the Friedman test at a reasonable significance level
we might now want to test where these differences lie exactly.

\begin{lstlisting}[language=R]
friedmanPostHocTestBMR(bmr, p.value = 0.1)
#> 
#>  Pairwise comparisons using Nemenyi multiple comparison test 
#>              with q approximation for unreplicated blocked data 
#> 
#> data:  mmce.test.mean and learner.id and task.id 
#> 
#>              lda   rpart
#> rpart        0.254 -    
#> randomForest 0.802 0.069
#> 
#> P value adjustment method: none
\end{lstlisting}

At this level of significance, we can reject the null hypothesis that
there exists no performance difference between the decision tree
(\href{http://www.rdocumentation.org/packages/rpart/functions/rpart.html}{rpart})
and the
\href{http://www.rdocumentation.org/packages/randomForest/functions/randomForest.html}{random
Forest}.

\paragraph{Critical differences
diagram}\label{critical-differences-diagram}

In order to visualize differently performing learners, a
\href{http://www.rdocumentation.org/packages/mlr/functions/plotCritDifferences.html}{critical
differences diagram} can be plotted, using either the Nemenyi test
(\lstinline!test = "nemenyi"!) or the Bonferroni-Dunn test
(\lstinline!test = "bd"!).

The mean rank of learners is displayed on the x-axis.

\begin{itemize}
\tightlist
\item
  Choosing \lstinline!test = "nemenyi"! compares all pairs of
  \href{http://www.rdocumentation.org/packages/mlr/functions/makeLearner.html}{Learner}s
  to each other, thus the output are groups of not significantly
  different learners. The diagram connects all groups of learners where
  the mean ranks do not differ by more than the critical differences.
  Learners that are not connected by a bar are significantly different,
  and the learner(s) with the lower mean rank can be considered
  ``better'' at the chosen significance level.
\item
  Choosing \lstinline!test = "bd"! performs a \emph{pairwise comparison
  with a baseline}. An interval which extends by the given
  \emph{critical difference} in both directions is drawn around the
  \href{http://www.rdocumentation.org/packages/mlr/functions/makeLearner.html}{Learner}
  chosen as baseline, though only comparisons with the baseline are
  possible. All learners within the interval are not significantly
  different, while the baseline can be considered better or worse than a
  given learner which is outside of the interval.
\end{itemize}

The critical difference \(\mathit{CD}\) is calculated by
\[\mathit{CD} = q_\alpha \cdot \sqrt{\frac{k(k+1)}{6N}},\] where \(N\)
denotes the number of tasks, \(k\) is the number of learners, and
\(q_\alpha\) comes from the studentized range statistic divided by
\(\sqrt{2}\). For details see
\href{http://www.jmlr.org/papers/volume7/demsar06a/demsar06a.pdf}{Demsar
(2006)}.

Function
\href{http://www.rdocumentation.org/packages/mlr/functions/generateCritDifferencesData.html}{generateCritDifferencesData}
does all necessary calculations while function
\href{http://www.rdocumentation.org/packages/mlr/functions/plotCritDifferences.html}{plotCritDifferences}
draws the plot. See the tutorial page about
\protect\hyperlink{visualization}{visualization} for details on data
generation and plotting functions.

\begin{lstlisting}[language=R]
### Nemenyi test
g = generateCritDifferencesData(bmr, p.value = 0.1, test = "nemenyi")
plotCritDifferences(g) + coord_cartesian(xlim = c(-1,5), ylim = c(0,2))
\end{lstlisting}

\includegraphics{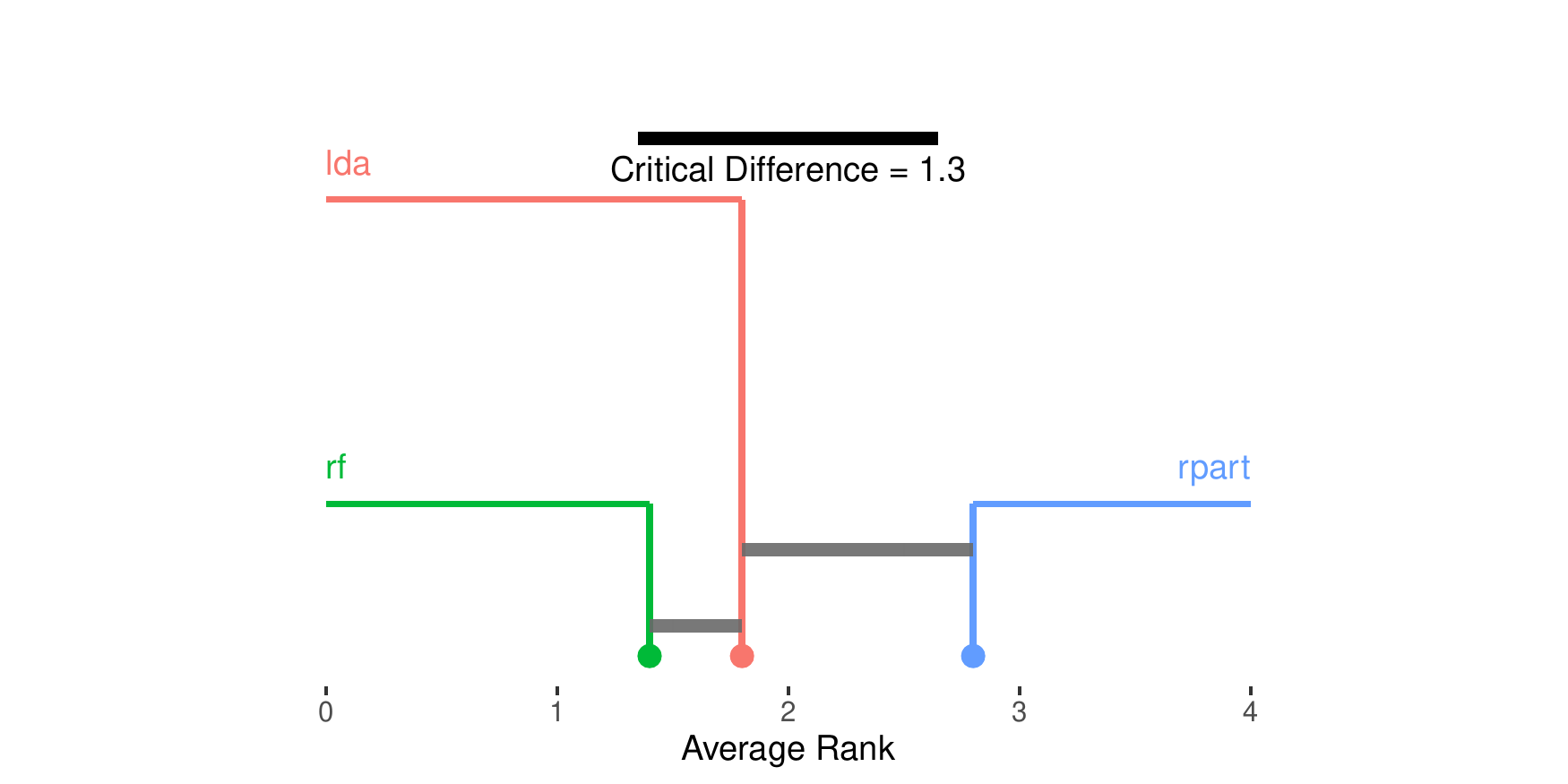}

\begin{lstlisting}[language=R]
### Bonferroni-Dunn test
g = generateCritDifferencesData(bmr, p.value = 0.1, test = "bd", baseline = "randomForest")
plotCritDifferences(g) + coord_cartesian(xlim = c(-1,5), ylim = c(0,2))
\end{lstlisting}

\includegraphics{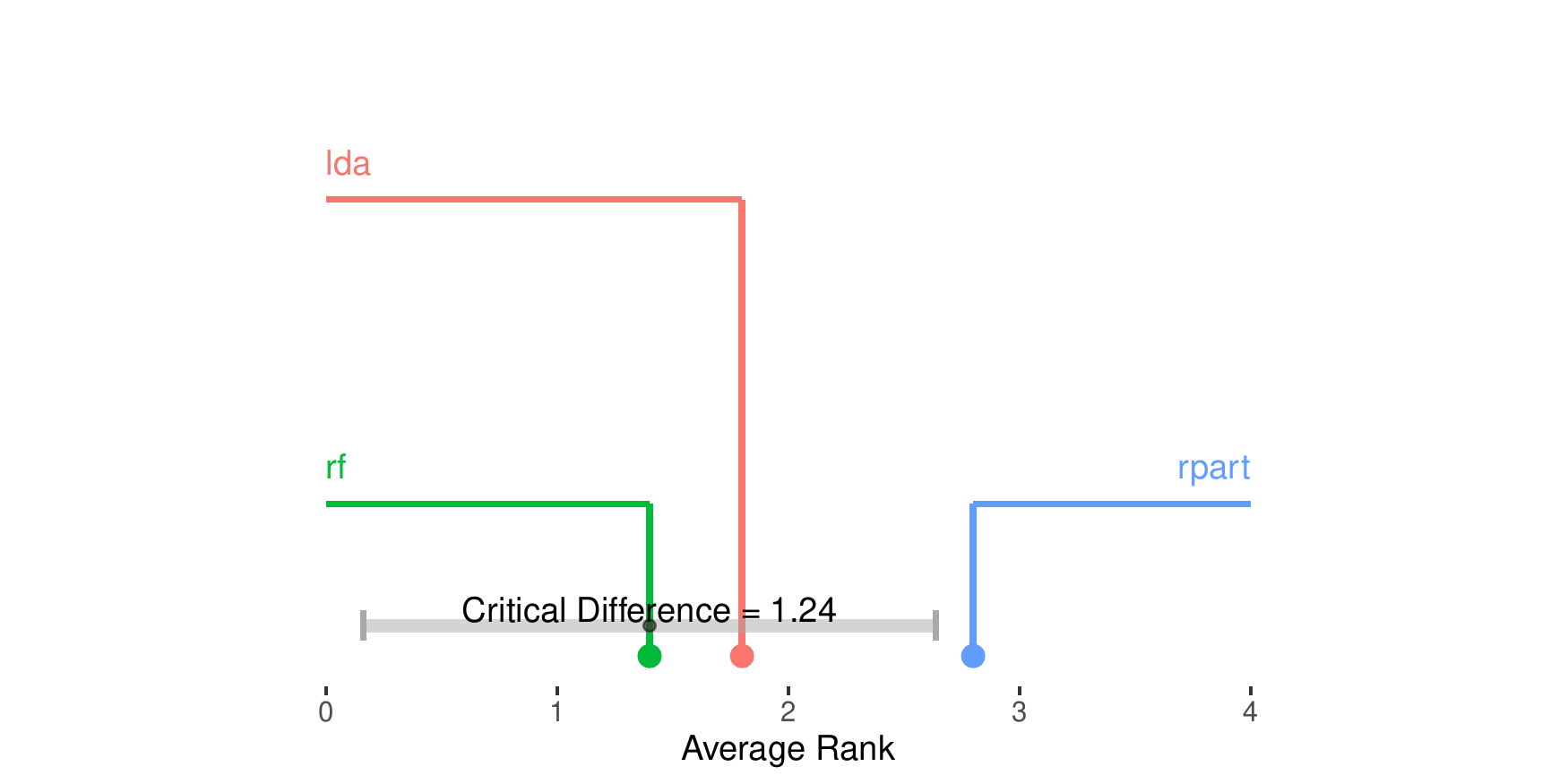}

\paragraph{Custom plots}\label{custom-plots}

You can easily generate your own visualizations by customizing the
\href{http://www.rdocumentation.org/packages/ggplot2/functions/ggplot.html}{ggplot}
objects returned by the plots above, retrieve the data from the
\href{http://www.rdocumentation.org/packages/ggplot2/functions/ggplot.html}{ggplot}
objects and use them as basis for your own plots, or rely on the
\href{http://www.rdocumentation.org/packages/base/functions/data.frame.html}{data.frame}s
returned by
\href{http://www.rdocumentation.org/packages/mlr/functions/getBMRPerformances.html}{getBMRPerformances}
or
\href{http://www.rdocumentation.org/packages/mlr/functions/getBMRAggrPerformances.html}{getBMRAggrPerformances}.
Here are some examples.

Instead of boxplots (as in
\href{http://www.rdocumentation.org/packages/mlr/functions/plotBMRBoxplots.html}{plotBMRBoxplots})
we could create density plots to show the performance values resulting
from individual resampling iterations.

\begin{lstlisting}[language=R]
perf = getBMRPerformances(bmr, as.df = TRUE)

### Density plots for two tasks
qplot(mmce, colour = learner.id, facets = . ~ task.id,
  data = perf[perf$task.id %in% c("iris-example", "Sonar-example"),], geom = "density") +
  theme(strip.text.x = element_text(size = 8))
\end{lstlisting}

\includegraphics{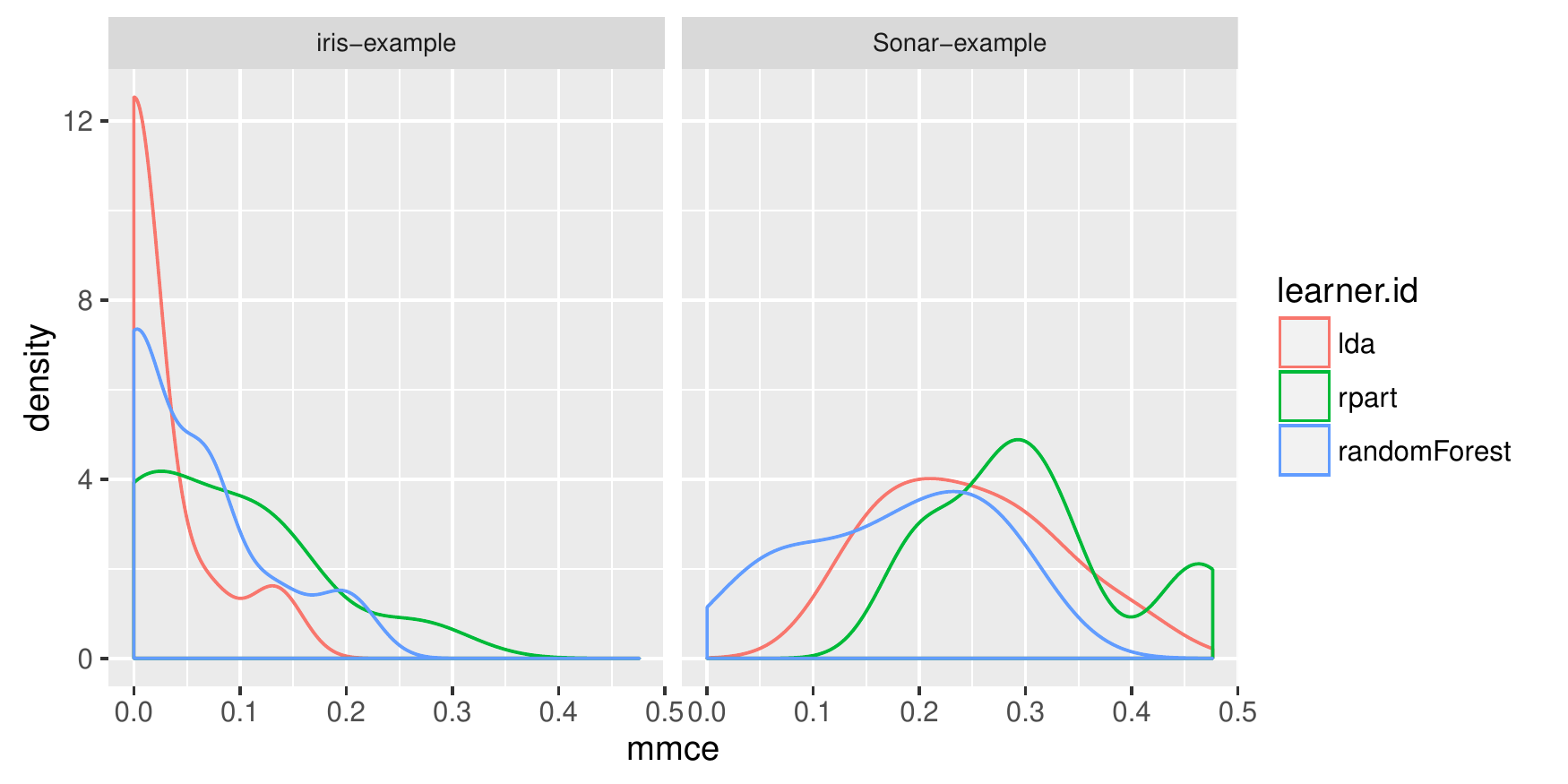}

In order to plot multiple performance measures in parallel,
\lstinline!perf! is reshaped to long format. Below we generate grouped
boxplots showing the error rate
(\protect\hyperlink{implemented-performance-measures}{mmce}) and the
training time
\protect\hyperlink{implemented-performance-measures}{timetrain}.

\begin{lstlisting}[language=R]
### Compare mmce and timetrain
df = reshape2::melt(perf, id.vars = c("task.id", "learner.id", "iter"))
df = df[df$variable != "ber",]
head(df)
#>        task.id learner.id iter variable     value
#> 1 iris-example        lda    1     mmce 0.0000000
#> 2 iris-example        lda    2     mmce 0.1333333
#> 3 iris-example        lda    3     mmce 0.0000000
#> 4 iris-example        lda    4     mmce 0.0000000
#> 5 iris-example        lda    5     mmce 0.0000000
#> 6 iris-example        lda    6     mmce 0.0000000

qplot(variable, value, data = df, colour = learner.id, geom = "boxplot",
  xlab = "measure", ylab = "performance") +
  facet_wrap(~ task.id, nrow = 2)
\end{lstlisting}

\includegraphics{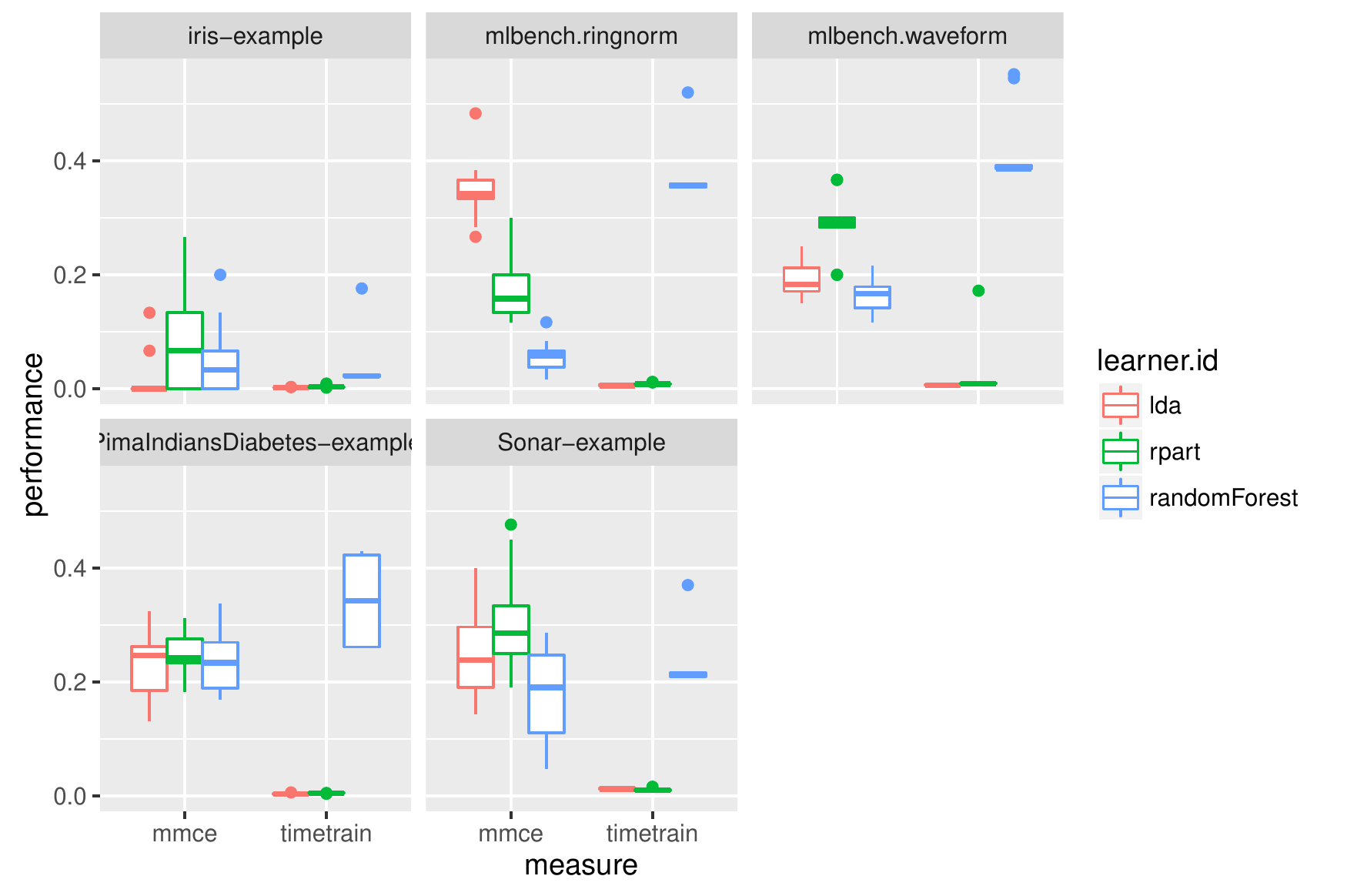}

It might also be useful to assess if learner performances in single
resampling iterations, i.e., in one fold, are related. This might help
to gain further insight, for example by having a closer look at train
and test sets from iterations where one learner performs exceptionally
well while another one is fairly bad. Moreover, this might be useful for
the construction of ensembles of learning algorithms. Below, function
\href{http://www.rdocumentation.org/packages/GGally/functions/ggpairs.html}{ggpairs}
from package
\href{http://www.rdocumentation.org/packages/GGally/}{GGally} is used to
generate a scatterplot matrix of mean misclassification errors
(\href{http://www.rdocumentation.org/packages/mlr/functions/measures.md.html}{mmce})
on the
\href{http://www.rdocumentation.org/packages/mlbench/functions/Sonar.html}{Sonar}
data set.

\begin{lstlisting}[language=R]
perf = getBMRPerformances(bmr, task.id = "Sonar-example", as.df = TRUE)
df = reshape2::melt(perf, id.vars = c("task.id", "learner.id", "iter"))
df = df[df$variable == "mmce",]
df = reshape2::dcast(df, task.id + iter ~ variable + learner.id)
head(df)
#>         task.id iter  mmce_lda mmce_rpart mmce_randomForest
#> 1 Sonar-example    1 0.2857143  0.2857143        0.14285714
#> 2 Sonar-example    2 0.2380952  0.2380952        0.23809524
#> 3 Sonar-example    3 0.3333333  0.2857143        0.28571429
#> 4 Sonar-example    4 0.2380952  0.3333333        0.04761905
#> 5 Sonar-example    5 0.1428571  0.2857143        0.19047619
#> 6 Sonar-example    6 0.4000000  0.4500000        0.25000000

GGally::ggpairs(df, 3:5)
\end{lstlisting}

\includegraphics{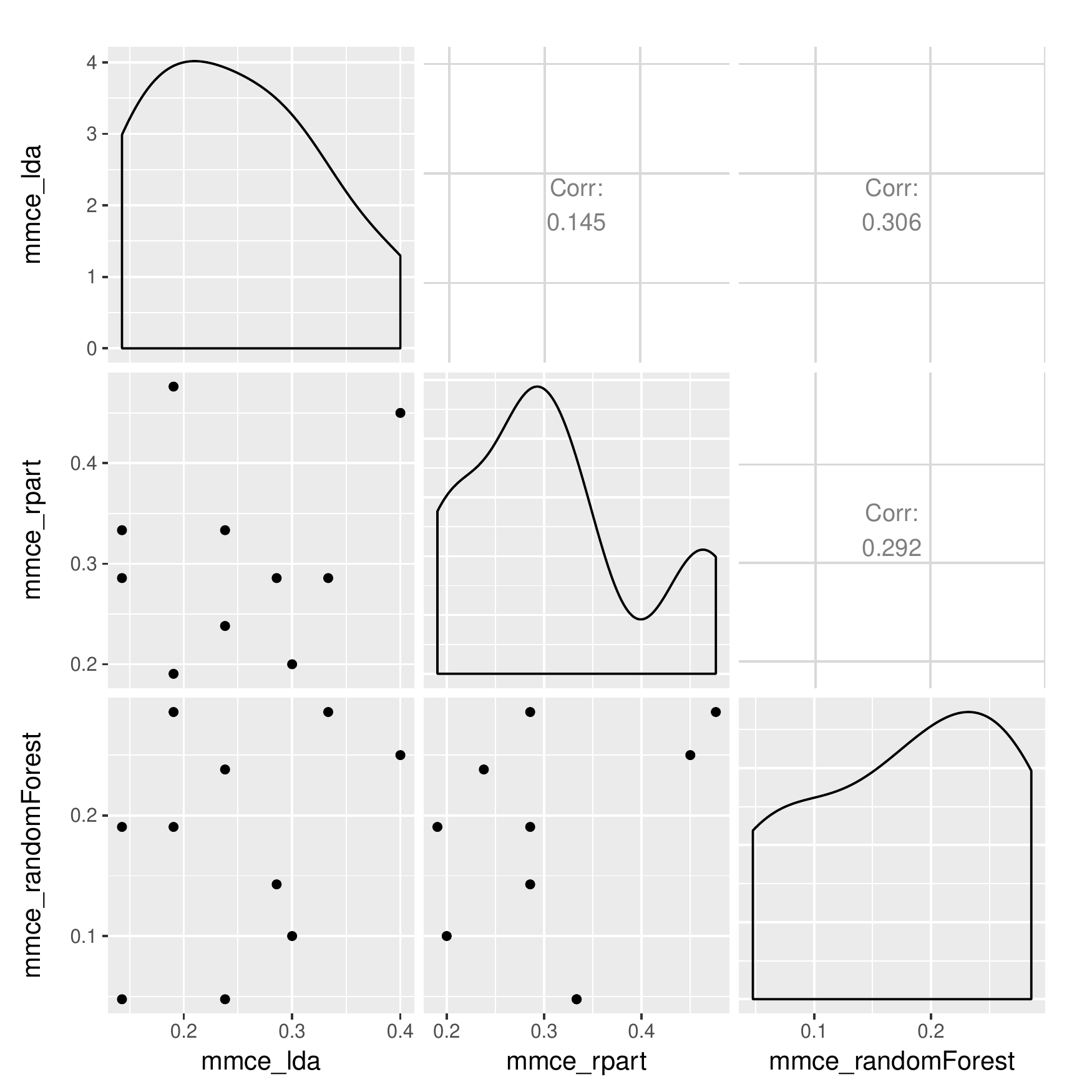}

\subsubsection{Further comments}\label{further-comments-1}

\begin{itemize}
\tightlist
\item
  Note that for supervised classification
  \href{http://www.rdocumentation.org/packages/mlr/}{mlr} offers some
  more plots that operate on
  \href{http://www.rdocumentation.org/packages/mlr/functions/BenchmarkResult.html}{BenchmarkResult}
  objects and allow you to compare the performance of learning
  algorithms. See for example the tutorial page on
  \protect\hyperlink{roc-analysis-and-performance-curves}{ROC curves}
  and functions
  \href{http://www.rdocumentation.org/packages/mlr/functions/generateThreshVsPerfData.html}{generateThreshVsPerfData},
  \href{http://www.rdocumentation.org/packages/mlr/functions/plotROCCurves.html}{plotROCCurves},
  and
  \href{http://www.rdocumentation.org/packages/mlr/functions/plotViperCharts.html}{plotViperCharts}
  as well as the page about
  \protect\hyperlink{classifier-calibration}{classifier calibration} and
  function
  \href{http://www.rdocumentation.org/packages/mlr/functions/generateCalibrationData.html}{generateCalibrationData}.
\item
  In the examples shown in this section we applied ``raw'' learning
  algorithms, but often things are more complicated. At the very least,
  many learners have hyperparameters that need to be tuned to get
  sensible results. Reliable performance estimates can be obtained by
  \protect\hyperlink{nested-resampling}{nested resampling}, i.e., by
  doing the tuning in an inner resampling loop while estimating the
  performance in an outer loop. Moreover, you might want to combine
  learners with pre-processing steps like imputation, scaling, outlier
  removal, dimensionality reduction or feature selection and so on. All
  this can be easily done using
  \href{http://www.rdocumentation.org/packages/mlr/}{mlr}'s wrapper
  functionality. The general principle is explained in the section about
  \href{http://www.rdocumentation.org/packages/mlr/functions/wrapper.md.html}{wrapped
  learners} in the Advanced part of this tutorial. There are also
  several sections devoted to common pre-processing steps.
\item
  Benchmark experiments can very quickly become computationally
  demanding. \href{http://www.rdocumentation.org/packages/mlr/}{mlr}
  offers some possibilities for
  \protect\hyperlink{parallelization}{parallelization}.
\end{itemize}

\hypertarget{parallelization}{\subsection{Parallelization}\label{parallelization}}

\textbf{R} by default does not make use of parallelization. With the
integration of
\href{http://www.rdocumentation.org/packages/parallelMap/}{parallelMap}
into \href{http://www.rdocumentation.org/packages/mlr/}{mlr}, it becomes
easy to activate the parallel computing capabilities already supported
by \href{http://www.rdocumentation.org/packages/mlr/}{mlr}.
\href{http://www.rdocumentation.org/packages/parallelMap/}{parallelMap}
supports all major parallelization backends: local multicore execution
using \href{http://www.rdocumentation.org/packages/parallel/}{parallel},
socket and MPI clusters using
\href{http://www.rdocumentation.org/packages/snow/}{snow}, makeshift
SSH-clusters using
\href{http://www.rdocumentation.org/packages/BatchJobs/}{BatchJobs} and
high performance computing clusters (managed by a scheduler like SLURM,
Torque/PBS, SGE or LSF) also using
\href{http://www.rdocumentation.org/packages/BatchJobs/}{BatchJobs}.

All you have to do is select a backend by calling one of the
\href{http://www.rdocumentation.org/packages/parallelMap/functions/parallelStart.html}{parallelStart*}
functions. The first loop
\href{http://www.rdocumentation.org/packages/mlr/}{mlr} encounters which
is marked as parallel executable will be automatically parallelized. It
is good practice to call
\href{http://www.rdocumentation.org/packages/parallelMap/functions/parallelStop.html}{parallelStop}
at the end of your script.

\begin{lstlisting}[language=R]
library("parallelMap")
parallelStartSocket(2)
#> Starting parallelization in mode=socket with cpus=2.

rdesc = makeResampleDesc("CV", iters = 3)
r = resample("classif.lda", iris.task, rdesc)
#> Exporting objects to slaves for mode socket: .mlr.slave.options
#> Mapping in parallel: mode = socket; cpus = 2; elements = 3.
#> [Resample] Result: mmce.test.mean=0.02

parallelStop()
#> Stopped parallelization. All cleaned up.
\end{lstlisting}

On Linux or Mac OS X, you may want to use
\href{http://www.rdocumentation.org/packages/parallelMap/functions/parallelStart.html}{parallelStartMulticore}
instead.

\subsubsection{Parallelization levels}\label{parallelization-levels}

We offer different parallelization levels for fine grained control over
the parallelization. E.g., if you do not want to parallelize the
\href{http://www.rdocumentation.org/packages/mlr/functions/benchmark.html}{benchmark}
function because it has only very few iterations but want to parallelize
the
\href{http://www.rdocumentation.org/packages/mlr/functions/resample.html}{resampling}
of each learner instead, you can specifically pass the \lstinline!level!
\lstinline!"mlr.resample"! to the
\href{http://www.rdocumentation.org/packages/parallelMap/functions/parallelStart.html}{parallelStart*}
function. Currently the following levels are supported:

\begin{lstlisting}[language=R]
parallelGetRegisteredLevels()
#> mlr: mlr.benchmark, mlr.resample, mlr.selectFeatures, mlr.tuneParams
\end{lstlisting}

Here is a brief explanation of what these levels do:

\begin{itemize}
\tightlist
\item
  \lstinline!"mlr.resample"!: Each resampling iteration (a train / test
  step) is a parallel job.
\item
  \lstinline!"mlr.benchmark"!: Each experiment ``run this learner on
  this data set'' is a parallel job.
\item
  \lstinline!"mlr.tuneParams"!: Each evaluation in hyperparameter space
  ``resample with these parameter settings'' is a parallel job. How many
  of these can be run independently in parallel, depends on the tuning
  algorithm. For grid search or random search this is no problem, but
  for other tuners it depends on how many points are produced in each
  iteration of the optimization. If a tuner works in a purely sequential
  fashion, we cannot work magic and the hyperparameter evaluation will
  also run sequentially. But note that you can still parallelize the
  underlying resampling.
\item
  \lstinline!"mlr.selectFeatures"!: Each evaluation in feature space
  ``resample with this feature subset'' is a parallel job. The same
  comments as for \lstinline!"mlr.tuneParams"! apply here.
\end{itemize}

\subsubsection{Custom learners and
parallelization}\label{custom-learners-and-parallelization}

If you have \protect\hyperlink{integrating-another-learner}{implemented
a custom learner yourself}, locally, you currently need to export this
to the slave. So if you see an error after calling, e.g., a parallelized
version of
\href{http://www.rdocumentation.org/packages/mlr/functions/resample.html}{resample}
like this:

\begin{lstlisting}
no applicable method for 'trainLearner' applied to an object of class <my_new_learner>
\end{lstlisting}

simply add the following line somewhere after calling
\href{http://www.rdocumentation.org/packages/parallelMap/functions/parallelStart.html}{parallelStart}.

\begin{lstlisting}
parallelExport("trainLearner.<my_new_learner>", "predictLearner.<my_new_learner>")
\end{lstlisting}

\subsubsection{The end}\label{the-end}

For further details, consult the
\href{https://github.com/berndbischl/parallelMap\#parallelmap}{parallelMap
tutorial} and
\href{http://www.rdocumentation.org/packages/parallelMap/}{help}.

\hypertarget{visualization}{\subsection{Visualization}\label{visualization}}

\subsubsection{Generation and plotting
functions}\label{generation-and-plotting-functions}

\href{http://www.rdocumentation.org/packages/mlr/}{mlr}'s visualization
capabilities rely on \emph{generation functions} which generate data for
plots, and \emph{plotting functions} which plot this output using either
\href{http://www.rdocumentation.org/packages/ggplot2/}{ggplot2} or
\href{http://www.rdocumentation.org/packages/ggvis/}{ggvis} (the latter
being currently experimental).

This separation allows users to easily make custom visualizations by
taking advantage of the generation functions. The only data
transformation that is handled inside plotting functions is reshaping.
The reshaped data is also accessible by calling the plotting functions
and then extracting the data from the
\href{http://www.rdocumentation.org/packages/ggplot2/functions/ggplot.html}{ggplot}
object.

The functions are named accordingly.

\begin{itemize}
\tightlist
\item
  Names of generation functions start with \lstinline!generate! and are
  followed by a title-case description of their
  \lstinline!FunctionPurpose!, followed by \lstinline!Data!, i.e.,
  \lstinline!generateFunctionPurposeData!. These functions output
  objects of class \lstinline!FunctionPurposeData!.
\item
  Plotting functions are prefixed by \lstinline!plot! followed by their
  purpose, i.e., \lstinline!plotFunctionPurpose!.
\item
  \href{http://www.rdocumentation.org/packages/ggvis/}{ggvis} plotting
  functions have an additional suffix \lstinline!GGVIS!, i.e.,
  \lstinline!plotFunctionPurposeGGVIS!.
\end{itemize}

\paragraph{Some examples}\label{some-examples}

In the example below we create a plot of classifier performance as
function of the decision threshold for the binary classification problem
\href{http://www.rdocumentation.org/packages/mlr/functions/sonar.task.html}{sonar.task}.
The generation function
\href{http://www.rdocumentation.org/packages/mlr/functions/generateThreshVsPerfData.html}{generateThreshVsPerfData}
creates an object of class
\href{http://www.rdocumentation.org/packages/mlr/functions/generateThreshVsPerfData.html}{ThreshVsPerfData}
which contains the data for the plot in slot \lstinline!$data!.

\begin{lstlisting}[language=R]
lrn = makeLearner("classif.lda", predict.type = "prob")
n = getTaskSize(sonar.task)
mod = train(lrn, task = sonar.task, subset = seq(1, n, by = 2))
pred = predict(mod, task = sonar.task, subset = seq(2, n, by = 2))
d = generateThreshVsPerfData(pred, measures = list(fpr, fnr, mmce))

class(d)
#> [1] "ThreshVsPerfData"

head(d$data)
#>         fpr       fnr      mmce  threshold
#> 1 1.0000000 0.0000000 0.4615385 0.00000000
#> 2 0.3541667 0.1964286 0.2692308 0.01010101
#> 3 0.3333333 0.2321429 0.2788462 0.02020202
#> 4 0.3333333 0.2321429 0.2788462 0.03030303
#> 5 0.3333333 0.2321429 0.2788462 0.04040404
#> 6 0.3125000 0.2321429 0.2692308 0.05050505
\end{lstlisting}

For plotting we can use the built-in
\href{http://www.rdocumentation.org/packages/mlr/}{mlr} function
\href{http://www.rdocumentation.org/packages/mlr/functions/plotThreshVsPerf.html}{plotThreshVsPerf}.

\begin{lstlisting}[language=R]
plotThreshVsPerf(d)
\end{lstlisting}

\includegraphics{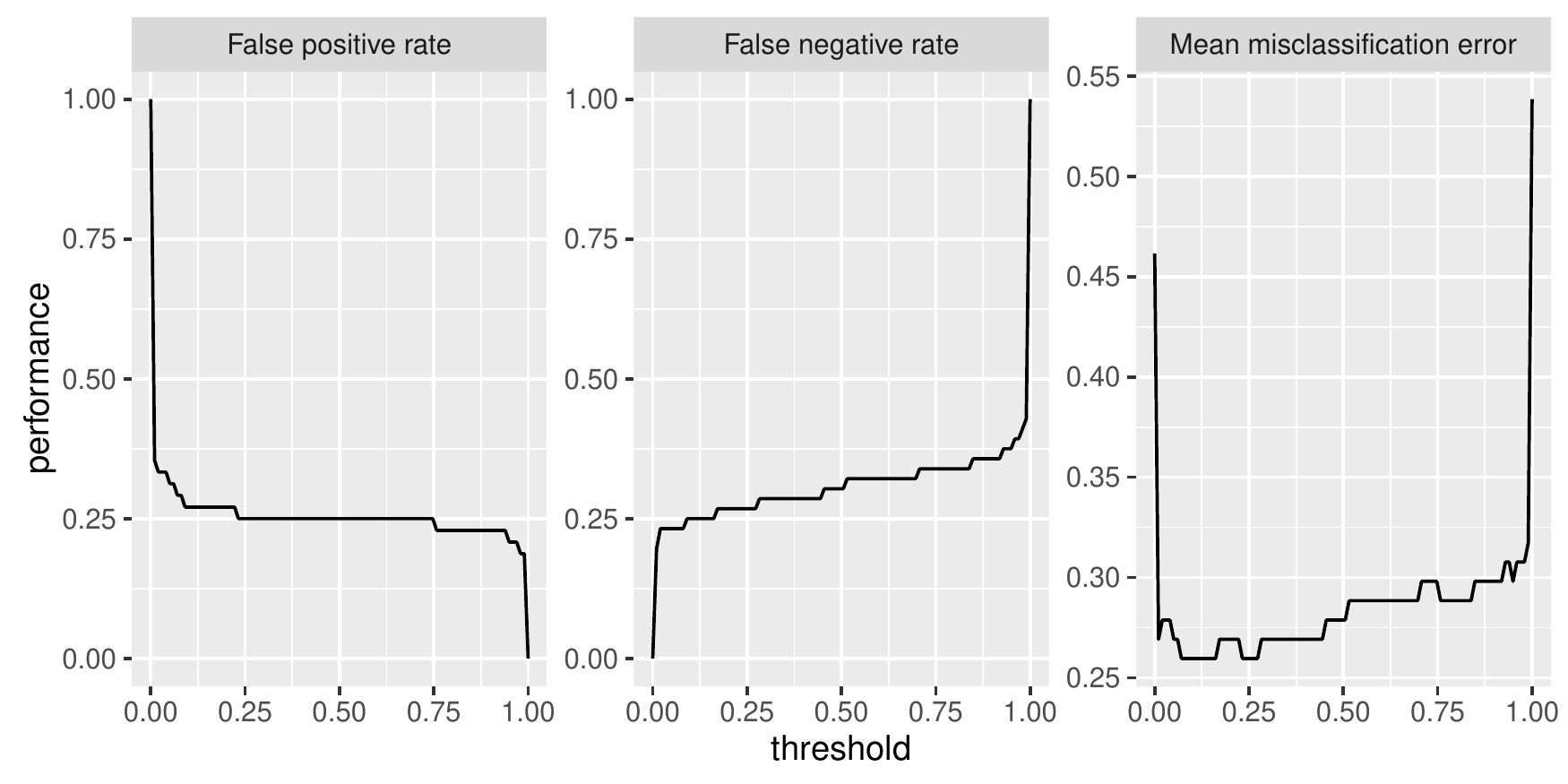}

Note that by default the
\href{http://www.rdocumentation.org/packages/mlr/functions/makeMeasure.html}{Measure}
\lstinline!name!s are used to annotate the panels.

\begin{lstlisting}[language=R]
fpr$name
#> [1] "False positive rate"

fpr$id
#> [1] "fpr"
\end{lstlisting}

This does not only apply to
\href{http://www.rdocumentation.org/packages/mlr/functions/plotThreshVsPerf.html}{plotThreshVsPerf},
but to other plot functions that show performance measures as well, for
example
\href{http://www.rdocumentation.org/packages/mlr/functions/plotLearningCurve.html}{plotLearningCurve}.
You can use the \lstinline!id!s instead of the names by setting
\lstinline!pretty.names = FALSE!.

\paragraph{Customizing plots}\label{customizing-plots}

As mentioned above it is easily possible to customize the built-in plots
or making your own visualizations from scratch based on the generated
data.

What will probably come up most often is changing labels and
annotations. Generally, this can be done by manipulating the
\href{http://www.rdocumentation.org/packages/ggplot2/functions/ggplot.html}{ggplot}
object, in this example the object returned by
\href{http://www.rdocumentation.org/packages/mlr/functions/plotThreshVsPerf.html}{plotThreshVsPerf},
using the usual
\href{http://www.rdocumentation.org/packages/ggplot2/}{ggplot2}
functions like
\href{http://www.rdocumentation.org/packages/ggplot2/functions/labs.html}{ylab}
or
\href{http://www.rdocumentation.org/packages/ggplot2/functions/labeller.html}{labeller}.
Moreover, you can change the underlying data, either \lstinline!d$data!
(resulting from
\href{http://www.rdocumentation.org/packages/mlr/functions/generateThreshVsPerfData.html}{generateThreshVsPerfData})
or the possibly reshaped data contained in the
\href{http://www.rdocumentation.org/packages/ggplot2/functions/ggplot.html}{ggplot}
object (resulting from
\href{http://www.rdocumentation.org/packages/mlr/functions/plotThreshVsPerf.html}{plotThreshVsPerf}),
most often by renaming columns or factor levels.

Below are two examples of how to alter the axis and panel labels of the
above plot.

Imagine you want to change the order of the panels and also are not
satisfied with the panel names, for example you find that ``Mean
misclassification error'' is too long and you prefer ``Error rate''
instead. Moreover, you want the error rate to be displayed first.

\begin{lstlisting}[language=R]
plt = plotThreshVsPerf(d, pretty.names = FALSE)

### Reshaped version of the underlying data d
head(plt$data)
#>    threshold measure performance
#> 1 0.00000000     fpr   1.0000000
#> 2 0.01010101     fpr   0.3541667
#> 3 0.02020202     fpr   0.3333333
#> 4 0.03030303     fpr   0.3333333
#> 5 0.04040404     fpr   0.3333333
#> 6 0.05050505     fpr   0.3125000

levels(plt$data$measure)
#> [1] "fpr"  "fnr"  "mmce"

### Rename and reorder factor levels
plt$data$measure = factor(plt$data$measure, levels = c("mmce", "fpr", "fnr"),
  labels = c("Error rate", "False positive rate", "False negative rate"))
plt = plt + xlab("Cutoff") + ylab("Performance")
plt
\end{lstlisting}

\includegraphics{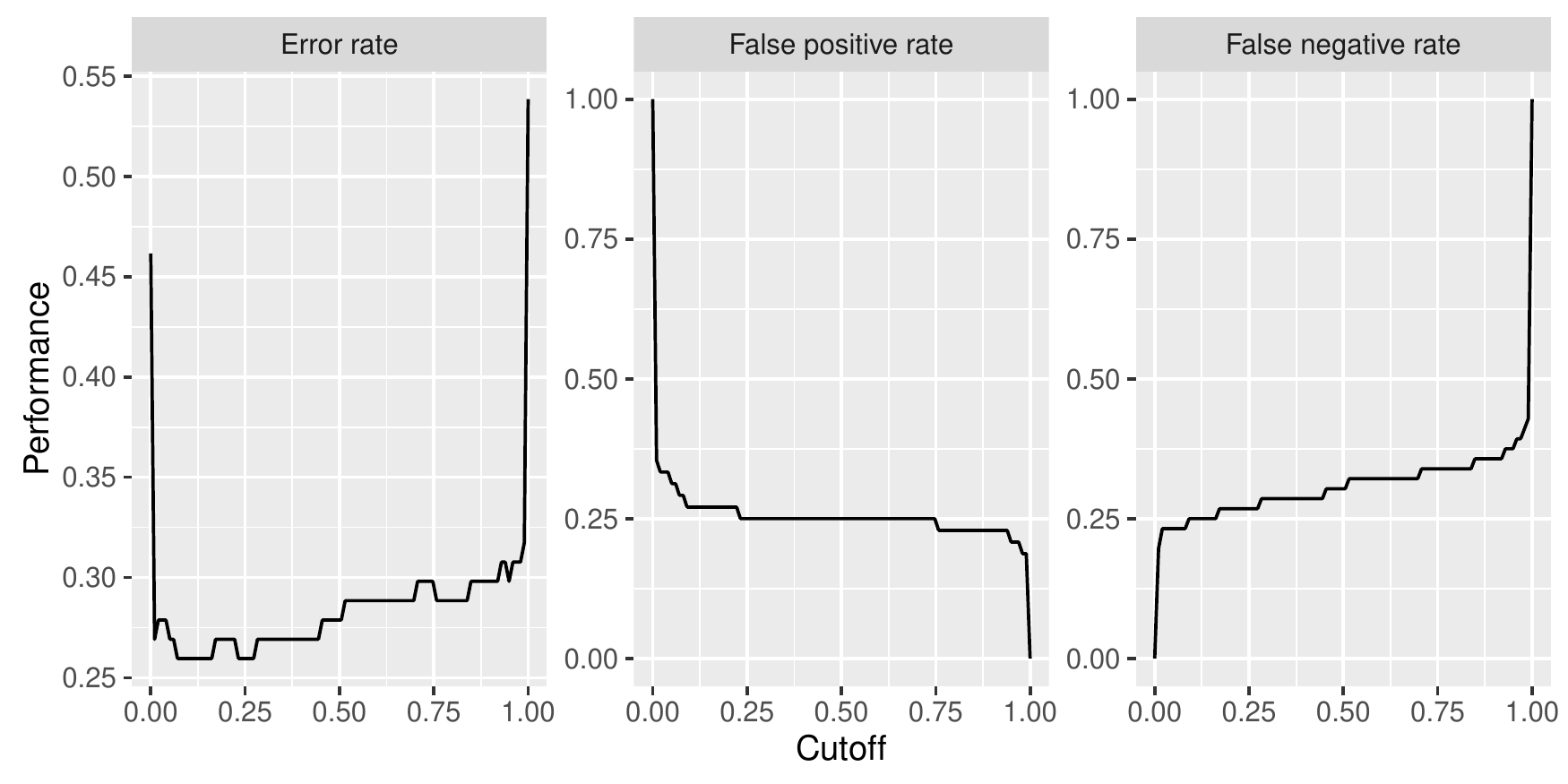}

Using the \href{ggplot2::labeller}{labeller} function requires calling
\href{http://www.rdocumentation.org/packages/ggplot2/functions/facet_wrap.html}{facet\_wrap}
(or
\href{http://www.rdocumentation.org/packages/ggplot2/functions/facet_grid.html}{facet\_grid}),
which can be useful if you want to change how the panels are positioned
(number of rows and columns) or influence the axis limits.

\begin{lstlisting}[language=R]
plt = plotThreshVsPerf(d, pretty.names = FALSE)

measure_names = c(
  fpr = "False positive rate",
  fnr = "False negative rate",
  mmce = "Error rate"
)
### Manipulate the measure names via the labeller function and
### arrange the panels in two columns and choose common axis limits for all panels
plt = plt + facet_wrap( ~ measure, labeller = labeller(measure = measure_names), ncol = 2)
plt = plt + xlab("Decision threshold") + ylab("Performance")
plt
\end{lstlisting}

\includegraphics{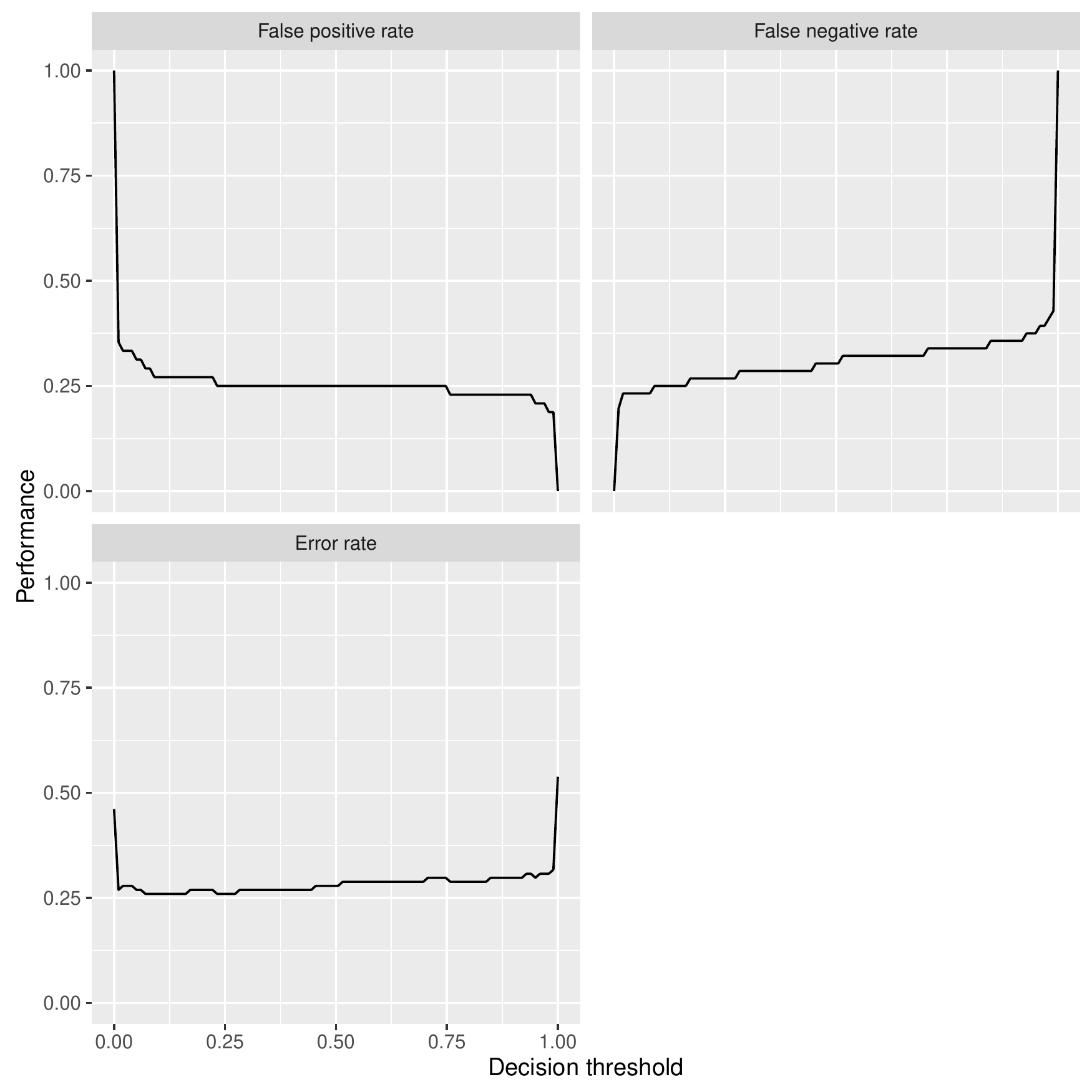}

Instead of using the built-in function
\href{http://www.rdocumentation.org/packages/mlr/functions/plotThreshVsPerf.html}{plotThreshVsPerf}
we could also manually create the plot based on the output of
\href{http://www.rdocumentation.org/packages/mlr/functions/generateThreshVsPerfData.html}{generateThreshVsPerfData}:
in this case to plot only one measure.

\begin{lstlisting}[language=R]
ggplot(d$data, aes(threshold, fpr)) + geom_line()
\end{lstlisting}

\includegraphics{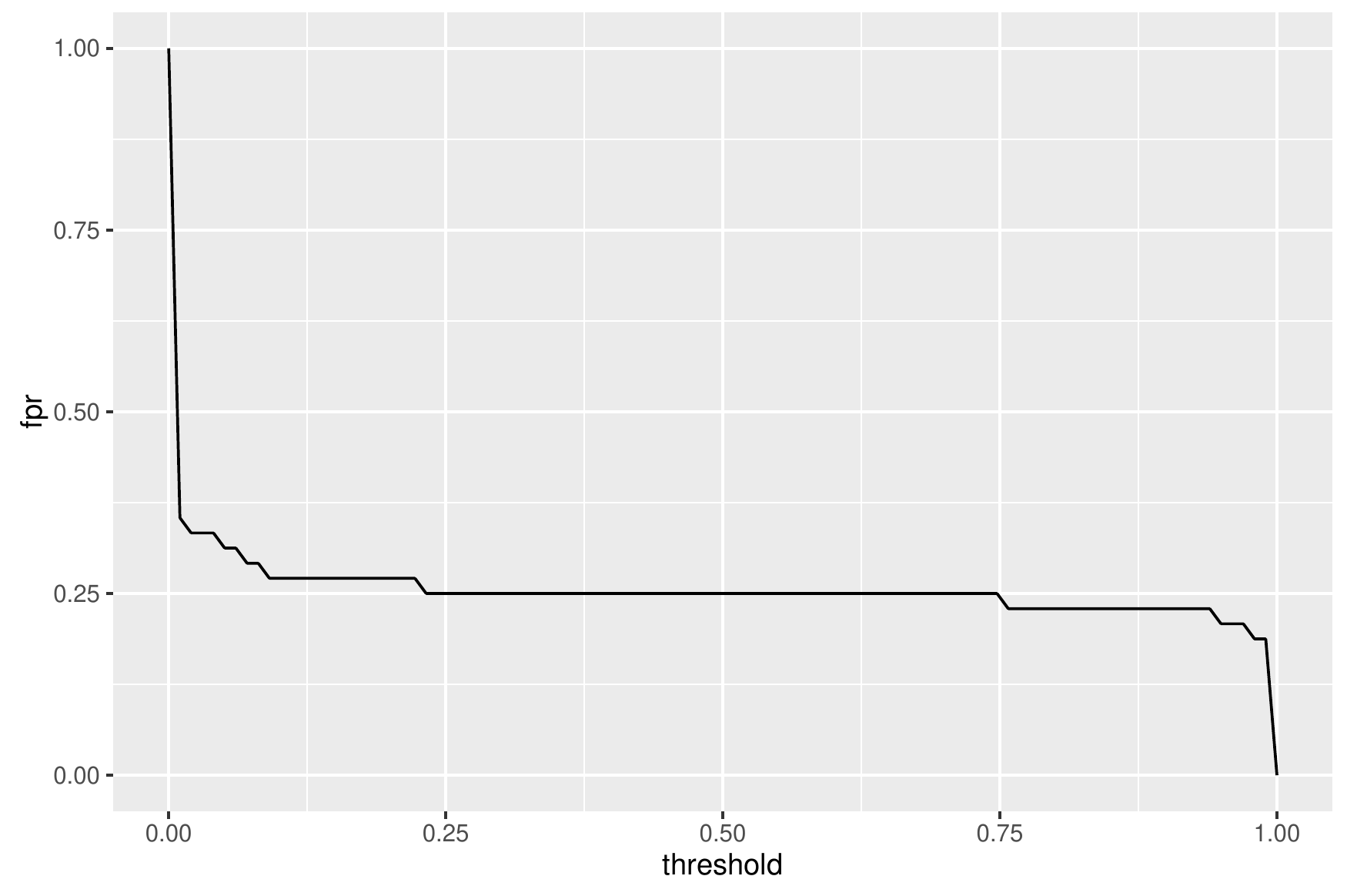}

The decoupling of generation and plotting functions is especially
practical if you prefer traditional
\href{http://www.rdocumentation.org/packages/graphics/}{graphics} or
\href{http://www.rdocumentation.org/packages/lattice/}{lattice}. Here is
a \href{http://www.rdocumentation.org/packages/lattice/}{lattice} plot
which gives a result similar to that of
\href{http://www.rdocumentation.org/packages/mlr/functions/plotThreshVsPerf.html}{plotThreshVsPerf}.

\begin{lstlisting}[language=R]
lattice::xyplot(fpr + fnr + mmce ~ threshold, data = d$data, type = "l", ylab = "performance",
  outer = TRUE, scales = list(relation = "free"),
  strip = strip.custom(factor.levels = sapply(d$measures, function(x) x$name),
    par.strip.text = list(cex = 0.8)))
\end{lstlisting}

\includegraphics{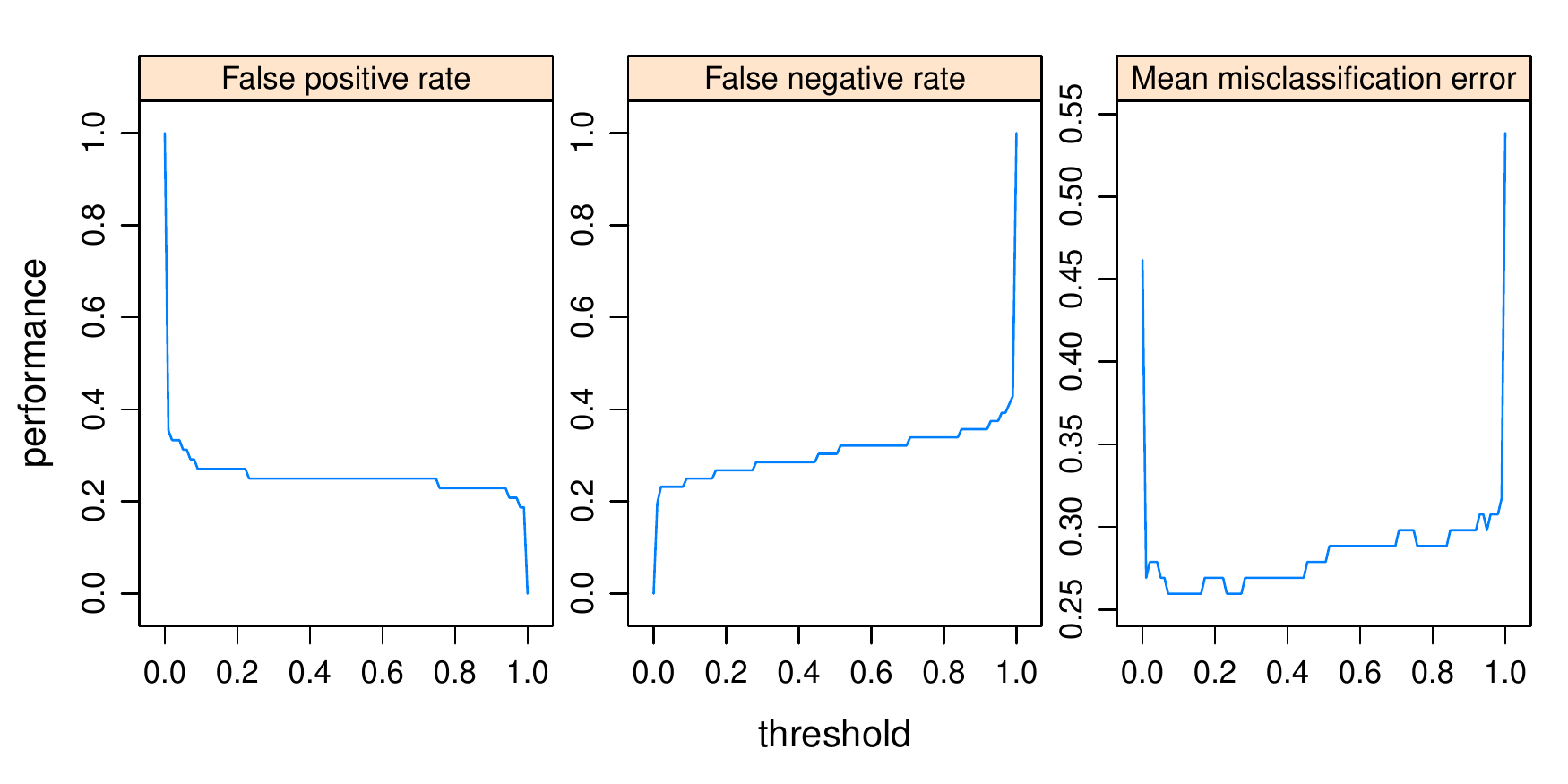}

Let's conclude with a brief look on a second example. Here we use
\href{http://www.rdocumentation.org/packages/mlr/functions/plotPartialDependence.html}{plotPartialDependence}
but extract the data from the
\href{http://www.rdocumentation.org/packages/ggplot2/functions/ggplot.html}{ggplot}
object \lstinline!plt!and use it to create a traditional
\href{http://www.rdocumentation.org/packages/graphics/functions/plot.html}{graphics::plot},
additional to the
\href{http://www.rdocumentation.org/packages/ggplot2/}{ggplot2} plot.

\begin{lstlisting}[language=R]
sonar = getTaskData(sonar.task)
pd = generatePartialDependenceData(mod, sonar, "V11")
plt = plotPartialDependence(pd)
head(plt$data)
#>   Class Probability Feature     Value
#> 1     M   0.2737158     V11 0.0289000
#> 2     M   0.3689970     V11 0.1072667
#> 3     M   0.4765742     V11 0.1856333
#> 4     M   0.5741233     V11 0.2640000
#> 5     M   0.6557857     V11 0.3423667
#> 6     M   0.7387962     V11 0.4207333

plt
\end{lstlisting}

\includegraphics{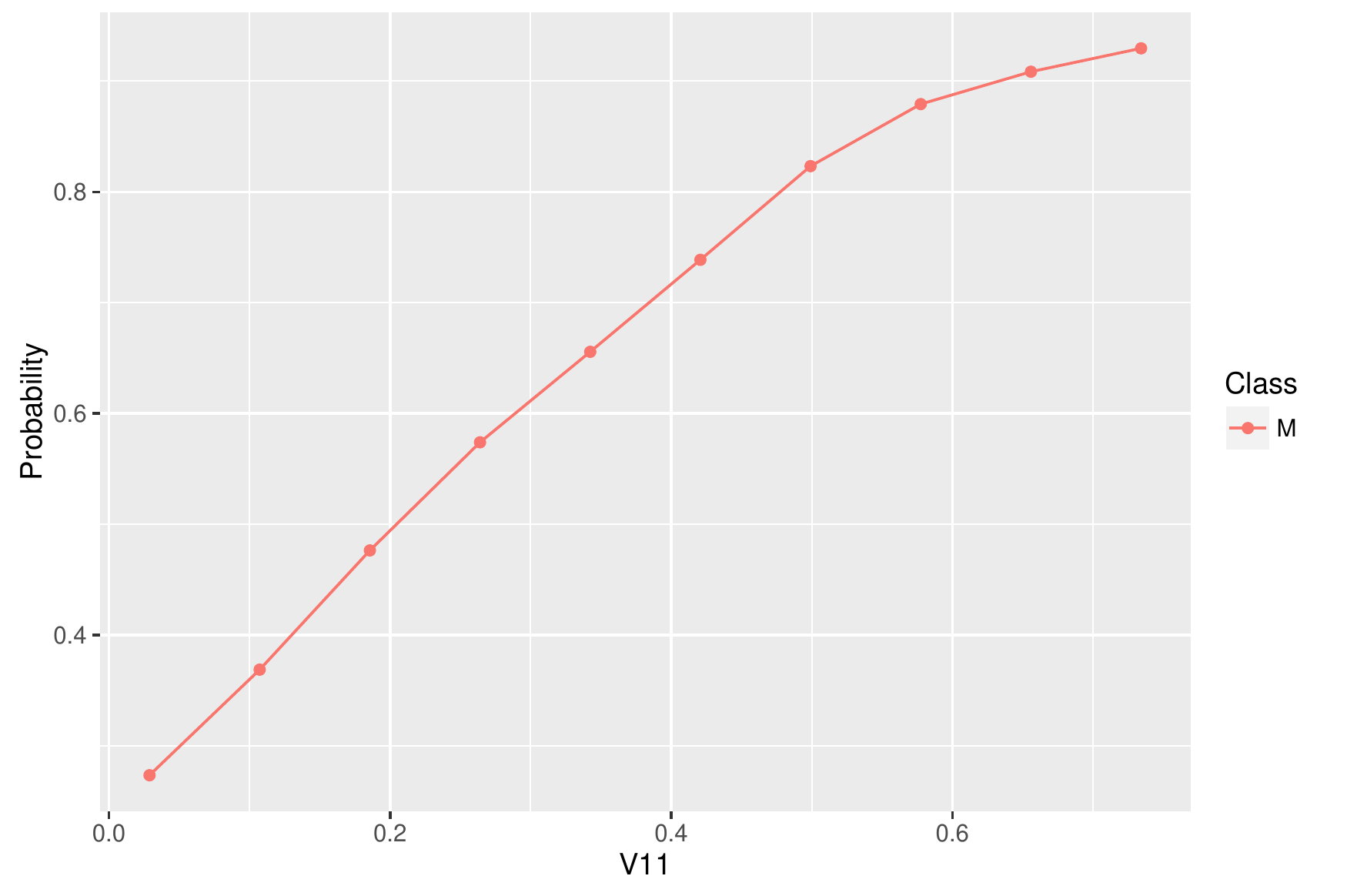}

\begin{lstlisting}[language=R]
plot(Probability ~ Value, data = plt$data, type = "b", xlab = plt$data$Feature[1])
\end{lstlisting}

\includegraphics{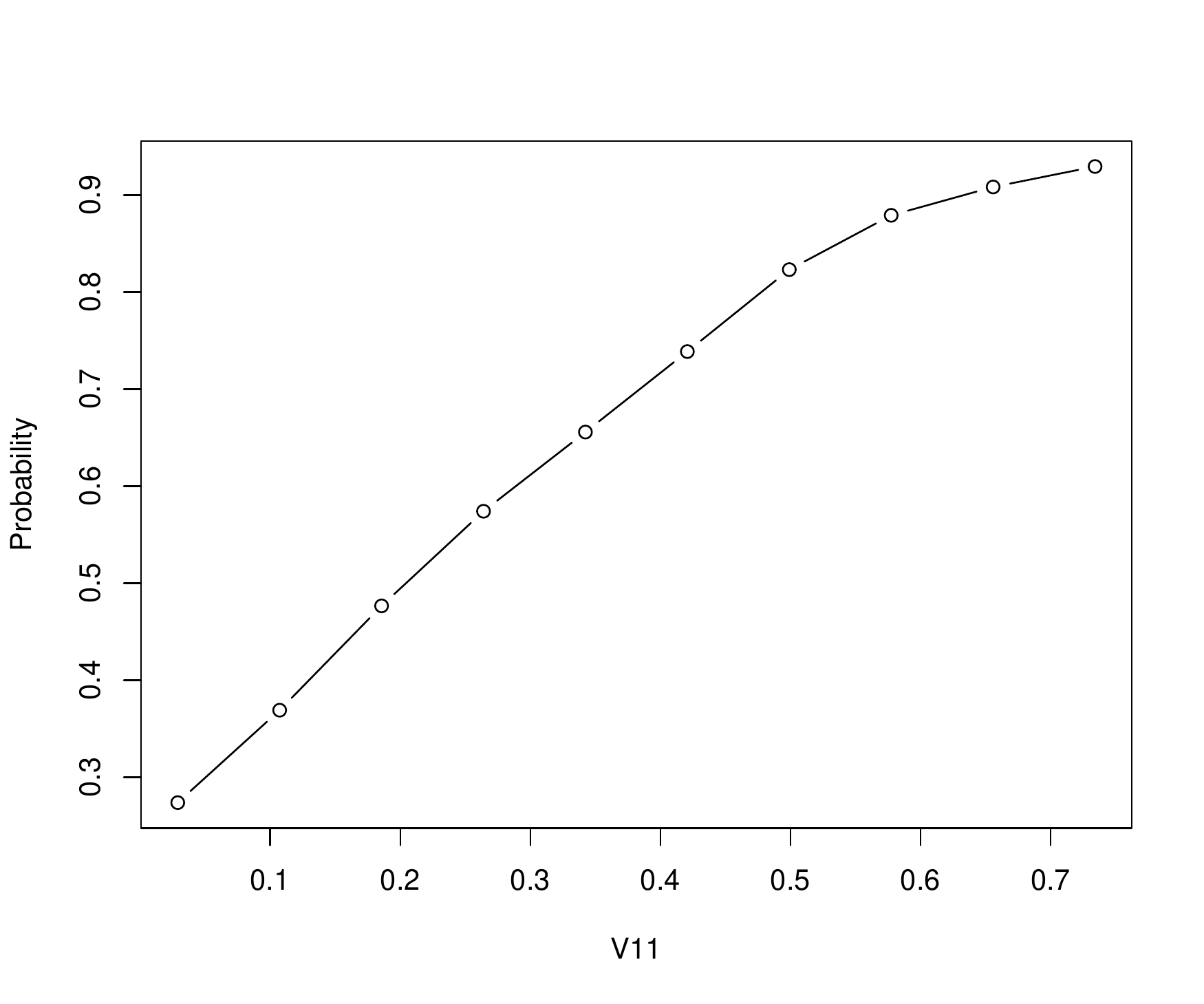}

\subsubsection{Available generation and plotting
functions}\label{available-generation-and-plotting-functions}

The table shows the currently available generation and plotting
functions. It also references tutorial pages that provide in depth
descriptions of the listed functions.

Note that some plots, e.g.,
\href{http://www.rdocumentation.org/packages/mlr/functions/plotTuneMultiCritResult.html}{plotTuneMultiCritResult}
are not described here since they lack a generation function. Both
\href{http://www.rdocumentation.org/packages/mlr/functions/plotThreshVsPerf.html}{plotThreshVsPerf}
and
\href{http://www.rdocumentation.org/packages/mlr/functions/plotROCCurves.html}{plotROCCurves}
operate on the result of
\href{http://www.rdocumentation.org/packages/mlr/functions/generateThreshVsPerfData.html}{generateThreshVsPerfData}.

The \href{http://www.rdocumentation.org/packages/ggvis/}{ggvis}
functions are experimental and are subject to change, though they should
work. Most generate interactive
\href{http://www.rdocumentation.org/packages/shiny/}{shiny}
applications, that automatically start and run locally.

\begin{longtable}[]{@{}llll@{}}
\toprule
\begin{minipage}[b]{0.20\columnwidth}\raggedright\strut
generation function\strut
\end{minipage} & \begin{minipage}[b]{0.16\columnwidth}\raggedright\strut
ggplot2 plotting function\strut
\end{minipage} & \begin{minipage}[b]{0.18\columnwidth}\raggedright\strut
ggvis plotting function\strut
\end{minipage} & \begin{minipage}[b]{0.35\columnwidth}\raggedright\strut
tutorial page\strut
\end{minipage}\tabularnewline
\midrule
\endhead
\begin{minipage}[t]{0.20\columnwidth}\raggedright\strut
\href{http://www.rdocumentation.org/packages/mlr/functions/generateThreshVsPerfData.html}{generateThreshVsPerfData}\strut
\end{minipage} & \begin{minipage}[t]{0.16\columnwidth}\raggedright\strut
\href{http://www.rdocumentation.org/packages/mlr/functions/plotThresVsPerf.html}{plotThresVsPerf}\strut
\end{minipage} & \begin{minipage}[t]{0.18\columnwidth}\raggedright\strut
\href{http://www.rdocumentation.org/packages/mlr/functions/plotThreshVsPerfGGVIS.html}{plotThreshVsPerfGGVIS}\strut
\end{minipage} & \begin{minipage}[t]{0.35\columnwidth}\raggedright\strut
\protect\hyperlink{evaluating-learner-performance}{Performance}\strut
\end{minipage}\tabularnewline
\begin{minipage}[t]{0.20\columnwidth}\raggedright\strut
\strut
\end{minipage} & \begin{minipage}[t]{0.16\columnwidth}\raggedright\strut
\href{http://www.rdocumentation.org/packages/mlr/functions/plotROCCurves.html}{plotROCCurves}\strut
\end{minipage} & \begin{minipage}[t]{0.18\columnwidth}\raggedright\strut
--\strut
\end{minipage} & \begin{minipage}[t]{0.35\columnwidth}\raggedright\strut
\protect\hyperlink{roc-analysis-and-performance-curves}{ROC
Analysis}\strut
\end{minipage}\tabularnewline
\begin{minipage}[t]{0.20\columnwidth}\raggedright\strut
\href{http://www.rdocumentation.org/packages/mlr/functions/generateCritDifferencesData.html}{generateCritDifferencesData}\strut
\end{minipage} & \begin{minipage}[t]{0.16\columnwidth}\raggedright\strut
\href{http://www.rdocumentation.org/packages/mlr/functions/plotCritDifferences.html}{plotCritDifferences}\strut
\end{minipage} & \begin{minipage}[t]{0.18\columnwidth}\raggedright\strut
--\strut
\end{minipage} & \begin{minipage}[t]{0.35\columnwidth}\raggedright\strut
\protect\hyperlink{benchmark-experiments}{Benchmark Experiments}\strut
\end{minipage}\tabularnewline
\begin{minipage}[t]{0.20\columnwidth}\raggedright\strut
\href{http://www.rdocumentation.org/packages/mlr/functions/generateHyperParsEffectData.html}{generateHyperParsEffectData}\strut
\end{minipage} & \begin{minipage}[t]{0.16\columnwidth}\raggedright\strut
\href{http://www.rdocumentation.org/packages/mlr/functions/plotHyperParsEffect.html}{plotHyperParsEffect}\strut
\end{minipage} & \begin{minipage}[t]{0.18\columnwidth}\raggedright\strut
\strut
\end{minipage} & \begin{minipage}[t]{0.35\columnwidth}\raggedright\strut
\protect\hyperlink{tuning-hyperparameters}{Tuning}\strut
\end{minipage}\tabularnewline
\begin{minipage}[t]{0.20\columnwidth}\raggedright\strut
\href{http://www.rdocumentation.org/packages/mlr/functions/generateFilterValuesData.html}{generateFilterValuesData}\strut
\end{minipage} & \begin{minipage}[t]{0.16\columnwidth}\raggedright\strut
\href{http://www.rdocumentation.org/packages/mlr/functions/plotFilterValues.html}{plotFilterValues}\strut
\end{minipage} & \begin{minipage}[t]{0.18\columnwidth}\raggedright\strut
\href{http://www.rdocumentation.org/packages/mlr/functions/plotFilterValuesGGVIS.html}{plotFilterValuesGGVIS}\strut
\end{minipage} & \begin{minipage}[t]{0.35\columnwidth}\raggedright\strut
\protect\hyperlink{feature-selection}{Feature Selection}\strut
\end{minipage}\tabularnewline
\begin{minipage}[t]{0.20\columnwidth}\raggedright\strut
\href{http://www.rdocumentation.org/packages/mlr/functions/generateLearningCurveData.html}{generateLearningCurveData}\strut
\end{minipage} & \begin{minipage}[t]{0.16\columnwidth}\raggedright\strut
\href{http://www.rdocumentation.org/packages/mlr/functions/plotLearningCurve.html}{plotLearningCurve}\strut
\end{minipage} & \begin{minipage}[t]{0.18\columnwidth}\raggedright\strut
\href{http://www.rdocumentation.org/packages/mlr/functions/plotLearningCurveGGVIS.html}{plotLearningCurveGGVIS}\strut
\end{minipage} & \begin{minipage}[t]{0.35\columnwidth}\raggedright\strut
\protect\hyperlink{learning-curve-analysis}{Learning Curves}\strut
\end{minipage}\tabularnewline
\begin{minipage}[t]{0.20\columnwidth}\raggedright\strut
\href{http://www.rdocumentation.org/packages/mlr/functions/generatePartialDependenceData.html}{generatePartialDependenceData}\strut
\end{minipage} & \begin{minipage}[t]{0.16\columnwidth}\raggedright\strut
\href{http://www.rdocumentation.org/packages/mlr/functions/plotPartialDependence.html}{plotPartialDependence}\strut
\end{minipage} & \begin{minipage}[t]{0.18\columnwidth}\raggedright\strut
\href{http://www.rdocumentation.org/packages/mlr/functions/plotPartialDependenceGGVIS.html}{plotPartialDependenceGGVIS}\strut
\end{minipage} & \begin{minipage}[t]{0.35\columnwidth}\raggedright\strut
\protect\hyperlink{exploring-learner-predictions}{Partial Dependence
Plots}\strut
\end{minipage}\tabularnewline
\begin{minipage}[t]{0.20\columnwidth}\raggedright\strut
\href{http://www.rdocumentation.org/packages/mlr/functions/generateFunctionalANOVAData.html}{generateFunctionalANOVAData}\strut
\end{minipage} & \begin{minipage}[t]{0.16\columnwidth}\raggedright\strut
\strut
\end{minipage} & \begin{minipage}[t]{0.18\columnwidth}\raggedright\strut
\strut
\end{minipage} & \begin{minipage}[t]{0.35\columnwidth}\raggedright\strut
\strut
\end{minipage}\tabularnewline
\begin{minipage}[t]{0.20\columnwidth}\raggedright\strut
\href{http://www.rdocumentation.org/packages/mlr/functions/generateCalibrationData.html}{generateCalibrationData}\strut
\end{minipage} & \begin{minipage}[t]{0.16\columnwidth}\raggedright\strut
\href{http://www.rdocumentation.org/packages/mlr/functions/plotCalibration.html}{plotCalibration}\strut
\end{minipage} & \begin{minipage}[t]{0.18\columnwidth}\raggedright\strut
--\strut
\end{minipage} & \begin{minipage}[t]{0.35\columnwidth}\raggedright\strut
\protect\hyperlink{classifier-calibration}{Classifier Calibration
Plots}\strut
\end{minipage}\tabularnewline
\bottomrule
\end{longtable}

\section{Advanced}\label{advanced}

\hypertarget{configuring-mlr}{\subsection{Configuring
mlr}\label{configuring-mlr}}

\href{http://www.rdocumentation.org/packages/mlr/}{mlr} is designed to
make usage errors due to typos or invalid parameter values as unlikely
as possible. Occasionally, you might want to break those barriers and
get full access, for example to reduce the amount of output on the
console or to turn off checks. For all available options simply refer to
the documentation of
\href{http://www.rdocumentation.org/packages/mlr/functions/configureMlr.html}{configureMlr}.
In the following we show some common use cases.

Generally, function
\href{http://www.rdocumentation.org/packages/mlr/functions/configureMlr.html}{configureMlr}
permits to set options globally for your current \textbf{R} session.

It is also possible to set options locally.

\begin{itemize}
\tightlist
\item
  All options referring to the behavior of learners (these are all
  options except \lstinline!show.info!) can be set for an individual
  learner via the \lstinline!config! argument of
  \href{http://www.rdocumentation.org/packages/mlr/functions/makeLearner.html}{makeLearner}.
  The local precedes the global configuration.
\item
  Some functions like
  \href{http://www.rdocumentation.org/packages/mlr/functions/resample.html}{resample},
  \href{http://www.rdocumentation.org/packages/mlr/functions/benchmark.html}{benchmark},
  \href{http://www.rdocumentation.org/packages/mlr/functions/selectFeatures.html}{selectFeatures},
  \href{http://www.rdocumentation.org/packages/mlr/functions/tuneParams.html}{tuneParams},
  and
  \href{http://www.rdocumentation.org/packages/mlr/functions/tuneParamsMultiCrit.html}{tuneParamsMultiCrit}
  have a \lstinline!show.info! flag that controls if progress messages
  are shown. The default value of \lstinline!show.info! can be set by
  \href{http://www.rdocumentation.org/packages/mlr/functions/configureMlr.html}{configureMlr}.
\end{itemize}

\subsubsection{Example: Reducing the output on the
console}\label{example-reducing-the-output-on-the-console}

You are bothered by all the output on the console like in this example?

\begin{lstlisting}[language=R]
rdesc = makeResampleDesc("Holdout")
r = resample("classif.multinom", iris.task, rdesc)
#> [Resample] holdout iter: 1
#> # weights:  18 (10 variable)
#> initial  value 109.861229 
#> iter  10 value 12.256619
#> iter  20 value 3.638740
#> iter  30 value 3.228628
#> iter  40 value 2.951100
#> iter  50 value 2.806521
#> iter  60 value 2.739076
#> iter  70 value 2.522206
#> iter  80 value 2.485225
#> iter  90 value 2.381397
#> iter 100 value 2.360602
#> final  value 2.360602 
#> stopped after 100 iterations
#> [Resample] Result: mmce.test.mean=0.02
\end{lstlisting}

You can suppress the output for this
\href{http://www.rdocumentation.org/packages/mlr/functions/makeLearner.html}{Learner}
and this
\href{http://www.rdocumentation.org/packages/mlr/functions/resample.html}{resample}
call as follows:

\begin{lstlisting}[language=R]
lrn = makeLearner("classif.multinom", config = list(show.learner.output = FALSE))
r = resample(lrn, iris.task, rdesc, show.info = FALSE)
\end{lstlisting}

(Note that
\href{http://www.rdocumentation.org/packages/nnet/functions/multinom.html}{multinom}
has a \lstinline!trace! switch that can alternatively be used to turn
off the progress messages.)

To globally suppress the output for all subsequent learners and calls to
\href{http://www.rdocumentation.org/packages/mlr/functions/resample.html}{resample},
\href{http://www.rdocumentation.org/packages/mlr/functions/benchmark.html}{benchmark}
etc. do the following:

\begin{lstlisting}[language=R]
configureMlr(show.learner.output = FALSE, show.info = FALSE)
r = resample("classif.multinom", iris.task, rdesc)
\end{lstlisting}

\subsubsection{Accessing and resetting the
configuration}\label{accessing-and-resetting-the-configuration}

Function
\href{http://www.rdocumentation.org/packages/mlr/functions/getMlrOptions.html}{getMlrOptions}
returns a
\href{http://www.rdocumentation.org/packages/base/functions/list.html}{list}
with the current configuration.

\begin{lstlisting}[language=R]
getMlrOptions()
#> $on.learner.error
#> [1] "stop"
#> 
#> $on.learner.warning
#> [1] "warn"
#> 
#> $on.par.out.of.bounds
#> [1] "stop"
#> 
#> $on.par.without.desc
#> [1] "stop"
#> 
#> $show.info
#> [1] FALSE
#> 
#> $show.learner.output
#> [1] FALSE
\end{lstlisting}

To restore the default configuration call
\href{http://www.rdocumentation.org/packages/mlr/functions/configureMlr.html}{configureMlr}
with an empty argument list.

\begin{lstlisting}[language=R]
configureMlr()
\end{lstlisting}

\begin{lstlisting}[language=R]
getMlrOptions()
#> $on.learner.error
#> [1] "stop"
#> 
#> $on.learner.warning
#> [1] "warn"
#> 
#> $on.par.out.of.bounds
#> [1] "stop"
#> 
#> $on.par.without.desc
#> [1] "stop"
#> 
#> $show.info
#> [1] TRUE
#> 
#> $show.learner.output
#> [1] TRUE
\end{lstlisting}

\subsubsection{Example: Turning off parameter
checking}\label{example-turning-off-parameter-checking}

It might happen that you want to set a parameter of a
\href{http://www.rdocumentation.org/packages/mlr/functions/makeLearner.html}{Learner},
but the parameter is not registered in the learner's
\href{http://www.rdocumentation.org/packages/ParamHelpers/functions/makeParamSet.html}{parameter
set} yet. In this case you might want to
\href{https://github.com/mlr-org/mlr\#get-in-touch}{contact us} or
\href{https://github.com/mlr-org/mlr/issues/new}{open an issue} as well!
But until the problem is fixed you can turn off
\href{http://www.rdocumentation.org/packages/mlr/}{mlr}'s parameter
checking. The parameter setting will then be passed to the underlying
function without further ado.

\begin{lstlisting}[language=R]
### Support Vector Machine with linear kernel and new parameter 'newParam'
lrn = makeLearner("classif.ksvm", kernel = "vanilladot", newParam = 3)
#> Error in setHyperPars2.Learner(learner, insert(par.vals, args)): classif.ksvm: Setting parameter newParam without available description object! 
#> Did you mean one of these hyperparameters instead: degree scaled kernel 
#> You can switch off this check by using configureMlr!

### Turn off parameter checking completely
configureMlr(on.par.without.desc = "quiet")
lrn = makeLearner("classif.ksvm", kernel = "vanilladot", newParam = 3)
train(lrn, iris.task)
#>  Setting default kernel parameters
#> Model for learner.id=classif.ksvm; learner.class=classif.ksvm
#> Trained on: task.id = iris-example; obs = 150; features = 4
#> Hyperparameters: fit=FALSE,kernel=vanilladot,newParam=3

### Option "quiet" also masks typos
lrn = makeLearner("classif.ksvm", kernl = "vanilladot")
train(lrn, iris.task)
#> Model for learner.id=classif.ksvm; learner.class=classif.ksvm
#> Trained on: task.id = iris-example; obs = 150; features = 4
#> Hyperparameters: fit=FALSE,kernl=vanilladot

### Alternatively turn off parameter checking, but still see warnings
configureMlr(on.par.without.desc = "warn")
lrn = makeLearner("classif.ksvm", kernl = "vanilladot", newParam = 3)
#> Warning in setHyperPars2.Learner(learner, insert(par.vals, args)): classif.ksvm: Setting parameter kernl without available description object! 
#> Did you mean one of these hyperparameters instead: kernel nu degree 
#> You can switch off this check by using configureMlr!
#> Warning in setHyperPars2.Learner(learner, insert(par.vals, args)): classif.ksvm: Setting parameter newParam without available description object! 
#> Did you mean one of these hyperparameters instead: degree scaled kernel 
#> You can switch off this check by using configureMlr!

train(lrn, iris.task)
#> Model for learner.id=classif.ksvm; learner.class=classif.ksvm
#> Trained on: task.id = iris-example; obs = 150; features = 4
#> Hyperparameters: fit=FALSE,kernl=vanilladot,newParam=3
\end{lstlisting}

\subsubsection{Example: Handling errors in a learning
method}\label{example-handling-errors-in-a-learning-method}

If a learning method throws an error the default behavior of
\href{http://www.rdocumentation.org/packages/mlr/}{mlr} is to generate
an exception as well. However, in some situations, for example if you
conduct a larger \protect\hyperlink{benchmark-experiments}{benchmark
study} with multiple data sets and learners, you usually don't want the
whole experiment stopped due to one error. You can prevent this using
the \lstinline!on.learner.error! option of
\href{http://www.rdocumentation.org/packages/mlr/functions/configureMlr.html}{configureMlr}.

\begin{lstlisting}[language=R]
### This call gives an error caused by the low number of observations in class "virginica"
train("classif.qda", task = iris.task, subset = 1:104)
#> Error in qda.default(x, grouping, ...): some group is too small for 'qda'
#> Timing stopped at: 0.003 0 0.002

### Get a warning instead of an error
configureMlr(on.learner.error = "warn")
mod = train("classif.qda", task = iris.task, subset = 1:104)
#> Warning in train("classif.qda", task = iris.task, subset = 1:104): Could not train learner classif.qda: Error in qda.default(x, grouping, ...) : 
#>   some group is too small for 'qda'

mod
#> Model for learner.id=classif.qda; learner.class=classif.qda
#> Trained on: task.id = iris-example; obs = 104; features = 4
#> Hyperparameters: 
#> Training failed: Error in qda.default(x, grouping, ...) : 
#>   some group is too small for 'qda'
#> 
#> Training failed: Error in qda.default(x, grouping, ...) : 
#>   some group is too small for 'qda'

### mod is an object of class FailureModel
isFailureModel(mod)
#> [1] TRUE

### Retrieve the error message
getFailureModelMsg(mod)
#> [1] "Error in qda.default(x, grouping, ...) : \n  some group is too small for 'qda'\n"

### predict and performance return NA's
pred = predict(mod, iris.task)
pred
#> Prediction: 150 observations
#> predict.type: response
#> threshold: 
#> time: NA
#>   id  truth response
#> 1  1 setosa     <NA>
#> 2  2 setosa     <NA>
#> 3  3 setosa     <NA>
#> 4  4 setosa     <NA>
#> 5  5 setosa     <NA>
#> 6  6 setosa     <NA>
#> ... (150 rows, 3 cols)

performance(pred)
#> mmce 
#>   NA
\end{lstlisting}

If \lstinline!on.learner.error = "warn"! a warning is issued instead of
an exception and an object of class
\href{http://www.rdocumentation.org/packages/mlr/functions/FailureModel.html}{FailureModel}
is created. You can extract the error message using function
\href{http://www.rdocumentation.org/packages/mlr/functions/getFailureModelMsg.html}{getFailureModelMsg}.
All further steps like prediction and performance calculation work and
return \lstinline!NA's!.

\hypertarget{wrapper}{\subsection{Wrapper}\label{wrapper}}

Wrappers can be employed to extend integrated
\href{http://www.rdocumentation.org/packages/mlr/functions/makeLearner.html}{learners}
with new functionality. The broad scope of operations and methods which
are implemented as wrappers underline the flexibility of the wrapping
approach:

\begin{itemize}
\tightlist
\item
  \protect\hyperlink{data-preprocessing}{Data preprocessing}
\item
  \protect\hyperlink{imputation-of-missing-values}{Imputation}
\item
  \protect\hyperlink{generic-bagging}{Bagging}
\item
  \protect\hyperlink{tuning-hyperparameters}{Tuning}
\item
  \protect\hyperlink{feature-selection}{Feature selection}
\item
  \protect\hyperlink{cost-sensitive-classification}{Cost-sensitive
  classification}
\item
  \protect\hyperlink{imbalanced-classification-problems}{Over- and
  undersampling} for imbalanced classification problems
\item
  \href{http://www.rdocumentation.org/packages/mlr/functions/makeMulticlassWrapper.html}{Multiclass
  extension} for binary-class learners
\item
  \protect\hyperlink{multilabel-classification}{Multilabel
  classification}
\end{itemize}

All these operations and methods have a few things in common: First,
they all wrap around
\href{http://www.rdocumentation.org/packages/mlr/}{mlr}
\href{http://www.rdocumentation.org/packages/mlr/functions/makeLearner.html}{learners}
and they return a new learner. Therefore learners can be wrapped
multiple times. Second, they are implemented using a \emph{train}
(pre-model hook) and \emph{predict} (post-model hook) method.

\subsubsection{Example: Bagging wrapper}\label{example-bagging-wrapper}

In this section we exemplary describe the bagging wrapper to create a
random forest which supports weights. To achieve that we combine several
decision trees from the
\href{http://www.rdocumentation.org/packages/rpart/}{rpart} package to
create our own custom random forest.

First, we create a weighted toy task.

\begin{lstlisting}[language=R]
data(iris)
task = makeClassifTask(data = iris, target = "Species", weights = as.integer(iris$Species))
\end{lstlisting}

Next, we use
\href{http://www.rdocumentation.org/packages/mlr/functions/makeBaggingWrapper.html}{makeBaggingWrapper}
to create the base learners and the bagged learner. We choose to set
equivalents of \lstinline!ntree! (100 base learners) and
\lstinline!mtry! (proportion of randomly selected features).

\begin{lstlisting}[language=R]
base.lrn = makeLearner("classif.rpart")
wrapped.lrn = makeBaggingWrapper(base.lrn, bw.iters = 100, bw.feats = 0.5)
print(wrapped.lrn)
#> Learner classif.rpart.bagged from package rpart
#> Type: classif
#> Name: ; Short name: 
#> Class: BaggingWrapper
#> Properties: twoclass,multiclass,missings,numerics,factors,ordered,prob,weights,featimp
#> Predict-Type: response
#> Hyperparameters: xval=0,bw.iters=100,bw.feats=0.5
\end{lstlisting}

As we can see in the output, the wrapped learner inherited all
properties from the base learner, especially the ``weights'' attribute
is still present. We can use this newly constructed learner like all
base learners, i.e.~we can use it in
\href{http://www.rdocumentation.org/packages/mlr/functions/train.html}{train},
\href{http://www.rdocumentation.org/packages/mlr/functions/benchmark.html}{benchmark},
\href{http://www.rdocumentation.org/packages/mlr/functions/resample.html}{resample},
etc.

\begin{lstlisting}[language=R]
benchmark(tasks = task, learners = list(base.lrn, wrapped.lrn))
#> Task: iris, Learner: classif.rpart
#> [Resample] cross-validation iter: 1
#> [Resample] cross-validation iter: 2
#> [Resample] cross-validation iter: 3
#> [Resample] cross-validation iter: 4
#> [Resample] cross-validation iter: 5
#> [Resample] cross-validation iter: 6
#> [Resample] cross-validation iter: 7
#> [Resample] cross-validation iter: 8
#> [Resample] cross-validation iter: 9
#> [Resample] cross-validation iter: 10
#> [Resample] Result: mmce.test.mean=0.0667
#> Task: iris, Learner: classif.rpart.bagged
#> [Resample] cross-validation iter: 1
#> [Resample] cross-validation iter: 2
#> [Resample] cross-validation iter: 3
#> [Resample] cross-validation iter: 4
#> [Resample] cross-validation iter: 5
#> [Resample] cross-validation iter: 6
#> [Resample] cross-validation iter: 7
#> [Resample] cross-validation iter: 8
#> [Resample] cross-validation iter: 9
#> [Resample] cross-validation iter: 10
#> [Resample] Result: mmce.test.mean=0.06
#>   task.id           learner.id mmce.test.mean
#> 1    iris        classif.rpart     0.06666667
#> 2    iris classif.rpart.bagged     0.06000000
\end{lstlisting}

That far we are quite happy with our new learner. But we hope for a
better performance by tuning some hyperparameters of both the decision
trees and bagging wrapper. Let's have a look at the available
hyperparameters of the fused learner:

\begin{lstlisting}[language=R]
getParamSet(wrapped.lrn)
#>                    Type len   Def   Constr Req Tunable Trafo
#> bw.iters        integer   -    10 1 to Inf   -    TRUE     -
#> bw.replace      logical   -  TRUE        -   -    TRUE     -
#> bw.size         numeric   -     -   0 to 1   -    TRUE     -
#> bw.feats        numeric   - 0.667   0 to 1   -    TRUE     -
#> minsplit        integer   -    20 1 to Inf   -    TRUE     -
#> minbucket       integer   -     - 1 to Inf   -    TRUE     -
#> cp              numeric   -  0.01   0 to 1   -    TRUE     -
#> maxcompete      integer   -     4 0 to Inf   -    TRUE     -
#> maxsurrogate    integer   -     5 0 to Inf   -    TRUE     -
#> usesurrogate   discrete   -     2    0,1,2   -    TRUE     -
#> surrogatestyle discrete   -     0      0,1   -    TRUE     -
#> maxdepth        integer   -    30  1 to 30   -    TRUE     -
#> xval            integer   -    10 0 to Inf   -   FALSE     -
#> parms           untyped   -     -        -   -    TRUE     -
\end{lstlisting}

We choose to tune the parameters \lstinline!minsplit! and
\lstinline!bw.feats! for the
\protect\hyperlink{implemented-performance-measures}{mmce} using a
\href{http://www.rdocumentation.org/packages/mlr/functions/TuneControl.html}{random
search} in a 3-fold CV:

\begin{lstlisting}[language=R]
ctrl = makeTuneControlRandom(maxit = 10)
rdesc = makeResampleDesc("CV", iters = 3)
par.set = makeParamSet(
  makeIntegerParam("minsplit", lower = 1, upper = 10),
  makeNumericParam("bw.feats", lower = 0.25, upper = 1)
)
tuned.lrn = makeTuneWrapper(wrapped.lrn, rdesc, mmce, par.set, ctrl)
print(tuned.lrn)
#> Learner classif.rpart.bagged.tuned from package rpart
#> Type: classif
#> Name: ; Short name: 
#> Class: TuneWrapper
#> Properties: numerics,factors,ordered,missings,weights,prob,twoclass,multiclass,featimp
#> Predict-Type: response
#> Hyperparameters: xval=0,bw.iters=100,bw.feats=0.5
\end{lstlisting}

Calling the train method of the newly constructed learner performs the
following steps:

\begin{enumerate}
\def\labelenumi{\arabic{enumi}.}
\tightlist
\item
  The tuning wrapper sets parameters for the underlying model in slot
  \lstinline!$next.learner! and calls its train method.
\item
  Next learner is the bagging wrapper. The passed down argument
  \lstinline!bw.feats! is used in the bagging wrapper training function,
  the argument \lstinline!minsplit! gets passed down to
  \lstinline!$next.learner!. The base wrapper function calls the base
  learner \lstinline!bw.iters! times and stores the resulting models.
\item
  The bagged models are evaluated using the mean
  \protect\hyperlink{implemented-performance-measures}{mmce} (default
  aggregation for this performance measure) and new parameters are
  selected using the tuning method.
\item
  This is repeated until the tuner terminates. Output is a tuned bagged
  learner.
\end{enumerate}

\begin{lstlisting}[language=R]
lrn = train(tuned.lrn, task = task)
#> [Tune] Started tuning learner classif.rpart.bagged for parameter set:
#>             Type len Def    Constr Req Tunable Trafo
#> minsplit integer   -   -   1 to 10   -    TRUE     -
#> bw.feats numeric   -   - 0.25 to 1   -    TRUE     -
#> With control class: TuneControlRandom
#> Imputation value: 1
#> [Tune-x] 1: minsplit=5; bw.feats=0.935
#> [Tune-y] 1: mmce.test.mean=0.0467; time: 0.1 min; memory: 179Mb use, 711Mb max
#> [Tune-x] 2: minsplit=9; bw.feats=0.675
#> [Tune-y] 2: mmce.test.mean=0.0467; time: 0.1 min; memory: 179Mb use, 711Mb max
#> [Tune-x] 3: minsplit=2; bw.feats=0.847
#> [Tune-y] 3: mmce.test.mean=0.0467; time: 0.1 min; memory: 179Mb use, 711Mb max
#> [Tune-x] 4: minsplit=4; bw.feats=0.761
#> [Tune-y] 4: mmce.test.mean=0.0467; time: 0.1 min; memory: 179Mb use, 711Mb max
#> [Tune-x] 5: minsplit=6; bw.feats=0.338
#> [Tune-y] 5: mmce.test.mean=0.0867; time: 0.0 min; memory: 179Mb use, 711Mb max
#> [Tune-x] 6: minsplit=1; bw.feats=0.637
#> [Tune-y] 6: mmce.test.mean=0.0467; time: 0.1 min; memory: 179Mb use, 711Mb max
#> [Tune-x] 7: minsplit=1; bw.feats=0.998
#> [Tune-y] 7: mmce.test.mean=0.0467; time: 0.1 min; memory: 179Mb use, 711Mb max
#> [Tune-x] 8: minsplit=4; bw.feats=0.698
#> [Tune-y] 8: mmce.test.mean=0.0467; time: 0.1 min; memory: 179Mb use, 711Mb max
#> [Tune-x] 9: minsplit=3; bw.feats=0.836
#> [Tune-y] 9: mmce.test.mean=0.0467; time: 0.1 min; memory: 179Mb use, 711Mb max
#> [Tune-x] 10: minsplit=10; bw.feats=0.529
#> [Tune-y] 10: mmce.test.mean=0.0533; time: 0.0 min; memory: 179Mb use, 711Mb max
#> [Tune] Result: minsplit=1; bw.feats=0.998 : mmce.test.mean=0.0467
print(lrn)
#> Model for learner.id=classif.rpart.bagged.tuned; learner.class=TuneWrapper
#> Trained on: task.id = iris; obs = 150; features = 4
#> Hyperparameters: xval=0,bw.iters=100,bw.feats=0.5
\end{lstlisting}

\hypertarget{data-preprocessing}{\subsection{Data
Preprocessing}\label{data-preprocessing}}

Data preprocessing refers to any transformation of the data done before
applying a learning algorithm. This comprises for example finding and
resolving inconsistencies, imputation of missing values, identifying,
removing or replacing outliers, discretizing numerical data or
generating numerical dummy variables for categorical data, any kind of
transformation like standardization of predictors or Box-Cox,
dimensionality reduction and feature extraction and/or selection.

\href{http://www.rdocumentation.org/packages/mlr/}{mlr} offers several
options for data preprocessing. Some of the following simple methods to
change a
\href{http://www.rdocumentation.org/packages/mlr/functions/Task.html}{Task}
(or
\href{http://www.rdocumentation.org/packages/base/functions/data.frame.html}{data.frame})
were already mentioned on the page about
\protect\hyperlink{learning-tasks}{learning tasks}:

\begin{itemize}
\tightlist
\item
  \href{http://www.rdocumentation.org/packages/mlr/functions/capLargeValues.html}{capLargeValues}:
  Convert large/infinite numeric values.
\item
  \href{http://www.rdocumentation.org/packages/mlr/functions/createDummyFeatures.html}{createDummyFeatures}:
  Generate dummy variables for factor features.
\item
  \href{http://www.rdocumentation.org/packages/mlr/functions/dropFeatures.html}{dropFeatures}:
  Remove selected features.
\item
  \href{http://www.rdocumentation.org/packages/mlr/functions/joinClassLevels.html}{joinClassLevels}:
  Only for classification: Merge existing classes to new, larger
  classes.
\item
  \href{http://www.rdocumentation.org/packages/mlr/functions/mergeSmallFactorLevels.html}{mergeSmallFactorLevels}:
  Merge infrequent levels of factor features.
\item
  \href{http://www.rdocumentation.org/packages/mlr/functions/normalizeFeatures.html}{normalizeFeatures}:
  Normalize features by different methods, e.g., standardization or
  scaling to a certain range.
\item
  \href{http://www.rdocumentation.org/packages/mlr/functions/removeConstantFeatures.html}{removeConstantFeatures}:
  Remove constant features.
\item
  \href{http://www.rdocumentation.org/packages/mlr/functions/subsetTask.html}{subsetTask}:
  Remove observations and/or features from a
  \href{http://www.rdocumentation.org/packages/mlr/functions/Task.html}{Task}.
\end{itemize}

Moreover, there are tutorial pages devoted to

\begin{itemize}
\tightlist
\item
  \protect\hyperlink{feature-selection}{Feature selection} and
\item
  \protect\hyperlink{imputation-of-missing-values}{Imputation of missing
  values}.
\end{itemize}

\subsubsection{Fusing learners with
preprocessing}\label{fusing-learners-with-preprocessing}

\href{http://www.rdocumentation.org/packages/mlr/}{mlr}'s wrapper
functionality permits to combine learners with preprocessing steps. This
means that the preprocessing ``belongs'' to the learner and is done any
time the learner is trained or predictions are made.

This is, on the one hand, very practical. You don't need to change any
data or learning
\href{http://www.rdocumentation.org/packages/mlr/functions/Task.html}{Task}s
and it's quite easy to combine different learners with different
preprocessing steps.

On the other hand this helps to avoid a common mistake in evaluating the
performance of a learner with preprocessing: Preprocessing is often seen
as completely independent of the later applied learning algorithms. When
estimating the performance of the a learner, e.g., by cross-validation
all preprocessing is done beforehand on the full data set and only
training/predicting the learner is done on the train/test sets.
Depending on what exactly is done as preprocessing this can lead to
overoptimistic results. For example if imputation by the mean is done on
the whole data set before evaluating the learner performance you are
using information from the test data during training, which can cause
overoptimistic performance results.

To clarify things one should distinguish between \emph{data-dependent}
and \emph{data-independent} preprocessing steps: Data-dependent steps in
some way learn from the data and give different results when applied to
different data sets. Data-independent steps always lead to the same
results. Clearly, correcting errors in the data or removing data columns
like Ids that should not be used for learning, is data-independent.
Imputation of missing values by the mean, as mentioned above, is
data-dependent. Imputation by a fixed constant, however, is not.

To get a honest estimate of learner performance combined with
preprocessing, all data-dependent preprocessing steps must be included
in the resampling. This is automatically done when fusing a learner with
preprocessing.

To this end \href{http://www.rdocumentation.org/packages/mlr/}{mlr}
provides two \protect\hyperlink{wrapper}{wrappers}:

\begin{itemize}
\tightlist
\item
  \href{http://www.rdocumentation.org/packages/mlr/functions/makePreprocWrapperCaret.html}{makePreprocWrapperCaret}
  is an interface to all preprocessing options offered by
  \href{http://www.rdocumentation.org/packages/caret/}{caret}'s
  \href{http://www.rdocumentation.org/packages/caret/functions/preProcess.html}{preProcess}
  function.
\item
  \href{http://www.rdocumentation.org/packages/mlr/functions/makePreprocWrapper.html}{makePreprocWrapper}
  permits to write your own custom preprocessing methods by defining the
  actions to be taken before training and before prediction.
\end{itemize}

As mentioned above the specified preprocessing steps then ``belong'' to
the wrapped
\href{http://www.rdocumentation.org/packages/mlr/functions/makeLearner.html}{Learner}.
In contrast to the preprocessing options listed above like
\href{http://www.rdocumentation.org/packages/mlr/functions/normalizeFeatures.html}{normalizeFeatures}

\begin{itemize}
\tightlist
\item
  the
  \href{http://www.rdocumentation.org/packages/mlr/functions/Task.html}{Task}
  itself remains unchanged,
\item
  the preprocessing is not done globally, i.e., for the whole data set,
  but for every pair of training/test data sets in, e.g., resampling,
\item
  any parameters controlling the preprocessing as, e.g., the percentage
  of outliers to be removed can be
  \protect\hyperlink{tuning-hyperparameters}{tuned} together with the
  base learner parameters.
\end{itemize}

We start with some examples for
\href{http://www.rdocumentation.org/packages/mlr/functions/makePreprocWrapperCaret.html}{makePreprocWrapperCaret}.

\subsubsection{Preprocessing with
makePreprocWrapperCaret}\label{preprocessing-with-makepreprocwrappercaret}

\href{http://www.rdocumentation.org/packages/mlr/functions/makePreprocWrapperCaret.html}{makePreprocWrapperCaret}
is an interface to
\href{http://www.rdocumentation.org/packages/caret/}{caret}'s
\href{http://www.rdocumentation.org/packages/caret/functions/preProcess.html}{preProcess}
function that provides many different options like imputation of missing
values, data transformations as scaling the features to a certain range
or Box-Cox and dimensionality reduction via Independent or Principal
Component Analysis. For all possible options see the help page of
function
\href{http://www.rdocumentation.org/packages/caret/functions/preProcess.html}{preProcess}.

Note that the usage of
\href{http://www.rdocumentation.org/packages/mlr/functions/makePreprocWrapperCaret.html}{makePreprocWrapperCaret}
is slightly different than that of
\href{http://www.rdocumentation.org/packages/caret/functions/preProcess.html}{preProcess}.

\begin{itemize}
\tightlist
\item
  \href{http://www.rdocumentation.org/packages/mlr/functions/makePreprocWrapperCaret.html}{makePreprocWrapperCaret}
  takes (almost) the same formal arguments as
  \href{\&caret:preProcess}{preProcess}, but their names are prefixed by
  \lstinline!ppc.!.
\item
  The only exception:
  \href{http://www.rdocumentation.org/packages/mlr/functions/makePreprocWrapperCaret.html}{makePreprocWrapperCaret}
  does not have a \lstinline!method! argument. Instead all preprocessing
  options that would be passed to
  \href{http://www.rdocumentation.org/packages/caret/functions/preProcess.html}{preProcess}'s
  \lstinline!method! argument are given as individual logical parameters
  to
  \href{http://www.rdocumentation.org/packages/mlr/functions/makePreprocWrapperCaret.html}{makePreprocWrapperCaret}.
\end{itemize}

For example the following call to
\href{http://www.rdocumentation.org/packages/caret/functions/preProcess.html}{preProcess}

\begin{lstlisting}[language=R]
preProcess(x, method = c("knnImpute", "pca"), pcaComp = 10)
\end{lstlisting}

with \lstinline!x! being a
\href{http://www.rdocumentation.org/packages/base/functions/matrix.html}{matrix}
or
\href{http://www.rdocumentation.org/packages/base/functions/data.frame.html}{data.frame}
would thus translate into

\begin{lstlisting}[language=R]
makePreprocWrapperCaret(learner, ppc.knnImpute = TRUE, ppc.pca = TRUE, ppc.pcaComp = 10)
\end{lstlisting}

where \lstinline!learner! is a
\href{http://www.rdocumentation.org/packages/mlr/}{mlr}
\href{http://www.rdocumentation.org/packages/mlr/functions/makeLearner.html}{Learner}
or the name of a learner class like \lstinline!"classif.lda"!.

If you enable multiple preprocessing options (like knn imputation and
principal component analysis above) these are executed in a certain
order detailed on the help page of function
\href{http://www.rdocumentation.org/packages/caret/functions/preProcess.html}{preProcess}.

In the following we show an example where principal components analysis
(PCA) is used for dimensionality reduction. This should never be applied
blindly, but can be beneficial with learners that get problems with high
dimensionality or those that can profit from rotating the data.

We consider the
\href{http://www.rdocumentation.org/packages/mlr/functions/sonar.task.html}{sonar.task},
which poses a binary classification problem with 208 observations and 60
features.

\begin{lstlisting}[language=R]
sonar.task
#> Supervised task: Sonar-example
#> Type: classif
#> Target: Class
#> Observations: 208
#> Features:
#> numerics  factors  ordered 
#>       60        0        0 
#> Missings: FALSE
#> Has weights: FALSE
#> Has blocking: FALSE
#> Classes: 2
#>   M   R 
#> 111  97 
#> Positive class: M
\end{lstlisting}

Below we fuse
\href{http://www.rdocumentation.org/packages/MASS/functions/qda.html}{quadratic
discriminant analysis} from package
\href{http://www.rdocumentation.org/packages/MASS/}{MASS} with a
principal components preprocessing step. The threshold is set to 0.9,
i.e., the principal components necessary to explain a cumulative
percentage of 90\% of the total variance are kept. The data are
automatically standardized prior to PCA.

\begin{lstlisting}[language=R]
lrn = makePreprocWrapperCaret("classif.qda", ppc.pca = TRUE, ppc.thresh = 0.9)
lrn
#> Learner classif.qda.preproc from package MASS
#> Type: classif
#> Name: ; Short name: 
#> Class: PreprocWrapperCaret
#> Properties: twoclass,multiclass,numerics,factors,prob
#> Predict-Type: response
#> Hyperparameters: ppc.BoxCox=FALSE,ppc.YeoJohnson=FALSE,ppc.expoTrans=FALSE,ppc.center=TRUE,ppc.scale=TRUE,ppc.range=FALSE,ppc.knnImpute=FALSE,ppc.bagImpute=FALSE,ppc.medianImpute=FALSE,ppc.pca=TRUE,ppc.ica=FALSE,ppc.spatialSign=FALSE,ppc.thresh=0.9,ppc.na.remove=TRUE,ppc.k=5,ppc.fudge=0.2,ppc.numUnique=3
\end{lstlisting}

The wrapped learner is trained on the
\href{http://www.rdocumentation.org/packages/mlr/functions/sonar.task.html}{sonar.task}.
By inspecting the underlying
\href{http://www.rdocumentation.org/packages/MASS/functions/qda.html}{qda}
model, we see that the first 22 principal components have been used for
training.

\begin{lstlisting}[language=R]
mod = train(lrn, sonar.task)
mod
#> Model for learner.id=classif.qda.preproc; learner.class=PreprocWrapperCaret
#> Trained on: task.id = Sonar-example; obs = 208; features = 60
#> Hyperparameters: ppc.BoxCox=FALSE,ppc.YeoJohnson=FALSE,ppc.expoTrans=FALSE,ppc.center=TRUE,ppc.scale=TRUE,ppc.range=FALSE,ppc.knnImpute=FALSE,ppc.bagImpute=FALSE,ppc.medianImpute=FALSE,ppc.pca=TRUE,ppc.ica=FALSE,ppc.spatialSign=FALSE,ppc.thresh=0.9,ppc.na.remove=TRUE,ppc.k=5,ppc.fudge=0.2,ppc.numUnique=3

getLearnerModel(mod)
#> Model for learner.id=classif.qda; learner.class=classif.qda
#> Trained on: task.id = Sonar-example; obs = 208; features = 22
#> Hyperparameters:

getLearnerModel(mod, more.unwrap = TRUE)
#> Call:
#> qda(f, data = getTaskData(.task, .subset, recode.target = "drop.levels"))
#> 
#> Prior probabilities of groups:
#>         M         R 
#> 0.5336538 0.4663462 
#> 
#> Group means:
#>          PC1        PC2        PC3         PC4         PC5         PC6
#> M  0.5976122 -0.8058235  0.9773518  0.03794232 -0.04568166 -0.06721702
#> R -0.6838655  0.9221279 -1.1184128 -0.04341853  0.05227489  0.07691845
#>          PC7         PC8        PC9       PC10        PC11          PC12
#> M  0.2278162 -0.01034406 -0.2530606 -0.1793157 -0.04084466 -0.0004789888
#> R -0.2606969  0.01183702  0.2895848  0.2051963  0.04673977  0.0005481212
#>          PC13       PC14        PC15        PC16        PC17        PC18
#> M -0.06138758 -0.1057137  0.02808048  0.05215865 -0.07453265  0.03869042
#> R  0.07024765  0.1209713 -0.03213333 -0.05968671  0.08528994 -0.04427460
#>          PC19         PC20        PC21         PC22
#> M -0.01192247  0.006098658  0.01263492 -0.001224809
#> R  0.01364323 -0.006978877 -0.01445851  0.001401586
\end{lstlisting}

Below the performances of
\href{http://www.rdocumentation.org/packages/MASS/functions/qda.html}{qda}
with and without PCA preprocessing are compared in a
\protect\hyperlink{benchmark-experiments}{benchmark experiment}. Note
that we use stratified resampling to prevent errors in
\href{http://www.rdocumentation.org/packages/MASS/functions/qda.html}{qda}
due to a too small number of observations from either class.

\begin{lstlisting}[language=R]
rin = makeResampleInstance("CV", iters = 3, stratify = TRUE, task = sonar.task)
res = benchmark(list(makeLearner("classif.qda"), lrn), sonar.task, rin, show.info = FALSE)
res
#>         task.id          learner.id mmce.test.mean
#> 1 Sonar-example         classif.qda      0.3941339
#> 2 Sonar-example classif.qda.preproc      0.2643202
\end{lstlisting}

PCA preprocessing in this case turns out to be really beneficial for the
performance of Quadratic Discriminant Analysis.

\paragraph{Joint tuning of preprocessing options and learner
parameters}\label{joint-tuning-of-preprocessing-options-and-learner-parameters}

Let's see if we can optimize this a bit. The threshold value of 0.9
above was chosen arbitrarily and led to 22 out of 60 principal
components. But maybe a lower or higher number of principal components
should be used. Moreover,
\href{http://www.rdocumentation.org/packages/MASS/functions/qda.html}{qda}
has several options that control how the class covariance matrices or
class probabilities are estimated.

Those preprocessing and learner parameters can be
\protect\hyperlink{tuning-hyperparameters}{tuned} jointly. Before doing
this let's first get an overview of all the parameters of the wrapped
learner using function
\href{http://www.rdocumentation.org/packages/mlr/functions/getParamSet.html}{getParamSet}.

\begin{lstlisting}[language=R]
getParamSet(lrn)
#>                      Type len     Def                      Constr Req
#> ppc.BoxCox        logical   -   FALSE                           -   -
#> ppc.YeoJohnson    logical   -   FALSE                           -   -
#> ppc.expoTrans     logical   -   FALSE                           -   -
#> ppc.center        logical   -    TRUE                           -   -
#> ppc.scale         logical   -    TRUE                           -   -
#> ppc.range         logical   -   FALSE                           -   -
#> ppc.knnImpute     logical   -   FALSE                           -   -
#> ppc.bagImpute     logical   -   FALSE                           -   -
#> ppc.medianImpute  logical   -   FALSE                           -   -
#> ppc.pca           logical   -   FALSE                           -   -
#> ppc.ica           logical   -   FALSE                           -   -
#> ppc.spatialSign   logical   -   FALSE                           -   -
#> ppc.thresh        numeric   -    0.95                    0 to Inf   -
#> ppc.pcaComp       integer   -       -                    1 to Inf   -
#> ppc.na.remove     logical   -    TRUE                           -   -
#> ppc.k             integer   -       5                    1 to Inf   -
#> ppc.fudge         numeric   -     0.2                    0 to Inf   -
#> ppc.numUnique     integer   -       3                    1 to Inf   -
#> ppc.n.comp        integer   -       -                    1 to Inf   -
#> method           discrete   -  moment            moment,mle,mve,t   -
#> nu                numeric   -       5                    2 to Inf   Y
#> predict.method   discrete   - plug-in plug-in,predictive,debiased   -
#>                  Tunable Trafo
#> ppc.BoxCox          TRUE     -
#> ppc.YeoJohnson      TRUE     -
#> ppc.expoTrans       TRUE     -
#> ppc.center          TRUE     -
#> ppc.scale           TRUE     -
#> ppc.range           TRUE     -
#> ppc.knnImpute       TRUE     -
#> ppc.bagImpute       TRUE     -
#> ppc.medianImpute    TRUE     -
#> ppc.pca             TRUE     -
#> ppc.ica             TRUE     -
#> ppc.spatialSign     TRUE     -
#> ppc.thresh          TRUE     -
#> ppc.pcaComp         TRUE     -
#> ppc.na.remove       TRUE     -
#> ppc.k               TRUE     -
#> ppc.fudge           TRUE     -
#> ppc.numUnique       TRUE     -
#> ppc.n.comp          TRUE     -
#> method              TRUE     -
#> nu                  TRUE     -
#> predict.method      TRUE     -
\end{lstlisting}

The parameters prefixed by \lstinline!ppc.! belong to preprocessing.
\lstinline!method!, \lstinline!nu! and \lstinline!predict.method! are
\href{http://www.rdocumentation.org/packages/MASS/functions/qda.html}{qda}
parameters.

Instead of tuning the PCA threshold (\lstinline!ppc.thresh!) we tune the
number of principal components (\lstinline!ppc.pcaComp!) directly.
Moreover, for
\href{http://www.rdocumentation.org/packages/MASS/functions/qda.html}{qda}
we try two different ways to estimate the posterior probabilities
(parameter \lstinline!predict.method!): the usual plug-in estimates and
unbiased estimates.

We perform a grid search and set the resolution to 10. This is for
demonstration. You might want to use a finer resolution.

\begin{lstlisting}[language=R]
ps = makeParamSet(
  makeIntegerParam("ppc.pcaComp", lower = 1, upper = getTaskNFeats(sonar.task)),
  makeDiscreteParam("predict.method", values = c("plug-in", "debiased"))
)
ctrl = makeTuneControlGrid(resolution = 10)
res = tuneParams(lrn, sonar.task, rin, par.set = ps, control = ctrl, show.info = FALSE)
res
#> Tune result:
#> Op. pars: ppc.pcaComp=8; predict.method=plug-in
#> mmce.test.mean=0.192

as.data.frame(res$opt.path)[1:3]
#>    ppc.pcaComp predict.method mmce.test.mean
#> 1            1        plug-in      0.4757074
#> 2            8        plug-in      0.1920635
#> 3           14        plug-in      0.2162871
#> 4           21        plug-in      0.2643202
#> 5           27        plug-in      0.2454106
#> 6           34        plug-in      0.2645273
#> 7           40        plug-in      0.2742581
#> 8           47        plug-in      0.3173223
#> 9           53        plug-in      0.3512767
#> 10          60        plug-in      0.3941339
#> 11           1       debiased      0.5336094
#> 12           8       debiased      0.2450656
#> 13          14       debiased      0.2403037
#> 14          21       debiased      0.2546584
#> 15          27       debiased      0.3075224
#> 16          34       debiased      0.3172533
#> 17          40       debiased      0.3125604
#> 18          47       debiased      0.2979986
#> 19          53       debiased      0.3079365
#> 20          60       debiased      0.3654244
\end{lstlisting}

There seems to be a preference for a lower number of principal
components (\textless{}27) for both \lstinline!"plug-in"! and
\lstinline!"debiased"! with \lstinline!"plug-in"! achieving slightly
lower error rates.

\subsubsection{Writing a custom preprocessing
wrapper}\label{writing-a-custom-preprocessing-wrapper}

If the options offered by
\href{http://www.rdocumentation.org/packages/mlr/functions/makePreprocWrapperCaret.html}{makePreprocWrapperCaret}
are not enough, you can write your own preprocessing wrapper using
function
\href{http://www.rdocumentation.org/packages/mlr/functions/makePreprocWrapper.html}{makePreprocWrapper}.

As described in the tutorial section about
\protect\hyperlink{wrapper}{wrapped learners} wrappers are implemented
using a \emph{train} and a \emph{predict} method. In case of
preprocessing wrappers these methods specify how to transform the data
before training and before prediction and are \emph{completely
user-defined}.

Below we show how to create a preprocessing wrapper that centers and
scales the data before training/predicting. Some learning methods as,
e.g., k nearest neighbors, support vector machines or neural networks
usually require scaled features. Many, but not all, have a built-in
scaling option where the training data set is scaled before model
fitting and the test data set is scaled accordingly, that is by using
the scaling parameters from the training stage, before making
predictions. In the following we show how to add a scaling option to a
\href{http://www.rdocumentation.org/packages/mlr/functions/makeLearner.html}{Learner}
by coupling it with function
\href{http://www.rdocumentation.org/packages/base/functions/scale.html}{scale}.

Note that we chose this simple example for demonstration.
Centering/scaling the data is also possible with
\href{http://www.rdocumentation.org/packages/mlr/functions/makePreprocWrapperCaret.html}{makePreprocWrapperCaret}.

\paragraph{Specifying the train
function}\label{specifying-the-train-function}

The \emph{train} function has to be a function with the following
arguments:

\begin{itemize}
\tightlist
\item
  \lstinline!data! is a
  \href{http://www.rdocumentation.org/packages/base/functions/data.frame.html}{data.frame}
  with columns for all features and the target variable.
\item
  \lstinline!target! is a string and denotes the name of the target
  variable in \lstinline!data!.
\item
  \lstinline!args! is a
  \href{http://www.rdocumentation.org/packages/base/functions/list.html}{list}
  of further arguments and parameters that influence the preprocessing.
\end{itemize}

It must return a
\href{http://www.rdocumentation.org/packages/base/functions/list.html}{list}
with elements \lstinline!$data! and \lstinline!$control!, where
\lstinline!$data! is the preprocessed data set and \lstinline!$control!
stores all information required to preprocess the data before
prediction.

The \emph{train} function for the scaling example is given below. It
calls
\href{http://www.rdocumentation.org/packages/base/functions/scale.html}{scale}
on the numerical features and returns the scaled training data and the
corresponding scaling parameters.

\lstinline!args! contains the \lstinline!center! and \lstinline!scale!
arguments of function
\href{http://www.rdocumentation.org/packages/base/functions/scale.html}{scale}
and slot \lstinline!$control! stores the scaling parameters to be used
in the prediction stage.

Regarding the latter note that the \lstinline!center! and
\lstinline!scale! arguments of
\href{http://www.rdocumentation.org/packages/base/functions/scale.html}{scale}
can be either a logical value or a numeric vector of length equal to the
number of the numeric columns in \lstinline!data!, respectively. If a
logical value was passed to \lstinline!args! we store the column means
and standard deviations/ root mean squares in the \lstinline!$center!
and \lstinline!$scale! slots of the returned \lstinline!$control!
object.

\begin{lstlisting}[language=R]
trainfun = function(data, target, args = list(center, scale)) {
  ## Identify numerical features
  cns = colnames(data)
  nums = setdiff(cns[sapply(data, is.numeric)], target)
  ## Extract numerical features from the data set and call scale
  x = as.matrix(data[, nums, drop = FALSE])
  x = scale(x, center = args$center, scale = args$scale)
  ## Store the scaling parameters in control
  ## These are needed to preprocess the data before prediction
  control = args
  if (is.logical(control$center) && control$center)
    control$center = attr(x, "scaled:center")
  if (is.logical(control$scale) && control$scale)
    control$scale = attr(x, "scaled:scale")
  ## Recombine the data
  data = data[, setdiff(cns, nums), drop = FALSE]
  data = cbind(data, as.data.frame(x))
  return(list(data = data, control = control))
}
\end{lstlisting}

\paragraph{Specifying the predict
function}\label{specifying-the-predict-function}

The \emph{predict} function has the following arguments:

\begin{itemize}
\tightlist
\item
  \lstinline!data! is a
  \href{http://www.rdocumentation.org/packages/base/functions/data.frame.html}{data.frame}
  containing \emph{only} feature values (as for prediction the target
  values naturally are not known).
\item
  \lstinline!target! is a string indicating the name of the target
  variable.
\item
  \lstinline!args! are the \lstinline!args! that were passed to the
  \emph{train} function.
\item
  \lstinline!control! is the object returned by the \emph{train}
  function.
\end{itemize}

It returns the preprocessed data.

In our scaling example the \emph{predict} function scales the numerical
features using the parameters from the training stage stored in
\lstinline!control!.

\begin{lstlisting}[language=R]
predictfun = function(data, target, args, control) {
  ## Identify numerical features
  cns = colnames(data)
  nums = cns[sapply(data, is.numeric)]
  ## Extract numerical features from the data set and call scale
  x = as.matrix(data[, nums, drop = FALSE])
  x = scale(x, center = control$center, scale = control$scale)
  ## Recombine the data
  data = data[, setdiff(cns, nums), drop = FALSE]  
  data = cbind(data, as.data.frame(x))
  return(data)
}
\end{lstlisting}

\paragraph{Creating the preprocessing
wrapper}\label{creating-the-preprocessing-wrapper}

Below we create a preprocessing wrapper with a
\href{http://www.rdocumentation.org/packages/nnet/functions/nnet.html}{regression
neural network} (which itself does not have a scaling option) as base
learner.

The \emph{train} and \emph{predict} functions defined above are passed
to
\href{http://www.rdocumentation.org/packages/mlr/functions/makePreprocWrapper.html}{makePreprocWrapper}
via the \lstinline!train! and \lstinline!predict! arguments.
\lstinline!par.vals! is a
\href{http://www.rdocumentation.org/packages/base/functions/list.html}{list}
of parameter values that is relayed to the \lstinline!args! argument of
the \emph{train} function.

\begin{lstlisting}[language=R]
lrn = makeLearner("regr.nnet", trace = FALSE, decay = 1e-02)
lrn = makePreprocWrapper(lrn, train = trainfun, predict = predictfun,
  par.vals = list(center = TRUE, scale = TRUE))
lrn
#> Learner regr.nnet.preproc from package nnet
#> Type: regr
#> Name: ; Short name: 
#> Class: PreprocWrapper
#> Properties: numerics,factors,weights
#> Predict-Type: response
#> Hyperparameters: size=3,trace=FALSE,decay=0.01
\end{lstlisting}

Let's compare the cross-validated mean squared error
(\protect\hyperlink{implemented-performance-measures}{mse}) on the
\href{http://www.rdocumentation.org/packages/mlbench/functions/BostonHousing.html}{Boston
Housing data set} with and without scaling.

\begin{lstlisting}[language=R]
rdesc = makeResampleDesc("CV", iters = 3)

r = resample(lrn, bh.task, resampling = rdesc, show.info = FALSE)
r
#> Resample Result
#> Task: BostonHousing-example
#> Learner: regr.nnet.preproc
#> Aggr perf: mse.test.mean=20.6
#> Runtime: 0.0987003

lrn = makeLearner("regr.nnet", trace = FALSE, decay = 1e-02)
r = resample(lrn, bh.task, resampling = rdesc, show.info = FALSE)
r
#> Resample Result
#> Task: BostonHousing-example
#> Learner: regr.nnet
#> Aggr perf: mse.test.mean=55.1
#> Runtime: 0.0740757
\end{lstlisting}

\paragraph{Joint tuning of preprocessing and learner
parameters}\label{joint-tuning-of-preprocessing-and-learner-parameters}

Often it's not clear which preprocessing options work best with a
certain learning algorithm. As already shown for the number of principal
components in
\href{http://www.rdocumentation.org/packages/mlr/functions/makePreprocWrapperCaret.html}{makePreprocWrapperCaret}
we can \protect\hyperlink{tuning-hyperparameters}{tune} them easily
together with other hyperparameters of the learner.

In our scaling example we can try if
\href{http://www.rdocumentation.org/packages/nnet/functions/nnet.html}{nnet}
works best with both centering and scaling the data or if it's better to
omit one of the two operations or do no preprocessing at all. In order
to tune \lstinline!center! and \lstinline!scale! we have to add
appropriate
\href{http://www.rdocumentation.org/packages/ParamHelpers/functions/LearnerParam.html}{LearnerParam}s
to the
\href{http://www.rdocumentation.org/packages/ParamHelpers/functions/ParamSet.html}{parameter
set} of the wrapped learner.

As mentioned above
\href{http://www.rdocumentation.org/packages/base/functions/scale.html}{scale}
allows for numeric and logical \lstinline!center! and \lstinline!scale!
arguments. As we want to use the latter option we declare
\lstinline!center! and \lstinline!scale! as logical learner parameters.

\begin{lstlisting}[language=R]
lrn = makeLearner("regr.nnet", trace = FALSE)
lrn = makePreprocWrapper(lrn, train = trainfun, predict = predictfun,
  par.set = makeParamSet(
    makeLogicalLearnerParam("center"),
    makeLogicalLearnerParam("scale")
  ),
  par.vals = list(center = TRUE, scale = TRUE))

lrn
#> Learner regr.nnet.preproc from package nnet
#> Type: regr
#> Name: ; Short name: 
#> Class: PreprocWrapper
#> Properties: numerics,factors,weights
#> Predict-Type: response
#> Hyperparameters: size=3,trace=FALSE,center=TRUE,scale=TRUE

getParamSet(lrn)
#>             Type len    Def      Constr Req Tunable Trafo
#> center   logical   -      -           -   -    TRUE     -
#> scale    logical   -      -           -   -    TRUE     -
#> size     integer   -      3    0 to Inf   -    TRUE     -
#> maxit    integer   -    100    1 to Inf   -    TRUE     -
#> linout   logical   -  FALSE           -   Y    TRUE     -
#> entropy  logical   -  FALSE           -   Y    TRUE     -
#> softmax  logical   -  FALSE           -   Y    TRUE     -
#> censored logical   -  FALSE           -   Y    TRUE     -
#> skip     logical   -  FALSE           -   -    TRUE     -
#> rang     numeric   -    0.7 -Inf to Inf   -    TRUE     -
#> decay    numeric   -      0    0 to Inf   -    TRUE     -
#> Hess     logical   -  FALSE           -   -    TRUE     -
#> trace    logical   -   TRUE           -   -   FALSE     -
#> MaxNWts  integer   -   1000    1 to Inf   -    TRUE     -
#> abstoll  numeric   - 0.0001 -Inf to Inf   -    TRUE     -
#> reltoll  numeric   -  1e-08 -Inf to Inf   -    TRUE     -
\end{lstlisting}

Now we do a simple grid search for the \lstinline!decay! parameter of
\href{http://www.rdocumentation.org/packages/nnet/functions/nnet.html}{nnet}
and the \lstinline!center! and \lstinline!scale! parameters.

\begin{lstlisting}[language=R]
rdesc = makeResampleDesc("Holdout")
ps = makeParamSet(
  makeDiscreteParam("decay", c(0, 0.05, 0.1)),
  makeLogicalParam("center"),
  makeLogicalParam("scale")
)
ctrl = makeTuneControlGrid()
res = tuneParams(lrn, bh.task, rdesc, par.set = ps, control = ctrl, show.info = FALSE)

res
#> Tune result:
#> Op. pars: decay=0.05; center=FALSE; scale=TRUE
#> mse.test.mean=14.8

as.data.frame(res$opt.path)
#>    decay center scale mse.test.mean dob eol error.message exec.time
#> 1      0   TRUE  TRUE      49.38128   1  NA          <NA>     0.038
#> 2   0.05   TRUE  TRUE      20.64761   2  NA          <NA>     0.045
#> 3    0.1   TRUE  TRUE      22.42986   3  NA          <NA>     0.050
#> 4      0  FALSE  TRUE      96.25474   4  NA          <NA>     0.022
#> 5   0.05  FALSE  TRUE      14.84306   5  NA          <NA>     0.047
#> 6    0.1  FALSE  TRUE      16.65383   6  NA          <NA>     0.044
#> 7      0   TRUE FALSE      40.51518   7  NA          <NA>     0.044
#> 8   0.05   TRUE FALSE      68.00069   8  NA          <NA>     0.044
#> 9    0.1   TRUE FALSE      55.42210   9  NA          <NA>     0.046
#> 10     0  FALSE FALSE      96.25474  10  NA          <NA>     0.022
#> 11  0.05  FALSE FALSE      56.25758  11  NA          <NA>     0.044
#> 12   0.1  FALSE FALSE      42.85529  12  NA          <NA>     0.045
\end{lstlisting}

\paragraph{Preprocessing wrapper
functions}\label{preprocessing-wrapper-functions}

If you have written a preprocessing wrapper that you might want to use
from time to time it's a good idea to encapsulate it in an own function
as shown below. If you think your preprocessing method is something
others might want to use as well and should be integrated into
\href{http://www.rdocumentation.org/packages/mlr/}{mlr} just
\href{https://github.com/mlr-org/mlr/issues}{contact us}.

\begin{lstlisting}[language=R]
makePreprocWrapperScale = function(learner, center = TRUE, scale = TRUE) {
  trainfun = function(data, target, args = list(center, scale)) {
    cns = colnames(data)
    nums = setdiff(cns[sapply(data, is.numeric)], target)
    x = as.matrix(data[, nums, drop = FALSE])
    x = scale(x, center = args$center, scale = args$scale)
    control = args
    if (is.logical(control$center) && control$center)
      control$center = attr(x, "scaled:center")
    if (is.logical(control$scale) && control$scale)
      control$scale = attr(x, "scaled:scale")
    data = data[, setdiff(cns, nums), drop = FALSE]
    data = cbind(data, as.data.frame(x))
    return(list(data = data, control = control))
  }
  predictfun = function(data, target, args, control) {
    cns = colnames(data)
    nums = cns[sapply(data, is.numeric)]
    x = as.matrix(data[, nums, drop = FALSE])
    x = scale(x, center = control$center, scale = control$scale)
    data = data[, setdiff(cns, nums), drop = FALSE]  
    data = cbind(data, as.data.frame(x))
    return(data)
  }
  makePreprocWrapper(
    learner,
    train = trainfun,
    predict = predictfun,
    par.set = makeParamSet(
      makeLogicalLearnerParam("center"),
      makeLogicalLearnerParam("scale")
    ),
    par.vals = list(center = center, scale = scale)
  )
}

lrn = makePreprocWrapperScale("classif.lda")
train(lrn, iris.task)
#> Model for learner.id=classif.lda.preproc; learner.class=PreprocWrapper
#> Trained on: task.id = iris-example; obs = 150; features = 4
#> Hyperparameters: center=TRUE,scale=TRUE
\end{lstlisting}

\hypertarget{imputation-of-missing-values}{\subsection{Imputation of
Missing Values}\label{imputation-of-missing-values}}

\href{http://www.rdocumentation.org/packages/mlr/}{mlr} provides several
imputation methods which are listed on the help page
\href{http://www.rdocumentation.org/packages/mlr/functions/imputations.html}{imputations}.
These include standard techniques as imputation by a constant value
(like a fixed constant, the mean, median or mode) and random numbers
(either from the empirical distribution of the feature under
consideration or a certain distribution family). Moreover, missing
values in one feature can be replaced based on the other features by
predictions from any supervised
\href{http://www.rdocumentation.org/packages/mlr/functions/makeLearner.html}{Learner}
integrated into \href{http://www.rdocumentation.org/packages/mlr/}{mlr}.

If your favourite option is not implemented in
\href{http://www.rdocumentation.org/packages/mlr/}{mlr} yet, you can
easily \protect\hyperlink{creating-an-imputation-method}{create your own
imputation method}.

Also note that some of the learning algorithms included in
\href{http://www.rdocumentation.org/packages/mlr/}{mlr} can deal with
missing values in a sensible way, i.e., other than simply deleting
observations with missing values. Those
\href{http://www.rdocumentation.org/packages/mlr/functions/makeLearner.html}{Learner}s
have the property \lstinline!"missings"! and thus can be identified
using
\href{http://www.rdocumentation.org/packages/mlr/functions/listLearners.html}{listLearners}.

\begin{lstlisting}[language=R]
### Regression learners that can deal with missing values
listLearners("regr", properties = "missings")[c("class", "package")]
#>                     class         package
#> 1         regr.blackboost    mboost,party
#> 2            regr.cforest           party
#> 3              regr.ctree           party
#> 4             regr.cubist          Cubist
#> 5                regr.gbm             gbm
#> 6    regr.randomForestSRC randomForestSRC
#> 7 regr.randomForestSRCSyn randomForestSRC
#> 8              regr.rpart           rpart
\end{lstlisting}

See also the list of \protect\hyperlink{integrated-learners}{integrated
learners} in the Appendix.

\subsubsection{Imputation and
reimputation}\label{imputation-and-reimputation}

Imputation can be done by function
\href{http://www.rdocumentation.org/packages/mlr/functions/impute.html}{impute}.
You can specify an imputation method for each feature individually or
for classes of features like numerics or factors. Moreover, you can
generate dummy variables that indicate which values are missing, also
either for classes of features or for individual features. These allow
to identify the patterns and reasons for missing data and permit to
treat imputed and observed values differently in a subsequent analysis.

Let's have a look at the
\href{http://www.rdocumentation.org/packages/datasets/functions/airquality.html}{airquality}
data set.

\begin{lstlisting}[language=R]
data(airquality)
summary(airquality)
#>      Ozone           Solar.R           Wind             Temp      
#>  Min.   :  1.00   Min.   :  7.0   Min.   : 1.700   Min.   :56.00  
#>  1st Qu.: 18.00   1st Qu.:115.8   1st Qu.: 7.400   1st Qu.:72.00  
#>  Median : 31.50   Median :205.0   Median : 9.700   Median :79.00  
#>  Mean   : 42.13   Mean   :185.9   Mean   : 9.958   Mean   :77.88  
#>  3rd Qu.: 63.25   3rd Qu.:258.8   3rd Qu.:11.500   3rd Qu.:85.00  
#>  Max.   :168.00   Max.   :334.0   Max.   :20.700   Max.   :97.00  
#>  NA's   :37       NA's   :7                                       
#>      Month            Day      
#>  Min.   :5.000   Min.   : 1.0  
#>  1st Qu.:6.000   1st Qu.: 8.0  
#>  Median :7.000   Median :16.0  
#>  Mean   :6.993   Mean   :15.8  
#>  3rd Qu.:8.000   3rd Qu.:23.0  
#>  Max.   :9.000   Max.   :31.0  
#> 
\end{lstlisting}

There are 37 \lstinline!NA's! in variable \lstinline!Ozone! (ozone
pollution) and 7 \lstinline!NA's! in variable \lstinline!Solar.R! (solar
radiation). For demonstration purposes we insert artificial
\lstinline!NA's! in column \lstinline!Wind! (wind speed) and coerce it
into a
\href{http://www.rdocumentation.org/packages/base/functions/factor.html}{factor}.

\begin{lstlisting}[language=R]
airq = airquality
ind = sample(nrow(airq), 10)
airq$Wind[ind] = NA
airq$Wind = cut(airq$Wind, c(0,8,16,24))
summary(airq)
#>      Ozone           Solar.R           Wind         Temp      
#>  Min.   :  1.00   Min.   :  7.0   (0,8]  :51   Min.   :56.00  
#>  1st Qu.: 18.00   1st Qu.:115.8   (8,16] :86   1st Qu.:72.00  
#>  Median : 31.50   Median :205.0   (16,24]: 6   Median :79.00  
#>  Mean   : 42.13   Mean   :185.9   NA's   :10   Mean   :77.88  
#>  3rd Qu.: 63.25   3rd Qu.:258.8                3rd Qu.:85.00  
#>  Max.   :168.00   Max.   :334.0                Max.   :97.00  
#>  NA's   :37       NA's   :7                                   
#>      Month            Day      
#>  Min.   :5.000   Min.   : 1.0  
#>  1st Qu.:6.000   1st Qu.: 8.0  
#>  Median :7.000   Median :16.0  
#>  Mean   :6.993   Mean   :15.8  
#>  3rd Qu.:8.000   3rd Qu.:23.0  
#>  Max.   :9.000   Max.   :31.0  
#> 
\end{lstlisting}

If you want to impute \lstinline!NA's! in all integer features (these
include \lstinline!Ozone! and \lstinline!Solar.R!) by the mean, in all
factor features (\lstinline!Wind!) by the mode and additionally generate
dummy variables for all integer features, you can do this as follows:

\begin{lstlisting}[language=R]
imp = impute(airq, classes = list(integer = imputeMean(), factor = imputeMode()),
  dummy.classes = "integer")
\end{lstlisting}

\href{http://www.rdocumentation.org/packages/mlr/functions/impute.html}{impute}
returns a
\href{http://www.rdocumentation.org/packages/base/functions/list.html}{list}
where slot \lstinline!$data! contains the imputed data set. Per default,
the dummy variables are factors with levels \lstinline!"TRUE"! and
\lstinline!"FALSE"!. It is also possible to create numeric zero-one
indicator variables.

\begin{lstlisting}[language=R]
head(imp$data, 10)
#>       Ozone  Solar.R    Wind Temp Month Day Ozone.dummy Solar.R.dummy
#> 1  41.00000 190.0000   (0,8]   67     5   1       FALSE         FALSE
#> 2  36.00000 118.0000   (0,8]   72     5   2       FALSE         FALSE
#> 3  12.00000 149.0000  (8,16]   74     5   3       FALSE         FALSE
#> 4  18.00000 313.0000  (8,16]   62     5   4       FALSE         FALSE
#> 5  42.12931 185.9315  (8,16]   56     5   5        TRUE          TRUE
#> 6  28.00000 185.9315  (8,16]   66     5   6       FALSE          TRUE
#> 7  23.00000 299.0000  (8,16]   65     5   7       FALSE         FALSE
#> 8  19.00000  99.0000  (8,16]   59     5   8       FALSE         FALSE
#> 9   8.00000  19.0000 (16,24]   61     5   9       FALSE         FALSE
#> 10 42.12931 194.0000  (8,16]   69     5  10        TRUE         FALSE
\end{lstlisting}

Slot \lstinline!$desc! is an
\href{http://www.rdocumentation.org/packages/mlr/functions/impute.html}{ImputationDesc}
object that stores all relevant information about the imputation. For
the current example this includes the means and the mode computed on the
non-missing data.

\begin{lstlisting}[language=R]
imp$desc
#> Imputation description
#> Target: 
#> Features: 6; Imputed: 6
#> impute.new.levels: TRUE
#> recode.factor.levels: TRUE
#> dummy.type: factor
\end{lstlisting}

The imputation description shows the name of the target variable (not
present), the number of features and the number of imputed features.
Note that the latter number refers to the features for which an
imputation method was specified (five integers plus one factor) and not
to the features actually containing \lstinline!NA's!.
\lstinline!dummy.type! indicates that the dummy variables are factors.
For details on \lstinline!impute.new.levels! and
\lstinline!recode.factor.levels! see the help page of function
\href{http://www.rdocumentation.org/packages/mlr/functions/impute.html}{impute}.

Let's have a look at another example involving a target variable. A
possible learning task associated with the
\href{http://www.rdocumentation.org/packages/datasets/functions/airquality.html}{airquality}
data is to predict the ozone pollution based on the meteorological
features. Since we do not want to use columns \lstinline!Day! and
\lstinline!Month! we remove them.

\begin{lstlisting}[language=R]
airq = subset(airq, select = 1:4)
\end{lstlisting}

The first 100 observations are used as training data set.

\begin{lstlisting}[language=R]
airq.train = airq[1:100,]
airq.test = airq[-c(1:100),]
\end{lstlisting}

In case of a supervised learning problem you need to pass the name of
the target variable to
\href{http://www.rdocumentation.org/packages/mlr/functions/impute.html}{impute}.
This prevents imputation and creation of a dummy variable for the target
variable itself and makes sure that the target variable is not used to
impute the features.

In contrast to the example above we specify imputation methods for
individual features instead of classes of features.

Missing values in \lstinline!Solar.R! are imputed by random numbers
drawn from the empirical distribution of the non-missing observations.

Function
\href{http://www.rdocumentation.org/packages/mlr/functions/imputations.html}{imputeLearner}
allows to use all supervised learning algorithms integrated into
\href{http://www.rdocumentation.org/packages/mlr/}{mlr} for imputation.
The type of the
\href{http://www.rdocumentation.org/packages/mlr/functions/makeLearner.html}{Learner}
(\lstinline!regr!, \lstinline!classif!) must correspond to the class of
the feature to be imputed. The missing values in \lstinline!Wind! are
replaced by the predictions of a classification tree
(\href{http://www.rdocumentation.org/packages/rpart/functions/rpart.html}{rpart}).
Per default, all available columns in \lstinline!airq.train! except the
target variable (\lstinline!Ozone!) and the variable to be imputed
(\lstinline!Wind!) are used as features in the classification tree, here
\lstinline!Solar.R! and \lstinline!Temp!. You can also select manually
which columns to use. Note that
\href{http://www.rdocumentation.org/packages/rpart/functions/rpart.html}{rpart}
can deal with missing feature values, therefore the \lstinline!NA's! in
column \lstinline!Solar.R! do not pose a problem.

\begin{lstlisting}[language=R]
imp = impute(airq.train, target = "Ozone", cols = list(Solar.R = imputeHist(),
  Wind = imputeLearner("classif.rpart")), dummy.cols = c("Solar.R", "Wind"))
summary(imp$data)
#>      Ozone           Solar.R            Wind         Temp      
#>  Min.   :  1.00   Min.   :  7.00   (0,8]  :34   Min.   :56.00  
#>  1st Qu.: 16.00   1st Qu.: 98.75   (8,16] :61   1st Qu.:69.00  
#>  Median : 34.00   Median :221.50   (16,24]: 5   Median :79.50  
#>  Mean   : 41.59   Mean   :191.54                Mean   :76.87  
#>  3rd Qu.: 63.00   3rd Qu.:274.25                3rd Qu.:84.00  
#>  Max.   :135.00   Max.   :334.00                Max.   :93.00  
#>  NA's   :31                                                    
#>  Solar.R.dummy Wind.dummy
#>  FALSE:93      FALSE:92  
#>  TRUE : 7      TRUE : 8  
#>                          
#>                          
#>                          
#>                          
#> 

imp$desc
#> Imputation description
#> Target: Ozone
#> Features: 3; Imputed: 2
#> impute.new.levels: TRUE
#> recode.factor.levels: TRUE
#> dummy.type: factor
\end{lstlisting}

The
\href{http://www.rdocumentation.org/packages/mlr/functions/impute.html}{ImputationDesc}
object can be used by function
\href{http://www.rdocumentation.org/packages/mlr/functions/reimpute.html}{reimpute}
to impute the test data set the same way as the training data.

\begin{lstlisting}[language=R]
airq.test.imp = reimpute(airq.test, imp$desc)
head(airq.test.imp)
#>   Ozone Solar.R   Wind Temp Solar.R.dummy Wind.dummy
#> 1   110     207  (0,8]   90         FALSE      FALSE
#> 2    NA     222 (8,16]   92         FALSE      FALSE
#> 3    NA     137 (8,16]   86         FALSE      FALSE
#> 4    44     192 (8,16]   86         FALSE      FALSE
#> 5    28     273 (8,16]   82         FALSE      FALSE
#> 6    65     157 (8,16]   80         FALSE      FALSE
\end{lstlisting}

Especially when evaluating a machine learning method by some resampling
technique you might want that
\href{http://www.rdocumentation.org/packages/mlr/functions/impute.html}{impute}/\href{http://www.rdocumentation.org/packages/mlr/functions/reimpute.html}{reimpute}
are called automatically each time before training/prediction. This can
be achieved by creating an imputation wrapper.

\subsubsection{Fusing a learner with
imputation}\label{fusing-a-learner-with-imputation}

You can couple a
\href{http://www.rdocumentation.org/packages/mlr/functions/makeLearner.html}{Learner}
with imputation by function
\href{http://www.rdocumentation.org/packages/mlr/functions/makeImputeWrapper.html}{makeImputeWrapper}
which basically has the same formal arguments as
\href{http://www.rdocumentation.org/packages/mlr/functions/impute.html}{impute}.
Like in the example above we impute \lstinline!Solar.R! by random
numbers from its empirical distribution, \lstinline!Wind! by the
predictions of a classification tree and generate dummy variables for
both features.

\begin{lstlisting}[language=R]
lrn = makeImputeWrapper("regr.lm", cols = list(Solar.R = imputeHist(),
  Wind = imputeLearner("classif.rpart")), dummy.cols = c("Solar.R", "Wind"))
lrn
#> Learner regr.lm.imputed from package stats
#> Type: regr
#> Name: ; Short name: 
#> Class: ImputeWrapper
#> Properties: numerics,factors,se,weights,missings
#> Predict-Type: response
#> Hyperparameters:
\end{lstlisting}

Before training the resulting
\href{http://www.rdocumentation.org/packages/mlr/functions/makeLearner.html}{Learner},
\href{http://www.rdocumentation.org/packages/mlr/functions/impute.html}{impute}
is applied to the training set. Before prediction
\href{http://www.rdocumentation.org/packages/mlr/functions/reimpute.html}{reimpute}
is called on the test set and the
\href{http://www.rdocumentation.org/packages/mlr/functions/impute.html}{ImputationDesc}
object from the training stage.

We again aim to predict the ozone pollution from the meteorological
variables. In order to create the
\href{http://www.rdocumentation.org/packages/mlr/functions/Task.html}{Task}
we need to delete observations with missing values in the target
variable.

\begin{lstlisting}[language=R]
airq = subset(airq, subset = !is.na(airq$Ozone))
task = makeRegrTask(data = airq, target = "Ozone")
\end{lstlisting}

In the following the 3-fold cross-validated
\protect\hyperlink{implemented-performance-measures}{mean squared error}
is calculated.

\begin{lstlisting}[language=R]
rdesc = makeResampleDesc("CV", iters = 3)
r = resample(lrn, task, resampling = rdesc, show.info = FALSE, models = TRUE)
r$aggr
#> mse.test.mean 
#>      524.3392
\end{lstlisting}

\begin{lstlisting}[language=R]
lapply(r$models, getLearnerModel, more.unwrap = TRUE)
#> [[1]]
#> 
#> Call:
#> stats::lm(formula = f, data = d)
#> 
#> Coefficients:
#>       (Intercept)            Solar.R         Wind(8,16]  
#>         -117.0954             0.0853           -27.6763  
#>       Wind(16,24]               Temp  Solar.R.dummyTRUE  
#>           -9.0988             2.0505           -27.4152  
#>    Wind.dummyTRUE  
#>            2.2535  
#> 
#> 
#> [[2]]
#> 
#> Call:
#> stats::lm(formula = f, data = d)
#> 
#> Coefficients:
#>       (Intercept)            Solar.R         Wind(8,16]  
#>         -94.84542            0.03936          -16.26255  
#>       Wind(16,24]               Temp  Solar.R.dummyTRUE  
#>          -7.00707            1.79513          -11.08578  
#>    Wind.dummyTRUE  
#>          -0.68340  
#> 
#> 
#> [[3]]
#> 
#> Call:
#> stats::lm(formula = f, data = d)
#> 
#> Coefficients:
#>       (Intercept)            Solar.R         Wind(8,16]  
#>         -57.30438            0.07426          -30.70737  
#>       Wind(16,24]               Temp  Solar.R.dummyTRUE  
#>         -18.25055            1.35898           -2.16654  
#>    Wind.dummyTRUE  
#>          -5.56400
\end{lstlisting}

A second possibility to fuse a learner with imputation is provided by
\href{http://www.rdocumentation.org/packages/mlr/functions/makePreprocWrapperCaret.html}{makePreprocWrapperCaret},
which is an interface to
\href{http://www.rdocumentation.org/packages/caret/}{caret}'s
\href{http://www.rdocumentation.org/packages/caret/functions/preProcess.html}{preProcess}
function.
\href{http://www.rdocumentation.org/packages/caret/functions/preProcess.html}{preProcess}
only works for numeric features and offers imputation by k-nearest
neighbors, bagged trees, and by the median.

\hypertarget{generic-bagging}{\subsection{Generic
Bagging}\label{generic-bagging}}

One reason why random forests perform so well is that they are using
bagging as a technique to gain more stability. But why do you want to
limit yourself to the classifiers already implemented in well known
random forests when it is really easy to build your own with
\href{http://www.rdocumentation.org/packages/mlr/}{mlr}?

Just bag an \href{http://www.rdocumentation.org/packages/mlr/}{mlr}
learner already
\href{http://www.rdocumentation.org/packages/mlr/functions/makeBaggingWrapper.html}{makeBaggingWrapper}.

As in a random forest, we need a
\href{http://www.rdocumentation.org/packages/mlr/functions/makeLearner.html}{Learner}
which is trained on a subset of the data during each iteration of the
bagging process. The subsets are chosen according to the parameters
given to
\href{http://www.rdocumentation.org/packages/mlr/functions/makeBaggingWrapper.html}{makeBaggingWrapper}:

\begin{itemize}
\tightlist
\item
  \lstinline!bw.iters! On how many subsets (samples) do we want to train
  our
  \href{http://www.rdocumentation.org/packages/mlr/functions/makeLearner.html}{Learner}?
\item
  \lstinline!bw.replace! Sample with replacement (also known as
  \emph{bootstrapping})?
\item
  \lstinline!bw.size! Percentage size of the samples. If
  \lstinline!bw.replace = TRUE!, \lstinline!bw.size = 1! is the default.
  This does not mean that one sample will contain all the observations
  as observations will occur multiple times in each sample.
\item
  \lstinline!bw.feats! Percentage size of randomly selected features for
  each iteration.
\end{itemize}

Of course we also need a
\href{http://www.rdocumentation.org/packages/mlr/functions/makeLearner.html}{Learner}
which we have to pass to
\href{http://www.rdocumentation.org/packages/mlr/functions/makeBaggingWrapper.html}{makeBaggingWrapper}.

\begin{lstlisting}[language=R]
lrn = makeLearner("classif.rpart")
bag.lrn = makeBaggingWrapper(lrn, bw.iters = 50, bw.replace = TRUE, bw.size = 0.8, bw.feats = 3/4)
\end{lstlisting}

Now we can compare the performance with and without bagging. First let's
try it without bagging:

\begin{lstlisting}[language=R]
rdesc = makeResampleDesc("CV", iters = 10)
r = resample(learner = lrn, task = sonar.task, resampling = rdesc, show.info = FALSE)
r$aggr
#> mmce.test.mean 
#>      0.2735714
\end{lstlisting}

And now with bagging:

\begin{lstlisting}[language=R]
rdesc = makeResampleDesc("CV", iters = 10)
result = resample(learner = bag.lrn, task = sonar.task, resampling = rdesc, show.info = FALSE)
result$aggr
#> mmce.test.mean 
#>      0.2069048
\end{lstlisting}

Training more learners takes more time, but can outperform pure learners
on noisy data with many features.

\subsubsection{Changing the type of
prediction}\label{changing-the-type-of-prediction}

In case of a \emph{classification} problem the predicted class labels
are determined by majority voting over the predictions of the individual
models. Additionally, posterior probabilities can be estimated as the
relative proportions of the predicted class labels. For this purpose you
have to change the predict type of the \emph{bagging learner} as
follows.

\begin{lstlisting}[language=R]
bag.lrn = setPredictType(bag.lrn, predict.type = "prob")
\end{lstlisting}

Note that it is not relevant if the \emph{base learner} itself can
predict probabilities and that for this reason the predict type of the
\emph{base learner} always has to be \lstinline!"response"!.

For \emph{regression} the mean value across predictions is computed.
Moreover, the standard deviation across predictions is estimated if the
predict type of the bagging learner is changed to \lstinline!"se"!.
Below, we give a small example for regression.

\begin{lstlisting}[language=R]
n = getTaskSize(bh.task)
train.inds = seq(1, n, 3)
test.inds  = setdiff(1:n, train.inds)
lrn = makeLearner("regr.rpart")
bag.lrn = makeBaggingWrapper(lrn)
bag.lrn = setPredictType(bag.lrn, predict.type = "se")
mod = train(learner = bag.lrn, task = bh.task, subset = train.inds)
\end{lstlisting}

With function
\href{http://www.rdocumentation.org/packages/mlr/functions/getLearnerModel.html}{getLearnerModel},
you can access the models fitted in the individual iterations.

\begin{lstlisting}[language=R]
head(getLearnerModel(mod), 2)
#> [[1]]
#> Model for learner.id=regr.rpart; learner.class=regr.rpart
#> Trained on: task.id = BostonHousing-example; obs = 169; features = 13
#> Hyperparameters: xval=0
#> 
#> [[2]]
#> Model for learner.id=regr.rpart; learner.class=regr.rpart
#> Trained on: task.id = BostonHousing-example; obs = 169; features = 13
#> Hyperparameters: xval=0
\end{lstlisting}

Predict the response and calculate the standard deviation:

\begin{lstlisting}[language=R]
pred = predict(mod, task = bh.task, subset = test.inds)
head(as.data.frame(pred))
#>   id truth response        se
#> 2  2  21.6 21.98377 1.2516733
#> 3  3  34.7 32.85076 1.0323484
#> 5  5  36.2 32.85076 1.0323484
#> 6  6  28.7 24.18223 0.6522748
#> 8  8  27.1 16.15829 2.7856383
#> 9  9  16.5 14.42388 3.7821479
\end{lstlisting}

In the column labelled \lstinline!se! the standard deviation for each
prediction is given.

Let's visualise this a bit using
\href{http://www.rdocumentation.org/packages/ggplot2/}{ggplot2}. Here we
plot the percentage of lower status of the population
(\lstinline!lstat!) against the prediction.

\begin{lstlisting}[language=R]
library("ggplot2")
library("reshape2")
data = cbind(as.data.frame(pred), getTaskData(bh.task, subset = test.inds))
g = ggplot(data, aes(x = lstat, y = response, ymin = response-se, ymax = response+se, col = age))
g + geom_point() + geom_linerange(alpha=0.5)
\end{lstlisting}

\includegraphics{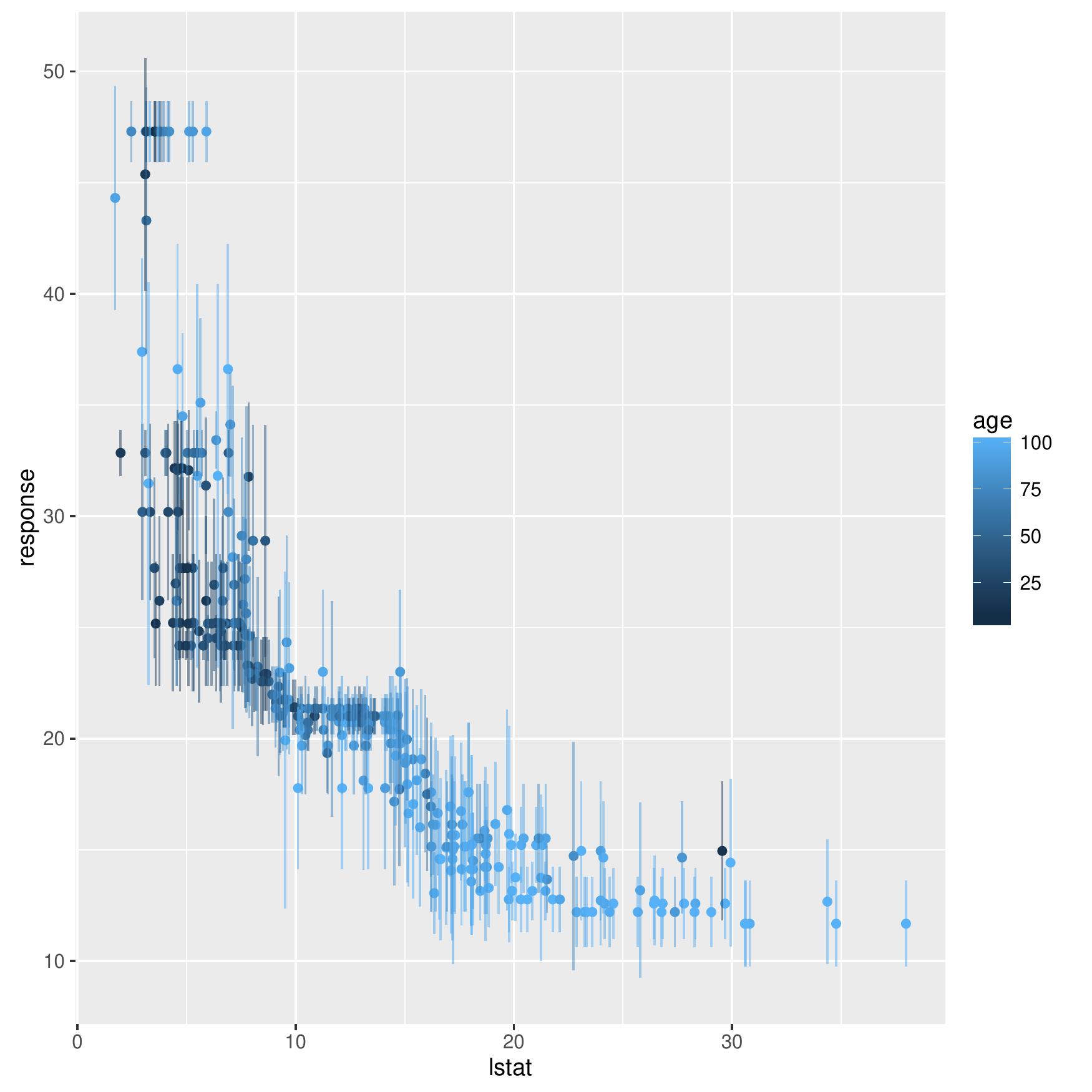}

\hypertarget{advanced-tuning}{\subsection{Advanced
Tuning}\label{advanced-tuning}}

\subsubsection{Iterated F-Racing for mixed spaces and
dependencies}\label{iterated-f-racing-for-mixed-spaces-and-dependencies}

The package supports a larger number of tuning algorithms, which can all
be looked up and selected via
\href{http://www.rdocumentation.org/packages/mlr/functions/TuneControl.html}{TuneControl}.
One of the cooler algorithms is iterated F-racing from the
\href{http://www.rdocumentation.org/packages/irace/}{irace} package
(technical description
\href{http://iridia.ulb.ac.be/IridiaTrSeries/link/IridiaTr2011-004.pdf}{here}).
This not only works for arbitrary parameter types (numeric, integer,
discrete, logical), but also for so-called dependent / hierarchical
parameters:

\begin{lstlisting}[language=R]
ps = makeParamSet(
  makeNumericParam("C", lower = -12, upper = 12, trafo = function(x) 2^x),
  makeDiscreteParam("kernel", values = c("vanilladot", "polydot", "rbfdot")),
  makeNumericParam("sigma", lower = -12, upper = 12, trafo = function(x) 2^x,
    requires = quote(kernel == "rbfdot")),
  makeIntegerParam("degree", lower = 2L, upper = 5L,
    requires = quote(kernel == "polydot"))
)
ctrl = makeTuneControlIrace(maxExperiments = 200L)
rdesc = makeResampleDesc("Holdout")
res = tuneParams("classif.ksvm", iris.task, rdesc, par.set = ps, control = ctrl, show.info = FALSE)
print(head(as.data.frame(res$opt.path)))
#>            C     kernel    sigma degree mmce.test.mean dob eol
#> 1   8.838138     rbfdot 3.947862     NA           0.18   1  NA
#> 2  -5.598352 vanilladot       NA     NA           0.12   1  NA
#> 3  -7.488611 vanilladot       NA     NA           0.36   1  NA
#> 4   4.267949    polydot       NA      3           0.08   1  NA
#> 5 -10.079158    polydot       NA      5           0.06   1  NA
#> 6 -10.643475 vanilladot       NA     NA           0.36   1  NA
#>   error.message exec.time
#> 1          <NA>     0.052
#> 2          <NA>     0.024
#> 3          <NA>     0.020
#> 4          <NA>     0.022
#> 5          <NA>     0.021
#> 6          <NA>     0.021
\end{lstlisting}

See how we made the kernel parameters like \lstinline!sigma! and
\lstinline!degree! dependent on the \lstinline!kernel! selection
parameters? This approach allows you to tune parameters of multiple
kernels at once, efficiently concentrating on the ones which work best
for your given data set.

\subsubsection{Tuning across whole model spaces with
ModelMultiplexer}\label{tuning-across-whole-model-spaces-with-modelmultiplexer}

We can now take the following example even one step further. If we use
the
\href{http://www.rdocumentation.org/packages/mlr/functions/makeModelMultiplexer.html}{ModelMultiplexer}
we can tune over different model classes at once, just as we did with
the SVM kernels above.

\begin{lstlisting}[language=R]
base.learners = list(
  makeLearner("classif.ksvm"),
  makeLearner("classif.randomForest")
)
lrn = makeModelMultiplexer(base.learners)
\end{lstlisting}

Function
\href{http://www.rdocumentation.org/packages/mlr/functions/makeModelMultiplexerParamSet.html}{makeModelMultiplexerParamSet}
offers a simple way to construct a parameter set for tuning: The
parameter names are prefixed automatically and the \lstinline!requires!
element is set, too, to make all parameters subordinate to
\lstinline!selected.learner!.

\begin{lstlisting}[language=R]
ps = makeModelMultiplexerParamSet(lrn,
  makeNumericParam("sigma", lower = -12, upper = 12, trafo = function(x) 2^x),
  makeIntegerParam("ntree", lower = 1L, upper = 500L)
)
print(ps)
#>                                Type len Def
#> selected.learner           discrete   -   -
#> classif.ksvm.sigma          numeric   -   -
#> classif.randomForest.ntree  integer   -   -
#>                                                       Constr Req Tunable
#> selected.learner           classif.ksvm,classif.randomForest   -    TRUE
#> classif.ksvm.sigma                                 -12 to 12   Y    TRUE
#> classif.randomForest.ntree                          1 to 500   Y    TRUE
#>                            Trafo
#> selected.learner               -
#> classif.ksvm.sigma             Y
#> classif.randomForest.ntree     -

rdesc = makeResampleDesc("CV", iters = 2L)
ctrl = makeTuneControlIrace(maxExperiments = 200L)
res = tuneParams(lrn, iris.task, rdesc, par.set = ps, control = ctrl, show.info = FALSE)
print(head(as.data.frame(res$opt.path)))
#>       selected.learner classif.ksvm.sigma classif.randomForest.ntree
#> 1         classif.ksvm           8.511120                         NA
#> 2         classif.ksvm           2.601238                         NA
#> 3 classif.randomForest                 NA                        435
#> 4 classif.randomForest                 NA                         18
#> 5         classif.ksvm          -1.884101                         NA
#> 6         classif.ksvm           4.388728                         NA
#>   mmce.test.mean dob eol error.message exec.time
#> 1      0.6466667   1  NA          <NA>     0.041
#> 2      0.1400000   1  NA          <NA>     0.038
#> 3      0.0400000   1  NA          <NA>     0.057
#> 4      0.0400000   1  NA          <NA>     0.031
#> 5      0.0400000   1  NA          <NA>     0.039
#> 6      0.3333333   1  NA          <NA>     0.041
\end{lstlisting}

\subsubsection{Multi-criteria evaluation and
optimization}\label{multi-criteria-evaluation-and-optimization}

During tuning you might want to optimize multiple, potentially
conflicting, performance measures simultaneously.

In the following example we aim to minimize both, the false positive and
the false negative rates
(\protect\hyperlink{implemented-performance-measures}{fpr} and
\protect\hyperlink{implemented-performance-measures}{fnr}). We again
tune the hyperparameters of an SVM (function
\href{http://www.rdocumentation.org/packages/kernlab/functions/ksvm.html}{ksvm})
with a radial basis kernel and use the
\href{http://www.rdocumentation.org/packages/mlr/functions/sonar.task.html}{sonar
classification task} for illustration. As search strategy we choose a
random search.

For all available multi-criteria tuning algorithms see
\href{http://www.rdocumentation.org/packages/mlr/functions/TuneMultiCritControl.html}{TuneMultiCritControl}.

\begin{lstlisting}[language=R]
ps = makeParamSet(
  makeNumericParam("C", lower = -12, upper = 12, trafo = function(x) 2^x),
  makeNumericParam("sigma", lower = -12, upper = 12, trafo = function(x) 2^x)
)
ctrl = makeTuneMultiCritControlRandom(maxit = 30L)
rdesc = makeResampleDesc("Holdout")
res = tuneParamsMultiCrit("classif.ksvm", task = sonar.task, resampling = rdesc, par.set = ps,
  measures = list(fpr, fnr), control = ctrl, show.info = FALSE)
res
#> Tune multicrit result:
#> Points on front: 3

head(as.data.frame(trafoOptPath(res$opt.path)))
#>              C        sigma fpr.test.mean fnr.test.mean dob eol
#> 1 6.731935e+01 1.324673e+03     1.0000000     0.0000000   1  NA
#> 2 4.719282e-02 7.660068e-04     1.0000000     0.0000000   2  NA
#> 3 7.004097e+00 1.211249e+01     1.0000000     0.0000000   3  NA
#> 4 1.207932e+00 6.096186e+00     1.0000000     0.0000000   4  NA
#> 5 5.203364e+00 2.781734e-03     0.1515152     0.1621622   5  NA
#> 6 5.638243e-04 7.956946e+02     1.0000000     0.0000000   6  NA
#>   error.message exec.time
#> 1          <NA>     0.034
#> 2          <NA>     0.034
#> 3          <NA>     0.035
#> 4          <NA>     0.035
#> 5          <NA>     0.033
#> 6          <NA>     0.036
\end{lstlisting}

The results can be visualized with function
\href{http://www.rdocumentation.org/packages/mlr/functions/plotTuneMultiCritResult.html}{plotTuneMultiCritResult}.
The plot shows the false positive and false negative rates for all
parameter settings evaluated during tuning. Points on the Pareto front
are slightly increased.

\begin{lstlisting}[language=R]
plotTuneMultiCritResult(res)
\end{lstlisting}

\includegraphics{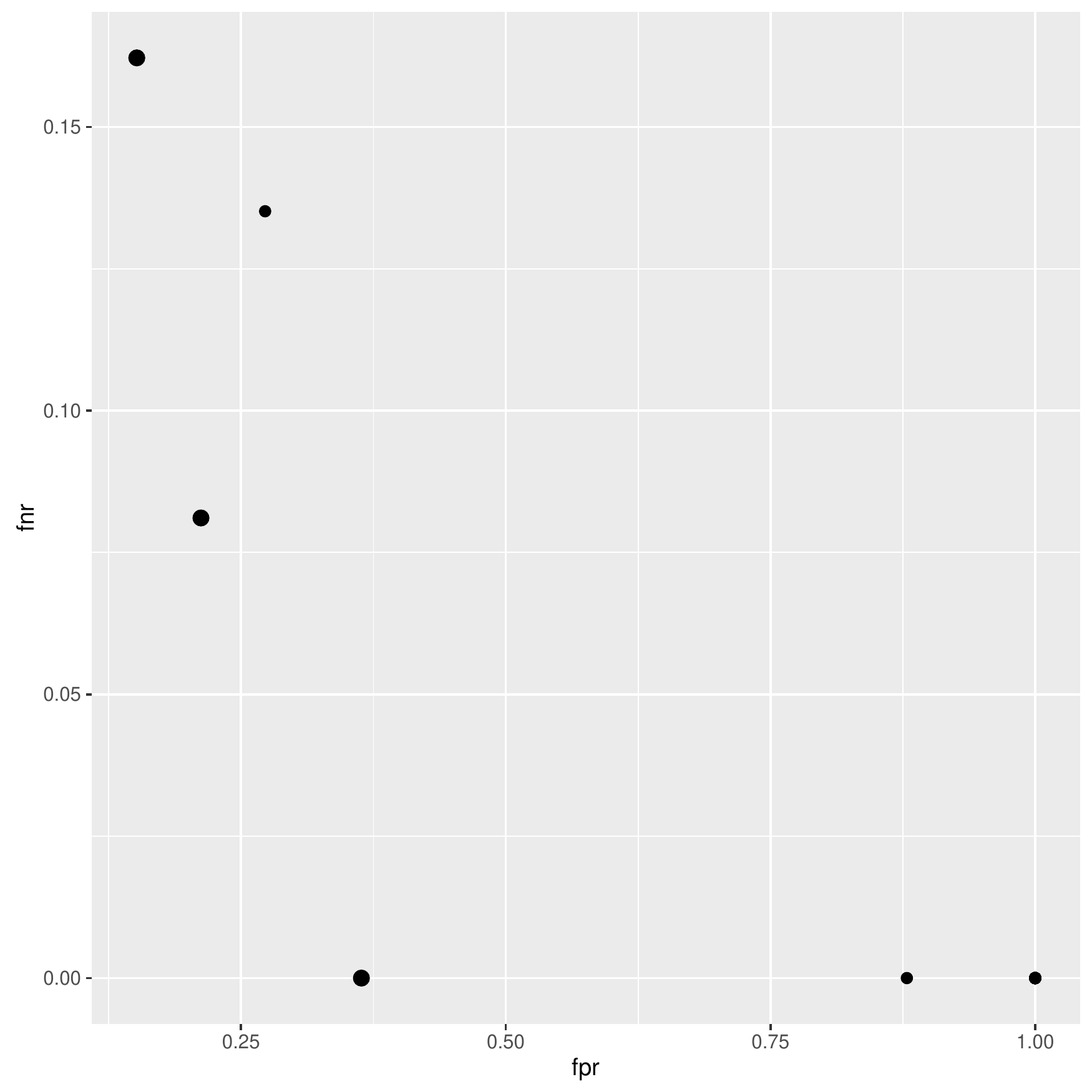}

\hypertarget{feature-selection}{\subsection{Feature
Selection}\label{feature-selection}}

Often, data sets include a large number of features. The technique of
extracting a subset of relevant features is called feature selection.
Feature selection can enhance the interpretability of the model, speed
up the learning process and improve the learner performance. There exist
different approaches to identify the relevant features.
\href{http://www.rdocumentation.org/packages/mlr/}{mlr} supports
\emph{filter} and \emph{wrapper methods}.

\subsubsection{Filter methods}\label{filter-methods}

Filter methods assign an importance value to each feature. Based on
these values the features can be ranked and a feature subset can be
selected.

\paragraph{Calculating the feature
importance}\label{calculating-the-feature-importance}

Different methods for calculating the feature importance are built into
\href{http://www.rdocumentation.org/packages/mlr/}{mlr}'s function
\href{http://www.rdocumentation.org/packages/mlr/functions/generateFilterValuesData.html}{generateFilterValuesData}
(\href{http://www.rdocumentation.org/packages/mlr/functions/getFilterValues.html}{getFilterValues}
has been deprecated in favor of
\href{http://www.rdocumentation.org/packages/mlr/functions/generateFilterValuesData.html}{generateFilterValuesData}.).
Currently, classification, regression and survival analysis tasks are
supported. A table showing all available methods can be found
\protect\hyperlink{integrated-filter-methods}{here}.

Function
\href{http://www.rdocumentation.org/packages/mlr/functions/generateFilterValuesData.html}{generateFilterValuesData}
requires the
\href{http://www.rdocumentation.org/packages/mlr/functions/Task.html}{Task}
and a character string specifying the filter method.

\begin{lstlisting}[language=R]
fv = generateFilterValuesData(iris.task, method = "information.gain")
fv
#> FilterValues:
#> Task: iris-example
#>           name    type information.gain
#> 1 Sepal.Length numeric        0.4521286
#> 2  Sepal.Width numeric        0.2672750
#> 3 Petal.Length numeric        0.9402853
#> 4  Petal.Width numeric        0.9554360
\end{lstlisting}

\lstinline!fv! is a
\href{http://www.rdocumentation.org/packages/mlr/functions/FilterValues.html}{FilterValues}
object and \lstinline!fv$data! contains a
\href{http://www.rdocumentation.org/packages/base/functions/data.frame.html}{data.frame}
that gives the importance values for all features. Optionally, a vector
of filter methods can be passed.

\begin{lstlisting}[language=R]
fv2 = generateFilterValuesData(iris.task, method = c("information.gain", "chi.squared"))
fv2$data
#>           name    type information.gain chi.squared
#> 1 Sepal.Length numeric        0.4521286   0.6288067
#> 2  Sepal.Width numeric        0.2672750   0.4922162
#> 3 Petal.Length numeric        0.9402853   0.9346311
#> 4  Petal.Width numeric        0.9554360   0.9432359
\end{lstlisting}

A bar plot of importance values for the individual features can be
obtained using function
\href{http://www.rdocumentation.org/packages/mlr/functions/plotFilterValues.html}{plotFilterValues}.

\begin{lstlisting}[language=R]
plotFilterValues(fv2)
\end{lstlisting}

\includegraphics{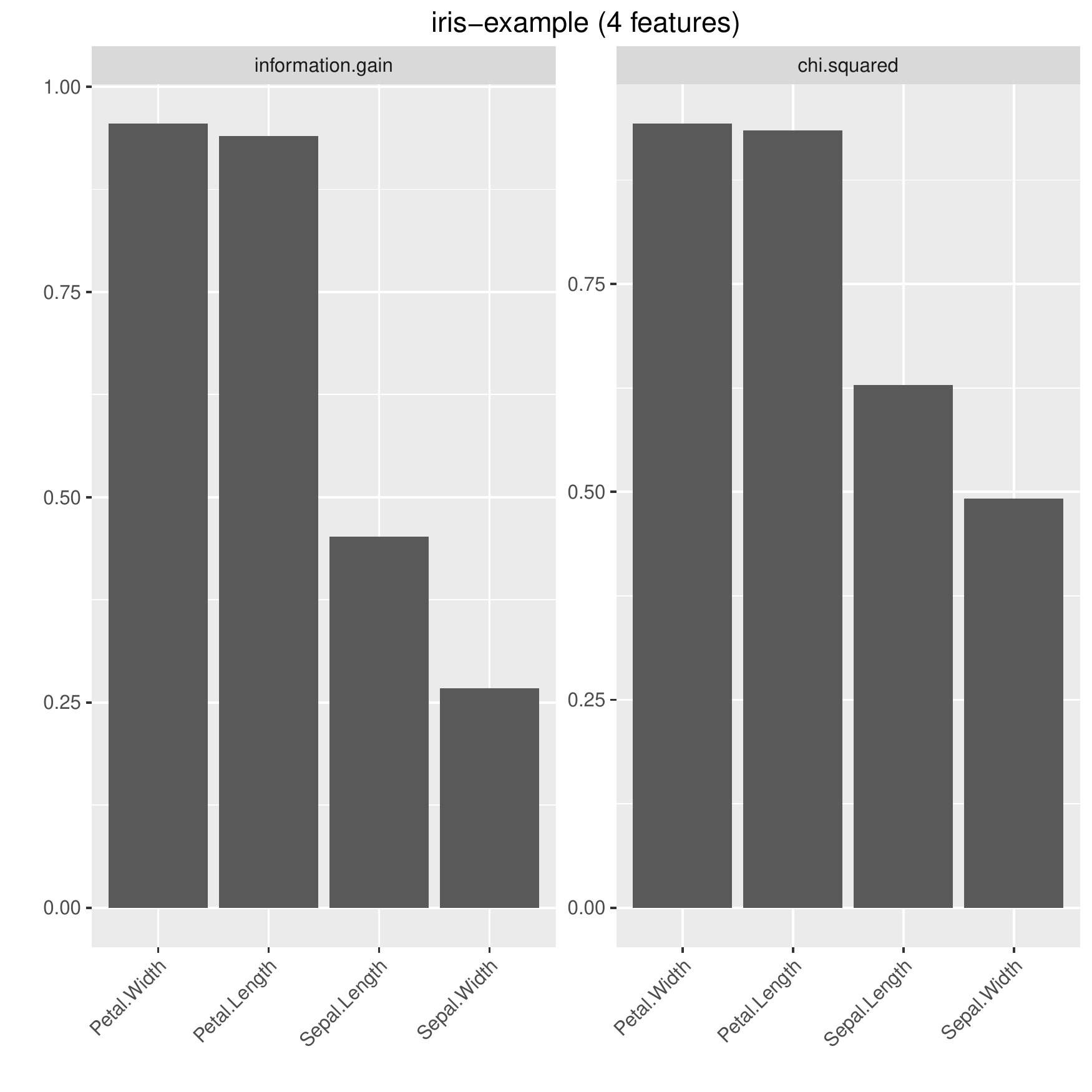}

By default
\href{http://www.rdocumentation.org/packages/mlr/functions/plotFilterValues.html}{plotFilterValues}
will create facetted subplots if multiple filter methods are passed as
input to
\href{http://www.rdocumentation.org/packages/mlr/functions/generateFilterValuesData.html}{generateFilterValuesData}.

There is also an experimental
\href{http://www.rdocumentation.org/packages/ggvis/}{ggvis} plotting
function,
\href{http://www.rdocumentation.org/packages/mlr/functions/plotFilterValuesGGVIS.html}{plotFilterValuesGGVIS}.
This takes the same arguments as
\href{http://www.rdocumentation.org/packages/mlr/functions/plotFilterValues.html}{plotFilterValues}
and produces a
\href{http://www.rdocumentation.org/packages/shiny/}{shiny} application
that allows the interactive selection of the displayed filter method,
the number of features selected, and the sorting method (e.g., ascending
or descending).

\begin{lstlisting}[language=R]
plotFilterValuesGGVIS(fv2)
\end{lstlisting}

According to the \lstinline!"information.gain"! measure,
\lstinline!Petal.Width! and \lstinline!Petal.Length! contain the most
information about the target variable \lstinline!Species!.

\paragraph{Selecting a feature subset}\label{selecting-a-feature-subset}

With \href{http://www.rdocumentation.org/packages/mlr/}{mlr}'s function
\href{http://www.rdocumentation.org/packages/mlr/functions/filterFeatures.html}{filterFeatures}
you can create a new
\href{http://www.rdocumentation.org/packages/mlr/functions/Task.html}{Task}
by leaving out features of lower importance.

There are several ways to select a feature subset based on feature
importance values:

\begin{itemize}
\tightlist
\item
  Keep a certain \emph{absolute number} (\lstinline!abs!) of features
  with highest importance.
\item
  Keep a certain \emph{percentage} (\lstinline!perc!) of features with
  highest importance.
\item
  Keep all features whose importance exceeds a certain \emph{threshold
  value} (\lstinline!threshold!).
\end{itemize}

Function
\href{http://www.rdocumentation.org/packages/mlr/functions/filterFeatures.html}{filterFeatures}
supports these three methods as shown in the following example.
Moreover, you can either specify the \lstinline!method! for calculating
the feature importance or you can use previously computed importance
values via argument \lstinline!fval!.

\begin{lstlisting}[language=R]
### Keep the 2 most important features
filtered.task = filterFeatures(iris.task, method = "information.gain", abs = 2)

### Keep the 25% most important features
filtered.task = filterFeatures(iris.task, fval = fv, perc = 0.25)

### Keep all features with importance greater than 0.5
filtered.task = filterFeatures(iris.task, fval = fv, threshold = 0.5)
filtered.task
#> Supervised task: iris-example
#> Type: classif
#> Target: Species
#> Observations: 150
#> Features:
#> numerics  factors  ordered 
#>        2        0        0 
#> Missings: FALSE
#> Has weights: FALSE
#> Has blocking: FALSE
#> Classes: 3
#>     setosa versicolor  virginica 
#>         50         50         50 
#> Positive class: NA
\end{lstlisting}

\paragraph{Fuse a learner with a filter
method}\label{fuse-a-learner-with-a-filter-method}

Often feature selection based on a filter method is part of the data
preprocessing and in a subsequent step a learning method is applied to
the filtered data. In a proper experimental setup you might want to
automate the selection of the features so that it can be part of the
validation method of your choice. A
\href{http://www.rdocumentation.org/packages/mlr/functions/makeLearner.html}{Learner}
can be fused with a filter method by function
\href{http://www.rdocumentation.org/packages/mlr/functions/makeFilterWrapper.html}{makeFilterWrapper}.
The resulting
\href{http://www.rdocumentation.org/packages/mlr/functions/makeLearner.html}{Learner}
has the additional class attribute
\href{http://www.rdocumentation.org/packages/mlr/functions/FilterWrapper.html}{FilterWrapper}.

In the following example we calculate the 10-fold cross-validated error
rate (\protect\hyperlink{implemented-performance-measures}{mmce}) of the
\href{http://www.rdocumentation.org/packages/FNN/functions/fnn.html}{k
nearest neighbor classifier} with preceding feature selection on the
\href{http://www.rdocumentation.org/packages/datasets/functions/iris.html}{iris}
data set. We use \lstinline!"information.gain"! as importance measure
and select the 2 features with highest importance. In each resampling
iteration feature selection is carried out on the corresponding training
data set before fitting the learner.

\begin{lstlisting}[language=R]
lrn = makeFilterWrapper(learner = "classif.fnn", fw.method = "information.gain", fw.abs = 2)
rdesc = makeResampleDesc("CV", iters = 10)
r = resample(learner = lrn, task = iris.task, resampling = rdesc, show.info = FALSE, models = TRUE)
r$aggr
#> mmce.test.mean 
#>           0.04
\end{lstlisting}

You may want to know which features have been used. Luckily, we have
called
\href{http://www.rdocumentation.org/packages/mlr/functions/resample.html}{resample}
with the argument \lstinline!models = TRUE!, which means that
\lstinline!r$models! contains a
\href{http://www.rdocumentation.org/packages/base/functions/list.html}{list}
of
\href{http://www.rdocumentation.org/packages/mlr/functions/makeWrappedModel.html}{models}
fitted in the individual resampling iterations. In order to access the
selected feature subsets we can call
\href{http://www.rdocumentation.org/packages/mlr/functions/getFilteredFeatures.html}{getFilteredFeatures}
on each model.

\begin{lstlisting}[language=R]
sfeats = sapply(r$models, getFilteredFeatures)
table(sfeats)
#> sfeats
#> Petal.Length  Petal.Width 
#>           10           10
\end{lstlisting}

The selection of features seems to be very stable. The features
\lstinline!Sepal.Length! and \lstinline!Sepal.Width! did not make it
into a single fold.

\paragraph{Tuning the size of the feature
subset}\label{tuning-the-size-of-the-feature-subset}

In the above examples the number/percentage of features to select or the
threshold value have been arbitrarily chosen. If filtering is a
preprocessing step before applying a learning method optimal values with
regard to the learner performance can be found by
\protect\hyperlink{tuning-hyperparameters}{tuning}.

In the following regression example we consider the
\href{http://www.rdocumentation.org/packages/mlbench/functions/BostonHousing.html}{BostonHousing}
data set. We use a
\href{http://www.rdocumentation.org/packages/stats/functions/lm.html}{linear
regression model} and determine the optimal percentage value for feature
selection such that the 3-fold cross-validated
\href{http://www.rdocumentation.org/packages/mlr/functions/mse.html}{mean
squared error} of the learner is minimal. As search strategy for tuning
a grid search is used.

\begin{lstlisting}[language=R]
lrn = makeFilterWrapper(learner = "regr.lm", fw.method = "chi.squared")
ps = makeParamSet(makeDiscreteParam("fw.perc", values = seq(0.2, 0.5, 0.05)))
rdesc = makeResampleDesc("CV", iters = 3)
res = tuneParams(lrn, task = bh.task, resampling = rdesc, par.set = ps,
  control = makeTuneControlGrid())
#> [Tune] Started tuning learner regr.lm.filtered for parameter set:
#>             Type len Def                         Constr Req Tunable Trafo
#> fw.perc discrete   -   - 0.2,0.25,0.3,0.35,0.4,0.45,0.5   -    TRUE     -
#> With control class: TuneControlGrid
#> Imputation value: Inf
#> [Tune-x] 1: fw.perc=0.2
#> [Tune-y] 1: mse.test.mean=40.6; time: 0.0 min; memory: 149Mb use, 667Mb max
#> [Tune-x] 2: fw.perc=0.25
#> [Tune-y] 2: mse.test.mean=40.6; time: 0.0 min; memory: 149Mb use, 667Mb max
#> [Tune-x] 3: fw.perc=0.3
#> [Tune-y] 3: mse.test.mean=37.1; time: 0.0 min; memory: 149Mb use, 667Mb max
#> [Tune-x] 4: fw.perc=0.35
#> [Tune-y] 4: mse.test.mean=35.8; time: 0.0 min; memory: 149Mb use, 667Mb max
#> [Tune-x] 5: fw.perc=0.4
#> [Tune-y] 5: mse.test.mean=35.8; time: 0.0 min; memory: 149Mb use, 667Mb max
#> [Tune-x] 6: fw.perc=0.45
#> [Tune-y] 6: mse.test.mean=27.4; time: 0.0 min; memory: 149Mb use, 667Mb max
#> [Tune-x] 7: fw.perc=0.5
#> [Tune-y] 7: mse.test.mean=27.4; time: 0.0 min; memory: 149Mb use, 667Mb max
#> [Tune] Result: fw.perc=0.5 : mse.test.mean=27.4
res
#> Tune result:
#> Op. pars: fw.perc=0.5
#> mse.test.mean=27.4
\end{lstlisting}

The performance of all percentage values visited during tuning is:

\begin{lstlisting}[language=R]
as.data.frame(res$opt.path)
#>   fw.perc mse.test.mean dob eol error.message exec.time
#> 1     0.2      40.59578   1  NA          <NA>     0.200
#> 2    0.25      40.59578   2  NA          <NA>     0.138
#> 3     0.3      37.05592   3  NA          <NA>     0.139
#> 4    0.35      35.83712   4  NA          <NA>     0.139
#> 5     0.4      35.83712   5  NA          <NA>     0.133
#> 6    0.45      27.39955   6  NA          <NA>     0.132
#> 7     0.5      27.39955   7  NA          <NA>     0.132
\end{lstlisting}

The optimal percentage and the corresponding performance can be accessed
as follows:

\begin{lstlisting}[language=R]
res$x
#> $fw.perc
#> [1] 0.5
res$y
#> mse.test.mean 
#>      27.39955
\end{lstlisting}

After tuning we can generate a new wrapped learner with the optimal
percentage value for further use.

\begin{lstlisting}[language=R]
lrn = makeFilterWrapper(learner = "regr.lm", fw.method = "chi.squared", fw.perc = res$x$fw.perc)
mod = train(lrn, bh.task)
mod
#> Model for learner.id=regr.lm.filtered; learner.class=FilterWrapper
#> Trained on: task.id = BostonHousing-example; obs = 506; features = 13
#> Hyperparameters: fw.method=chi.squared,fw.perc=0.5

getFilteredFeatures(mod)
#> [1] "crim"  "zn"    "rm"    "dis"   "rad"   "lstat"
\end{lstlisting}

Here is another example using
\protect\hyperlink{tuning-hyperparameters}{multi-criteria tuning}. We
consider
\href{http://www.rdocumentation.org/packages/MASS/functions/lda.html}{linear
discriminant analysis} with precedent feature selection based on the
Chi-squared statistic of independence (\lstinline!"chi.squared"!) on the
\href{http://www.rdocumentation.org/packages/mlbench/functions/sonar.html}{Sonar}
data set and tune the threshold value. During tuning both, the false
positive and the false negative rate
(\protect\hyperlink{implemented-performance-measures}{fpr} and
\protect\hyperlink{implemented-performance-measures}{fnr}), are
minimized. As search strategy we choose a random search (see
\href{http://www.rdocumentation.org/packages/mlr/functions/TuneMultiCritControl.html}{makeTuneMultiCritControlRandom}).

\begin{lstlisting}[language=R]
lrn = makeFilterWrapper(learner = "classif.lda", fw.method = "chi.squared")
ps = makeParamSet(makeNumericParam("fw.threshold", lower = 0.1, upper = 0.9))
rdesc = makeResampleDesc("CV", iters = 10)
res = tuneParamsMultiCrit(lrn, task = sonar.task, resampling = rdesc, par.set = ps,
  measures = list(fpr, fnr), control = makeTuneMultiCritControlRandom(maxit = 50L),
  show.info = FALSE)
res
#> Tune multicrit result:
#> Points on front: 13
head(as.data.frame(res$opt.path))
#>   fw.threshold fpr.test.mean fnr.test.mean dob eol error.message exec.time
#> 1    0.4892321     0.3092818     0.2639033   1  NA          <NA>     1.255
#> 2    0.2481696     0.2045499     0.2319697   2  NA          <NA>     1.288
#> 3    0.7691875     0.5128000     0.3459740   3  NA          <NA>     1.207
#> 4    0.1470133     0.2045499     0.2319697   4  NA          <NA>     1.268
#> 5    0.5958241     0.5028216     0.5239538   5  NA          <NA>     1.216
#> 6    0.6892421     0.6323959     0.4480808   6  NA          <NA>     1.196
\end{lstlisting}

The results can be visualized with function
\href{http://www.rdocumentation.org/packages/mlr/functions/plotTuneMultiCritResult.html}{plotTuneMultiCritResult}.
The plot shows the false positive and false negative rates for all
parameter values visited during tuning. The size of the points on the
Pareto front is slightly increased.

\begin{lstlisting}[language=R]
plotTuneMultiCritResult(res)
\end{lstlisting}

\includegraphics{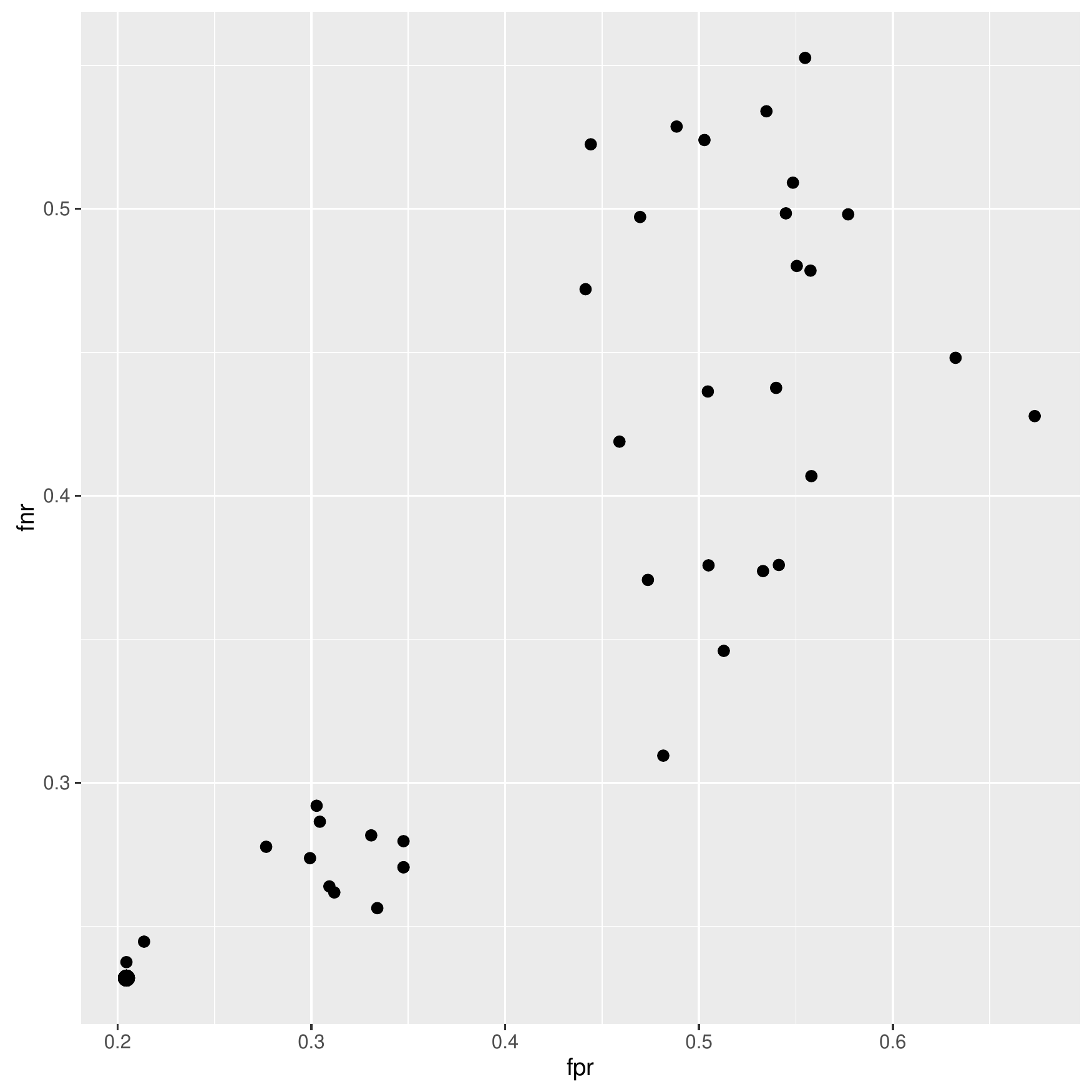}

\subsubsection{Wrapper methods}\label{wrapper-methods}

Wrapper methods use the performance of a learning algorithm to assess
the usefulness of a feature set. In order to select a feature subset a
learner is trained repeatedly on different feature subsets and the
subset which leads to the best learner performance is chosen.

In order to use the wrapper approach we have to decide:

\begin{itemize}
\tightlist
\item
  How to assess the performance: This involves choosing a performance
  measure that serves as feature selection criterion and a resampling
  strategy.
\item
  Which learning method to use.
\item
  How to search the space of possible feature subsets.
\end{itemize}

The search strategy is defined by functions following the naming
convention \lstinline!makeFeatSelControl<search_strategy!. The following
search strategies are available:

\begin{itemize}
\tightlist
\item
  Exhaustive search
  (\href{http://www.rdocumentation.org/packages/mlr/functions/FeatSelControl.html}{makeFeatSelControlExhaustive}),
\item
  Genetic algorithm
  (\href{http://www.rdocumentation.org/packages/mlr/functions/FeatSelControl.html}{makeFeatSelControlGA}),
\item
  Random search
  (\href{http://www.rdocumentation.org/packages/mlr/functions/FeatSelControl.html}{makeFeatSelControlRandom}),
\item
  Deterministic forward or backward search
  (\href{http://www.rdocumentation.org/packages/mlr/functions/FeatSelControl.html}{makeFeatSelControlSequential}).
\end{itemize}

\paragraph{Select a feature subset}\label{select-a-feature-subset}

Feature selection can be conducted with function
\href{http://www.rdocumentation.org/packages/mlr/functions/selectFeatures.html}{selectFeatures}.

In the following example we perform an exhaustive search on the
\href{http://www.rdocumentation.org/packages/TH.data/functions/wpbc.html}{Wisconsin
Prognostic Breast Cancer} data set. As learning method we use the
\href{http://www.rdocumentation.org/packages/survival/functions/coxph.html}{Cox
proportional hazards model}. The performance is assessed by the holdout
estimate of the concordance index
(\protect\hyperlink{implemented-performance-measures}{cindex}).

\begin{lstlisting}[language=R]
### Specify the search strategy
ctrl = makeFeatSelControlRandom(maxit = 20L)
ctrl
#> FeatSel control: FeatSelControlRandom
#> Same resampling instance: TRUE
#> Imputation value: <worst>
#> Max. features: <not used>
#> Max. iterations: 20
#> Tune threshold: FALSE
#> Further arguments: prob=0.5
\end{lstlisting}

\lstinline!ctrl! is a
\href{http://www.rdocumentation.org/packages/mlr/functions/FeatSelControl.html}{FeatSelControl}
object that contains information about the search strategy and potential
parameter values.

\begin{lstlisting}[language=R]
### Resample description
rdesc = makeResampleDesc("Holdout")

### Select features
sfeats = selectFeatures(learner = "surv.coxph", task = wpbc.task, resampling = rdesc,
  control = ctrl, show.info = FALSE)
sfeats
#> FeatSel result:
#> Features (17): mean_radius, mean_area, mean_smoothness, mean_concavepoints, mean_symmetry, mean_fractaldim, SE_texture, SE_perimeter, SE_smoothness, SE_compactness, SE_concavity, SE_concavepoints, worst_area, worst_compactness, worst_concavepoints, tsize, pnodes
#> cindex.test.mean=0.714
\end{lstlisting}

\lstinline!sfeats!is a
\href{http://www.rdocumentation.org/packages/mlr/functions/selectFeatures.html}{FeatSelResult}
object. The selected features and the corresponding performance can be
accessed as follows:

\begin{lstlisting}[language=R]
sfeats$x
#>  [1] "mean_radius"         "mean_area"           "mean_smoothness"    
#>  [4] "mean_concavepoints"  "mean_symmetry"       "mean_fractaldim"    
#>  [7] "SE_texture"          "SE_perimeter"        "SE_smoothness"      
#> [10] "SE_compactness"      "SE_concavity"        "SE_concavepoints"   
#> [13] "worst_area"          "worst_compactness"   "worst_concavepoints"
#> [16] "tsize"               "pnodes"
sfeats$y
#> cindex.test.mean 
#>         0.713799
\end{lstlisting}

In a second example we fit a simple linear regression model to the
\href{http://www.rdocumentation.org/packages/mlbench/functions/BostonHousing.html}{BostonHousing}
data set and use a sequential search to find a feature set that
minimizes the mean squared error
(\protect\hyperlink{implemented-performance-measures}{mse}).
\lstinline!method = "sfs"! indicates that we want to conduct a
sequential forward search where features are added to the model until
the performance cannot be improved anymore. See the documentation page
\href{http://www.rdocumentation.org/packages/mlr/functions/FeatSelControl.html}{makeFeatSelControlSequential}
for other available sequential search methods. The search is stopped if
the improvement is smaller than \lstinline!alpha = 0.02!.

\begin{lstlisting}[language=R]
### Specify the search strategy
ctrl = makeFeatSelControlSequential(method = "sfs", alpha = 0.02)

### Select features
rdesc = makeResampleDesc("CV", iters = 10)
sfeats = selectFeatures(learner = "regr.lm", task = bh.task, resampling = rdesc, control = ctrl,
  show.info = FALSE)
sfeats
#> FeatSel result:
#> Features (11): crim, zn, chas, nox, rm, dis, rad, tax, ptratio, b, lstat
#> mse.test.mean=23.7
\end{lstlisting}

Further information about the sequential feature selection process can
be obtained by function
\href{http://www.rdocumentation.org/packages/mlr/functions/analyzeFeatSelResult.html}{analyzeFeatSelResult}.

\begin{lstlisting}[language=R]
analyzeFeatSelResult(sfeats)
#> Features         : 11
#> Performance      : mse.test.mean=23.7
#> crim, zn, chas, nox, rm, dis, rad, tax, ptratio, b, lstat
#> 
#> Path to optimum:
#> - Features:    0  Init   :                       Perf = 84.831  Diff: NA  *
#> - Features:    1  Add    : lstat                 Perf = 38.894  Diff: 45.936  *
#> - Features:    2  Add    : rm                    Perf = 31.279  Diff: 7.6156  *
#> - Features:    3  Add    : ptratio               Perf = 28.108  Diff: 3.1703  *
#> - Features:    4  Add    : dis                   Perf = 27.48  Diff: 0.62813  *
#> - Features:    5  Add    : nox                   Perf = 26.079  Diff: 1.4008  *
#> - Features:    6  Add    : b                     Perf = 25.563  Diff: 0.51594  *
#> - Features:    7  Add    : chas                  Perf = 25.132  Diff: 0.43097  *
#> - Features:    8  Add    : zn                    Perf = 24.792  Diff: 0.34018  *
#> - Features:    9  Add    : rad                   Perf = 24.599  Diff: 0.19327  *
#> - Features:   10  Add    : tax                   Perf = 24.082  Diff: 0.51706  *
#> - Features:   11  Add    : crim                  Perf = 23.732  Diff: 0.35  *
#> 
#> Stopped, because no improving feature was found.
\end{lstlisting}

\paragraph{Fuse a learner with feature
selection}\label{fuse-a-learner-with-feature-selection}

A
\href{http://www.rdocumentation.org/packages/mlr/functions/makeLearner.html}{Learner}
can be fused with a feature selection strategy (i.e., a search strategy,
a performance measure and a resampling strategy) by function
\href{http://www.rdocumentation.org/packages/mlr/functions/makeFeatSelWrapper.html}{makeFeatSelWrapper}.
During training features are selected according to the specified
selection scheme. Then, the learner is trained on the selected feature
subset.

\begin{lstlisting}[language=R]
rdesc = makeResampleDesc("CV", iters = 3)
lrn = makeFeatSelWrapper("surv.coxph", resampling = rdesc,
  control = makeFeatSelControlRandom(maxit = 10), show.info = FALSE)
mod = train(lrn, task = wpbc.task)
mod
#> Model for learner.id=surv.coxph.featsel; learner.class=FeatSelWrapper
#> Trained on: task.id = wpbc-example; obs = 194; features = 32
#> Hyperparameters:
\end{lstlisting}

The result of the feature selection can be extracted by function
\href{http://www.rdocumentation.org/packages/mlr/functions/getFeatSelResult.html}{getFeatSelResult}.

\begin{lstlisting}[language=R]
sfeats = getFeatSelResult(mod)
sfeats
#> FeatSel result:
#> Features (19): mean_radius, mean_texture, mean_perimeter, mean_area, mean_smoothness, mean_compactness, mean_concavepoints, mean_fractaldim, SE_compactness, SE_concavity, SE_concavepoints, SE_symmetry, worst_texture, worst_perimeter, worst_area, worst_concavepoints, worst_symmetry, tsize, pnodes
#> cindex.test.mean=0.631
\end{lstlisting}

The selected features are:

\begin{lstlisting}[language=R]
sfeats$x
#>  [1] "mean_radius"         "mean_texture"        "mean_perimeter"     
#>  [4] "mean_area"           "mean_smoothness"     "mean_compactness"   
#>  [7] "mean_concavepoints"  "mean_fractaldim"     "SE_compactness"     
#> [10] "SE_concavity"        "SE_concavepoints"    "SE_symmetry"        
#> [13] "worst_texture"       "worst_perimeter"     "worst_area"         
#> [16] "worst_concavepoints" "worst_symmetry"      "tsize"              
#> [19] "pnodes"
\end{lstlisting}

The 5-fold cross-validated performance of the learner specified above
can be computed as follows:

\begin{lstlisting}[language=R]
out.rdesc = makeResampleDesc("CV", iters = 5)

r = resample(learner = lrn, task = wpbc.task, resampling = out.rdesc, models = TRUE,
  show.info = FALSE)
r$aggr
#> cindex.test.mean 
#>         0.632357
\end{lstlisting}

The selected feature sets in the individual resampling iterations can be
extracted as follows:

\begin{lstlisting}[language=R]
lapply(r$models, getFeatSelResult)
#> [[1]]
#> FeatSel result:
#> Features (18): mean_texture, mean_area, mean_smoothness, mean_compactness, mean_concavity, mean_symmetry, SE_radius, SE_compactness, SE_concavity, SE_concavepoints, SE_fractaldim, worst_radius, worst_smoothness, worst_compactness, worst_concavity, worst_symmetry, tsize, pnodes
#> cindex.test.mean=0.66
#> 
#> [[2]]
#> FeatSel result:
#> Features (12): mean_area, mean_compactness, mean_symmetry, mean_fractaldim, SE_perimeter, SE_area, SE_concavity, SE_symmetry, worst_texture, worst_smoothness, worst_fractaldim, tsize
#> cindex.test.mean=0.652
#> 
#> [[3]]
#> FeatSel result:
#> Features (14): mean_compactness, mean_symmetry, mean_fractaldim, SE_radius, SE_perimeter, SE_smoothness, SE_concavity, SE_concavepoints, SE_fractaldim, worst_concavity, worst_concavepoints, worst_symmetry, worst_fractaldim, pnodes
#> cindex.test.mean=0.607
#> 
#> [[4]]
#> FeatSel result:
#> Features (18): mean_radius, mean_texture, mean_perimeter, mean_compactness, mean_concavity, SE_texture, SE_area, SE_smoothness, SE_concavity, SE_symmetry, SE_fractaldim, worst_radius, worst_compactness, worst_concavepoints, worst_symmetry, worst_fractaldim, tsize, pnodes
#> cindex.test.mean=0.653
#> 
#> [[5]]
#> FeatSel result:
#> Features (14): mean_radius, mean_texture, mean_compactness, mean_concavepoints, mean_symmetry, SE_texture, SE_compactness, SE_symmetry, SE_fractaldim, worst_radius, worst_smoothness, worst_compactness, worst_concavity, pnodes
#> cindex.test.mean=0.626
\end{lstlisting}

\hypertarget{nested-resampling}{\subsection{Nested
Resampling}\label{nested-resampling}}

In order to obtain honest performance estimates for a learner all parts
of the model building like preprocessing and model selection steps
should be included in the resampling, i.e., repeated for every pair of
training/test data. For steps that themselves require resampling like
\protect\hyperlink{tuning-hyperparameters}{parameter tuning} or
\protect\hyperlink{feature-selection}{feature selection} (via the
wrapper approach) this results in two nested resampling loops.

\includegraphics{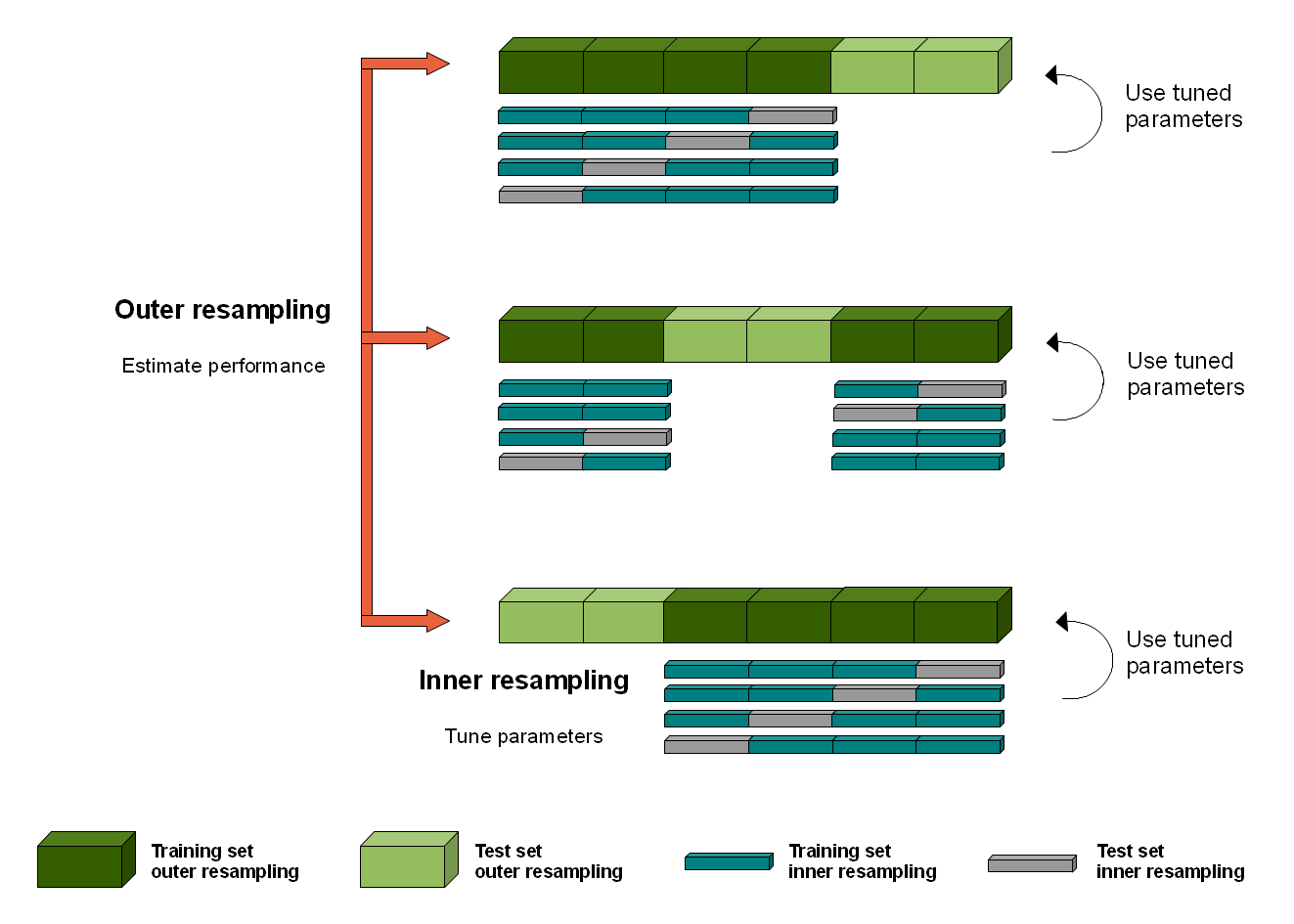}

The graphic above illustrates nested resampling for parameter tuning
with 3-fold cross-validation in the outer and 4-fold cross-validation in
the inner loop.

In the outer resampling loop, we have three pairs of training/test sets.
On each of these outer training sets parameter tuning is done, thereby
executing the inner resampling loop. This way, we get one set of
selected hyperparameters for each outer training set. Then the learner
is fitted on each outer training set using the corresponding selected
hyperparameters and its performance is evaluated on the outer test sets.

In \href{http://www.rdocumentation.org/packages/mlr/}{mlr}, you can get
nested resampling for free without programming any looping by using the
\protect\hyperlink{wrapper}{wrapper functionality}. This works as
follows:

\begin{enumerate}
\def\labelenumi{\arabic{enumi}.}
\tightlist
\item
  Generate a wrapped
  \href{http://www.rdocumentation.org/packages/mlr/functions/makeLearner.html}{Learner}
  via function
  \href{http://www.rdocumentation.org/packages/mlr/functions/makeTuneWrapper.html}{makeTuneWrapper}
  or
  \href{http://www.rdocumentation.org/packages/mlr/functions/makeFeatSelWrapper.html}{makeFeatSelWrapper}.
  Specify the inner resampling strategy using their
  \lstinline!resampling! argument.
\item
  Call function
  \href{http://www.rdocumentation.org/packages/mlr/functions/resample.html}{resample}
  (see also the section about
  \protect\hyperlink{resampling}{resampling}) and pass the outer
  resampling strategy to its \lstinline!resampling! argument.
\end{enumerate}

You can freely combine different inner and outer resampling strategies.

The outer strategy can be a resample description
(\href{http://www.rdocumentation.org/packages/mlr/functions/makeResampleDesc.html}{ResampleDesc})
or a resample instance
(\href{http://www.rdocumentation.org/packages/mlr/functions/makeResampleInstance.html}{ResampleInstance}).
A common setup is prediction and performance evaluation on a fixed outer
test set. This can be achieved by using function
\href{http://www.rdocumentation.org/packages/mlr/functions/makeFixedHoldoutInstance.html}{makeFixedHoldoutInstance}
to generate the outer
\href{http://www.rdocumentation.org/packages/mlr/functions/makeResampleInstance.html}{ResampleInstance}.

The inner resampling strategy should preferably be a
\href{http://www.rdocumentation.org/packages/mlr/functions/makeResampleDesc.html}{ResampleDesc},
as the sizes of the outer training sets might differ. Per default, the
inner resample description is instantiated once for every outer training
set. This way during tuning/feature selection all parameter or feature
sets are compared on the same inner training/test sets to reduce
variance. You can also turn this off using the
\lstinline!same.resampling.instance! argument of
\href{http://www.rdocumentation.org/packages/mlr/functions/TuneControl.html}{makeTuneControl*}
or
\href{http://www.rdocumentation.org/packages/mlr/functions/FeatSelControl.html}{makeFeatSelControl*}.

Nested resampling is computationally expensive. For this reason in the
examples shown below we use relatively small search spaces and a low
number of resampling iterations. In practice, you normally have to
increase both. As this is computationally intensive you might want to
have a look at section
\protect\hyperlink{parallelization}{parallelization}.

\subsubsection{Tuning}\label{tuning}

As you might recall from the tutorial page about
\protect\hyperlink{tuning-hyperparameters}{tuning}, you need to define a
search space by function
\href{http://www.rdocumentation.org/packages/ParamHelpers/functions/makeParamSet.html}{makeParamSet},
a search strategy by
\href{http://www.rdocumentation.org/packages/mlr/functions/TuneControl.html}{makeTuneControl*},
and a method to evaluate hyperparameter settings (i.e., the inner
resampling strategy and a performance measure).

Below is a classification example. We evaluate the performance of a
support vector machine
(\href{http://www.rdocumentation.org/packages/kernlab/functions/ksvm.html}{ksvm})
with tuned cost parameter \lstinline!C! and RBF kernel parameter
\lstinline!sigma!. We use 3-fold cross-validation in the outer and
subsampling with 2 iterations in the inner loop. For tuning a grid
search is used to find the hyperparameters with lowest error rate
(\protect\hyperlink{implemented-performance-measures}{mmce} is the
default measure for classification). The wrapped
\href{http://www.rdocumentation.org/packages/mlr/functions/makeLearner.html}{Learner}
is generated by calling
\href{http://www.rdocumentation.org/packages/mlr/functions/makeTuneWrapper.html}{makeTuneWrapper}.

Note that in practice the parameter set should be larger. A common
recommendation is \lstinline!2^(-12:12)! for both \lstinline!C! and
\lstinline!sigma!.

\begin{lstlisting}[language=R]
### Tuning in inner resampling loop
ps = makeParamSet(
  makeDiscreteParam("C", values = 2^(-2:2)),
  makeDiscreteParam("sigma", values = 2^(-2:2))
)
ctrl = makeTuneControlGrid()
inner = makeResampleDesc("Subsample", iters = 2)
lrn = makeTuneWrapper("classif.ksvm", resampling = inner, par.set = ps, control = ctrl, show.info = FALSE)

### Outer resampling loop
outer = makeResampleDesc("CV", iters = 3)
r = resample(lrn, iris.task, resampling = outer, extract = getTuneResult, show.info = FALSE)

r
#> Resample Result
#> Task: iris-example
#> Learner: classif.ksvm.tuned
#> Aggr perf: mmce.test.mean=0.0533
#> Runtime: 17.5504
\end{lstlisting}

You can obtain the error rates on the 3 outer test sets by:

\begin{lstlisting}[language=R]
r$measures.test
#>   iter mmce
#> 1    1 0.02
#> 2    2 0.06
#> 3    3 0.08
\end{lstlisting}

\paragraph{Accessing the tuning
result}\label{accessing-the-tuning-result-1}

We have kept the results of the tuning for further evaluations. For
example one might want to find out, if the best obtained configurations
vary for the different outer splits. As storing entire models may be
expensive (but possible by setting \lstinline!models = TRUE!) we used
the \lstinline!extract! option of
\href{http://www.rdocumentation.org/packages/mlr/functions/resample.html}{resample}.
Function
\href{http://www.rdocumentation.org/packages/mlr/functions/getTuneResult.html}{getTuneResult}
returns, among other things, the optimal hyperparameter values and the
\href{http://www.rdocumentation.org/packages/ParamHelpers/functions/OptPath.html}{optimization
path} for each iteration of the outer resampling loop. Note that the
performance values shown when printing \lstinline!r$extract! are the
aggregated performances resulting from inner resampling on the outer
training set for the best hyperparameter configurations (not to be
confused with \lstinline!r$measures.test! shown above).

\begin{lstlisting}[language=R]
r$extract
#> [[1]]
#> Tune result:
#> Op. pars: C=2; sigma=0.25
#> mmce.test.mean=0.0147
#> 
#> [[2]]
#> Tune result:
#> Op. pars: C=4; sigma=0.25
#> mmce.test.mean=   0
#> 
#> [[3]]
#> Tune result:
#> Op. pars: C=4; sigma=0.25
#> mmce.test.mean=0.0735

names(r$extract[[1]])
#> [1] "learner"   "control"   "x"         "y"         "threshold" "opt.path"
\end{lstlisting}

We can compare the optimal parameter settings obtained in the 3
resampling iterations. As you can see, the optimal configuration usually
depends on the data. You may be able to identify a \emph{range} of
parameter settings that achieve good performance though, e.g., the
values for \lstinline!C! should be at least 1 and the values for
\lstinline!sigma! should be between 0 and 1.

With function
\href{http://www.rdocumentation.org/packages/mlr/functions/getNestedTuneResultsOptPathDf.html}{getNestedTuneResultsOptPathDf}
you can extract the optimization paths for the 3 outer cross-validation
iterations for further inspection and analysis. These are stacked in one
\href{http://www.rdocumentation.org/packages/base/functions/data.frame.html}{data.frame}
with column \lstinline!iter! indicating the resampling iteration.

\begin{lstlisting}[language=R]
opt.paths = getNestedTuneResultsOptPathDf(r)
head(opt.paths, 10)
#>       C sigma mmce.test.mean dob eol error.message exec.time iter
#> 1  0.25  0.25     0.05882353   1  NA          <NA>     0.034    1
#> 2   0.5  0.25     0.04411765   2  NA          <NA>     0.036    1
#> 3     1  0.25     0.04411765   3  NA          <NA>     0.034    1
#> 4     2  0.25     0.01470588   4  NA          <NA>     0.034    1
#> 5     4  0.25     0.05882353   5  NA          <NA>     0.034    1
#> 6  0.25   0.5     0.05882353   6  NA          <NA>     0.035    1
#> 7   0.5   0.5     0.01470588   7  NA          <NA>     0.034    1
#> 8     1   0.5     0.02941176   8  NA          <NA>     0.035    1
#> 9     2   0.5     0.01470588   9  NA          <NA>     0.035    1
#> 10    4   0.5     0.05882353  10  NA          <NA>     0.035    1
\end{lstlisting}

Below we visualize the \lstinline!opt.path!s for the 3 outer resampling
iterations.

\begin{lstlisting}[language=R]
g = ggplot(opt.paths, aes(x = C, y = sigma, fill = mmce.test.mean))
g + geom_tile() + facet_wrap(~ iter)
\end{lstlisting}

\includegraphics{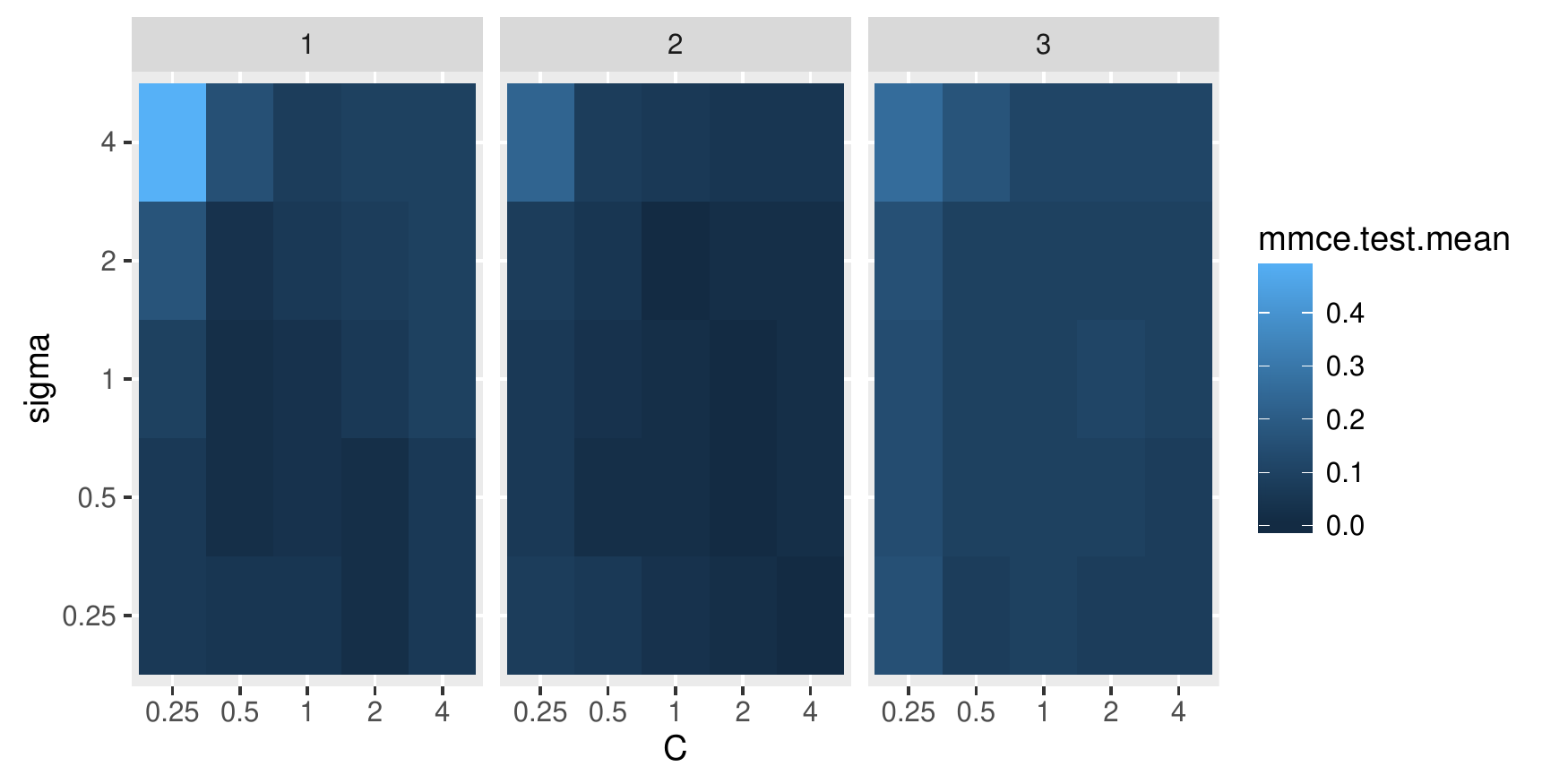}

Another useful function is
\href{http://www.rdocumentation.org/packages/mlr/functions/getNestedTuneResultsX.html}{getNestedTuneResultsX},
which extracts the best found hyperparameter settings for each outer
resampling iteration.

\begin{lstlisting}[language=R]
getNestedTuneResultsX(r)
#>   C sigma
#> 1 2  0.25
#> 2 4  0.25
#> 3 4  0.25
\end{lstlisting}

\subsubsection{Feature selection}\label{feature-selection-1}

As you might recall from the section about
\protect\hyperlink{feature-selection}{feature selection},
\href{http://www.rdocumentation.org/packages/mlr/}{mlr} supports the
filter and the wrapper approach.

\paragraph{Wrapper methods}\label{wrapper-methods-1}

Wrapper methods use the performance of a learning algorithm to assess
the usefulness of a feature set. In order to select a feature subset a
learner is trained repeatedly on different feature subsets and the
subset which leads to the best learner performance is chosen.

For feature selection in the inner resampling loop, you need to choose a
search strategy (function
\href{http://www.rdocumentation.org/packages/mlr/functions/FeatSelControl.html}{makeFeatSelControl*}),
a performance measure and the inner resampling strategy. Then use
function
\href{http://www.rdocumentation.org/packages/mlr/functions/makeFeatSelWrapper.html}{makeFeatSelWrapper}
to bind everything together.

Below we use sequential forward selection with linear regression on the
\href{http://www.rdocumentation.org/packages/mlbench/functions/BostonHousing.html}{BostonHousing}
data set
(\href{http://www.rdocumentation.org/packages/mlr/functions/bh.task.html}{bh.task}).

\begin{lstlisting}[language=R]
### Feature selection in inner resampling loop
inner = makeResampleDesc("CV", iters = 3)
lrn = makeFeatSelWrapper("regr.lm", resampling = inner,
  control = makeFeatSelControlSequential(method = "sfs"), show.info = FALSE)

### Outer resampling loop
outer = makeResampleDesc("Subsample", iters = 2)
r = resample(learner = lrn, task = bh.task, resampling = outer, extract = getFeatSelResult,
  show.info = FALSE)

r
#> Resample Result
#> Task: BostonHousing-example
#> Learner: regr.lm.featsel
#> Aggr perf: mse.test.mean=31.7
#> Runtime: 39.7649

r$measures.test
#>   iter      mse
#> 1    1 35.08611
#> 2    2 28.31215
\end{lstlisting}

\subparagraph{Accessing the selected
features}\label{accessing-the-selected-features}

The result of the feature selection can be extracted by function
\href{http://www.rdocumentation.org/packages/mlr/functions/getFeatSelResult.html}{getFeatSelResult}.
It is also possible to keep whole
\href{http://www.rdocumentation.org/packages/mlr/functions/makeWrappedModel.html}{models}
by setting \lstinline!models = TRUE! when calling
\href{http://www.rdocumentation.org/packages/mlr/functions/resample.html}{resample}.

\begin{lstlisting}[language=R]
r$extract
#> [[1]]
#> FeatSel result:
#> Features (10): crim, zn, indus, nox, rm, dis, rad, tax, ptratio, lstat
#> mse.test.mean=20.2
#> 
#> [[2]]
#> FeatSel result:
#> Features (9): zn, nox, rm, dis, rad, tax, ptratio, b, lstat
#> mse.test.mean=22.6

### Selected features in the first outer resampling iteration
r$extract[[1]]$x
#>  [1] "crim"    "zn"      "indus"   "nox"     "rm"      "dis"     "rad"    
#>  [8] "tax"     "ptratio" "lstat"

### Resampled performance of the selected feature subset on the first inner training set
r$extract[[1]]$y
#> mse.test.mean 
#>      20.15939
\end{lstlisting}

As for tuning, you can extract the optimization paths. The resulting
\href{http://www.rdocumentation.org/packages/base/functions/data.frame.html}{data.frame}s
contain, among others, binary columns for all features, indicating if
they were included in the linear regression model, and the corresponding
performances.

\begin{lstlisting}[language=R]
opt.paths = lapply(r$extract, function(x) as.data.frame(x$opt.path))
head(opt.paths[[1]])
#>   crim zn indus chas nox rm age dis rad tax ptratio b lstat mse.test.mean
#> 1    0  0     0    0   0  0   0   0   0   0       0 0     0      80.33019
#> 2    1  0     0    0   0  0   0   0   0   0       0 0     0      65.95316
#> 3    0  1     0    0   0  0   0   0   0   0       0 0     0      69.15417
#> 4    0  0     1    0   0  0   0   0   0   0       0 0     0      55.75473
#> 5    0  0     0    1   0  0   0   0   0   0       0 0     0      80.48765
#> 6    0  0     0    0   1  0   0   0   0   0       0 0     0      63.06724
#>   dob eol error.message exec.time
#> 1   1   2          <NA>     0.017
#> 2   2   2          <NA>     0.027
#> 3   2   2          <NA>     0.027
#> 4   2   2          <NA>     0.027
#> 5   2   2          <NA>     0.031
#> 6   2   2          <NA>     0.026
\end{lstlisting}

An easy-to-read version of the optimization path for sequential feature
selection can be obtained with function
\href{http://www.rdocumentation.org/packages/mlr/functions/analyzeFeatSelResult.html}{analyzeFeatSelResult}.

\begin{lstlisting}[language=R]
analyzeFeatSelResult(r$extract[[1]])
#> Features         : 10
#> Performance      : mse.test.mean=20.2
#> crim, zn, indus, nox, rm, dis, rad, tax, ptratio, lstat
#> 
#> Path to optimum:
#> - Features:    0  Init   :                       Perf = 80.33  Diff: NA  *
#> - Features:    1  Add    : lstat                 Perf = 36.451  Diff: 43.879  *
#> - Features:    2  Add    : rm                    Perf = 27.289  Diff: 9.1623  *
#> - Features:    3  Add    : ptratio               Perf = 24.004  Diff: 3.2849  *
#> - Features:    4  Add    : nox                   Perf = 23.513  Diff: 0.49082  *
#> - Features:    5  Add    : dis                   Perf = 21.49  Diff: 2.023  *
#> - Features:    6  Add    : crim                  Perf = 21.12  Diff: 0.37008  *
#> - Features:    7  Add    : indus                 Perf = 20.82  Diff: 0.29994  *
#> - Features:    8  Add    : rad                   Perf = 20.609  Diff: 0.21054  *
#> - Features:    9  Add    : tax                   Perf = 20.209  Diff: 0.40059  *
#> - Features:   10  Add    : zn                    Perf = 20.159  Diff: 0.049441  *
#> 
#> Stopped, because no improving feature was found.
\end{lstlisting}

\paragraph{Filter methods with tuning}\label{filter-methods-with-tuning}

Filter methods assign an importance value to each feature. Based on
these values you can select a feature subset by either keeping all
features with importance higher than a certain threshold or by keeping a
fixed number or percentage of the highest ranking features. Often,
neither the theshold nor the number or percentage of features is known
in advance and thus tuning is necessary.

In the example below the threshold value (\lstinline!fw.threshold!) is
tuned in the inner resampling loop. For this purpose the base
\href{http://www.rdocumentation.org/packages/mlr/functions/makeLearner.html}{Learner}
\lstinline!"regr.lm"! is wrapped two times. First,
\href{http://www.rdocumentation.org/packages/mlr/functions/makeFilterWrapper.html}{makeFilterWrapper}
is used to fuse linear regression with a feature filtering preprocessing
step. Then a tuning step is added by
\href{http://www.rdocumentation.org/packages/mlr/functions/makeTuneWrapper.html}{makeTuneWrapper}.

\begin{lstlisting}[language=R]
### Tuning of the percentage of selected filters in the inner loop
lrn = makeFilterWrapper(learner = "regr.lm", fw.method = "chi.squared")
ps = makeParamSet(makeDiscreteParam("fw.threshold", values = seq(0, 1, 0.2)))
ctrl = makeTuneControlGrid()
inner = makeResampleDesc("CV", iters = 3)
lrn = makeTuneWrapper(lrn, resampling = inner, par.set = ps, control = ctrl, show.info = FALSE)

### Outer resampling loop
outer = makeResampleDesc("CV", iters = 3)
r = resample(learner = lrn, task = bh.task, resampling = outer, models = TRUE, show.info = FALSE)
r
#> Resample Result
#> Task: BostonHousing-example
#> Learner: regr.lm.filtered.tuned
#> Aggr perf: mse.test.mean=25.4
#> Runtime: 6.16262
\end{lstlisting}

\subparagraph{Accessing the selected features and optimal
percentage}\label{accessing-the-selected-features-and-optimal-percentage}

In the above example we kept the complete
\href{http://www.rdocumentation.org/packages/mlr/functions/makeWrappedModel.html}{model}s.

Below are some examples that show how to extract information from the
\href{http://www.rdocumentation.org/packages/mlr/functions/makeWrappedModel.html}{model}s.

\begin{lstlisting}[language=R]
r$models
#> [[1]]
#> Model for learner.id=regr.lm.filtered.tuned; learner.class=TuneWrapper
#> Trained on: task.id = BostonHousing-example; obs = 337; features = 13
#> Hyperparameters: fw.method=chi.squared
#> 
#> [[2]]
#> Model for learner.id=regr.lm.filtered.tuned; learner.class=TuneWrapper
#> Trained on: task.id = BostonHousing-example; obs = 338; features = 13
#> Hyperparameters: fw.method=chi.squared
#> 
#> [[3]]
#> Model for learner.id=regr.lm.filtered.tuned; learner.class=TuneWrapper
#> Trained on: task.id = BostonHousing-example; obs = 337; features = 13
#> Hyperparameters: fw.method=chi.squared
\end{lstlisting}

The result of the feature selection can be extracted by function
\href{http://www.rdocumentation.org/packages/mlr/functions/getFilteredFeatures.html}{getFilteredFeatures}.
Almost always all 13 features are selected.

\begin{lstlisting}[language=R]
lapply(r$models, function(x) getFilteredFeatures(x$learner.model$next.model))
#> [[1]]
#>  [1] "crim"    "zn"      "indus"   "chas"    "nox"     "rm"      "age"    
#>  [8] "dis"     "rad"     "tax"     "ptratio" "b"       "lstat"  
#> 
#> [[2]]
#>  [1] "crim"    "zn"      "indus"   "nox"     "rm"      "age"     "dis"    
#>  [8] "rad"     "tax"     "ptratio" "b"       "lstat"  
#> 
#> [[3]]
#>  [1] "crim"    "zn"      "indus"   "chas"    "nox"     "rm"      "age"    
#>  [8] "dis"     "rad"     "tax"     "ptratio" "b"       "lstat"
\end{lstlisting}

Below the
\href{http://www.rdocumentation.org/packages/mlr/functions/TuneResult.html}{tune
results} and
\href{http://www.rdocumentation.org/packages/ParamHelpers/functions/OptPath.html}{optimization
paths} are accessed.

\begin{lstlisting}[language=R]
res = lapply(r$models, getTuneResult)
res
#> [[1]]
#> Tune result:
#> Op. pars: fw.threshold=0
#> mse.test.mean=24.9
#> 
#> [[2]]
#> Tune result:
#> Op. pars: fw.threshold=0.4
#> mse.test.mean=27.2
#> 
#> [[3]]
#> Tune result:
#> Op. pars: fw.threshold=0
#> mse.test.mean=19.7

opt.paths = lapply(res, function(x) as.data.frame(x$opt.path))
opt.paths[[1]]
#>   fw.threshold mse.test.mean dob eol error.message exec.time
#> 1            0      24.89160   1  NA          <NA>     0.148
#> 2          0.2      25.18817   2  NA          <NA>     0.149
#> 3          0.4      25.18817   3  NA          <NA>     0.141
#> 4          0.6      32.15930   4  NA          <NA>     0.139
#> 5          0.8      90.89848   5  NA          <NA>     0.131
#> 6            1      90.89848   6  NA          <NA>     0.131
\end{lstlisting}

\subsubsection{Benchmark experiments}\label{benchmark-experiments-1}

In a benchmark experiment multiple learners are compared on one or
several tasks (see also the section about
\protect\hyperlink{benchmark-experiments}{benchmarking}). Nested
resampling in benchmark experiments is achieved the same way as in
resampling:

\begin{itemize}
\tightlist
\item
  First, use
  \href{http://www.rdocumentation.org/packages/mlr/functions/makeTuneWrapper.html}{makeTuneWrapper}
  or
  \href{http://www.rdocumentation.org/packages/mlr/functions/makeFeatSelWrapper.html}{makeFeatSelWrapper}
  to generate wrapped
  \href{http://www.rdocumentation.org/packages/mlr/functions/makeLearner.html}{Learner}s
  with the inner resampling strategies of your choice.
\item
  Second, call
  \href{http://www.rdocumentation.org/packages/mlr/functions/benchmark.html}{benchmark}
  and specify the outer resampling strategies for all tasks.
\end{itemize}

The inner resampling strategies should be
\href{http://www.rdocumentation.org/packages/mlr/functions/makeResampleDesc.html}{resample
descriptions}. You can use different inner resampling strategies for
different wrapped learners. For example it might be practical to do
fewer subsampling or bootstrap iterations for slower learners.

If you have larger benchmark experiments you might want to have a look
at the section about
\protect\hyperlink{parallelization}{parallelization}.

As mentioned in the section about
\protect\hyperlink{benchmark-experiments}{benchmark experiments} you can
also use different resampling strategies for different learning tasks by
passing a
\href{http://www.rdocumentation.org/packages/base/functions/list.html}{list}
of resampling descriptions or instances to
\href{http://www.rdocumentation.org/packages/mlr/functions/benchmark.html}{benchmark}.

We will see three examples to show different benchmark settings:

\begin{enumerate}
\def\labelenumi{\arabic{enumi}.}
\tightlist
\item
  Two data sets + two classification algorithms + tuning
\item
  One data set + two regression algorithms + feature selection
\item
  One data set + two regression algorithms + feature filtering + tuning
\end{enumerate}

\paragraph{Example 1: Two tasks, two learners,
tuning}\label{example-1-two-tasks-two-learners-tuning}

Below is a benchmark experiment with two data sets,
\href{http://www.rdocumentation.org/packages/datasets/functions/iris.html}{iris}
and
\href{http://www.rdocumentation.org/packages/mlbench/functions/sonar.html}{sonar},
and two
\href{http://www.rdocumentation.org/packages/mlr/functions/makeLearner.html}{Learner}s,
\href{http://www.rdocumentation.org/packages/kernlab/functions/ksvm.html}{ksvm}
and
\href{http://www.rdocumentation.org/packages/kknn/functions/kknn.html}{kknn},
that are both tuned.

As inner resampling strategies we use holdout for
\href{http://www.rdocumentation.org/packages/kernlab/functions/ksvm.html}{ksvm}
and subsampling with 3 iterations for
\href{http://www.rdocumentation.org/packages/kknn/functions/kknn.html}{kknn}.
As outer resampling strategies we take holdout for the
\href{http://www.rdocumentation.org/packages/datasets/functions/iris.html}{iris}
and bootstrap with 2 iterations for the
\href{http://www.rdocumentation.org/packages/mlbench/functions/sonar.html}{sonar}
data
(\href{http://www.rdocumentation.org/packages/mlr/functions/sonar.task.html}{sonar.task}).
We consider the accuracy
(\protect\hyperlink{implemented-performance-measures}{acc}), which is
used as tuning criterion, and also calculate the balanced error rate
(\protect\hyperlink{implemented-performance-measures}{ber}).

\begin{lstlisting}[language=R]
### List of learning tasks
tasks = list(iris.task, sonar.task)

### Tune svm in the inner resampling loop
ps = makeParamSet(
  makeDiscreteParam("C", 2^(-1:1)),
  makeDiscreteParam("sigma", 2^(-1:1)))
ctrl = makeTuneControlGrid()
inner = makeResampleDesc("Holdout")
lrn1 = makeTuneWrapper("classif.ksvm", resampling = inner, par.set = ps, control = ctrl,
  show.info = FALSE)

### Tune k-nearest neighbor in inner resampling loop
ps = makeParamSet(makeDiscreteParam("k", 3:5))
ctrl = makeTuneControlGrid()
inner = makeResampleDesc("Subsample", iters = 3)
lrn2 = makeTuneWrapper("classif.kknn", resampling = inner, par.set = ps, control = ctrl,
  show.info = FALSE)

### Learners
lrns = list(lrn1, lrn2)

### Outer resampling loop
outer = list(makeResampleDesc("Holdout"), makeResampleDesc("Bootstrap", iters = 2))
res = benchmark(lrns, tasks, outer, measures = list(acc, ber), show.info = FALSE)
res
#>         task.id         learner.id acc.test.mean ber.test.mean
#> 1  iris-example classif.ksvm.tuned     0.9400000    0.05882353
#> 2  iris-example classif.kknn.tuned     0.9200000    0.08683473
#> 3 Sonar-example classif.ksvm.tuned     0.5289307    0.50000000
#> 4 Sonar-example classif.kknn.tuned     0.8077080    0.19549714
\end{lstlisting}

The
\href{http://www.rdocumentation.org/packages/base/functions/print.html}{print}
method for the
\href{http://www.rdocumentation.org/packages/mlr/functions/BenchmarkResult.html}{BenchmarkResult}
shows the aggregated performances from the outer resampling loop.

As you might recall,
\href{http://www.rdocumentation.org/packages/mlr/}{mlr} offers several
accessor function to extract information from the benchmark result.
These are listed on the help page of
\href{http://www.rdocumentation.org/packages/mlr/functions/BenchmarkResult.html}{BenchmarkResult}
and many examples are shown on the tutorial page about
\protect\hyperlink{benchmark-experiments}{benchmark experiments}.

The performance values in individual outer resampling runs can be
obtained by
\href{http://www.rdocumentation.org/packages/mlr/functions/getBMRPerformances.html}{getBMRPerformances}.
Note that, since we used different outer resampling strategies for the
two tasks, the number of rows per task differ.

\begin{lstlisting}[language=R]
getBMRPerformances(res, as.df = TRUE)
#>         task.id         learner.id iter       acc        ber
#> 1  iris-example classif.ksvm.tuned    1 0.9400000 0.05882353
#> 2  iris-example classif.kknn.tuned    1 0.9200000 0.08683473
#> 3 Sonar-example classif.ksvm.tuned    1 0.5373134 0.50000000
#> 4 Sonar-example classif.ksvm.tuned    2 0.5205479 0.50000000
#> 5 Sonar-example classif.kknn.tuned    1 0.8208955 0.18234767
#> 6 Sonar-example classif.kknn.tuned    2 0.7945205 0.20864662
\end{lstlisting}

The results from the parameter tuning can be obtained through function
\href{http://www.rdocumentation.org/packages/mlr/functions/getBMRTuneResults.html}{getBMRTuneResults}.

\begin{lstlisting}[language=R]
getBMRTuneResults(res)
#> $`iris-example`
#> $`iris-example`$classif.ksvm.tuned
#> $`iris-example`$classif.ksvm.tuned[[1]]
#> Tune result:
#> Op. pars: C=0.5; sigma=0.5
#> mmce.test.mean=0.0588
#> 
#> 
#> $`iris-example`$classif.kknn.tuned
#> $`iris-example`$classif.kknn.tuned[[1]]
#> Tune result:
#> Op. pars: k=3
#> mmce.test.mean=0.049
#> 
#> 
#> 
#> $`Sonar-example`
#> $`Sonar-example`$classif.ksvm.tuned
#> $`Sonar-example`$classif.ksvm.tuned[[1]]
#> Tune result:
#> Op. pars: C=1; sigma=2
#> mmce.test.mean=0.343
#> 
#> $`Sonar-example`$classif.ksvm.tuned[[2]]
#> Tune result:
#> Op. pars: C=2; sigma=0.5
#> mmce.test.mean= 0.2
#> 
#> 
#> $`Sonar-example`$classif.kknn.tuned
#> $`Sonar-example`$classif.kknn.tuned[[1]]
#> Tune result:
#> Op. pars: k=4
#> mmce.test.mean=0.11
#> 
#> $`Sonar-example`$classif.kknn.tuned[[2]]
#> Tune result:
#> Op. pars: k=3
#> mmce.test.mean=0.0667
\end{lstlisting}

As for several other accessor functions a clearer representation as
\href{http://www.rdocumentation.org/packages/base/functions/data.frame.html}{data.frame}
can be achieved by setting \lstinline!as.df = TRUE!.

\begin{lstlisting}[language=R]
getBMRTuneResults(res, as.df = TRUE)
#>         task.id         learner.id iter   C sigma mmce.test.mean  k
#> 1  iris-example classif.ksvm.tuned    1 0.5   0.5     0.05882353 NA
#> 2  iris-example classif.kknn.tuned    1  NA    NA     0.04901961  3
#> 3 Sonar-example classif.ksvm.tuned    1 1.0   2.0     0.34285714 NA
#> 4 Sonar-example classif.ksvm.tuned    2 2.0   0.5     0.20000000 NA
#> 5 Sonar-example classif.kknn.tuned    1  NA    NA     0.10952381  4
#> 6 Sonar-example classif.kknn.tuned    2  NA    NA     0.06666667  3
\end{lstlisting}

It is also possible to extract the tuning results for individual tasks
and learners and, as shown in earlier examples, inspect the
\href{http://www.rdocumentation.org/packages/ParamHelpers/functions/OptPath.html}{optimization
path}.

\begin{lstlisting}[language=R]
tune.res = getBMRTuneResults(res, task.ids = "Sonar-example", learner.ids = "classif.ksvm.tuned",
  as.df = TRUE)
tune.res
#>         task.id         learner.id iter C sigma mmce.test.mean
#> 1 Sonar-example classif.ksvm.tuned    1 1   2.0      0.3428571
#> 2 Sonar-example classif.ksvm.tuned    2 2   0.5      0.2000000

getNestedTuneResultsOptPathDf(res$results[["Sonar-example"]][["classif.ksvm.tuned"]])
\end{lstlisting}

\paragraph{Example 2: One task, two learners, feature
selection}\label{example-2-one-task-two-learners-feature-selection}

Let's see how we can do \protect\hyperlink{feature-selection}{feature
selection} in a benchmark experiment:

\begin{lstlisting}[language=R]
### Feature selection in inner resampling loop
ctrl = makeFeatSelControlSequential(method = "sfs")
inner = makeResampleDesc("Subsample", iters = 2)
lrn = makeFeatSelWrapper("regr.lm", resampling = inner, control = ctrl, show.info = FALSE)

### Learners
lrns = list(makeLearner("regr.rpart"), lrn)

### Outer resampling loop
outer = makeResampleDesc("Subsample", iters = 2)
res = benchmark(tasks = bh.task, learners = lrns, resampling = outer, show.info = FALSE)

res
#>                 task.id      learner.id mse.test.mean
#> 1 BostonHousing-example      regr.rpart      25.86232
#> 2 BostonHousing-example regr.lm.featsel      25.07465
\end{lstlisting}

The selected features can be extracted by function
\href{http://www.rdocumentation.org/packages/mlr/functions/getBMRFeatSelResults.html}{getBMRFeatSelResults}.

\begin{lstlisting}[language=R]
getBMRFeatSelResults(res)
#> $`BostonHousing-example`
#> $`BostonHousing-example`$regr.rpart
#> NULL
#> 
#> $`BostonHousing-example`$regr.lm.featsel
#> $`BostonHousing-example`$regr.lm.featsel[[1]]
#> FeatSel result:
#> Features (8): crim, zn, chas, nox, rm, dis, ptratio, lstat
#> mse.test.mean=26.7
#> 
#> $`BostonHousing-example`$regr.lm.featsel[[2]]
#> FeatSel result:
#> Features (10): crim, zn, nox, rm, dis, rad, tax, ptratio, b, lstat
#> mse.test.mean=24.3
\end{lstlisting}

You can access results for individual learners and tasks and inspect
them further.

\begin{lstlisting}[language=R]
feats = getBMRFeatSelResults(res, learner.id = "regr.lm.featsel")
feats = feats$`BostonHousing-example`$`regr.lm.featsel`

### Selected features in the first outer resampling iteration
feats[[1]]$x
#> [1] "crim"    "zn"      "chas"    "nox"     "rm"      "dis"     "ptratio"
#> [8] "lstat"

### Resampled performance of the selected feature subset on the first inner training set
feats[[1]]$y
#> mse.test.mean 
#>      26.72574
\end{lstlisting}

As for tuning, you can extract the optimization paths. The resulting
\href{http://www.rdocumentation.org/packages/base/functions/data.frame.html}{data.frame}s
contain, among others, binary columns for all features, indicating if
they were included in the linear regression model, and the corresponding
performances.
\href{http://www.rdocumentation.org/packages/mlr/functions/analyzeFeatSelResult.html}{analyzeFeatSelResult}
gives a clearer overview.

\begin{lstlisting}[language=R]
opt.paths = lapply(feats, function(x) as.data.frame(x$opt.path))
head(opt.paths[[1]])
#>   crim zn indus chas nox rm age dis rad tax ptratio b lstat mse.test.mean
#> 1    0  0     0    0   0  0   0   0   0   0       0 0     0      90.16159
#> 2    1  0     0    0   0  0   0   0   0   0       0 0     0      82.85880
#> 3    0  1     0    0   0  0   0   0   0   0       0 0     0      79.55202
#> 4    0  0     1    0   0  0   0   0   0   0       0 0     0      70.02071
#> 5    0  0     0    1   0  0   0   0   0   0       0 0     0      86.93409
#> 6    0  0     0    0   1  0   0   0   0   0       0 0     0      76.32457
#>   dob eol error.message exec.time
#> 1   1   2          <NA>     0.014
#> 2   2   2          <NA>     0.021
#> 3   2   2          <NA>     0.021
#> 4   2   2          <NA>     0.020
#> 5   2   2          <NA>     0.022
#> 6   2   2          <NA>     0.020

analyzeFeatSelResult(feats[[1]])
#> Features         : 8
#> Performance      : mse.test.mean=26.7
#> crim, zn, chas, nox, rm, dis, ptratio, lstat
#> 
#> Path to optimum:
#> - Features:    0  Init   :                       Perf = 90.162  Diff: NA  *
#> - Features:    1  Add    : lstat                 Perf = 42.646  Diff: 47.515  *
#> - Features:    2  Add    : ptratio               Perf = 34.52  Diff: 8.1263  *
#> - Features:    3  Add    : rm                    Perf = 30.454  Diff: 4.066  *
#> - Features:    4  Add    : dis                   Perf = 29.405  Diff: 1.0495  *
#> - Features:    5  Add    : nox                   Perf = 28.059  Diff: 1.3454  *
#> - Features:    6  Add    : chas                  Perf = 27.334  Diff: 0.72499  *
#> - Features:    7  Add    : zn                    Perf = 26.901  Diff: 0.43296  *
#> - Features:    8  Add    : crim                  Perf = 26.726  Diff: 0.17558  *
#> 
#> Stopped, because no improving feature was found.
\end{lstlisting}

\paragraph{Example 3: One task, two learners, feature filtering with
tuning}\label{example-3-one-task-two-learners-feature-filtering-with-tuning}

Here is a minimal example for feature filtering with tuning of the
feature subset size.

\begin{lstlisting}[language=R]
### Feature filtering with tuning in the inner resampling loop
lrn = makeFilterWrapper(learner = "regr.lm", fw.method = "chi.squared")
ps = makeParamSet(makeDiscreteParam("fw.abs", values = seq_len(getTaskNFeats(bh.task))))
ctrl = makeTuneControlGrid()
inner = makeResampleDesc("CV", iter = 2)
lrn = makeTuneWrapper(lrn, resampling = inner, par.set = ps, control = ctrl,
  show.info = FALSE)

### Learners
lrns = list(makeLearner("regr.rpart"), lrn)

### Outer resampling loop
outer = makeResampleDesc("Subsample", iter = 3)
res = benchmark(tasks = bh.task, learners = lrns, resampling = outer, show.info = FALSE)

res
#>                 task.id             learner.id mse.test.mean
#> 1 BostonHousing-example             regr.rpart      22.11687
#> 2 BostonHousing-example regr.lm.filtered.tuned      23.76666
\end{lstlisting}

\begin{lstlisting}[language=R]
### Performances on individual outer test data sets
getBMRPerformances(res, as.df = TRUE)
#>                 task.id             learner.id iter      mse
#> 1 BostonHousing-example             regr.rpart    1 23.55486
#> 2 BostonHousing-example             regr.rpart    2 20.03453
#> 3 BostonHousing-example             regr.rpart    3 22.76121
#> 4 BostonHousing-example regr.lm.filtered.tuned    1 27.51086
#> 5 BostonHousing-example regr.lm.filtered.tuned    2 24.87820
#> 6 BostonHousing-example regr.lm.filtered.tuned    3 18.91091
\end{lstlisting}

\subsection{Cost-Sensitive
Classification}\label{cost-sensitive-classification-1}

In \emph{regular classification} the aim is to minimize the
misclassification rate and thus all types of misclassification errors
are deemed equally severe. A more general setting is
\emph{cost-sensitive classification} where the costs caused by different
kinds of errors are not assumed to be equal and the objective is to
minimize the expected costs.

In case of \emph{class-dependent costs} the costs depend on the true and
predicted class label. The costs \(c(k, l)\) for predicting class \(k\)
if the true label is \(l\) are usually organized into a \(K \times K\)
cost matrix where \(K\) is the number of classes. Naturally, it is
assumed that the cost of predicting the correct class label \(y\) is
minimal (that is \(c(y, y) \leq c(k, y)\) for all \(k = 1,\ldots,K\)).

A further generalization of this scenario are \emph{example-dependent
misclassification costs} where each example \((x, y)\) is coupled with
an individual cost vector of length \(K\). Its \(k\)-th component
expresses the cost of assigning \(x\) to class \(k\). A real-world
example is fraud detection where the costs do not only depend on the
true and predicted status fraud/non-fraud, but also on the amount of
money involved in each case. Naturally, the cost of predicting the true
class label \(y\) is assumed to be minimum. The true class labels are
redundant information, as they can be easily inferred from the cost
vectors. Moreover, given the cost vector, the expected costs do not
depend on the true class label \(y\). The classification problem is
therefore completely defined by the feature values \(x\) and the
corresponding cost vectors.

In the following we show ways to handle cost-sensitive classification
problems in \href{http://www.rdocumentation.org/packages/mlr/}{mlr}.
Some of the functionality is currently experimental, and there may be
changes in the future.

\subsubsection{Class-dependent misclassification
costs}\label{class-dependent-misclassification-costs}

There are some classification methods that can accomodate
misclassification costs directly. One example is
\href{http://www.rdocumentation.org/packages/rpart/functions/rpart.html}{rpart}.

Alternatively, we can use cost-insensitive methods and manipulate the
predictions or the training data in order to take misclassification
costs into account.
\href{http://www.rdocumentation.org/packages/mlr/}{mlr} supports
\emph{thresholding} and \emph{rebalancing}.

\begin{enumerate}
\def\labelenumi{\arabic{enumi}.}
\item
  \textbf{Thresholding}: The thresholds used to turn posterior
  probabilities into class labels are chosen such that the costs are
  minimized. This requires a
  \href{http://www.rdocumentation.org/packages/mlr/functions/makeLearner.html}{Learner}
  that can predict posterior probabilities. During training the costs
  are not taken into account.
\item
  \textbf{Rebalancing}: The idea is to change the proportion of the
  classes in the training data set in order to account for costs during
  training, either by \emph{weighting} or by \emph{sampling}.
  Rebalancing does not require that the
  \href{http://www.rdocumentation.org/packages/mlr/functions/makeLearner.html}{Learner}
  can predict probabilities.

  \begin{enumerate}
  \def\labelenumii{\roman{enumii}.}
  \item
    For \emph{weighting} we need a
    \href{http://www.rdocumentation.org/packages/mlr/functions/makeLearner.html}{Learner}
    that supports class weights or observation weights.
  \item
    If the
    \href{http://www.rdocumentation.org/packages/mlr/functions/makeLearner.html}{Learner}
    cannot deal with weights the proportion of classes can be changed by
    \emph{over-} and \emph{undersampling}.
  \end{enumerate}
\end{enumerate}

We start with binary classification problems and afterwards deal with
multi-class problems.

\paragraph{Binary classification
problems}\label{binary-classification-problems}

The positive and negative classes are labeled \(1\) and \(-1\),
respectively, and we consider the following cost matrix where the rows
indicate true classes and the columns predicted classes:

\begin{longtable}[]{@{}ccc@{}}
\toprule
true/pred. & \(+1\) & \(-1\)\tabularnewline
\(+1\) & \(c(+1,+1)\) & \(c(-1,+1)\)\tabularnewline
\(-1\) & \(c(+1,-1)\) & \(c(-1,-1)\)\tabularnewline
\bottomrule
\end{longtable}

Often, the diagonal entries are zero or the cost matrix is rescaled to
achieve zeros in the diagonal (see for example
\href{http://machinelearning.org/archive/icml2008/papers/150.pdf}{O'Brien
et al, 2008}).

A well-known cost-sensitive classification problem is posed by the
\href{http://www.rdocumentation.org/packages/caret/functions/GermanCredit.html}{German
Credit data set} (see also the
\href{https://archive.ics.uci.edu/ml/datasets/Statlog+(German+Credit+Data)}{UCI
Machine Learning Repository}). The corresponding cost matrix (though
\href{http://www.cs.iastate.edu/~honavar/elkan.pdf}{Elkan (2001)} argues
that this matrix is economically unreasonable) is given as:

\begin{longtable}[]{@{}lcc@{}}
\toprule
true/pred. & Bad & Good\tabularnewline
Bad & 0 & 5\tabularnewline
Good & 1 & 0\tabularnewline
\bottomrule
\end{longtable}

As in the table above, the rows indicate true and the columns predicted
classes.

In case of class-dependent costs it is sufficient to generate an
ordinary
\href{http://www.rdocumentation.org/packages/mlr/functions/Task.html}{ClassifTask}.
A
\href{http://www.rdocumentation.org/packages/mlr/functions/Task.html}{CostSensTask}
is only needed if the costs are example-dependent. In the \textbf{R}
code below we create the
\href{http://www.rdocumentation.org/packages/mlr/functions/Task.html}{ClassifTask},
remove two constant features from the data set and generate the cost
matrix. Per default, Bad is the positive class.

\begin{lstlisting}[language=R]
data(GermanCredit, package = "caret")
credit.task = makeClassifTask(data = GermanCredit, target = "Class")
credit.task = removeConstantFeatures(credit.task)
#> Removing 2 columns: Purpose.Vacation,Personal.Female.Single

credit.task
#> Supervised task: GermanCredit
#> Type: classif
#> Target: Class
#> Observations: 1000
#> Features:
#> numerics  factors  ordered 
#>       59        0        0 
#> Missings: FALSE
#> Has weights: FALSE
#> Has blocking: FALSE
#> Classes: 2
#>  Bad Good 
#>  300  700 
#> Positive class: Bad

costs = matrix(c(0, 1, 5, 0), 2)
colnames(costs) = rownames(costs) = getTaskClassLevels(credit.task)
costs
#>      Bad Good
#> Bad    0    5
#> Good   1    0
\end{lstlisting}

\subparagraph{1. Thresholding}\label{thresholding}

We start by fitting a
\href{http://www.rdocumentation.org/packages/nnet/functions/multinom.html}{logistic
regression model} to the
\href{http://www.rdocumentation.org/packages/caret/functions/GermanCredit.html}{German
credit data set} and predict posterior probabilities.

\begin{lstlisting}[language=R]
### Train and predict posterior probabilities
lrn = makeLearner("classif.multinom", predict.type = "prob", trace = FALSE)
mod = train(lrn, credit.task)
pred = predict(mod, task = credit.task)
pred
#> Prediction: 1000 observations
#> predict.type: prob
#> threshold: Bad=0.50,Good=0.50
#> time: 0.01
#>   id truth   prob.Bad prob.Good response
#> 1  1  Good 0.03525092 0.9647491     Good
#> 2  2   Bad 0.63222363 0.3677764      Bad
#> 3  3  Good 0.02807414 0.9719259     Good
#> 4  4  Good 0.25182703 0.7481730     Good
#> 5  5   Bad 0.75193275 0.2480673      Bad
#> 6  6  Good 0.26230149 0.7376985     Good
#> ... (1000 rows, 5 cols)
\end{lstlisting}

The default thresholds for both classes are 0.5. But according to the
cost matrix we should predict class Good only if we are very sure that
Good is indeed the correct label. Therefore we should increase the
threshold for class Good and decrease the threshold for class Bad.

i. Theoretical thresholding

The theoretical threshold for the \emph{positive} class can be
calculated from the cost matrix as
\[t^* = \frac{c(+1,-1) - c(-1,-1)}{c(+1,-1) - c(+1,+1) + c(-1,+1) - c(-1,-1)}.\]
For more details see
\href{http://www.cs.iastate.edu/~honavar/elkan.pdf}{Elkan (2001)}.

Below the theoretical threshold for the
\href{http://www.rdocumentation.org/packages/caret/functions/GermanCredit.html}{German
credit example} is calculated and used to predict class labels. Since
the diagonal of the cost matrix is zero the formula given above
simplifies accordingly.

\begin{lstlisting}[language=R]
### Calculate the theoretical threshold for the positive class
th = costs[2,1]/(costs[2,1] + costs[1,2])
th
#> [1] 0.1666667
\end{lstlisting}

As you may recall you can change thresholds in
\href{http://www.rdocumentation.org/packages/mlr/}{mlr} either before
training by using the \lstinline!predict.threshold! option of
\href{http://www.rdocumentation.org/packages/mlr/functions/makeLearner.html}{makeLearner}
or after prediction by calling
\href{http://www.rdocumentation.org/packages/mlr/functions/setThreshold.html}{setThreshold}
on the
\href{http://www.rdocumentation.org/packages/mlr/functions/Prediction.html}{Prediction}
object.

As we already have a prediction we use the
\href{http://www.rdocumentation.org/packages/mlr/functions/setThreshold.html}{setThreshold}
function. It returns an altered
\href{http://www.rdocumentation.org/packages/mlr/functions/Prediction.html}{Prediction}
object with class predictions for the theoretical threshold.

\begin{lstlisting}[language=R]
### Predict class labels according to the theoretical threshold
pred.th = setThreshold(pred, th)
pred.th
#> Prediction: 1000 observations
#> predict.type: prob
#> threshold: Bad=0.17,Good=0.83
#> time: 0.01
#>   id truth   prob.Bad prob.Good response
#> 1  1  Good 0.03525092 0.9647491     Good
#> 2  2   Bad 0.63222363 0.3677764      Bad
#> 3  3  Good 0.02807414 0.9719259     Good
#> 4  4  Good 0.25182703 0.7481730      Bad
#> 5  5   Bad 0.75193275 0.2480673      Bad
#> 6  6  Good 0.26230149 0.7376985      Bad
#> ... (1000 rows, 5 cols)
\end{lstlisting}

In order to calculate the average costs over the entire data set we
first need to create a new performance
\href{http://www.rdocumentation.org/packages/mlr/functions/makeMeasure.html}{Measure}.
This can be done through function
\href{http://www.rdocumentation.org/packages/mlr/functions/makeCostMeasure.html}{makeCostMeasure}.
It is expected that the rows of the cost matrix indicate true and the
columns predicted class labels.

\begin{lstlisting}[language=R]
credit.costs = makeCostMeasure(id = "credit.costs", name = "Credit costs", costs = costs,
  best = 0, worst = 5)
credit.costs
#> Name: Credit costs
#> Performance measure: credit.costs
#> Properties: classif,classif.multi,req.pred,req.truth,predtype.response,predtype.prob
#> Minimize: TRUE
#> Best: 0; Worst: 5
#> Aggregated by: test.mean
#> Note:
\end{lstlisting}

Then the average costs can be computed by function
\href{http://www.rdocumentation.org/packages/mlr/functions/performance.html}{performance}.
Below we compare the average costs and the error rate
(\protect\hyperlink{implemented-performance-measures}{mmce}) of the
learning algorithm with both default thresholds 0.5 and theoretical
thresholds.

\begin{lstlisting}[language=R]
### Performance with default thresholds 0.5
performance(pred, measures = list(credit.costs, mmce))
#> credit.costs         mmce 
#>        0.774        0.214

### Performance with theoretical thresholds
performance(pred.th, measures = list(credit.costs, mmce))
#> credit.costs         mmce 
#>        0.478        0.346
\end{lstlisting}

These performance values may be overly optimistic as we used the same
data set for training and prediction, and resampling strategies should
be preferred. In the \textbf{R} code below we make use of the
\lstinline!predict.threshold! argument of
\href{http://www.rdocumentation.org/packages/mlr/functions/makeLearner.html}{makeLearner}
to set the threshold before doing a 3-fold cross-validation on the
\href{http://www.rdocumentation.org/packages/mlr/functions/credit.task.html}{credit.task}.
Note that we create a
\href{http://www.rdocumentation.org/packages/mlr/functions/makeResampleInstance.html}{ResampleInstance}
(\lstinline!rin!) that is used throughout the next several code chunks
to get comparable performance values.

\begin{lstlisting}[language=R]
### Cross-validated performance with theoretical thresholds
rin = makeResampleInstance("CV", iters = 3, task = credit.task)
lrn = makeLearner("classif.multinom", predict.type = "prob", predict.threshold = th, trace = FALSE)
r = resample(lrn, credit.task, resampling = rin, measures = list(credit.costs, mmce), show.info = FALSE)
r
#> Resample Result
#> Task: GermanCredit
#> Learner: classif.multinom
#> Aggr perf: credit.costs.test.mean=0.558,mmce.test.mean=0.362
#> Runtime: 0.16464
\end{lstlisting}

If we are also interested in the cross-validated performance for the
default threshold values we can call
\href{http://www.rdocumentation.org/packages/mlr/functions/setThreshold.html}{setThreshold}
on the
\href{http://www.rdocumentation.org/packages/mlr/functions/ResamplePrediction.html}{resample
prediction} \lstinline!r$pred!.

\begin{lstlisting}[language=R]
### Cross-validated performance with default thresholds
performance(setThreshold(r$pred, 0.5), measures = list(credit.costs, mmce))
#> credit.costs         mmce 
#>    0.8521695    0.2480205
\end{lstlisting}

Theoretical thresholding is only reliable if the predicted posterior
probabilities are correct. If there is bias the thresholds have to be
shifted accordingly.

Useful in this regard is function
\href{http://www.rdocumentation.org/packages/mlr/functions/plotThreshVsPerf.html}{plotThreshVsPerf}
that you can use to plot the average costs as well as any other
performance measure versus possible threshold values for the positive
class in \([0,1]\). The underlying data is generated by
\href{http://www.rdocumentation.org/packages/mlr/functions/generateThreshVsPerfData.html}{generateThreshVsPerfData}.

The following plots show the cross-validated costs and error rate
(\protect\hyperlink{implemented-performance-measures}{mmce}). The
theoretical threshold \lstinline!th! calculated above is indicated by
the vertical line. As you can see from the left-hand plot the
theoretical threshold seems a bit large.

\begin{lstlisting}[language=R]
d = generateThreshVsPerfData(r, measures = list(credit.costs, mmce))
plotThreshVsPerf(d, mark.th = th)
\end{lstlisting}

\includegraphics{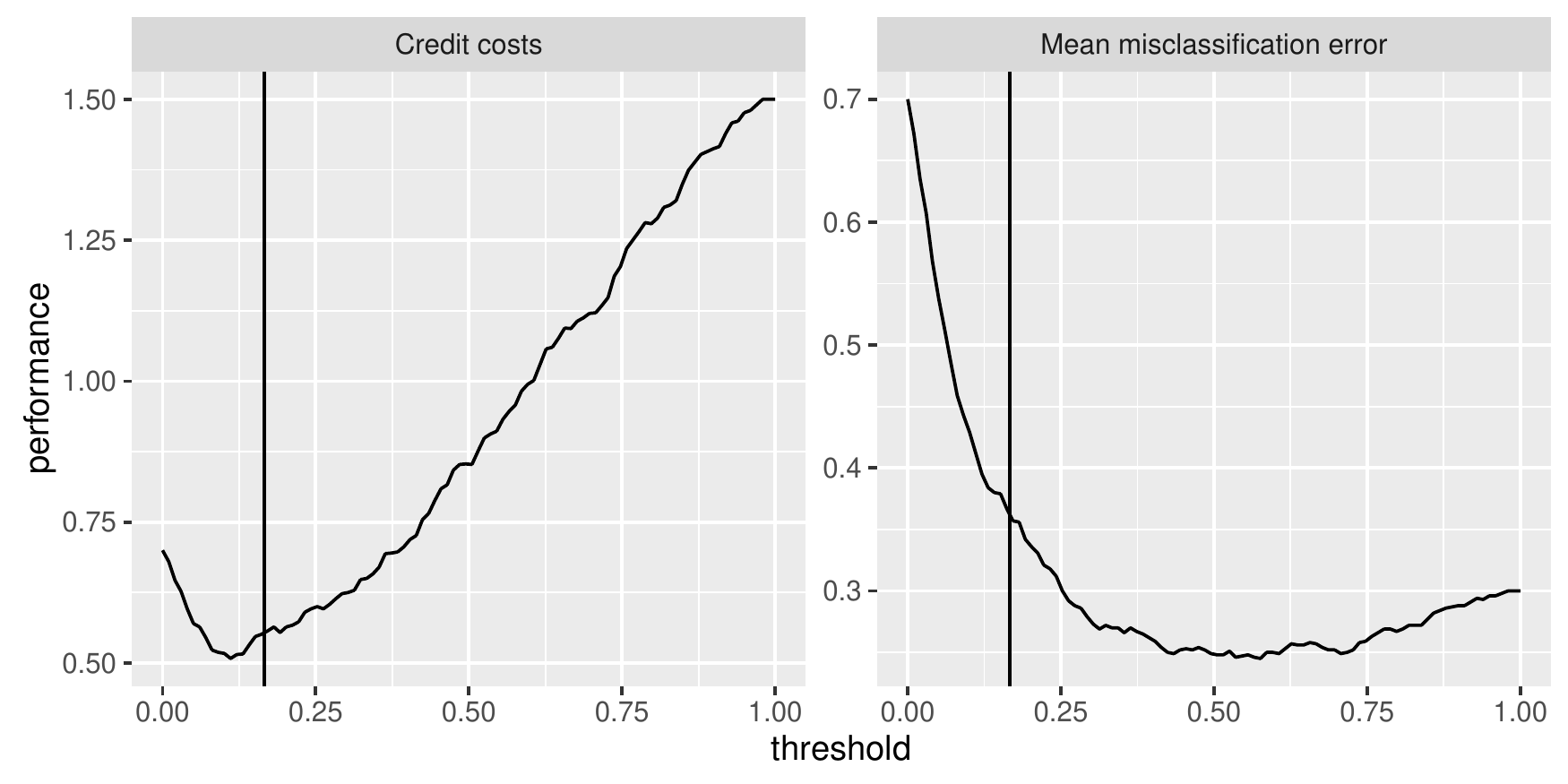}

ii. Empirical thresholding

The idea of \emph{empirical thresholding} (see
\href{http://sun0.cs.uca.edu/~ssheng/papers/AAAI06a.pdf}{Sheng and Ling,
2006}) is to select cost-optimal threshold values for a given learning
method based on the training data. In contrast to \emph{theoretical
thresholding} it suffices if the estimated posterior probabilities are
order-correct.

In order to determine optimal threshold values you can use
\href{http://www.rdocumentation.org/packages/mlr/}{mlr}'s function
\href{http://www.rdocumentation.org/packages/mlr/functions/tuneThreshold.html}{tuneThreshold}.
As tuning the threshold on the complete training data set can lead to
overfitting, you should use resampling strategies. Below we perform
3-fold cross-validation and use
\href{http://www.rdocumentation.org/packages/mlr/functions/tuneThreshold.html}{tuneThreshold}
to calculate threshold values with lowest average costs over the 3 test
data sets.

\begin{lstlisting}[language=R]
lrn = makeLearner("classif.multinom", predict.type = "prob", trace = FALSE)

### 3-fold cross-validation
r = resample(lrn, credit.task, resampling = rin, measures = list(credit.costs, mmce), show.info = FALSE)
r
#> Resample Result
#> Task: GermanCredit
#> Learner: classif.multinom
#> Aggr perf: credit.costs.test.mean=0.852,mmce.test.mean=0.248
#> Runtime: 0.165735

### Tune the threshold based on the predicted probabilities on the 3 test data sets
tune.res = tuneThreshold(pred = r$pred, measure = credit.costs)
tune.res
#> $th
#> [1] 0.1115426
#> 
#> $perf
#> credit.costs 
#>     0.507004
\end{lstlisting}

\href{http://www.rdocumentation.org/packages/mlr/functions/tuneThreshold.html}{tuneThreshold}
returns the optimal threshold value for the positive class and the
corresponding performance. As expected the tuned threshold is smaller
than the theoretical threshold.

\subparagraph{2. Rebalancing}\label{rebalancing}

In order to minimize the average costs, observations from the less
costly class should be given higher importance during training. This can
be achieved by \emph{weighting} the classes, provided that the learner
under consideration has a `class weights' or an `observation weights'
argument. To find out which learning methods support either type of
weights have a look at the \protect\hyperlink{integrated-learners}{list
of integrated learners} in the Appendix or use
\href{http://www.rdocumentation.org/packages/mlr/functions/listLearners.html}{listLearners}.

\begin{lstlisting}[language=R]
### Learners that accept observation weights
listLearners("classif", properties = "weights")[c("class", "package")]
#>                         class         package
#> 1              classif.avNNet            nnet
#> 2            classif.binomial           stats
#> 3          classif.blackboost    mboost,party
#> 4                 classif.C50             C50
#> 5             classif.cforest           party
#> 6               classif.ctree           party
#> 7            classif.cvglmnet          glmnet
#> 8          classif.extraTrees      extraTrees
#> 9                 classif.gbm             gbm
#> 10           classif.glmboost          mboost
#> 11             classif.glmnet          glmnet
#> 12   classif.h2o.deeplearning             h2o
#> 13            classif.h2o.glm             h2o
#> 14             classif.logreg           stats
#> 15           classif.multinom            nnet
#> 16               classif.nnet            nnet
#> 17                classif.plr         stepPlr
#> 18             classif.probit           stats
#> 19    classif.randomForestSRC randomForestSRC
#> 20 classif.randomForestSRCSyn randomForestSRC
#> 21              classif.rpart           rpart
#> 22            classif.xgboost         xgboost

### Learners that can deal with class weights
listLearners("classif", properties = "class.weights")[c("class", "package")]
#>                            class      package
#> 1                   classif.ksvm      kernlab
#> 2       classif.LiblineaRL1L2SVC    LiblineaR
#> 3      classif.LiblineaRL1LogReg    LiblineaR
#> 4       classif.LiblineaRL2L1SVC    LiblineaR
#> 5      classif.LiblineaRL2LogReg    LiblineaR
#> 6         classif.LiblineaRL2SVC    LiblineaR
#> 7 classif.LiblineaRMultiClassSVC    LiblineaR
#> 8           classif.randomForest randomForest
#> 9                    classif.svm        e1071
\end{lstlisting}

Alternatively, \emph{over- and undersampling} techniques can be used.

i. Weighting

Just as \emph{theoretical thresholds}, \emph{theoretical weights} can be
calculated from the cost matrix. If \(t\) indicates the target threshold
and \(t_0\) the original threshold for the positive class the proportion
of observations in the positive class has to be multiplied by
\[\frac{1-t}{t} \frac{t_0}{1-t_0}.\] Alternatively, the proportion of
observations in the negative class can be multiplied by the inverse. A
proof is given by
\href{http://www.cs.iastate.edu/~honavar/elkan.pdf}{Elkan (2001)}.

In most cases, the original threshold is \(t_0 = 0.5\) and thus the
second factor vanishes. If additionally the target threshold \(t\)
equals the theoretical threshold \(t^*\) the proportion of observations
in the positive class has to be multiplied by
\[\frac{1-t^*}{t^*} = \frac{c(-1,+1) - c(+1,+1)}{c(+1,-1) - c(-1,-1)}.\]

For the \href{\&caret:GermanCredit}{credit example} the theoretical
threshold corresponds to a weight of 5 for the positive class.

\begin{lstlisting}[language=R]
### Weight for positive class corresponding to theoretical treshold
w = (1 - th)/th
w
#> [1] 5
\end{lstlisting}

A unified and convenient way to assign class weights to a
\href{http://www.rdocumentation.org/packages/mlr/functions/makeLearner.html}{Learner}
(and tune them) is provided by function
\href{http://www.rdocumentation.org/packages/mlr/functions/makeWeightedClassesWrapper.html}{makeWeightedClassesWrapper}.
The class weights are specified using argument \lstinline!wcw.weight!.
For learners that support observation weights a suitable weight vector
is then generated internally during training or resampling. If the
learner can deal with class weights, the weights are basically passed on
to the appropriate learner parameter. The advantage of using the wrapper
in this case is the unified way to specify the class weights.

Below is an example using learner \lstinline!"classif.multinom"!
(\href{http://www.rdocumentation.org/packages/nnet/functions/multinom.html}{multinom}
from package \href{http://www.rdocumentation.org/packages/nnet/}{nnet})
which accepts observation weights. For binary classification problems it
is sufficient to specify the weight \lstinline!w! for the positive
class. The negative class then automatically receives weight 1.

\begin{lstlisting}[language=R]
### Weighted learner
lrn = makeLearner("classif.multinom", trace = FALSE)
lrn = makeWeightedClassesWrapper(lrn, wcw.weight = w)
lrn
#> Learner weightedclasses.classif.multinom from package nnet
#> Type: classif
#> Name: ; Short name: 
#> Class: WeightedClassesWrapper
#> Properties: twoclass,multiclass,numerics,factors,prob
#> Predict-Type: response
#> Hyperparameters: trace=FALSE,wcw.weight=5

r = resample(lrn, credit.task, rin, measures = list(credit.costs, mmce), show.info = FALSE)
r
#> Resample Result
#> Task: GermanCredit
#> Learner: weightedclasses.classif.multinom
#> Aggr perf: credit.costs.test.mean=0.526,mmce.test.mean=0.346
#> Runtime: 0.2037
\end{lstlisting}

For classification methods like \lstinline!"classif.ksvm"! (the support
vector machine
\href{http://www.rdocumentation.org/packages/kernlab/functions/ksvm.html}{ksvm}
in package
\href{http://www.rdocumentation.org/packages/kernlab/}{kernlab}) that
support class weights you can pass them directly.

\begin{lstlisting}[language=R]
lrn = makeLearner("classif.ksvm", class.weights = c(Bad = w, Good = 1))
\end{lstlisting}

Or, more conveniently, you can again use
\href{http://www.rdocumentation.org/packages/mlr/functions/makeWeightedClassesWrapper.html}{makeWeightedClassesWrapper}.

\begin{lstlisting}[language=R]
lrn = makeWeightedClassesWrapper("classif.ksvm", wcw.weight = w)
r = resample(lrn, credit.task, rin, measures = list(credit.costs, mmce), show.info = FALSE)
r
#> Resample Result
#> Task: GermanCredit
#> Learner: weightedclasses.classif.ksvm
#> Aggr perf: credit.costs.test.mean=0.575,mmce.test.mean=0.311
#> Runtime: 0.293009
\end{lstlisting}

Just like the theoretical threshold, the theoretical weights may not
always be suitable, therefore you can tune the weight for the positive
class as shown in the following example. Calculating the theoretical
weight beforehand may help to narrow down the search interval.

\begin{lstlisting}[language=R]
lrn = makeLearner("classif.multinom", trace = FALSE)
lrn = makeWeightedClassesWrapper(lrn)
ps = makeParamSet(makeDiscreteParam("wcw.weight", seq(4, 12, 0.5)))
ctrl = makeTuneControlGrid()
tune.res = tuneParams(lrn, credit.task, resampling = rin, par.set = ps,
  measures = list(credit.costs, mmce), control = ctrl, show.info = FALSE)
tune.res
#> Tune result:
#> Op. pars: wcw.weight=7.5
#> credit.costs.test.mean=0.501,mmce.test.mean=0.381

as.data.frame(tune.res$opt.path)[1:3]
#>    wcw.weight credit.costs.test.mean mmce.test.mean
#> 1           4              0.5650291      0.3330127
#> 2         4.5              0.5550251      0.3430167
#> 3           5              0.5260320      0.3460197
#> 4         5.5              0.5130070      0.3530147
#> 5           6              0.5160100      0.3640137
#> 6         6.5              0.5160160      0.3720157
#> 7           7              0.5040250      0.3760167
#> 8         7.5              0.5010040      0.3810038
#> 9           8              0.5100130      0.3900128
#> 10        8.5              0.5100070      0.3940108
#> 11          9              0.5110080      0.4030078
#> 12        9.5              0.5160130      0.4080128
#> 13         10              0.5260140      0.4180138
#> 14       10.5              0.5240060      0.4200098
#> 15         11              0.5319991      0.4280029
#> 16       11.5              0.5289901      0.4330019
#> 17         12              0.5249801      0.4369999
\end{lstlisting}

ii. Over- and undersampling

If the
\href{http://www.rdocumentation.org/packages/mlr/functions/makeLearner.html}{Learner}
supports neither observation nor class weights the proportions of the
classes in the training data can be changed by over- or undersampling.

In the
\href{http://www.rdocumentation.org/packages/caret/functions/GermanCredit.html}{GermanCredit
data set} the positive class Bad should receive a theoretical weight of
\lstinline!w = (1 - th)/th = 5!. This can be achieved by oversampling
class Bad with a \lstinline!rate! of 5 or by undersampling class Good
with a \lstinline!rate! of 1/5 (using functions
\href{http://www.rdocumentation.org/packages/mlr/functions/oversample.html}{oversample}
or
\href{http://www.rdocumentation.org/packages/mlr/functions/oversample.html}{undersample}).

\begin{lstlisting}[language=R]
credit.task.over = oversample(credit.task, rate = w, cl = "Bad")
lrn = makeLearner("classif.multinom", trace = FALSE)
mod = train(lrn, credit.task.over)
pred = predict(mod, task = credit.task)
performance(pred, measures = list(credit.costs, mmce))
#> credit.costs         mmce 
#>        0.439        0.323
\end{lstlisting}

Note that in the above example the learner was trained on the
oversampled task \lstinline!credit.task.over!. In order to get the
training performance on the original task predictions were calculated
for \lstinline!credit.task!.

We usually prefer resampled performance values, but simply calling
\href{http://www.rdocumentation.org/packages/mlr/functions/resample.html}{resample}
on the oversampled task does not work since predictions have to be based
on the original task. The solution is to create a wrapped
\href{http://www.rdocumentation.org/packages/mlr/functions/makeLearner.html}{Learner}
via function
\href{http://www.rdocumentation.org/packages/mlr/functions/makeUndersampleWrapper.html}{makeOversampleWrapper}.
Internally,
\href{http://www.rdocumentation.org/packages/mlr/functions/oversample.html}{oversample}
is called before training, but predictions are done on the original
data.

\begin{lstlisting}[language=R]
lrn = makeLearner("classif.multinom", trace = FALSE)
lrn = makeOversampleWrapper(lrn, osw.rate = w, osw.cl = "Bad")
lrn
#> Learner classif.multinom.oversampled from package mlr,nnet
#> Type: classif
#> Name: ; Short name: 
#> Class: OversampleWrapper
#> Properties: numerics,factors,weights,prob,twoclass,multiclass
#> Predict-Type: response
#> Hyperparameters: trace=FALSE,osw.rate=5,osw.cl=Bad

r = resample(lrn, credit.task, rin, measures = list(credit.costs, mmce), show.info = FALSE)
r
#> Resample Result
#> Task: GermanCredit
#> Learner: classif.multinom.oversampled
#> Aggr perf: credit.costs.test.mean=0.535,mmce.test.mean=0.351
#> Runtime: 0.330427
\end{lstlisting}

Of course, we can also tune the oversampling rate. For this purpose we
again have to create an
\href{http://www.rdocumentation.org/packages/mlr/functions/makeUndersampleWrapper.html}{OversampleWrapper}.
Optimal values for parameter \lstinline!osw.rate! can be obtained using
function
\href{http://www.rdocumentation.org/packages/mlr/functions/tuneParams.html}{tuneParams}.

\begin{lstlisting}[language=R]
lrn = makeLearner("classif.multinom", trace = FALSE)
lrn = makeOversampleWrapper(lrn, osw.cl = "Bad")
ps = makeParamSet(makeDiscreteParam("osw.rate", seq(3, 7, 0.25)))
ctrl = makeTuneControlGrid()
tune.res = tuneParams(lrn, credit.task, rin, par.set = ps, measures = list(credit.costs, mmce),
  control = ctrl, show.info = FALSE)
tune.res
#> Tune result:
#> Op. pars: osw.rate=6.25
#> credit.costs.test.mean=0.507,mmce.test.mean=0.355
\end{lstlisting}

\paragraph{Multi-class problems}\label{multi-class-problems}

We consider the
\href{http://www.rdocumentation.org/packages/mlbench/functions/mlbench.waveform.html}{waveform}
data set from package
\href{http://www.rdocumentation.org/packages/mlbench/}{mlbench} and add
an artificial cost matrix:

\begin{longtable}[]{@{}lccc@{}}
\toprule
true/pred. & 1 & 2 & 3\tabularnewline
1 & 0 & 30 & 80\tabularnewline
2 & 5 & 0 & 4\tabularnewline
3 & 10 & 8 & 0\tabularnewline
\bottomrule
\end{longtable}

We start by creating the
\href{http://www.rdocumentation.org/packages/mlr/functions/Task.html}{Task},
the cost matrix and the corresponding performance measure.

\begin{lstlisting}[language=R]
### Task
df = mlbench::mlbench.waveform(500)
wf.task = makeClassifTask(id = "waveform", data = as.data.frame(df), target = "classes")

### Cost matrix
costs = matrix(c(0, 5, 10, 30, 0, 8, 80, 4, 0), 3)
colnames(costs) = rownames(costs) = getTaskClassLevels(wf.task)

### Performance measure
wf.costs = makeCostMeasure(id = "wf.costs", name = "Waveform costs", costs = costs,
  best = 0, worst = 10)
\end{lstlisting}

In the multi-class case, both, \emph{thresholding} and
\emph{rebalancing} correspond to cost matrices of a certain structure
where \(c(k,l) = c(l)\) for \(k\), \(l = 1, \ldots, K\), \(k \neq l\).
This condition means that the cost of misclassifying an observation is
independent of the predicted class label (see
\href{http://homes.cs.washington.edu/~pedrod/papers/kdd99.pdf}{Domingos,
1999}). Given a cost matrix of this type, theoretical thresholds and
weights can be derived in a similar manner as in the binary case.
Obviously, the cost matrix given above does not have this special
structure.

\subparagraph{1. Thresholding}\label{thresholding-1}

Given a vector of positive threshold values as long as the number of
classes \(K\), the predicted probabilities for all classes are adjusted
by dividing them by the corresponding threshold value. Then the class
with the highest adjusted probability is predicted. This way, as in the
binary case, classes with a low threshold are preferred to classes with
a larger threshold.

Again this can be done by function
\href{http://www.rdocumentation.org/packages/mlr/functions/setThreshold.html}{setThreshold}
as shown in the following example (or alternatively by the
\lstinline!predict.threshold! option of
\href{http://www.rdocumentation.org/packages/mlr/functions/makeLearner.html}{makeLearner}).
Note that the threshold vector needs to have names that correspond to
the class labels.

\begin{lstlisting}[language=R]
lrn = makeLearner("classif.rpart", predict.type = "prob")
rin = makeResampleInstance("CV", iters = 3, task = wf.task)
r = resample(lrn, wf.task, rin, measures = list(wf.costs, mmce), show.info = FALSE)
r
#> Resample Result
#> Task: waveform
#> Learner: classif.rpart
#> Aggr perf: wf.costs.test.mean=7.02,mmce.test.mean=0.262
#> Runtime: 0.0421195

### Calculate thresholds as 1/(average costs of true classes)
th = 2/rowSums(costs)
names(th) = getTaskClassLevels(wf.task)
th
#>          1          2          3 
#> 0.01818182 0.22222222 0.11111111

pred.th = setThreshold(r$pred, threshold = th)
performance(pred.th, measures = list(wf.costs, mmce))
#>  wf.costs      mmce 
#> 5.0372268 0.3502393
\end{lstlisting}

The threshold vector \lstinline!th! in the above example is chosen
according to the average costs of the true classes 55, 4.5 and 9. More
exactly, \lstinline!th! corresponds to an artificial cost matrix of the
structure mentioned above with off-diagonal elements
\(c(2,1) = c(3,1) = 55\), \(c(1,2) = c(3,2) = 4.5\) and
\(c(1,3) = c(2,3) = 9\). This threshold vector may be not optimal but
leads to smaller total costs on the data set than the default.

ii. Empirical thresholding

As in the binary case it is possible to tune the threshold vector using
function
\href{http://www.rdocumentation.org/packages/mlr/functions/tuneThreshold.html}{tuneThreshold}.
Since the scaling of the threshold vector does not change the predicted
class labels
\href{http://www.rdocumentation.org/packages/mlr/functions/tuneThreshold.html}{tuneThreshold}
returns threshold values that lie in {[}0,1{]} and sum to unity.

\begin{lstlisting}[language=R]
tune.res = tuneThreshold(pred = r$pred, measure = wf.costs)
tune.res
#> $th
#>          1          2          3 
#> 0.01447413 0.35804444 0.62748143 
#> 
#> $perf
#> [1] 4.544369
\end{lstlisting}

For comparison we show the standardized version of the theoretically
motivated threshold vector chosen above.

\begin{lstlisting}[language=R]
th/sum(th)
#>          1          2          3 
#> 0.05172414 0.63218391 0.31609195
\end{lstlisting}

\subparagraph{2. Rebalancing}\label{rebalancing-1}

i. Weighting

In the multi-class case you have to pass a vector of weights as long as
the number of classes \(K\) to function
\href{http://www.rdocumentation.org/packages/mlr/functions/makeWeightedClassesWrapper.html}{makeWeightedClassesWrapper}.
The weight vector can be tuned using function
\href{http://www.rdocumentation.org/packages/mlr/functions/tuneParams.html}{tuneParams}.

\begin{lstlisting}[language=R]
lrn = makeLearner("classif.multinom", trace = FALSE)
lrn = makeWeightedClassesWrapper(lrn)

ps = makeParamSet(makeNumericVectorParam("wcw.weight", len = 3, lower = 0, upper = 1))
ctrl = makeTuneControlRandom()

tune.res = tuneParams(lrn, wf.task, resampling = rin, par.set = ps,
  measures = list(wf.costs, mmce), control = ctrl, show.info = FALSE)
tune.res
#> Tune result:
#> Op. pars: wcw.weight=0.836,0.225,0.05
#> wf.costs.test.mean=3.18,mmce.test.mean=0.194
\end{lstlisting}

\subsubsection{Example-dependent misclassification
costs}\label{example-dependent-misclassification-costs}

In case of example-dependent costs we have to create a special
\href{http://www.rdocumentation.org/packages/mlr/functions/Task.html}{Task}
via function
\href{http://www.rdocumentation.org/packages/mlr/functions/makeCostSensTask.html}{makeCostSensTask}.
For this purpose the feature values \(x\) and an \(n \times K\)
\lstinline!cost! matrix that contains the cost vectors for all \(n\)
examples in the data set are required.

We use the
\href{http://www.rdocumentation.org/packages/datasets/functions/iris.html}{iris}
data and generate an artificial cost matrix (see
\href{http://dx.doi.org/10.1145/1102351.1102358}{Beygelzimer et al.,
2005}).

\begin{lstlisting}[language=R]
df = iris
cost = matrix(runif(150 * 3, 0, 2000), 150) * (1 - diag(3))[df$Species,] + runif(150, 0, 10)
colnames(cost) = levels(iris$Species)
rownames(cost) = rownames(iris)
df$Species = NULL

costsens.task = makeCostSensTask(id = "iris", data = df, cost = cost)
costsens.task
#> Supervised task: iris
#> Type: costsens
#> Observations: 150
#> Features:
#> numerics  factors  ordered 
#>        4        0        0 
#> Missings: FALSE
#> Has blocking: FALSE
#> Classes: 3
#> setosa, versicolor, virginica
\end{lstlisting}

\href{http://www.rdocumentation.org/packages/mlr/}{mlr} provides several
\protect\hyperlink{wrapper}{wrappers} to turn regular classification or
regression methods into
\href{http://www.rdocumentation.org/packages/mlr/functions/makeLearner.html}{Learner}s
that can deal with example-dependent costs.

\begin{itemize}
\tightlist
\item
  \href{http://www.rdocumentation.org/packages/mlr/functions/makeCostSensClassifWrapper.html}{makeCostSensClassifWrapper}
  (wraps a classification
  \href{http://www.rdocumentation.org/packages/mlr/functions/makeLearner.html}{Learner}):
  This is a naive approach where the costs are coerced into class labels
  by choosing the class label with minimum cost for each example. Then a
  regular classification method is used.
\item
  \href{http://www.rdocumentation.org/packages/mlr/functions/makeCostSensRegrWrapper.html}{makeCostSensRegrWrapper}
  (wraps a regression
  \href{http://www.rdocumentation.org/packages/mlr/functions/makeLearner.html}{Learner}):
  An individual regression model is fitted for the costs of each class.
  In the prediction step first the costs are predicted for all classes
  and then the class with the lowest predicted costs is selected.
\item
  \href{http://www.rdocumentation.org/packages/mlr/functions/makeCostSensWeightedPairsWrapper.html}{makeCostSensWeightedPairsWrapper}
  (wraps a classification
  \href{http://www.rdocumentation.org/packages/mlr/functions/makeLearner.html}{Learner}):
  This is also known as \emph{cost-sensitive one-vs-one} (CS-OVO) and
  the most sophisticated of the currently supported methods. For each
  pair of classes, a binary classifier is fitted. For each observation
  the class label is defined as the element of the pair with minimal
  costs. During fitting, the observations are weighted with the absolute
  difference in costs. Prediction is performed by simple voting.
\end{itemize}

In the following example we use the third method. We create the wrapped
\href{http://www.rdocumentation.org/packages/mlr/functions/makeLearner.html}{Learner}
and train it on the
\href{http://www.rdocumentation.org/packages/mlr/functions/Task.html}{CostSensTask}
defined above.

\begin{lstlisting}[language=R]
lrn = makeLearner("classif.multinom", trace = FALSE)
lrn = makeCostSensWeightedPairsWrapper(lrn)
lrn
#> Learner costsens.classif.multinom from package nnet
#> Type: costsens
#> Name: ; Short name: 
#> Class: CostSensWeightedPairsWrapper
#> Properties: twoclass,multiclass,numerics,factors
#> Predict-Type: response
#> Hyperparameters: trace=FALSE

mod = train(lrn, costsens.task)
mod
#> Model for learner.id=costsens.classif.multinom; learner.class=CostSensWeightedPairsWrapper
#> Trained on: task.id = iris; obs = 150; features = 4
#> Hyperparameters: trace=FALSE
\end{lstlisting}

The models corresponding to the individual pairs can be accessed by
function
\href{http://www.rdocumentation.org/packages/mlr/functions/getLearnerModel.html}{getLearnerModel}.

\begin{lstlisting}[language=R]
getLearnerModel(mod)
#> [[1]]
#> Model for learner.id=classif.multinom; learner.class=classif.multinom
#> Trained on: task.id = feats; obs = 150; features = 4
#> Hyperparameters: trace=FALSE
#> 
#> [[2]]
#> Model for learner.id=classif.multinom; learner.class=classif.multinom
#> Trained on: task.id = feats; obs = 150; features = 4
#> Hyperparameters: trace=FALSE
#> 
#> [[3]]
#> Model for learner.id=classif.multinom; learner.class=classif.multinom
#> Trained on: task.id = feats; obs = 150; features = 4
#> Hyperparameters: trace=FALSE
\end{lstlisting}

\href{http://www.rdocumentation.org/packages/mlr/}{mlr} provides some
performance measures for example-specific cost-sensitive classification.
In the following example we calculate the mean costs of the predicted
class labels
(\protect\hyperlink{implemented-performance-measures}{meancosts}) and
the misclassification penalty
(\protect\hyperlink{implemented-performance-measures}{mcp}). The latter
measure is the average difference between the costs caused by the
predicted class labels, i.e.,
\protect\hyperlink{implemented-performance-measures}{meancosts}, and the
costs resulting from choosing the class with lowest cost for each
observation. In order to compute these measures the costs for the test
observations are required and therefore the
\href{http://www.rdocumentation.org/packages/mlr/functions/Task.html}{Task}
has to be passed to
\href{http://www.rdocumentation.org/packages/mlr/functions/performance.html}{performance}.

\begin{lstlisting}[language=R]
pred = predict(mod, task = costsens.task)
pred
#> Prediction: 150 observations
#> predict.type: response
#> threshold: 
#> time: 0.03
#>   id response
#> 1  1   setosa
#> 2  2   setosa
#> 3  3   setosa
#> 4  4   setosa
#> 5  5   setosa
#> 6  6   setosa
#> ... (150 rows, 2 cols)

performance(pred, measures = list(meancosts, mcp), task = costsens.task)
#> meancosts       mcp 
#>  129.9553  124.7782
\end{lstlisting}

\hypertarget{imbalanced-classification-problems}{\subsection{Imbalanced
Classification Problems}\label{imbalanced-classification-problems}}

In case of \emph{binary classification} strongly imbalanced classes
often lead to unsatisfactory results regarding the prediction of new
observations, especially for the small class. In this context
\emph{imbalanced classes} simply means that the number of observations
of one class (usu. positive or majority class) by far exceeds the number
of observations of the other class (usu. negative or minority class).
This setting can be observed fairly often in practice and in various
disciplines like credit scoring, fraud detection, medical diagnostics or
churn management.

Most classification methods work best when the number of observations
per class are roughly equal. The problem with \emph{imbalanced classes}
is that because of the dominance of the majority class classifiers tend
to ignore cases of the minority class as noise and therefore predict the
majority class far more often. In order to lay more weight on the cases
of the minority class, there are numerous correction methods which
tackle the \emph{imbalanced classification problem}. These methods can
generally be divided into \emph{cost- and sampling-based approaches}.
Below all methods supported by
\href{http://www.rdocumentation.org/packages/mlr/}{mlr} are introduced.

\subsubsection{Sampling-based
approaches}\label{sampling-based-approaches}

The basic idea of \emph{sampling methods} is to simply adjust the
proportion of the classes in order to increase the weight of the
minority class observations within the model.

The \emph{sampling-based approaches} can be divided further into three
different categories:

\begin{enumerate}
\def\labelenumi{\arabic{enumi}.}
\item
  \textbf{Undersampling methods}: Elimination of randomly chosen cases
  of the majority class to decrease their effect on the classifier. All
  cases of the minority class are kept.
\item
  \textbf{Oversampling methods}: Generation of additional cases (copies,
  artificial observations) of the minority class to increase their
  effect on the classifier. All cases of the majority class are kept.
\item
  \textbf{Hybrid methods}: Mixture of under- and oversampling
  strategies.
\end{enumerate}

All these methods directly access the underlying data and ``rearrange''
it. In this way the sampling is done as part of the
\emph{preprocesssing} and can therefore be combined with every
appropriate classifier.

\href{http://www.rdocumentation.org/packages/mlr/}{mlr} currently
supports the first two approaches.

\subsubsection{(Simple) over- and
undersampling}\label{simple-over--and-undersampling}

As mentioned above \emph{undersampling} always refers to the majority
class, while \emph{oversampling} affects the minority class. By the use
of \emph{undersampling}, randomly chosen observations of the majority
class are eliminated. Through (simple) \emph{oversampling} all
observations of the minority class are considered at least once when
fitting the model. In addition, exact copies of minority class cases are
created by random sampling with repetitions.

First, let's take a look at the effect for a classification
\protect\hyperlink{learning-tasks}{task}. Based on a simulated
\href{http://www.rdocumentation.org/packages/mlr/functions/Task.html}{ClassifTask}
with imbalanced classes two new tasks (\lstinline!task.over!,
\lstinline!task.under!) are created via
\href{http://www.rdocumentation.org/packages/mlr/}{mlr} functions
\href{http://www.rdocumentation.org/packages/mlr/functions/oversample.html}{oversample}
and
\href{http://www.rdocumentation.org/packages/mlr/functions/oversample.html}{undersample},
respectively.

\begin{lstlisting}[language=R]
data.imbal.train = rbind(
  data.frame(x = rnorm(100, mean = 1), class = "A"),
  data.frame(x = rnorm(5000, mean = 2), class = "B")
)
task = makeClassifTask(data = data.imbal.train, target = "class")
task.over = oversample(task, rate = 8)
task.under = undersample(task, rate = 1/8)

table(getTaskTargets(task))
#> 
#>    A    B 
#>  100 5000

table(getTaskTargets(task.over))
#> 
#>    A    B 
#>  800 5000

table(getTaskTargets(task.under))
#> 
#>   A   B 
#> 100 625
\end{lstlisting}

Please note that the \emph{undersampling rate} has to be between 0 and
1, where 1 means no undersampling and 0.5 implies a reduction of the
majority class size to 50 percent. Correspondingly, the
\emph{oversampling rate} must be greater or equal to 1, where 1 means no
oversampling and 2 would result in doubling the minority class size.

As a result the
\protect\hyperlink{evaluating-learner-performance}{performance} should
improve if the model is applied to new data.

\begin{lstlisting}[language=R]
lrn = makeLearner("classif.rpart", predict.type = "prob")
mod = train(lrn, task)
mod.over = train(lrn, task.over)
mod.under = train(lrn, task.under)
data.imbal.test = rbind(
  data.frame(x = rnorm(10, mean = 1), class = "A"),
  data.frame(x = rnorm(500, mean = 2), class = "B")
)

performance(predict(mod, newdata = data.imbal.test), measures = list(mmce, ber, auc))
#>       mmce        ber        auc 
#> 0.01960784 0.50000000 0.50000000

performance(predict(mod.over, newdata = data.imbal.test), measures = list(mmce, ber, auc))
#>       mmce        ber        auc 
#> 0.04509804 0.41500000 0.58500000

performance(predict(mod.under, newdata = data.imbal.test), measures = list(mmce, ber, auc))
#>       mmce        ber        auc 
#> 0.05098039 0.41800000 0.70550000
\end{lstlisting}

In this case the \emph{performance measure} has to be considered very
carefully. As the \emph{misclassification rate}
(\protect\hyperlink{implemented-performance-measures}{mmce}) evaluates
the overall accuracy of the predictions, the \emph{balanced error rate}
(\protect\hyperlink{implemented-performance-measures}{ber}) and
\emph{area under the ROC Curve}
(\protect\hyperlink{implemented-performance-measures}{auc}) might be
more suitable here, as the misclassifications within each class are
separately taken into account.

\paragraph{Over- and undersampling
wrappers}\label{over--and-undersampling-wrappers}

Alternatively, \href{http://www.rdocumentation.org/packages/mlr/}{mlr}
also offers the integration of over- and undersampling via a
\protect\hyperlink{wrapper}{wrapper approach}. This way over- and
undersampling can be applied to already existing
\protect\hyperlink{learners}{learners} to extend their functionality.

The example given above is repeated once again, but this time with
extended learners instead of modified tasks (see
\href{http://www.rdocumentation.org/packages/mlr/functions/makeUndersampleWrapper.html}{makeOversampleWrapper}
and
\href{http://www.rdocumentation.org/packages/mlr/functions/makeUndersampleWrapper.html}{makeUndersampleWrapper}).
Just like before the \emph{undersampling rate} has to be between 0 and
1, while the \emph{oversampling rate} has a lower boundary of 1.

\begin{lstlisting}[language=R]
lrn.over = makeOversampleWrapper(lrn, osw.rate = 8)
lrn.under = makeUndersampleWrapper(lrn, usw.rate = 1/8)
mod = train(lrn, task)
mod.over = train(lrn.over, task)
mod.under = train(lrn.under, task)

performance(predict(mod, newdata = data.imbal.test), measures = list(mmce, ber, auc))
#>       mmce        ber        auc 
#> 0.01960784 0.50000000 0.50000000

performance(predict(mod.over, newdata = data.imbal.test), measures = list(mmce, ber, auc))
#>       mmce        ber        auc 
#> 0.03333333 0.40900000 0.72020000

performance(predict(mod.under, newdata = data.imbal.test), measures = list(mmce, ber, auc))
#>       mmce        ber        auc 
#> 0.04509804 0.41500000 0.71660000
\end{lstlisting}

\paragraph{Extensions to oversampling}\label{extensions-to-oversampling}

Two extensions to (simple) oversampling are available in
\href{http://www.rdocumentation.org/packages/mlr/}{mlr}.

\subparagraph{1. SMOTE (Synthetic Minority Oversampling
Technique)}\label{smote-synthetic-minority-oversampling-technique}

As the duplicating of the minority class observations can lead to
overfitting, within \emph{SMOTE} the ``new cases'' are constructed in a
different way. For each new observation, one randomly chosen minority
class observation as well as one of its \emph{randomly chosen next
neighbours} are interpolated, so that finally a new \emph{artificial
observation} of the minority class is created. The
\href{http://www.rdocumentation.org/packages/mlr/functions/smote.html}{smote}
function in \href{http://www.rdocumentation.org/packages/mlr/}{mlr}
handles numeric as well as factor features, as the gower distance is
used for nearest neighbour calculation. The factor level of the new
artificial case is sampled from the given levels of the two input
observations.

Analogous to oversampling, \emph{SMOTE preprocessing} is possible via
modification of the task.

\begin{lstlisting}[language=R]
task.smote = smote(task, rate = 8, nn = 5)
table(getTaskTargets(task))
#> 
#>    A    B 
#>  100 5000

table(getTaskTargets(task.smote))
#> 
#>    A    B 
#>  800 5000
\end{lstlisting}

Alternatively, a new wrapped learner can be created via
\href{http://www.rdocumentation.org/packages/mlr/functions/makeSMOTEWrapper.html}{makeSMOTEWrapper}.

\begin{lstlisting}[language=R]
lrn.smote = makeSMOTEWrapper(lrn, sw.rate = 8, sw.nn = 5)
mod.smote = train(lrn.smote, task)
performance(predict(mod.smote, newdata = data.imbal.test), measures = list(mmce, ber, auc))
#>       mmce        ber        auc 
#> 0.04509804 0.41500000 0.71660000
\end{lstlisting}

By default the number of nearest neighbours considered within the
algorithm is set to 5.

\subparagraph{2. Overbagging}\label{overbagging}

Another extension of oversampling consists in the combination of
sampling with the \protect\hyperlink{generic-bagging}{bagging approach}.
For each iteration of the bagging process, minority class observations
are oversampled with a given rate in \lstinline!obw.rate!. The majority
class cases can either all be taken into account for each iteration
(\lstinline!obw.maxcl = "all"!) or bootstrapped with replacement to
increase variability between training data sets during iterations
(\lstinline!obw.maxcl = "boot"!).

The construction of the \textbf{Overbagging Wrapper} works similar to
\href{http://www.rdocumentation.org/packages/mlr/functions/makeBaggingWrapper.html}{makeBaggingWrapper}.
First an existing
\href{http://www.rdocumentation.org/packages/mlr/}{mlr} learner has to
be passed to
\href{http://www.rdocumentation.org/packages/mlr/functions/makeOverBaggingWrapper.html}{makeOverBaggingWrapper}.
The number of iterations or fitted models can be set via
\lstinline!obw.iters!.

\begin{lstlisting}[language=R]
lrn = makeLearner("classif.rpart", predict.type = "response")
obw.lrn = makeOverBaggingWrapper(lrn, obw.rate = 8, obw.iters = 3)
\end{lstlisting}

For \emph{binary classification} the prediction is based on majority
voting to create a discrete label. Corresponding probabilities are
predicted by considering the proportions of all the predicted labels.
Please note that the benefit of the sampling process is \emph{highly
dependent} on the specific learner as shown in the following example.

First, let's take a look at the tree learner with and without
overbagging:

\begin{lstlisting}[language=R]
lrn = setPredictType(lrn, "prob")
rdesc = makeResampleDesc("CV", iters = 5)
r1 = resample(learner = lrn, task = task, resampling = rdesc, show.info = FALSE,
  measures = list(mmce, ber, auc))
r1$aggr
#> mmce.test.mean  ber.test.mean  auc.test.mean 
#>     0.01960784     0.50000000     0.50000000

obw.lrn = setPredictType(obw.lrn, "prob")
r2 = resample(learner = obw.lrn, task = task, resampling = rdesc, show.info = FALSE,
  measures = list(mmce, ber, auc))
r2$aggr
#> mmce.test.mean  ber.test.mean  auc.test.mean 
#>     0.04470588     0.43611719     0.58535862
\end{lstlisting}

Now let's consider a \emph{random forest} as initial learner:

\begin{lstlisting}[language=R]
lrn = makeLearner("classif.randomForest")
obw.lrn = makeOverBaggingWrapper(lrn, obw.rate = 8, obw.iters = 3)

lrn = setPredictType(lrn, "prob")
r1 = resample(learner = lrn, task = task, resampling = rdesc, show.info = FALSE,
  measures = list(mmce, ber, auc))
r1$aggr
#> mmce.test.mean  ber.test.mean  auc.test.mean 
#>     0.03509804     0.46089748     0.58514212

obw.lrn = setPredictType(obw.lrn, "prob")
r2 = resample(learner = obw.lrn, task = task, resampling = rdesc, show.info = FALSE,
  measures = list(mmce, ber, auc))
r2$aggr
#> mmce.test.mean  ber.test.mean  auc.test.mean 
#>     0.04098039     0.45961754     0.54926842
\end{lstlisting}

While \emph{overbagging} slighty improves the performance of the
\emph{decision tree}, the auc decreases in the second example when
additional overbagging is applied. As the \emph{random forest} itself is
already a strong learner (and a bagged one as well), a further bagging
step isn't very helpful here and usually won't improve the model.

\subsubsection{Cost-based approaches}\label{cost-based-approaches}

In contrast to sampling, \emph{cost-based approaches} usually require
particular learners, which can deal with different \emph{class-dependent
costs} (\protect\hyperlink{cost-sensitive-classification}{Cost-Sensitive
Classification}).

\paragraph{Weighted classes wrapper}\label{weighted-classes-wrapper}

Another approach independent of the underlying classifier is to assign
the costs as \emph{class weights}, so that each observation receives a
weight, depending on the class it belongs to. Similar to the
sampling-based approaches, the effect of the minority class observations
is thereby increased simply by a higher weight of these instances and
vice versa for majority class observations.

In this way every learner which supports weights can be extended through
the \protect\hyperlink{wrapper}{wrapper approach}. If the learner does
not have a direct parameter for class weights, but supports observation
weights, the weights depending on the class are internally set in the
wrapper.

\begin{lstlisting}[language=R]
lrn = makeLearner("classif.logreg")
wcw.lrn = makeWeightedClassesWrapper(lrn, wcw.weight = 0.01)
\end{lstlisting}

For binary classification, the single number passed to the classifier
corresponds to the weight of the positive / majority class, while the
negative / minority class receives a weight of 1. So actually, no real
costs are used within this approach, but the cost ratio is taken into
account.

If the underlying learner already has a parameter for class weighting
(e.g., \lstinline!class.weights! in \lstinline!"classif.ksvm"!), the
\lstinline!wcw.weight! is basically passed to the specific class
weighting parameter.

\begin{lstlisting}[language=R]
lrn = makeLearner("classif.ksvm")
wcw.lrn = makeWeightedClassesWrapper(lrn, wcw.weight = 0.01)
\end{lstlisting}

\hypertarget{roc-analysis-and-performance-curves}{\subsection{ROC
Analysis and Performance
Curves}\label{roc-analysis-and-performance-curves}}

For binary scoring classifiers a \emph{threshold} (or \emph{cutoff})
value controls how predicted posterior probabilities are converted into
class labels. ROC curves and other performance plots serve to visualize
and analyse the relationship between one or two performance measures and
the threshold.

This page is mainly devoted to \emph{receiver operating characteristic}
(ROC) curves that plot the \emph{true positive rate} (sensitivity) on
the vertical axis against the \emph{false positive rate} (1 -
specificity, fall-out) on the horizontal axis for all possible threshold
values. Creating other performance plots like \emph{lift charts} or
\emph{precision/recall graphs} works analogously and is shown briefly.

In addition to performance visualization ROC curves are helpful in

\begin{itemize}
\tightlist
\item
  determining an optimal decision threshold for given class prior
  probabilities and misclassification costs (for alternatives see also
  the pages about
  \protect\hyperlink{cost-sensitive-classification}{cost-sensitive
  classification} and
  \protect\hyperlink{imbalanced-classification-problems}{imbalanced
  classification problems} in this tutorial),
\item
  identifying regions where one classifier outperforms another and
  building suitable multi-classifier systems,
\item
  obtaining calibrated estimates of the posterior probabilities.
\end{itemize}

For more information see the tutorials and introductory papers by
\href{http://binf.gmu.edu/mmasso/ROC101.pdf}{Fawcett (2004)},
\href{https://ccrma.stanford.edu/workshops/mir2009/references/ROCintro.pdf}{Fawcett
(2006)} as well as
\href{http://www.cs.bris.ac.uk/~flach/ICML04tutorial/index.html}{Flach
(ICML 2004)}.

In many applications as, e.g., diagnostic tests or spam detection, there
is uncertainty about the class priors or the misclassification costs at
the time of prediction, for example because it's hard to quantify the
costs or because costs and class priors vary over time. Under these
circumstances the classifier is expected to work well for a whole range
of decision thresholds and the area under the ROC curve (AUC) provides a
scalar performance measure for comparing and selecting classifiers.
\href{http://www.rdocumentation.org/packages/mlr/}{mlr} provides the AUC
for binary classification
(\protect\hyperlink{implemented-performance-measures}{auc} based on
package
\href{http://www.rdocumentation.org/packages/ROCR/functions/performance.html}{ROCR})
and also several generalizations of the AUC to the multi-class case
(e.g.,
\protect\hyperlink{implemented-performance-measures}{multiclass.au1p},
\protect\hyperlink{implemented-performance-measures}{multiclass.au1u}
based on
\href{https://www.math.ucdavis.edu/~saito/data/roc/ferri-class-perf-metrics.pdf}{Ferri
et al. (2009)}).

\href{http://www.rdocumentation.org/packages/mlr/}{mlr} offers three
ways to plot ROC and other performance curves.

\begin{enumerate}
\def\labelenumi{\arabic{enumi}.}
\tightlist
\item
  Function
  \href{http://www.rdocumentation.org/packages/mlr/functions/plotROCCurves.html}{plotROCCurves}
  can, based on the output of
  \href{http://www.rdocumentation.org/packages/mlr/functions/generateThreshVsPerfData.html}{generateThreshVsPerfData},
  plot performance curves for any pair of
  \protect\hyperlink{implemented-performance-measures}{performance
  measures} available in
  \href{http://www.rdocumentation.org/packages/mlr/}{mlr}.
\item
  \href{http://www.rdocumentation.org/packages/mlr/}{mlr} offers an
  interface to package
  \href{http://www.rdocumentation.org/packages/ROCR/}{ROCR} through
  function
  \href{http://www.rdocumentation.org/packages/mlr/functions/asROCRPrediction.html}{asROCRPrediction}.
\item
  \href{http://www.rdocumentation.org/packages/mlr/}{mlr}'s function
  \href{http://www.rdocumentation.org/packages/mlr/functions/plotViperCharts.html}{plotViperCharts}
  provides an interface to \href{http://viper.ijs.si}{ViperCharts}.
\end{enumerate}

With \href{http://www.rdocumentation.org/packages/mlr/}{mlr} version 2.8
functions \lstinline!generateROCRCurvesData!,
\lstinline!plotROCRCurves!, and \lstinline!plotROCRCurvesGGVIS! were
deprecated.

Below are some examples that demonstrate the three possible ways. Note
that you can only use \protect\hyperlink{learners}{learners} that are
capable of predicting probabilities. Have a look at the
\protect\hyperlink{integrated-learners}{learner table in the Appendix}
or run
\lstinline!listLearners("classif", properties = c("twoclass", "prob"))!
to get a list of all learners that support this.

\subsubsection{Performance plots with
plotROCCurves}\label{performance-plots-with-plotroccurves}

As you might recall
\href{http://www.rdocumentation.org/packages/mlr/functions/generateThreshVsPerfData.html}{generateThreshVsPerfData}
calculates one or several performance measures for a sequence of
decision thresholds from 0 to 1. It provides S3 methods for objects of
class
\href{http://www.rdocumentation.org/packages/mlr/functions/Prediction.html}{Prediction},
\href{http://www.rdocumentation.org/packages/mlr/functions/ResampleResult.html}{ResampleResult}
and
\href{http://www.rdocumentation.org/packages/mlr/functions/BenchmarkResult.html}{BenchmarkResult}
(resulting from
\href{http://www.rdocumentation.org/packages/mlr/functions/predict.WrappedModel.html}{predict},
\href{http://www.rdocumentation.org/packages/mlr/functions/resample.html}{resample}
or
\href{http://www.rdocumentation.org/packages/mlr/functions/benchmark.html}{benchmark}).
\href{http://www.rdocumentation.org/packages/mlr/functions/plotROCCurves.html}{plotROCCurves}
plots the result of
\href{http://www.rdocumentation.org/packages/mlr/functions/generateThreshVsPerfData.html}{generateThreshVsPerfData}
using \href{http://www.rdocumentation.org/packages/ggplot2/}{ggplot2}.

\paragraph{Example 1: Single
predictions}\label{example-1-single-predictions}

We consider the
\href{http://www.rdocumentation.org/packages/mlbench/functions/Sonar.html}{Sonar}
data set from package
\href{http://www.rdocumentation.org/packages/mlbench/}{mlbench}, which
poses a binary classification problem
(\href{http://www.rdocumentation.org/packages/mlr/functions/sonar.task.html}{sonar.task})
and apply
\href{http://www.rdocumentation.org/packages/MASS/functions/lda.html}{linear
discriminant analysis}.

\begin{lstlisting}[language=R]
n = getTaskSize(sonar.task)
train.set = sample(n, size = round(2/3 * n))
test.set = setdiff(seq_len(n), train.set)

lrn1 = makeLearner("classif.lda", predict.type = "prob")
mod1 = train(lrn1, sonar.task, subset = train.set)
pred1 = predict(mod1, task = sonar.task, subset = test.set)
\end{lstlisting}

Since we want to plot ROC curves we calculate the false and true
positive rates
(\protect\hyperlink{implemented-performance-measures}{fpr} and
\protect\hyperlink{implemented-performance-measures}{tpr}).
Additionally, we also compute error rates
(\protect\hyperlink{implemented-performance-measures}{mmce}).

\begin{lstlisting}[language=R]
df = generateThreshVsPerfData(pred1, measures = list(fpr, tpr, mmce))
\end{lstlisting}

\href{http://www.rdocumentation.org/packages/mlr/functions/generateThreshVsPerfData.html}{generateThreshVsPerfData}
returns an object of class
\href{http://www.rdocumentation.org/packages/mlr/functions/generateThreshVsPerfData.html}{ThreshVsPerfData}
which contains the performance values in the \lstinline!$data! element.

Per default,
\href{http://www.rdocumentation.org/packages/mlr/functions/plotROCCurves.html}{plotROCCurves}
plots the performance values of the first two measures passed to
\href{http://www.rdocumentation.org/packages/mlr/functions/generateThreshVsPerfData.html}{generateThreshVsPerfData}.
The first is shown on the x-axis, the second on the y-axis. Moreover, a
diagonal line that represents the performance of a random classifier is
added. You can remove the diagonal by setting
\lstinline!diagonal = FALSE!.

\begin{lstlisting}[language=R]
plotROCCurves(df)
\end{lstlisting}

The corresponding area under curve
(\protect\hyperlink{implemented-performance-measures}{auc}) can be
calculated as usual by calling
\href{http://www.rdocumentation.org/packages/mlr/functions/performance.html}{performance}.

\begin{lstlisting}[language=R]
performance(pred1, auc)
#>      auc 
#> 0.847973
\end{lstlisting}

\href{http://www.rdocumentation.org/packages/mlr/functions/plotROCCurves.html}{plotROCCurves}
always requires a pair of performance measures that are plotted against
each other. If you want to plot individual measures versus the decision
threshold you can use function
\href{http://www.rdocumentation.org/packages/mlr/functions/plotThreshVsPerf.html}{plotThreshVsPerf}.

\begin{lstlisting}[language=R]
plotThreshVsPerf(df)
\end{lstlisting}

\includegraphics{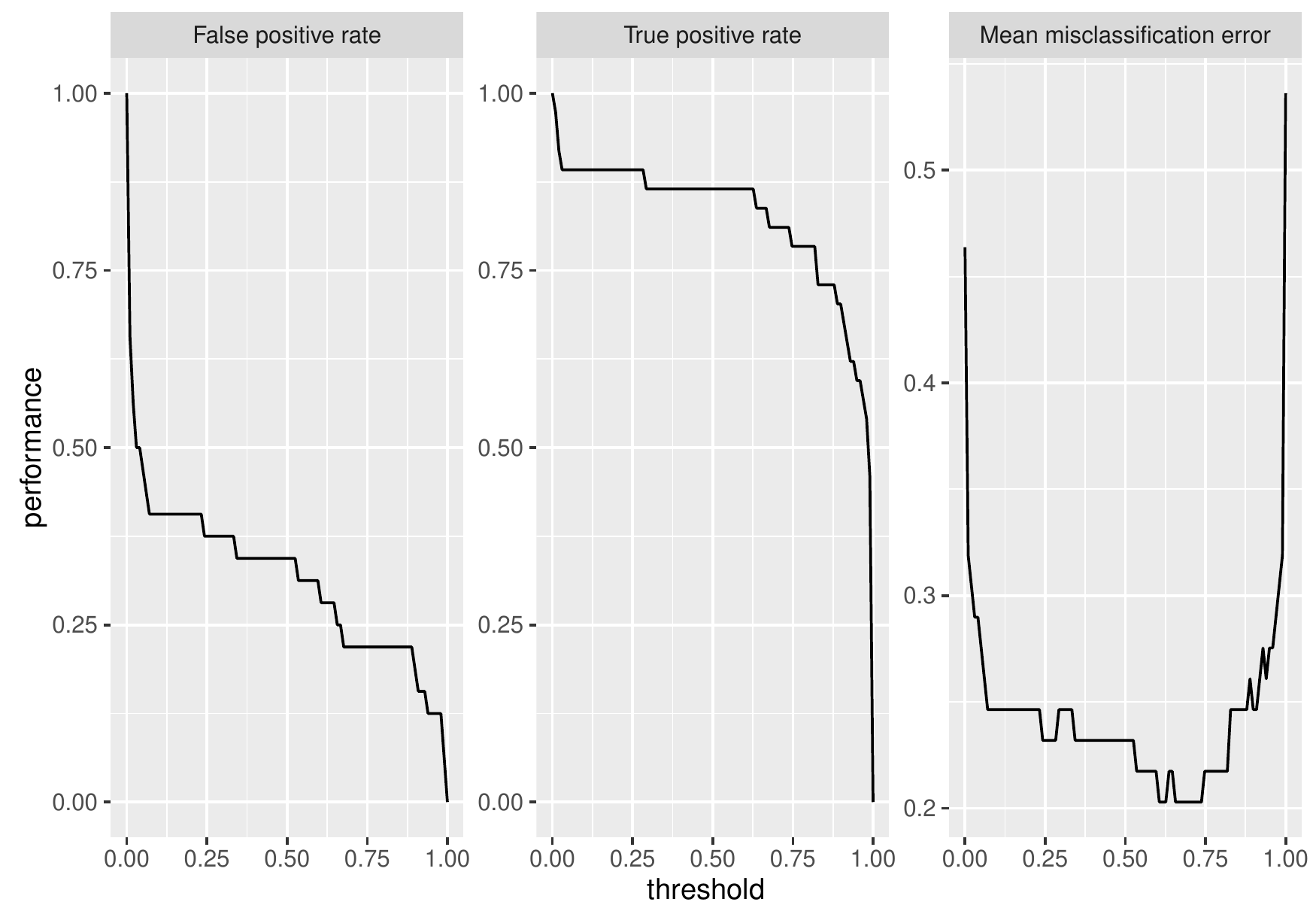}

Additional to
\href{http://www.rdocumentation.org/packages/MASS/functions/lda.html}{linear
discriminant analysis} we try a support vector machine with RBF kernel
(\href{http://www.rdocumentation.org/packages/kernlab/functions/ksvm.html}{ksvm}).

\begin{lstlisting}[language=R]
lrn2 = makeLearner("classif.ksvm", predict.type = "prob")
mod2 = train(lrn2, sonar.task, subset = train.set)
pred2 = predict(mod2, task = sonar.task, subset = test.set)
\end{lstlisting}

In order to compare the performance of the two learners you might want
to display the two corresponding ROC curves in one plot. For this
purpose just pass a named
\href{http://www.rdocumentation.org/packages/base/functions/list.html}{list}
of
\href{http://www.rdocumentation.org/packages/mlr/functions/Prediction.html}{Prediction}s
to
\href{http://www.rdocumentation.org/packages/mlr/functions/generateThreshVsPerfData.html}{generateThreshVsPerfData}.

\begin{lstlisting}[language=R]
df = generateThreshVsPerfData(list(lda = pred1, ksvm = pred2), measures = list(fpr, tpr))
plotROCCurves(df)
\end{lstlisting}

\includegraphics{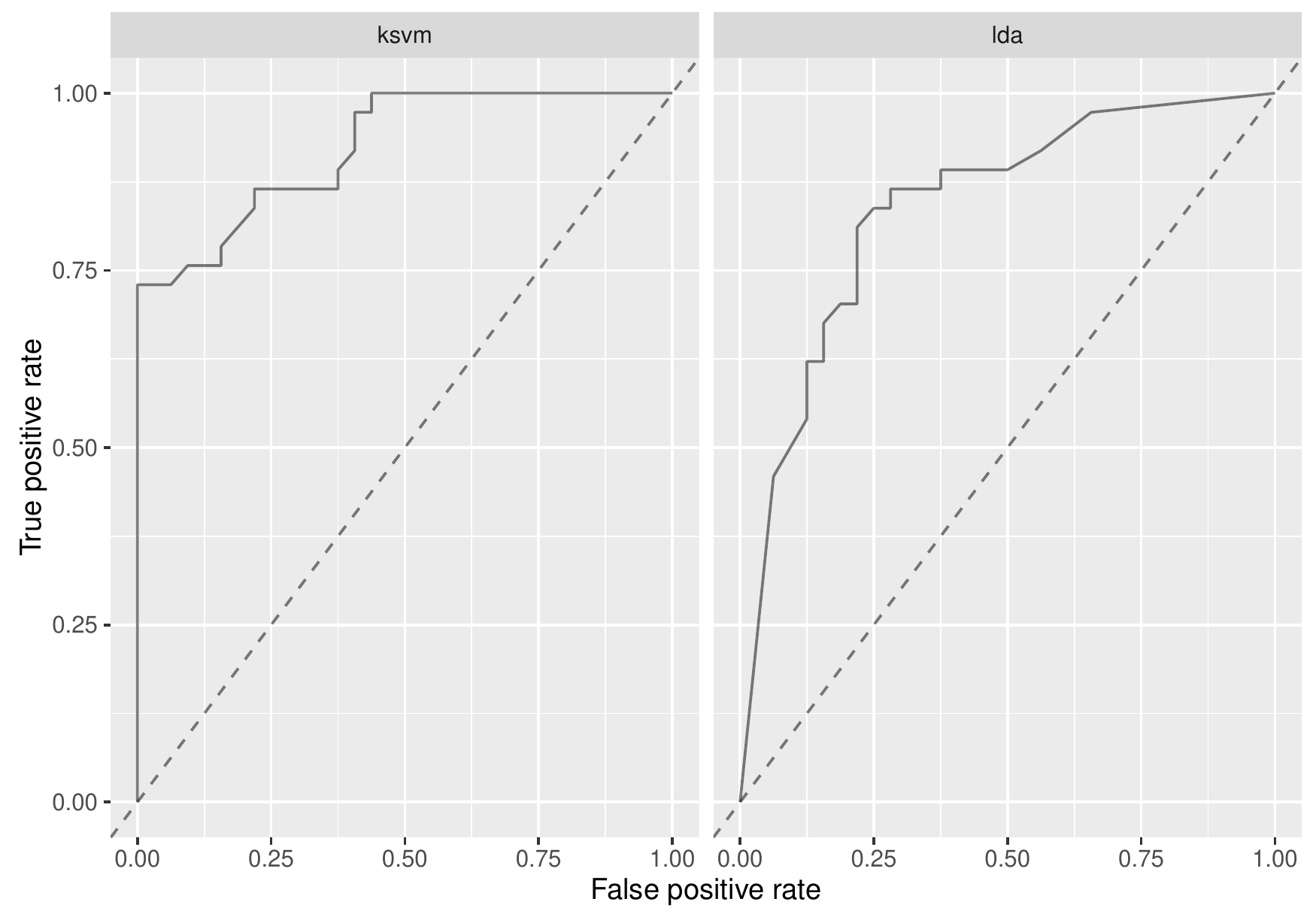}

It's clear from the plot above that
\href{http://www.rdocumentation.org/packages/kernlab/functions/ksvm.html}{ksvm}
has a slightly higher AUC than
\href{http://www.rdocumentation.org/packages/MASS/functions/lda.html}{lda}.

\begin{lstlisting}[language=R]
performance(pred2, auc)
#>       auc 
#> 0.9214527
\end{lstlisting}

Based on the \lstinline!$data! member of \lstinline!df! you can easily
generate custom plots. Below the curves for the two learners are
superposed.

\begin{lstlisting}[language=R]
qplot(x = fpr, y = tpr, color = learner, data = df$data, geom = "path")
\end{lstlisting}

It is easily possible to generate other performance plots by passing the
appropriate performance measures to
\href{http://www.rdocumentation.org/packages/mlr/functions/generateThreshVsPerfData.html}{generateThreshVsPerfData}
and
\href{http://www.rdocumentation.org/packages/mlr/functions/plotROCCurves.html}{plotROCCurves}.
Below, we generate a \emph{precision/recall graph} (precision = positive
predictive value = ppv, recall = tpr) and a
\emph{sensitivity/specificity plot} (sensitivity = tpr, specificity =
tnr).

\begin{lstlisting}[language=R]
df = generateThreshVsPerfData(list(lda = pred1, ksvm = pred2), measures = list(ppv, tpr, tnr))

### Precision/recall graph
plotROCCurves(df, measures = list(tpr, ppv), diagonal = FALSE)
#> Warning: Removed 1 rows containing missing values (geom_path).

### Sensitivity/specificity plot
plotROCCurves(df, measures = list(tnr, tpr), diagonal = FALSE)
\end{lstlisting}

\paragraph{Example 2: Benchmark
experiment}\label{example-2-benchmark-experiment}

The analysis in the example above can be improved a little. Instead of
writing individual code for training/prediction of each learner, which
can become tedious very quickly, we can use function
\href{http://www.rdocumentation.org/packages/mlr/functions/benchmark.html}{benchmark}
(see also \protect\hyperlink{benchmark-experiments}{Benchmark
Experiments}) and, ideally, the support vector machine should have been
\protect\hyperlink{tuning-hyperparameters}{tuned}.

We again consider the
\href{http://www.rdocumentation.org/packages/mlbench/functions/Sonar.html}{Sonar}
data set and apply
\href{http://www.rdocumentation.org/packages/MASS/functions/lda.html}{lda}
as well as
\href{http://www.rdocumentation.org/packages/kernlab/functions/ksvm.html}{ksvm}.
We first generate a
\href{http://www.rdocumentation.org/packages/mlr/functions/makeTuneWrapper.html}{tuning
wrapper} for
\href{http://www.rdocumentation.org/packages/kernlab/functions/ksvm.html}{ksvm}.
The cost parameter is tuned on a (for demonstration purposes small)
parameter grid. We assume that we are interested in a good performance
over the complete threshold range and therefore tune with regard to the
\protect\hyperlink{implemented-performance-measures}{auc}. The error
rate (\protect\hyperlink{implemented-performance-measures}{mmce}) for a
threshold value of 0.5 is reported as well.

\begin{lstlisting}[language=R]
### Tune wrapper for ksvm
rdesc.inner = makeResampleDesc("Holdout")
ms = list(auc, mmce)
ps = makeParamSet(
  makeDiscreteParam("C", 2^(-1:1))
)
ctrl = makeTuneControlGrid()
lrn2 = makeTuneWrapper(lrn2, rdesc.inner, ms, ps, ctrl, show.info = FALSE)
\end{lstlisting}

Below the actual benchmark experiment is conducted. As resampling
strategy we use 5-fold cross-validation and again calculate the
\protect\hyperlink{implemented-performance-measures}{auc} as well as the
error rate (for a threshold/cutoff value of 0.5).

\begin{lstlisting}[language=R]
### Benchmark experiment
lrns = list(lrn1, lrn2)
rdesc.outer = makeResampleDesc("CV", iters = 5)

bmr = benchmark(lrns, tasks = sonar.task, resampling = rdesc.outer, measures = ms, show.info = FALSE)
bmr
#>         task.id         learner.id auc.test.mean mmce.test.mean
#> 1 Sonar-example        classif.lda     0.7835442      0.2592334
#> 2 Sonar-example classif.ksvm.tuned     0.9454418      0.1390244
\end{lstlisting}

Calling
\href{http://www.rdocumentation.org/packages/mlr/functions/generateThreshVsPerfData.html}{generateThreshVsPerfData}
and
\href{http://www.rdocumentation.org/packages/mlr/functions/plotROCCurves.html}{plotROCCurves}
on the
\href{http://www.rdocumentation.org/packages/mlr/functions/BenchmarkResult.html}{benchmark
result} produces a plot with ROC curves for all learners in the
experiment.

\begin{lstlisting}[language=R]
df = generateThreshVsPerfData(bmr, measures = list(fpr, tpr, mmce))
plotROCCurves(df)
\end{lstlisting}

\includegraphics{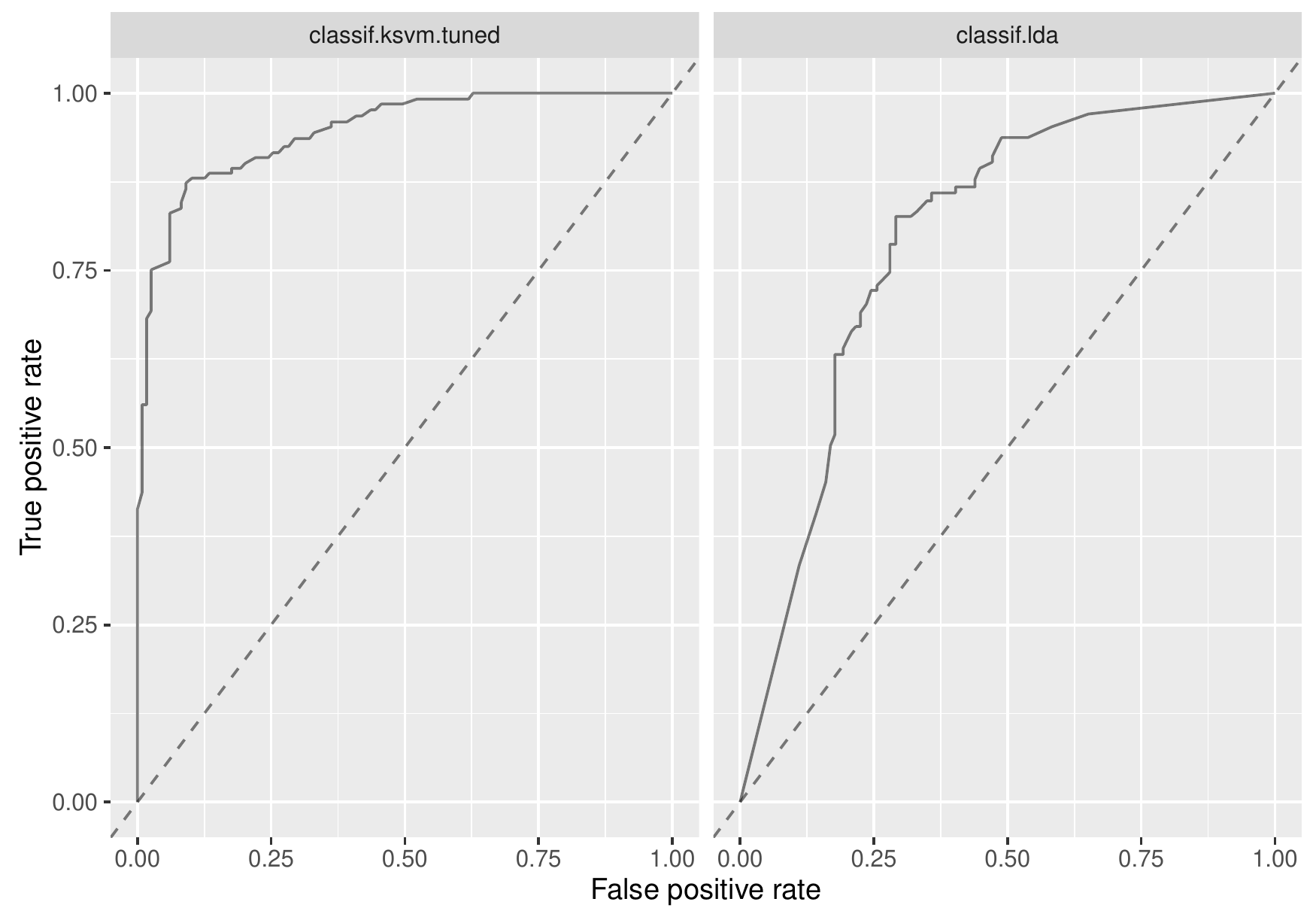}

Per default,
\href{http://www.rdocumentation.org/packages/mlr/functions/generateThreshVsPerfData.html}{generateThreshVsPerfData}
calculates aggregated performances according to the chosen resampling
strategy (5-fold cross-validation) and aggregation scheme
(\href{http://www.rdocumentation.org/packages/mlr/functions/aggregations.html}{test.mean})
for each threshold in the sequence. This way we get
\emph{threshold-averaged} ROC curves.

If you want to plot the individual ROC curves for each resample
iteration set \lstinline!aggregate = FALSE!.

\begin{lstlisting}[language=R]
df = generateThreshVsPerfData(bmr, measures = list(fpr, tpr, mmce), aggregate = FALSE)
plotROCCurves(df)
\end{lstlisting}

\includegraphics{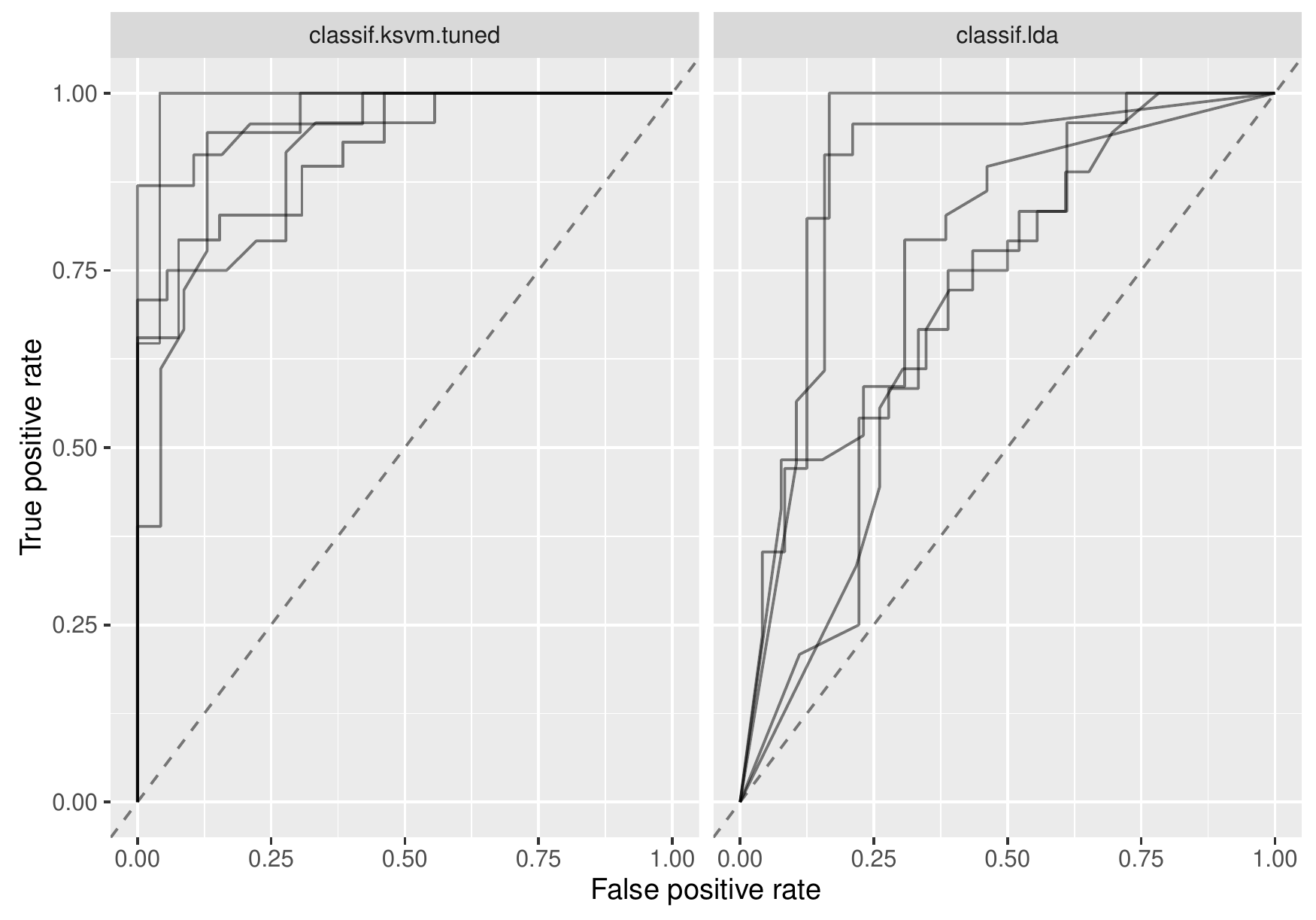}

The same applies for
\href{http://www.rdocumentation.org/packages/mlr/functions/plotThreshVsPerf.html}{plotThreshVsPerf}.

\begin{lstlisting}[language=R]
plotThreshVsPerf(df) +
  theme(strip.text.x = element_text(size = 7))
\end{lstlisting}

\includegraphics{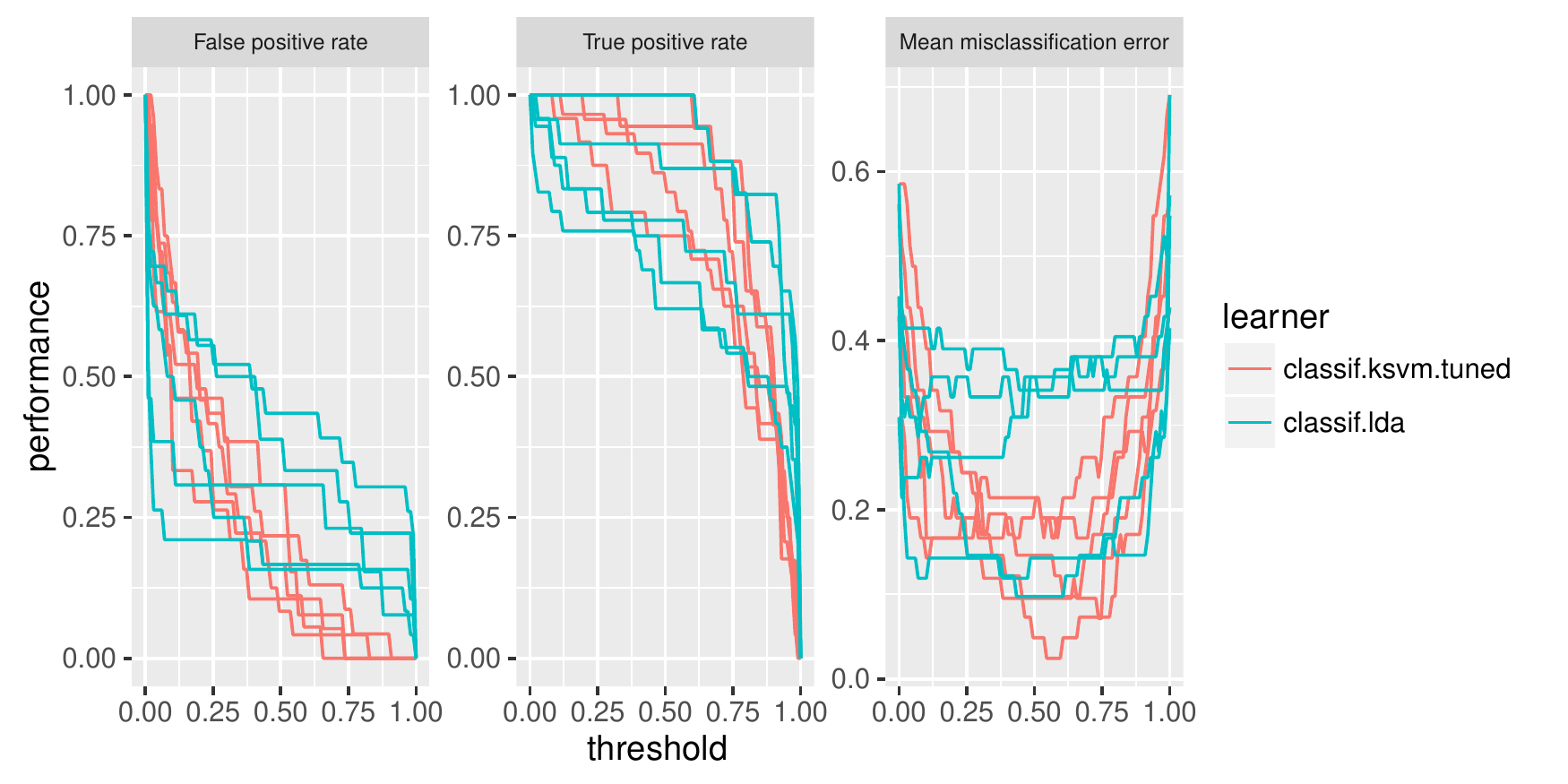}

An alternative to averaging is to just merge the 5 test folds and draw a
single ROC curve. Merging can be achieved by manually changing the
\href{http://www.rdocumentation.org/packages/base/functions/class.html}{class}
attribute of the prediction objects from
\href{http://www.rdocumentation.org/packages/mlr/functions/ResamplePrediction.html}{ResamplePrediction}
to
\href{http://www.rdocumentation.org/packages/mlr/functions/Prediction.html}{Prediction}.

Below, the predictions are extracted from the
\href{http://www.rdocumentation.org/packages/mlr/functions/BenchmarkResult.html}{BenchmarkResult}
via function
\href{http://www.rdocumentation.org/packages/mlr/functions/getBMRPredictions.html}{getBMRPredictions},
the
\href{http://www.rdocumentation.org/packages/base/functions/class.html}{class}
is changed and the ROC curves are created.

Averaging methods are normally preferred (cp.
\href{https://ccrma.stanford.edu/workshops/mir2009/references/ROCintro.pdf}{Fawcett,
2006}), as they permit to assess the variability, which is needed to
properly compare classifier performance.

\begin{lstlisting}[language=R]
### Extract predictions
preds = getBMRPredictions(bmr)[[1]]

### Change the class attribute
preds2 = lapply(preds, function(x) {class(x) = "Prediction"; return(x)})

### Draw ROC curves
df = generateThreshVsPerfData(preds2, measures = list(fpr, tpr, mmce))
plotROCCurves(df)
\end{lstlisting}

\includegraphics{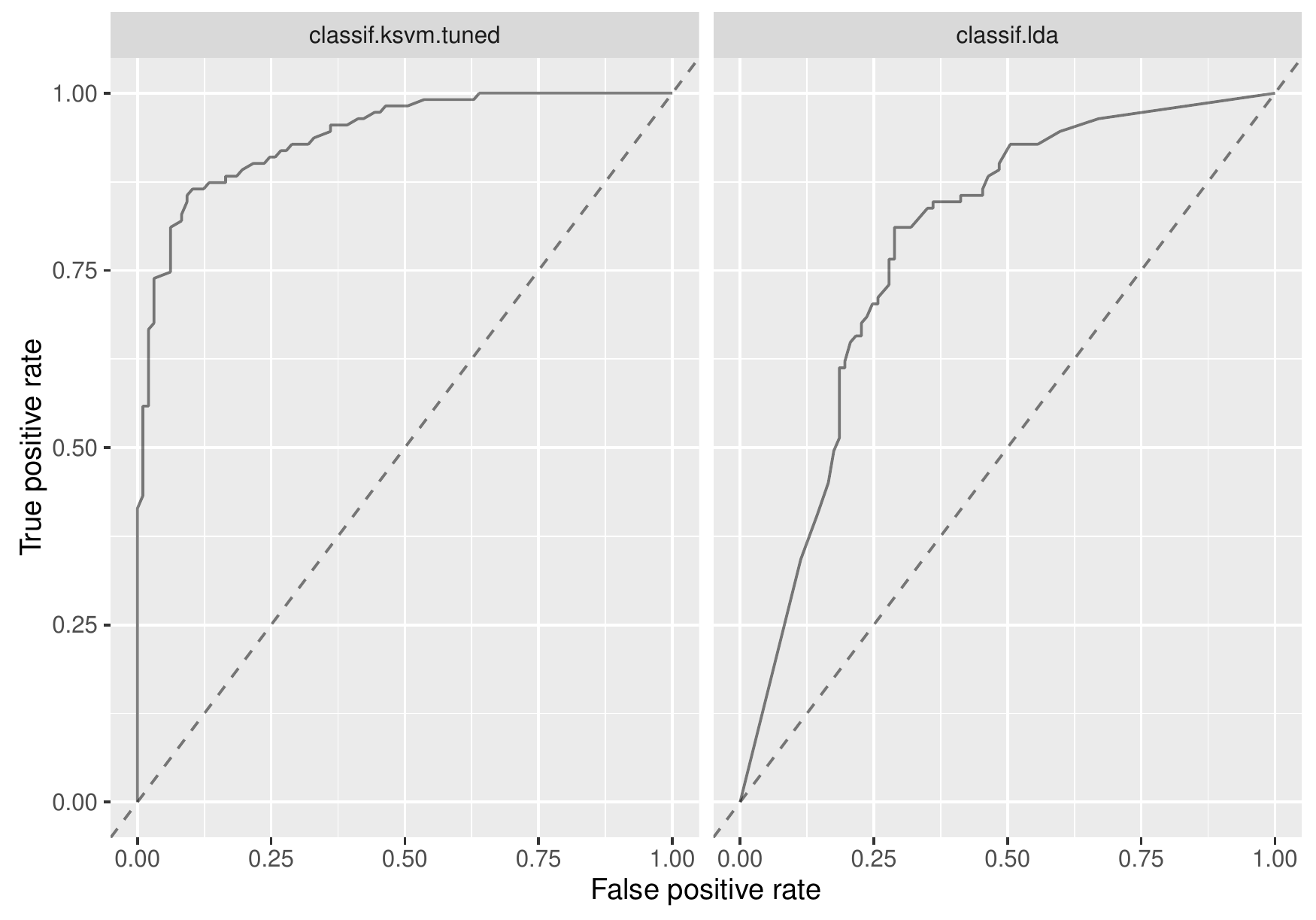}

Again, you can easily create other standard evaluation plots by passing
the appropriate performance measures to
\href{http://www.rdocumentation.org/packages/mlr/functions/generateThreshVsPerfData.html}{generateThreshVsPerfData}
and
\href{http://www.rdocumentation.org/packages/mlr/functions/plotROCCurves.html}{plotROCCurves}.

\subsubsection{Performance plots with
asROCRPrediction}\label{performance-plots-with-asrocrprediction}

Drawing performance plots with package
\href{http://www.rdocumentation.org/packages/ROCR/}{ROCR} works through
three basic commands:

\begin{enumerate}
\def\labelenumi{\arabic{enumi}.}
\tightlist
\item
  \href{http://www.rdocumentation.org/packages/ROCR/functions/prediction.html}{ROCR::prediction}:
  Create a \href{http://www.rdocumentation.org/packages/ROCR/}{ROCR}
  \href{\&ROCR::prediction-class}{prediction} object.
\item
  \href{http://www.rdocumentation.org/packages/ROCR/functions/performance.html}{ROCR::performance}:
  Calculate one or more performance measures for the given prediction
  object.
\item
  \href{\&ROCR::plot-methods}{ROCR::plot}: Generate the performance
  plot.
\end{enumerate}

\href{http://www.rdocumentation.org/packages/mlr/}{mlr}'s function
\href{http://www.rdocumentation.org/packages/mlr/functions/asROCRPrediction.html}{asROCRPrediction}
converts an \href{http://www.rdocumentation.org/packages/mlr/}{mlr}
\href{http://www.rdocumentation.org/packages/mlr/functions/Prediction.html}{Prediction}
object to a \href{http://www.rdocumentation.org/packages/ROCR/}{ROCR}
\href{\&ROCR::prediction-class}{prediction} object, so you can easily
generate performance plots by doing steps 2. and 3. yourself.
\href{http://www.rdocumentation.org/packages/ROCR/}{ROCR}'s
\href{\&ROCR::plot-methods}{plot} method has some nice features which
are not (yet) available in
\href{http://www.rdocumentation.org/packages/mlr/functions/plotROCCurves.html}{plotROCCurves},
for example plotting the convex hull of the ROC curves. Some examples
are shown below.

\paragraph{Example 1: Single predictions
(continued)}\label{example-1-single-predictions-continued}

We go back to out first example where we trained and predicted
\href{http://www.rdocumentation.org/packages/MASS/functions/lda.html}{lda}
on the
\href{http://www.rdocumentation.org/packages/mlr/functions/sonar.task.html}{sonar
classification task}.

\begin{lstlisting}[language=R]
n = getTaskSize(sonar.task)
train.set = sample(n, size = round(2/3 * n))
test.set = setdiff(seq_len(n), train.set)

### Train and predict linear discriminant analysis
lrn1 = makeLearner("classif.lda", predict.type = "prob")
mod1 = train(lrn1, sonar.task, subset = train.set)
pred1 = predict(mod1, task = sonar.task, subset = test.set)
\end{lstlisting}

Below we use
\href{http://www.rdocumentation.org/packages/mlr/functions/asROCRPrediction.html}{asROCRPrediction}
to convert the lda prediction, let
\href{http://www.rdocumentation.org/packages/ROCR/}{ROCR} calculate the
true and false positive rate and plot the ROC curve.

\begin{lstlisting}[language=R]
### Convert prediction
ROCRpred1 = asROCRPrediction(pred1)

### Calculate true and false positive rate
ROCRperf1 = ROCR::performance(ROCRpred1, "tpr", "fpr")

### Draw ROC curve
ROCR::plot(ROCRperf1)
\end{lstlisting}

Below is the same ROC curve, but we make use of some more graphical
parameters: The ROC curve is color-coded by the threshold and selected
threshold values are printed on the curve. Additionally, the convex hull
(black broken line) of the ROC curve is drawn.

\begin{lstlisting}[language=R]
### Draw ROC curve
ROCR::plot(ROCRperf1, colorize = TRUE, print.cutoffs.at = seq(0.1, 0.9, 0.1), lwd = 2)

### Draw convex hull of ROC curve
ch = ROCR::performance(ROCRpred1, "rch")
ROCR::plot(ch, add = TRUE, lty = 2)
\end{lstlisting}

\paragraph{Example 2: Benchmark experiments
(continued)}\label{example-2-benchmark-experiments-continued}

We again consider the benchmark experiment conducted earlier. We first
extract the predictions by
\href{http://www.rdocumentation.org/packages/mlr/functions/getBMRPredictions.html}{getBMRPredictions}
and then convert them via function
\href{http://www.rdocumentation.org/packages/mlr/functions/asROCRPrediction.html}{asROCRPrediction}.

\begin{lstlisting}[language=R]
### Extract predictions
preds = getBMRPredictions(bmr)[[1]]

### Convert predictions
ROCRpreds = lapply(preds, asROCRPrediction)

### Calculate true and false positive rate
ROCRperfs = lapply(ROCRpreds, function(x) ROCR::performance(x, "tpr", "fpr"))
\end{lstlisting}

We draw the vertically averaged ROC curves (solid lines) as well as the
ROC curves for the individual resampling iterations (broken lines).
Moreover, standard error bars are plotted for selected true positive
rates (0.1, 0.2, \ldots{}, 0.9). See
\href{http://www.rdocumentation.org/packages/ROCR/}{ROCR}'s
\href{\&ROCR::plot-methods}{plot} function for details.

\begin{lstlisting}[language=R]
### lda average ROC curve
plot(ROCRperfs[[1]], col = "blue", avg = "vertical", spread.estimate = "stderror",
  show.spread.at = seq(0.1, 0.8, 0.1), plotCI.col = "blue", plotCI.lwd = 2, lwd = 2)
### lda individual ROC curves
plot(ROCRperfs[[1]], col = "blue", lty = 2, lwd = 0.25, add = TRUE)

### ksvm average ROC curve
plot(ROCRperfs[[2]], col = "red", avg = "vertical", spread.estimate = "stderror",
  show.spread.at = seq(0.1, 0.6, 0.1), plotCI.col = "red", plotCI.lwd = 2, lwd = 2, add = TRUE)
### ksvm individual ROC curves
plot(ROCRperfs[[2]], col = "red", lty = 2, lwd = 0.25, add = TRUE)

legend("bottomright", legend = getBMRLearnerIds(bmr), lty = 1, lwd = 2, col = c("blue", "red"))
\end{lstlisting}

In order to create other evaluation plots like \emph{precision/recall
graphs} you just have to change the performance measures when calling
\href{http://www.rdocumentation.org/packages/ROCR/functions/performance.html}{ROCR::performance}.
(Note that you have to use the measures provided by
\href{http://www.rdocumentation.org/packages/ROCR/}{ROCR} listed
\href{http://www.rdocumentation.org/packages/ROCR/functions/performance.html}{here}
and not \href{http://www.rdocumentation.org/packages/mlr/}{mlr}'s
performance measures.)

\begin{lstlisting}[language=R]
### Extract and convert predictions
preds = getBMRPredictions(bmr)[[1]]
ROCRpreds = lapply(preds, asROCRPrediction)

### Calculate precision and recall
ROCRperfs = lapply(ROCRpreds, function(x) ROCR::performance(x, "prec", "rec"))

### Draw performance plot
plot(ROCRperfs[[1]], col = "blue", avg = "threshold")
plot(ROCRperfs[[2]], col = "red", avg = "threshold", add = TRUE)
legend("bottomleft", legend = getBMRLearnerIds(bmr), lty = 1, col = c("blue", "red"))
\end{lstlisting}

If you want to plot a performance measure versus the threshold, specify
only one measure when calling
\href{http://www.rdocumentation.org/packages/ROCR/functions/performance.html}{ROCR::performance}.
Below the average accuracy over the 5 cross-validation iterations is
plotted against the threshold. Moreover, boxplots for certain threshold
values (0.1, 0.2, \ldots{}, 0.9) are drawn.

\begin{lstlisting}[language=R]
### Extract and convert predictions
preds = getBMRPredictions(bmr)[[1]]
ROCRpreds = lapply(preds, asROCRPrediction)

### Calculate accuracy
ROCRperfs = lapply(ROCRpreds, function(x) ROCR::performance(x, "acc"))

### Plot accuracy versus threshold
plot(ROCRperfs[[1]], avg = "vertical", spread.estimate = "boxplot", lwd = 2, col = "blue",
  show.spread.at = seq(0.1, 0.9, 0.1), ylim = c(0,1), xlab = "Threshold")
\end{lstlisting}

\subsubsection{Viper charts}\label{viper-charts}

\href{http://www.rdocumentation.org/packages/mlr/}{mlr} also supports
\href{http://viper.ijs.si/}{ViperCharts} for plotting ROC and other
performance curves. Like
\href{http://www.rdocumentation.org/packages/mlr/functions/generateThreshVsPerfData.html}{generateThreshVsPerfData}
it has S3 methods for objects of class
\href{http://www.rdocumentation.org/packages/mlr/functions/Prediction.html}{Prediction},
\href{http://www.rdocumentation.org/packages/mlr/functions/ResampleResult.html}{ResampleResult}
and
\href{http://www.rdocumentation.org/packages/mlr/functions/BenchmarkResult.html}{BenchmarkResult}.
Below plots for the benchmark experiment (Example 2) are generated.

\begin{lstlisting}[language=R]
z = plotViperCharts(bmr, chart = "rocc", browse = FALSE)
\end{lstlisting}

You can see the plot created this way
\href{http://viper.ijs.si/chart/roc/4e519f6d-2fc6-4f38-a9d7-e3f70df03102/}{here}.
Note that besides ROC curves you get several other plots like lift
charts or cost curves. For details, see
\href{http://www.rdocumentation.org/packages/mlr/functions/plotViperCharts.html}{plotViperCharts}.

\subsection{Multilabel
Classification}\label{multilabel-classification-1}

Multilabel classification is a classification problem where multiple
target labels can be assigned to each observation instead of only one
like in multiclass classification.

Two different approaches exist for multilabel classification.
\emph{Problem transformation methods} try to transform the multilabel
classification into binary or multiclass classification problems.
\emph{Algorithm adaptation methods} adapt multiclass algorithms so they
can be applied directly to the problem.

\subsubsection{Creating a task}\label{creating-a-task}

The first thing you have to do for multilabel classification in
\href{http://www.rdocumentation.org/packages/mlr/}{mlr} is to get your
data in the right format. You need a
\href{http://www.rdocumentation.org/packages/base/functions/data.frame.html}{data.frame}
which consists of the features and a logical vector for each label which
indicates if the label is present in the observation or not. After that
you can create a
\href{http://www.rdocumentation.org/packages/mlr/functions/Task.html}{MultilabelTask}
like a normal
\href{http://www.rdocumentation.org/packages/mlr/functions/Task.html}{ClassifTask}.
Instead of one target name you have to specify a vector of targets which
correspond to the names of logical variables in the
\href{http://www.rdocumentation.org/packages/base/functions/data.frame.html}{data.frame}.
In the following example we get the yeast data frame from the already
existing
\href{http://www.rdocumentation.org/packages/mlr/functions/yeast.task.html}{yeast.task},
extract the 14 label names and create the task again.

\begin{lstlisting}[language=R]
yeast = getTaskData(yeast.task)
labels = colnames(yeast)[1:14]
yeast.task = makeMultilabelTask(id = "multi", data = yeast, target = labels)
yeast.task
#> Supervised task: multi
#> Type: multilabel
#> Target: label1,label2,label3,label4,label5,label6,label7,label8,label9,label10,label11,label12,label13,label14
#> Observations: 2417
#> Features:
#> numerics  factors  ordered 
#>      103        0        0 
#> Missings: FALSE
#> Has weights: FALSE
#> Has blocking: FALSE
#> Classes: 14
#>  label1  label2  label3  label4  label5  label6  label7  label8  label9 
#>     762    1038     983     862     722     597     428     480     178 
#> label10 label11 label12 label13 label14 
#>     253     289    1816    1799      34
\end{lstlisting}

\subsubsection{Constructing a learner}\label{constructing-a-learner-1}

Multilabel classification in
\href{http://www.rdocumentation.org/packages/mlr/}{mlr} can currently be
done in two ways:

\begin{itemize}
\item
  Algorithm adaptation methods: Treat the whole problem with a specific
  algorithm.
\item
  Problem transformation methods: Transform the problem, so that simple
  binary classification algorithms can be applied.
\end{itemize}

\paragraph{Algorithm adaptation
methods}\label{algorithm-adaptation-methods}

Currently the available algorithm adaptation methods in \textbf{R} are
the multivariate random forest in the
\href{http://www.rdocumentation.org/packages/randomForestSRC/}{randomForestSRC}
package and the random ferns multilabel algorithm in the
\href{http://www.rdocumentation.org/packages/rFerns/}{rFerns} package.
You can create the learner for these algorithms like in multiclass
classification problems.

\begin{lstlisting}[language=R]
lrn.rfsrc = makeLearner("multilabel.randomForestSRC")
lrn.rFerns = makeLearner("multilabel.rFerns")
lrn.rFerns
#> Learner multilabel.rFerns from package rFerns
#> Type: multilabel
#> Name: Random ferns; Short name: rFerns
#> Class: multilabel.rFerns
#> Properties: numerics,factors,ordered
#> Predict-Type: response
#> Hyperparameters:
\end{lstlisting}

\paragraph{Problem transformation
methods}\label{problem-transformation-methods}

For generating a wrapped multilabel learner first create a binary (or
multiclass) classification learner with
\href{http://www.rdocumentation.org/packages/mlr/functions/makeLearner.html}{makeLearner}.
Afterwards apply a function like
\href{http://www.rdocumentation.org/packages/mlr/functions/makeMultilabelBinaryRelevanceWrapper.html}{makeMultilabelBinaryRelevanceWrapper},
\href{http://www.rdocumentation.org/packages/mlr/functions/makeMultilabelClassifierChainsWrapper.html}{makeMultilabelClassifierChainsWrapper},
\href{http://www.rdocumentation.org/packages/mlr/functions/makeMultilabelNestedStackingWrapper.html}{makeMultilabelNestedStackingWrapper},
\href{http://www.rdocumentation.org/packages/mlr/functions/makeMultilabelDBRWrapper.html}{makeMultilabelDBRWrapper}
or
\href{http://www.rdocumentation.org/packages/mlr/functions/makeMultilabelStackingWrapper.html}{makeMultilabelStackingWrapper}
on the learner to convert it to a learner that uses the respective
problem transformation method.

You can also generate a binary relevance learner directly, as you can
see in the example.

\begin{lstlisting}[language=R]
lrn.br = makeLearner("classif.rpart", predict.type = "prob")
lrn.br = makeMultilabelBinaryRelevanceWrapper(lrn.br)
lrn.br
#> Learner multilabel.classif.rpart from package rpart
#> Type: multilabel
#> Name: ; Short name: 
#> Class: MultilabelBinaryRelevanceWrapper
#> Properties: numerics,factors,ordered,missings,weights,prob,twoclass,multiclass
#> Predict-Type: prob
#> Hyperparameters: xval=0

lrn.br2 = makeMultilabelBinaryRelevanceWrapper("classif.rpart")
lrn.br2
#> Learner multilabel.classif.rpart from package rpart
#> Type: multilabel
#> Name: ; Short name: 
#> Class: MultilabelBinaryRelevanceWrapper
#> Properties: numerics,factors,ordered,missings,weights,prob,twoclass,multiclass
#> Predict-Type: response
#> Hyperparameters: xval=0
\end{lstlisting}

The different methods are shortly described in the following.

\subparagraph{Binary relevance}\label{binary-relevance}

This problem transformation method converts the multilabel problem to
binary classification problems for each label and applies a simple
binary classificator on these. In
\href{http://www.rdocumentation.org/packages/mlr/}{mlr} this can be done
by converting your binary learner to a wrapped binary relevance
multilabel learner.

\subparagraph{Classifier chains}\label{classifier-chains}

Trains consecutively the labels with the input data. The input data in
each step is augmented by the already trained labels (with the real
observed values). Therefore an order of the labels has to be specified.
At prediction time the labels are predicted in the same order as while
training. The required labels in the input data are given by the
previous done prediction of the respective label.

\subparagraph{Nested stacking}\label{nested-stacking}

Same as classifier chains, but the labels in the input data are not the
real ones, but estimations of the labels obtained by the already trained
learners.

\subparagraph{Dependent binary
relevance}\label{dependent-binary-relevance}

Each label is trained with the real observed values of all other labels.
In prediction phase for a label the other necessary labels are obtained
in a previous step by a base learner like the binary relevance method.

\subparagraph{Stacking}\label{stacking}

Same as the dependent binary relevance method, but in the training phase
the labels used as input for each label are obtained by the binary
relevance method.

\subsubsection{Train}\label{train}

You can
\href{http://www.rdocumentation.org/packages/mlr/functions/train.html}{train}
a model as usual with a multilabel learner and a multilabel task as
input. You can also pass \lstinline!subset! and \lstinline!weights!
arguments if the learner supports this.

\begin{lstlisting}[language=R]
mod = train(lrn.br, yeast.task)
mod = train(lrn.br, yeast.task, subset = 1:1500, weights = rep(1/1500, 1500))
mod
#> Model for learner.id=multilabel.classif.rpart; learner.class=MultilabelBinaryRelevanceWrapper
#> Trained on: task.id = multi; obs = 1500; features = 103
#> Hyperparameters: xval=0

mod2 = train(lrn.rfsrc, yeast.task, subset = 1:100)
mod2
#> Model for learner.id=multilabel.randomForestSRC; learner.class=multilabel.randomForestSRC
#> Trained on: task.id = multi; obs = 100; features = 103
#> Hyperparameters: na.action=na.impute
\end{lstlisting}

\subsubsection{Predict}\label{predict}

Prediction can be done as usual in
\href{http://www.rdocumentation.org/packages/mlr/}{mlr} with
\href{http://www.rdocumentation.org/packages/mlr/functions/predict.WrappedModel.html}{predict}
and by passing a trained model and either the task to the
\lstinline!task! argument or some new data to the \lstinline!newdata!
argument. As always you can specify a \lstinline!subset! of the data
which should be predicted.

\begin{lstlisting}[language=R]
pred = predict(mod, task = yeast.task, subset = 1:10)
pred = predict(mod, newdata = yeast[1501:1600,])
names(as.data.frame(pred))
#>  [1] "truth.label1"     "truth.label2"     "truth.label3"    
#>  [4] "truth.label4"     "truth.label5"     "truth.label6"    
#>  [7] "truth.label7"     "truth.label8"     "truth.label9"    
#> [10] "truth.label10"    "truth.label11"    "truth.label12"   
#> [13] "truth.label13"    "truth.label14"    "prob.label1"     
#> [16] "prob.label2"      "prob.label3"      "prob.label4"     
#> [19] "prob.label5"      "prob.label6"      "prob.label7"     
#> [22] "prob.label8"      "prob.label9"      "prob.label10"    
#> [25] "prob.label11"     "prob.label12"     "prob.label13"    
#> [28] "prob.label14"     "response.label1"  "response.label2" 
#> [31] "response.label3"  "response.label4"  "response.label5" 
#> [34] "response.label6"  "response.label7"  "response.label8" 
#> [37] "response.label9"  "response.label10" "response.label11"
#> [40] "response.label12" "response.label13" "response.label14"

pred2 = predict(mod2, task = yeast.task)
names(as.data.frame(pred2))
#>  [1] "id"               "truth.label1"     "truth.label2"    
#>  [4] "truth.label3"     "truth.label4"     "truth.label5"    
#>  [7] "truth.label6"     "truth.label7"     "truth.label8"    
#> [10] "truth.label9"     "truth.label10"    "truth.label11"   
#> [13] "truth.label12"    "truth.label13"    "truth.label14"   
#> [16] "response.label1"  "response.label2"  "response.label3" 
#> [19] "response.label4"  "response.label5"  "response.label6" 
#> [22] "response.label7"  "response.label8"  "response.label9" 
#> [25] "response.label10" "response.label11" "response.label12"
#> [28] "response.label13" "response.label14"
\end{lstlisting}

Depending on the chosen \lstinline!predict.type! of the learner you get
true and predicted values and possibly probabilities for each class
label. These can be extracted by the usual accessor functions
\href{http://www.rdocumentation.org/packages/mlr/functions/getPredictionTruth.html}{getPredictionTruth},
\href{http://www.rdocumentation.org/packages/mlr/functions/getPredictionResponse.html}{getPredictionResponse}
and
\href{http://www.rdocumentation.org/packages/mlr/functions/getPredictionProbabilities.html}{getPredictionProbabilities}.

\subsubsection{Performance}\label{performance}

The performance of your prediction can be assessed via function
\href{http://www.rdocumentation.org/packages/mlr/functions/performance.html}{performance}.
You can specify via the \lstinline!measures! argument which
\protect\hyperlink{implemented-performance-measures}{measure(s)} to
calculate. The default measure for multilabel classification is the
Hamming loss
(\protect\hyperlink{implemented-performance-measures}{multilabel.hamloss}).
All available measures for multilabel classification can be shown by
\href{http://www.rdocumentation.org/packages/mlr/functions/listMeasures.html}{listMeasures}
and found in the
\protect\hyperlink{implemented-performance-measures}{table of
performance measures} and the
\href{http://www.rdocumentation.org/packages/mlr/functions/measures.html}{measures}
documentation page.

\begin{lstlisting}[language=R]
performance(pred)
#> multilabel.hamloss 
#>          0.2257143

performance(pred2, measures = list(multilabel.subset01, multilabel.hamloss, multilabel.acc,
  multilabel.f1, timepredict))
#> multilabel.subset01  multilabel.hamloss      multilabel.acc 
#>           0.8663633           0.2049471           0.4637509 
#>       multilabel.f1         timepredict 
#>           0.5729926           1.0800000

listMeasures("multilabel")
#>  [1] "multilabel.f1"       "multilabel.subset01" "multilabel.tpr"     
#>  [4] "multilabel.ppv"      "multilabel.acc"      "timeboth"           
#>  [7] "timepredict"         "multilabel.hamloss"  "featperc"           
#> [10] "timetrain"
\end{lstlisting}

\subsubsection{Resampling}\label{resampling-1}

For evaluating the overall performance of the learning algorithm you can
do some \protect\hyperlink{resampling}{resampling}. As usual you have to
define a resampling strategy, either via
\href{http://www.rdocumentation.org/packages/mlr/functions/makeResampleDesc.html}{makeResampleDesc}
or
\href{http://www.rdocumentation.org/packages/mlr/functions/makeResampleInstance.html}{makeResampleInstance}.
After that you can run the
\href{http://www.rdocumentation.org/packages/mlr/functions/resample.html}{resample}
function. Below the default measure Hamming loss is calculated.

\begin{lstlisting}[language=R]
rdesc = makeResampleDesc(method = "CV", stratify = FALSE, iters = 3)
r = resample(learner = lrn.br, task = yeast.task, resampling = rdesc, show.info = FALSE)
r
#> Resample Result
#> Task: multi
#> Learner: multilabel.classif.rpart
#> Aggr perf: multilabel.hamloss.test.mean=0.225
#> Runtime: 4.2915

r = resample(learner = lrn.rFerns, task = yeast.task, resampling = rdesc, show.info = FALSE)
r
#> Resample Result
#> Task: multi
#> Learner: multilabel.rFerns
#> Aggr perf: multilabel.hamloss.test.mean=0.473
#> Runtime: 0.395229
\end{lstlisting}

\subsubsection{Binary performance}\label{binary-performance}

If you want to calculate a binary performance measure like, e.g., the
\protect\hyperlink{implemented-performance-measures}{accuracy}, the
\protect\hyperlink{implemented-performance-measures}{mmce} or the
\protect\hyperlink{implemented-performance-measures}{auc} for each
label, you can use function
\href{http://www.rdocumentation.org/packages/mlr/functions/getMultilabelBinaryPerformances.html}{getMultilabelBinaryPerformances}.
You can apply this function to any multilabel prediction, e.g., also on
the resample multilabel prediction. For calculating the
\protect\hyperlink{implemented-performance-measures}{auc} you need
predicted probabilities.

\begin{lstlisting}[language=R]
getMultilabelBinaryPerformances(pred, measures = list(acc, mmce, auc))
#>         acc.test.mean mmce.test.mean auc.test.mean
#> label1           0.75           0.25     0.6321925
#> label2           0.64           0.36     0.6547917
#> label3           0.68           0.32     0.7118227
#> label4           0.69           0.31     0.6764835
#> label5           0.73           0.27     0.6676923
#> label6           0.70           0.30     0.6417739
#> label7           0.81           0.19     0.5968750
#> label8           0.73           0.27     0.5164474
#> label9           0.89           0.11     0.4688458
#> label10          0.86           0.14     0.3996463
#> label11          0.85           0.15     0.5000000
#> label12          0.76           0.24     0.5330667
#> label13          0.75           0.25     0.5938610
#> label14          1.00           0.00            NA

getMultilabelBinaryPerformances(r$pred, measures = list(acc, mmce))
#>         acc.test.mean mmce.test.mean
#> label1     0.69383533      0.3061647
#> label2     0.58254034      0.4174597
#> label3     0.70211005      0.2978899
#> label4     0.71369466      0.2863053
#> label5     0.70831609      0.2916839
#> label6     0.60488209      0.3951179
#> label7     0.54447662      0.4555234
#> label8     0.53289201      0.4671080
#> label9     0.30906082      0.6909392
#> label10    0.44683492      0.5531651
#> label11    0.45676458      0.5432354
#> label12    0.52916839      0.4708316
#> label13    0.53702938      0.4629706
#> label14    0.01406703      0.9859330
\end{lstlisting}

\hypertarget{learning-curve-analysis}{\subsection{Learning Curve
Analysis}\label{learning-curve-analysis}}

To analyse how the increase of observations in the training set improves
the performance of a learner the \emph{learning curve} is an appropriate
visual tool. The experiment is conducted with an increasing subsample
size and the performance is measured. In the plot the x-axis represents
the relative subsample size whereas the y-axis represents the
performance.

Note that this function internally uses
\href{http://www.rdocumentation.org/packages/mlr/functions/benchmark.html}{benchmark}
in combination with
\href{http://www.rdocumentation.org/packages/mlr/functions/makeDownsampleWrapper.html}{makeDownsampleWrapper},
so for every run new observations are drawn. Thus the results are noisy.
To reduce noise increase the number of resampling iterations. You can
define the resampling method in the \lstinline!resampling! argument of
\href{http://www.rdocumentation.org/packages/mlr/functions/generateLearningCurveData.html}{generateLearningCurveData}.
It is also possible to pass a
\href{http://www.rdocumentation.org/packages/mlr/functions/makeResampleInstance.html}{ResampleInstance}
(which is a result of
\href{http://www.rdocumentation.org/packages/mlr/functions/makeResampleInstance.html}{makeResampleInstance})
to make resampling consistent for all passed learners and each step of
increasing the number of observations.

\subsubsection{Plotting the learning
curve}\label{plotting-the-learning-curve}

The \href{http://www.rdocumentation.org/packages/mlr/}{mlr} function
\href{http://www.rdocumentation.org/packages/mlr/functions/generateLearningCurveData.html}{generateLearningCurveData}
can generate the data for \emph{learning curves} for multiple
\protect\hyperlink{integrated-learners}{learners} and multiple
\protect\hyperlink{implemented-performance-measures}{performance
measures} at once. With
\href{http://www.rdocumentation.org/packages/mlr/functions/plotLearningCurve.html}{plotLearningCurve}
the result of
\href{http://www.rdocumentation.org/packages/mlr/functions/generateLearningCurveData.html}{generateLearningCurveData}
can be plotted using
\href{http://www.rdocumentation.org/packages/ggplot2/}{ggplot2}.
\href{http://www.rdocumentation.org/packages/mlr/functions/plotLearningCurve.html}{plotLearningCurve}
has an argument \lstinline!facet! which can be either ``measure'' or
``learner''. By default \lstinline!facet = "measure"! and facetted
subplots are created for each measure input to
\href{http://www.rdocumentation.org/packages/mlr/functions/generateLearningCurveData.html}{generateLearningCurveData}.
If \lstinline!facet = "measure"! learners are mapped to color, and vice
versa.

\begin{lstlisting}[language=R]
r = generateLearningCurveData(
  learners = list("classif.rpart", "classif.knn"),
  task = sonar.task,
  percs = seq(0.1, 1, by = 0.2),
  measures = list(tp, fp, tn, fn),
  resampling = makeResampleDesc(method = "CV", iters = 5),
  show.info = FALSE)
plotLearningCurve(r)
\end{lstlisting}

\includegraphics{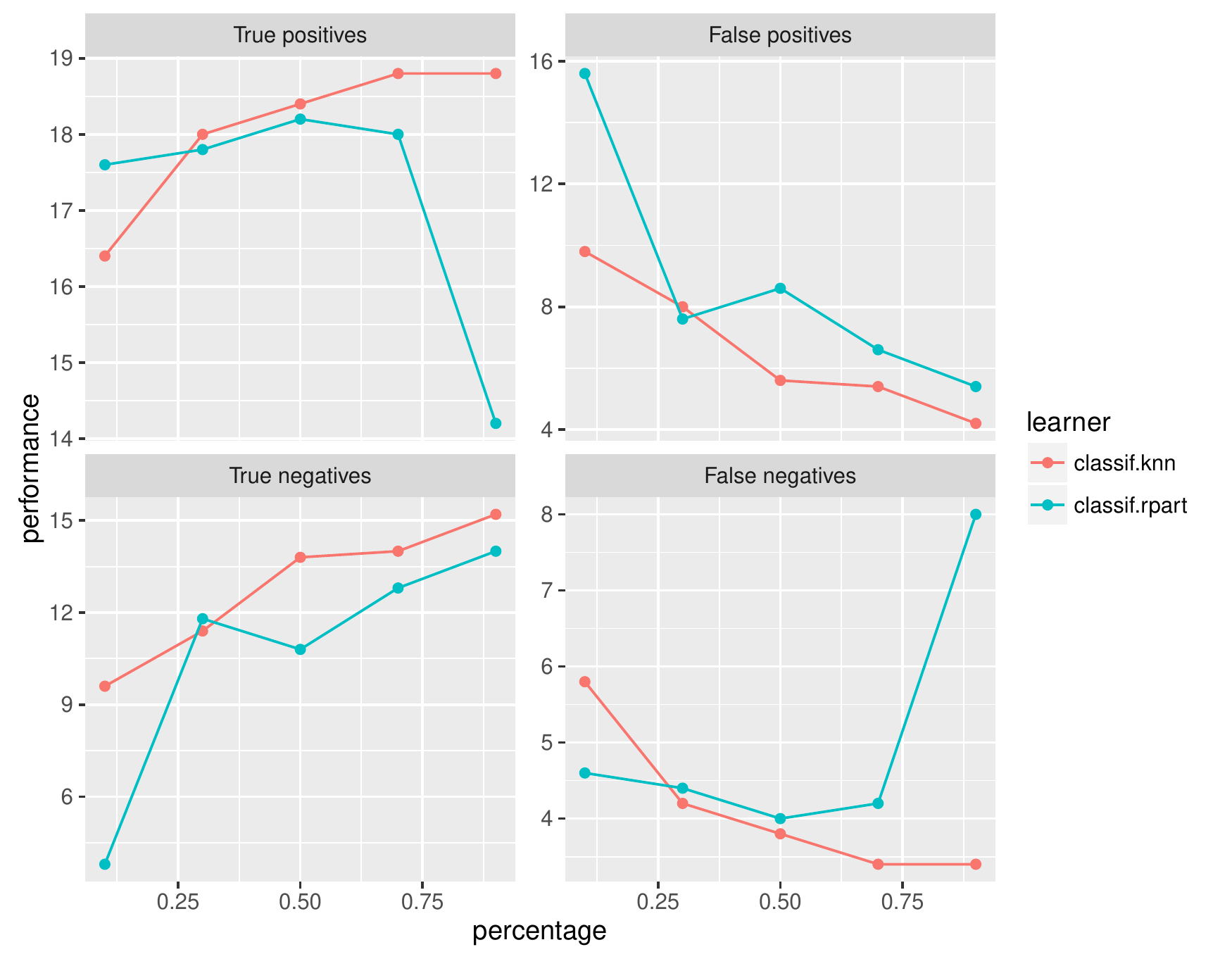}

What happens in
\href{http://www.rdocumentation.org/packages/mlr/functions/generateLearningCurveData.html}{generateLearningCurveData}
is the following: Each learner will be internally wrapped in a
\href{http://www.rdocumentation.org/packages/mlr/functions/makeDownsampleWrapper.html}{DownsampleWrapper}.
To measure the performance at the first step of \lstinline!percs!, say
\lstinline!0.1!, first the data will be split into a \emph{training} and
a \emph{test set} according to the given \emph{resampling strategy}.
Then a random sample containing 10\% of the observations of the
\emph{training set} will be drawn and used to train the learner. The
performance will be measured on the \emph{complete test set}. These
steps will be repeated as defined by the given \emph{resampling method}
and for each value of \lstinline!percs!.

In the first example a simplified usage of the \lstinline!learners!
argument was used, so that it's sufficient to give the \emph{name}. It
is also possible to create a learner the usual way and even to mix it.
Make sure to give different \lstinline!id!s in this case.

\begin{lstlisting}[language=R]
lrns = list(
  makeLearner(cl = "classif.ksvm", id = "ksvm1" , sigma = 0.2, C = 2),
  makeLearner(cl = "classif.ksvm", id = "ksvm2" , sigma = 0.1, C = 1),
  "classif.randomForest"
)
rin = makeResampleDesc(method = "CV", iters = 5)
lc = generateLearningCurveData(learners = lrns, task = sonar.task,
  percs = seq(0.1, 1, by = 0.1), measures = acc,
  resampling = rin, show.info = FALSE)
plotLearningCurve(lc)
\end{lstlisting}

\includegraphics{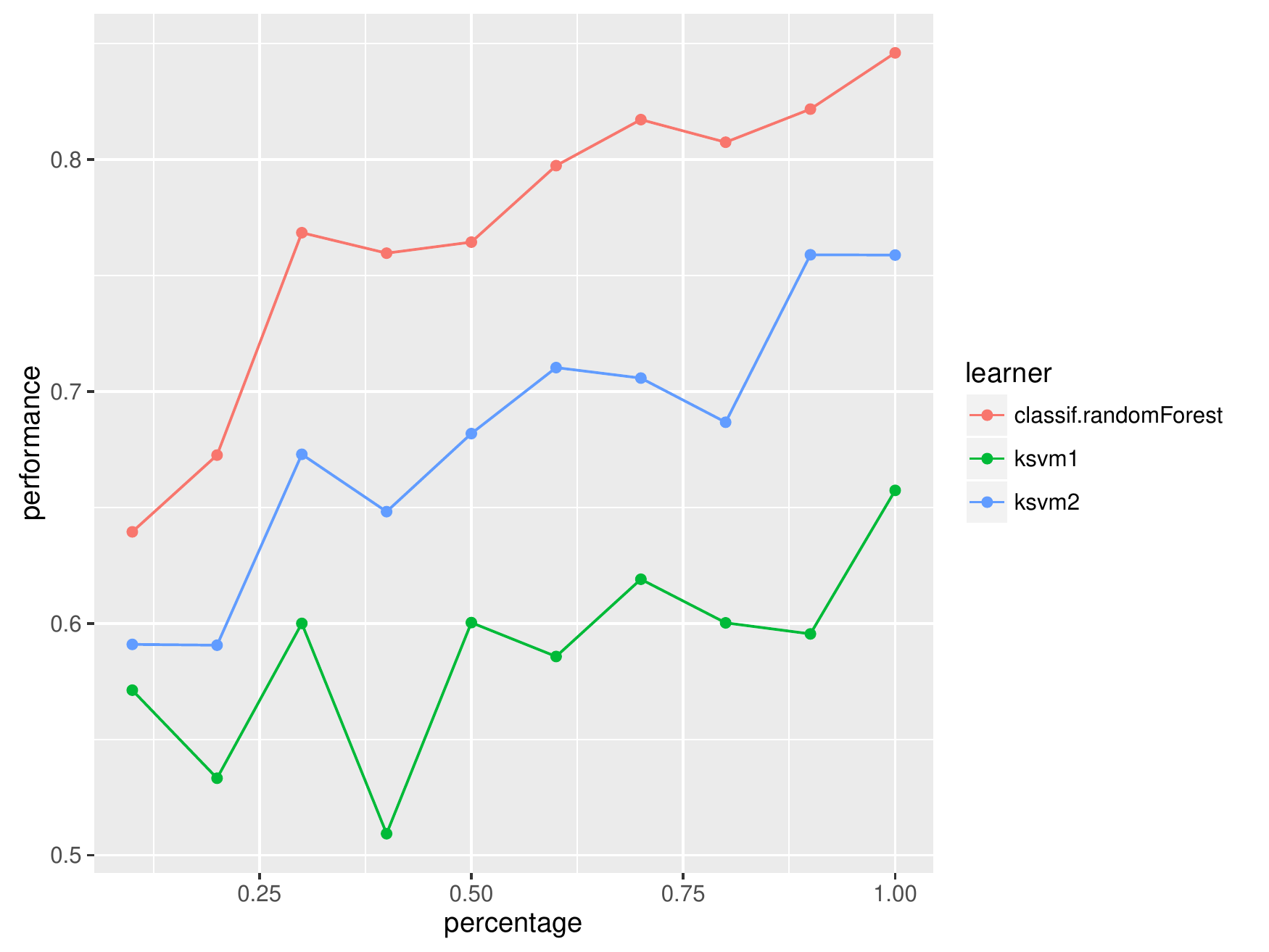}

We can display performance on the train set as well as the test set:

\begin{lstlisting}[language=R]
rin2 = makeResampleDesc(method = "CV", iters = 5, predict = "both")
lc2 = generateLearningCurveData(learners = lrns, task = sonar.task,
  percs = seq(0.1, 1, by = 0.1),
  measures = list(acc,setAggregation(acc, train.mean)), resampling = rin2,
  show.info = FALSE)
plotLearningCurve(lc2, facet = "learner")
\end{lstlisting}

\includegraphics{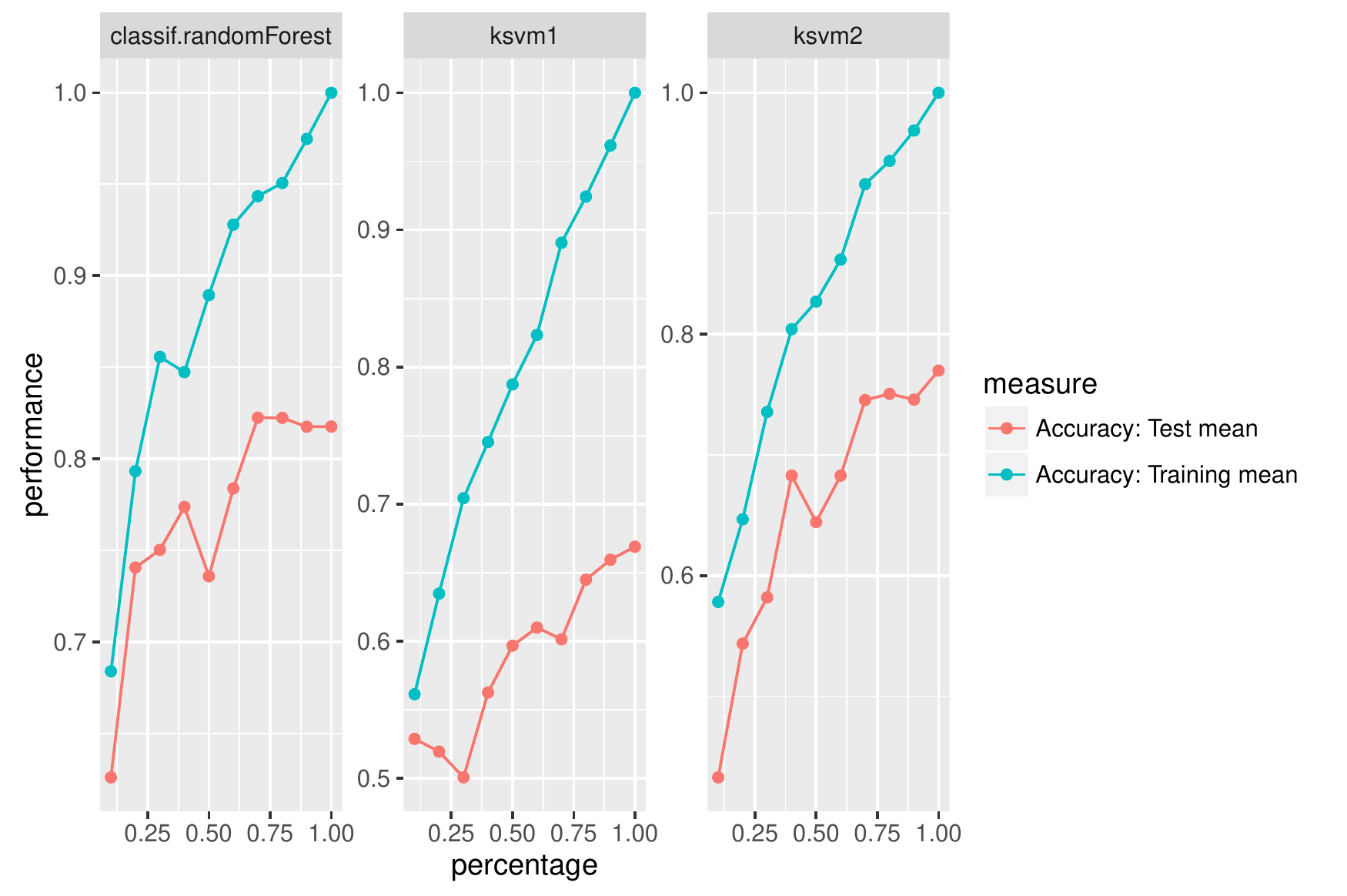}

There is also an experimental
\href{http://www.rdocumentation.org/packages/ggvis/}{ggvis} plotting
function,
\href{http://www.rdocumentation.org/packages/mlr/functions/plotLearningCurveGGVIS.html}{plotLearningCurveGGVIS}.
Instead of the \lstinline!facet! argument to
\href{http://www.rdocumentation.org/packages/mlr/functions/plotLearningCurve.html}{plotLearningCurve}
there is an argument \lstinline!interactive! which plays a similar role.
As subplots are not available in
\href{http://www.rdocumentation.org/packages/ggvis/}{ggvis}, measures or
learners are mapped to an interactive sidebar which allows selection of
the displayed measures or learners. The other feature is mapped to
color.

\begin{lstlisting}[language=R]
plotLearningCurveGGVIS(r, interactive = "measure")
\end{lstlisting}

\hypertarget{exploring-learner-predictions}{\subsection{Exploring
Learner Predictions}\label{exploring-learner-predictions}}

Learners use features to learn a prediction function and make
predictions, but the effect of those features is often not apparent.
\href{http://www.rdocumentation.org/packages/mlr/}{mlr} can estimate the
partial dependence of a learned function on a subset of the feature
space using
\href{http://www.rdocumentation.org/packages/mlr/functions/generatePartialDependenceData.html}{generatePartialDependenceData}.

Partial dependence plots reduce the potentially high dimensional
function estimated by the learner, and display a marginalized version of
this function in a lower dimensional space. For example suppose
\(Y = f(X) + \epsilon\), where \(\mathbb{E}[\epsilon|X] = 0\). With
\((X, Y)\) pairs drawn independently from this statistical model, a
learner may estimate \(\hat{f}\), which, if \(X\) is high dimensional,
can be uninterpretable. Suppose we want to approximate the relationship
between some subset of \(X\). We partition \(X\) into two sets, \(X_s\)
and \(X_c\) such that \(X = X_s \cup X_c\), where \(X_s\) is a subset of
\(X\) of interest.

The partial dependence of \(f\) on \(X_s\) is

\[f_{X_s} = \mathbb{E}_{X_c}f(X_s, X_c).\]

\(X_c\) is integrated out. We use the following estimator:

\[\hat{f}_{X_s} = \frac{1}{N} \sum_{i = 1}^N \hat{f}(X_s, x_{ic}).\]

The individual conditional expectation of an observation can also be
estimated using the above algorithm absent the averaging, giving
\(\hat{f}^{(i)}_{X_s}\). This allows the discovery of features of
\(\hat{f}\) that may be obscured by an aggregated summary of
\(\hat{f}\).

The partial derivative of the partial dependence function,
\(\frac{\partial \hat{f}_{X_s}}{\partial X_s}\), and the individual
conditional expectation function,
\(\frac{\partial \hat{f}^{(i)}_{X_s}}{\partial X_s}\), can also be
computed. For regression and survival tasks the partial derivative of a
single feature \(X_s\) is the gradient of the partial dependence
function, and for classification tasks where the learner can output
class probabilities the Jacobian. Note that if the learner produces
discontinuous partial dependence (e.g., piecewise constant functions
such as decision trees, ensembles of decision trees, etc.) the
derivative will be 0 (where the function is not changing) or trending
towards positive or negative infinity (at the discontinuities where the
derivative is undefined). Plotting the partial dependence function of
such learners may give the impression that the function is not
discontinuous because the prediction grid is not composed of all
discontinuous points in the predictor space. This results in a line
interpolating that makes the function appear to be piecewise linear
(where the derivative would be defined except at the boundaries of each
piece).

The partial derivative can be informative regarding the additivity of
the learned function in certain features. If \(\hat{f}^{(i)}_{X_s}\) is
an additive function in a feature \(X_s\), then its partial derivative
will not depend on any other features (\(X_c\)) that may have been used
by the learner. Variation in the estimated partial derivative indicates
that there is a region of interaction between \(X_s\) and \(X_c\) in
\(\hat{f}\). Similarly, instead of using the mean to estimate the
expected value of the function at different values of \(X_s\), instead
computing the variance can highlight regions of interaction between
\(X_s\) and \(X_c\).

See \href{http://arxiv.org/abs/1309.6392}{Goldstein, Kapelner, Bleich,
and Pitkin (2014)} for more details and their package
\href{http://www.rdocumentation.org/packages/ICEbox/}{ICEbox} for the
original implementation. The algorithm works for any supervised learner
with classification, regression, and survival tasks.

\subsubsection{Generating partial
dependences}\label{generating-partial-dependences}

Our implementation, following
\href{http://www.rdocumentation.org/packages/mlr/}{mlr}'s
\protect\hyperlink{visualization}{visualization} pattern, consists of
the above mentioned function
\href{http://www.rdocumentation.org/packages/mlr/functions/generatePartialDependenceData.html}{generatePartialDependenceData},
as well as two visualization functions,
\href{http://www.rdocumentation.org/packages/mlr/functions/plotPartialDependence.html}{plotPartialDependence}
and
\href{http://www.rdocumentation.org/packages/mlr/functions/plotPartialDependenceGGVIS.html}{plotPartialDependenceGGVIS}.
The former generates input (objects of class
\href{http://www.rdocumentation.org/packages/mlr/functions/PartialDependenceData.html}{PartialDependenceData})
for the latter.

The first step executed by
\href{http://www.rdocumentation.org/packages/mlr/functions/generatePartialDependenceData.html}{generatePartialDependenceData}
is to generate a feature grid for every element of the character vector
\lstinline!features! passed. The data are given by the \lstinline!input!
argument, which can be a
\href{http://www.rdocumentation.org/packages/mlr/functions/Task.html}{Task}
or a
\href{http://www.rdocumentation.org/packages/base/functions/data.frame.html}{data.frame}.
The feature grid can be generated in several ways. A uniformly spaced
grid of length \lstinline!gridsize! (default 10) from the empirical
minimum to the empirical maximum is created by default, but arguments
\lstinline!fmin! and \lstinline!fmax! may be used to override the
empirical default (the lengths of \lstinline!fmin! and \lstinline!fmax!
must match the length of \lstinline!features!). Alternatively the
feature data can be resampled, either by using a bootstrap or by
subsampling.

\begin{lstlisting}[language=R]
lrn.classif = makeLearner("classif.ksvm", predict.type = "prob")
fit.classif = train(lrn.classif, iris.task)
pd = generatePartialDependenceData(fit.classif, iris.task, "Petal.Width")
pd
#> PartialDependenceData
#> Task: iris-example
#> Features: Petal.Width
#> Target: setosa, versicolor, virginica
#> Derivative: FALSE
#> Interaction: FALSE
#> Individual: FALSE
#>    Class Probability Petal.Width
#> 1 setosa   0.4983925   0.1000000
#> 2 setosa   0.4441165   0.3666667
#> 3 setosa   0.3808075   0.6333333
#> 4 setosa   0.3250243   0.9000000
#> 5 setosa   0.2589014   1.1666667
#> 6 setosa   0.1870692   1.4333333
#> ... (30 rows, 3 cols)
\end{lstlisting}

As noted above, \(X_s\) does not have to be unidimensional. If it is
not, the \lstinline!interaction! flag must be set to \lstinline!TRUE!.
Then the individual feature grids are combined using the Cartesian
product, and the estimator above is applied, producing the partial
dependence for every combination of unique feature values. If the
\lstinline!interaction! flag is \lstinline!FALSE! (the default) then by
default \(X_s\) is assumed unidimensional, and partial dependencies are
generated for each feature separately. The resulting output when
\lstinline!interaction = FALSE! has a column for each feature, and
\lstinline!NA! where the feature was not used.

\begin{lstlisting}[language=R]
pd.lst = generatePartialDependenceData(fit.classif, iris.task, c("Petal.Width", "Petal.Length"), FALSE)
head(pd.lst$data)
#>    Class Probability Petal.Width Petal.Length
#> 1 setosa   0.4983925   0.1000000           NA
#> 2 setosa   0.4441165   0.3666667           NA
#> 3 setosa   0.3808075   0.6333333           NA
#> 4 setosa   0.3250243   0.9000000           NA
#> 5 setosa   0.2589014   1.1666667           NA
#> 6 setosa   0.1870692   1.4333333           NA

tail(pd.lst$data)
#>        Class Probability Petal.Width Petal.Length
#> 55 virginica   0.2006336          NA     3.622222
#> 56 virginica   0.3114545          NA     4.277778
#> 57 virginica   0.4404613          NA     4.933333
#> 58 virginica   0.6005358          NA     5.588889
#> 59 virginica   0.7099841          NA     6.244444
#> 60 virginica   0.7242584          NA     6.900000
\end{lstlisting}

\begin{lstlisting}[language=R]
pd.int = generatePartialDependenceData(fit.classif, iris.task, c("Petal.Width", "Petal.Length"), TRUE)
pd.int
#> PartialDependenceData
#> Task: iris-example
#> Features: Petal.Width, Petal.Length
#> Target: setosa, versicolor, virginica
#> Derivative: FALSE
#> Interaction: TRUE
#> Individual: FALSE
#>    Class Probability Petal.Width Petal.Length
#> 1 setosa   0.6885025   0.1000000            1
#> 2 setosa   0.6824560   0.3666667            1
#> 3 setosa   0.6459476   0.6333333            1
#> 4 setosa   0.5750861   0.9000000            1
#> 5 setosa   0.4745925   1.1666667            1
#> 6 setosa   0.3749285   1.4333333            1
#> ... (300 rows, 4 cols)
\end{lstlisting}

At each step in the estimation of \(\hat{f}_{X_s}\) a set of predictions
of length \(N\) is generated. By default the mean prediction is used.
For classification where \lstinline!predict.type = "prob"! this entails
the mean class probabilities. However, other summaries of the
predictions may be used. For regression and survival tasks the function
used here must either return one number or three, and, if the latter,
the numbers must be sorted lowest to highest. For classification tasks
the function must return a number for each level of the target feature.

As noted, the \lstinline!fun! argument can be a function which returns
three numbers (sorted low to high) for a regression task. This allows
further exploration of relative feature importance. If a feature is
relatively important, the bounds are necessarily tighter because the
feature accounts for more of the variance of the predictions, i.e., it
is ``used'' more by the learner. More directly setting
\lstinline!fun = var! identifies regions of interaction between \(X_s\)
and \(X_c\).

\begin{lstlisting}[language=R]
lrn.regr = makeLearner("regr.ksvm")
fit.regr = train(lrn.regr, bh.task)
pd.regr = generatePartialDependenceData(fit.regr, bh.task, "lstat", fun = median)
pd.regr
#> PartialDependenceData
#> Task: BostonHousing-example
#> Features: lstat
#> Target: medv
#> Derivative: FALSE
#> Interaction: FALSE
#> Individual: FALSE
#>       medv     lstat
#> 1 24.69031  1.730000
#> 2 23.72479  5.756667
#> 3 22.34841  9.783333
#> 4 20.78817 13.810000
#> 5 19.76183 17.836667
#> 6 19.33115 21.863333
#> ... (10 rows, 2 cols)
\end{lstlisting}

\begin{lstlisting}[language=R]
pd.ci = generatePartialDependenceData(fit.regr, bh.task, "lstat",
  fun = function(x) quantile(x, c(.25, .5, .75)))
pd.ci
#> PartialDependenceData
#> Task: BostonHousing-example
#> Features: lstat
#> Target: medv
#> Derivative: FALSE
#> Interaction: FALSE
#> Individual: FALSE
#>       medv     lstat    lower    upper
#> 1 24.69031  1.730000 21.36068 29.75615
#> 2 23.72479  5.756667 20.80590 28.02338
#> 3 22.34841  9.783333 20.06507 25.22291
#> 4 20.78817 13.810000 18.55592 23.68100
#> 5 19.76183 17.836667 16.52737 22.98520
#> 6 19.33115 21.863333 15.14425 22.12766
#> ... (10 rows, 4 cols)
\end{lstlisting}

\begin{lstlisting}[language=R]
pd.classif = generatePartialDependenceData(fit.classif, iris.task, "Petal.Length", fun = median)
pd.classif
#> PartialDependenceData
#> Task: iris-example
#> Features: Petal.Length
#> Target: setosa, versicolor, virginica
#> Derivative: FALSE
#> Interaction: FALSE
#> Individual: FALSE
#>    Class Probability Petal.Length
#> 1 setosa  0.31008788     1.000000
#> 2 setosa  0.24271454     1.655556
#> 3 setosa  0.17126036     2.311111
#> 4 setosa  0.09380787     2.966667
#> 5 setosa  0.04579912     3.622222
#> 6 setosa  0.02455344     4.277778
#> ... (30 rows, 3 cols)
\end{lstlisting}

In addition to bounds based on a summary of the distribution of the
conditional expectation of each observation, learners which can estimate
the variance of their predictions can also be used. The argument
\lstinline!bounds! is a numeric vector of length two which is added (so
the first number should be negative) to the point prediction to produce
a confidence interval for the partial dependence. The default is the
.025 and .975 quantiles of the Gaussian distribution.

\begin{lstlisting}[language=R]
fit.se = train(makeLearner("regr.randomForest", predict.type = "se"), bh.task)
pd.se = generatePartialDependenceData(fit.se, bh.task, c("lstat", "crim"))
head(pd.se$data)
#>       medv     lstat crim    lower    upper
#> 1 31.02186  1.730000   NA 27.65357 34.39015
#> 2 25.94429  5.756667   NA 23.43079 28.45779
#> 3 23.52758  9.783333   NA 21.23661 25.81856
#> 4 22.05223 13.810000   NA 20.30446 23.80000
#> 5 20.44293 17.836667   NA 18.72603 22.15982
#> 6 19.80143 21.863333   NA 18.04932 21.55353

tail(pd.se$data)
#>        medv lstat     crim    lower    upper
#> 15 21.65846    NA 39.54849 19.50827 23.80866
#> 16 21.64409    NA 49.43403 19.49704 23.79114
#> 17 21.63038    NA 59.31957 19.48054 23.78023
#> 18 21.61514    NA 69.20512 19.46092 23.76936
#> 19 21.61969    NA 79.09066 19.46819 23.77119
#> 20 21.61987    NA 88.97620 19.46843 23.77130
\end{lstlisting}

As previously mentioned if the aggregation function is not used, i.e.,
it is the identity, then the conditional expectation of
\(\hat{f}^{(i)}_{X_s}\) is estimated. If \lstinline!individual = TRUE!
then
\href{http://www.rdocumentation.org/packages/mlr/functions/generatePartialDependenceData.html}{generatePartialDependenceData}
returns \(n\) partial dependence estimates made at each point in the
prediction grid constructed from the features.

\begin{lstlisting}[language=R]
pd.ind.regr = generatePartialDependenceData(fit.regr, bh.task, "lstat", individual = TRUE)
pd.ind.regr
#> PartialDependenceData
#> Task: BostonHousing-example
#> Features: lstat
#> Target: medv
#> Derivative: FALSE
#> Interaction: FALSE
#> Individual: TRUE
#> Predictions centered: FALSE
#>       medv     lstat idx
#> 1 25.66995  1.730000   1
#> 2 24.71747  5.756667   1
#> 3 23.64157  9.783333   1
#> 4 22.70812 13.810000   1
#> 5 22.00059 17.836667   1
#> 6 21.46195 21.863333   1
#> ... (5060 rows, 3 cols)
\end{lstlisting}

The resulting output, particularly the element \lstinline!data! in the
returned object, has an additional column \lstinline!idx! which gives
the index of the observation to which the row pertains.

For classification tasks this index references both the class and the
observation index.

\begin{lstlisting}[language=R]
pd.ind.classif = generatePartialDependenceData(fit.classif, iris.task, "Petal.Length", individual = TRUE)
pd.ind.classif
#> PartialDependenceData
#> Task: iris-example
#> Features: Petal.Length
#> Target: setosa, versicolor, virginica
#> Derivative: FALSE
#> Interaction: FALSE
#> Individual: TRUE
#> Predictions centered: FALSE
#>    Class Probability Petal.Length      idx
#> 1 setosa   0.9814053            1 1.setosa
#> 2 setosa   0.9747355            1 2.setosa
#> 3 setosa   0.9815516            1 3.setosa
#> 4 setosa   0.9795761            1 4.setosa
#> 5 setosa   0.9806494            1 5.setosa
#> 6 setosa   0.9758763            1 6.setosa
#> ... (4500 rows, 4 cols)
\end{lstlisting}

Individual estimates of partial dependence can also be centered by
predictions made at all \(n\) observations for a particular point in the
prediction grid created by the features. This is controlled by the
argument \lstinline!center! which is a list of the same length as the
length of the \lstinline!features! argument and contains the values of
the \lstinline!features! desired.

\begin{lstlisting}[language=R]
iris = getTaskData(iris.task)
pd.ind.classif = generatePartialDependenceData(fit.classif, iris.task, "Petal.Length", individual = TRUE,
  center = list("Petal.Length" = min(iris$Petal.Length)))
\end{lstlisting}

Partial derivatives can also be computed for individual partial
dependence estimates and aggregate partial dependence. This is
restricted to a single feature at a time. The derivatives of individual
partial dependence estimates can be useful in finding regions of
interaction between the feature for which the derivative is estimated
and the features excluded.

\begin{lstlisting}[language=R]
pd.regr.der = generatePartialDependenceData(fit.regr, bh.task, "lstat", derivative = TRUE)
head(pd.regr.der$data)
#>         medv     lstat
#> 1 -0.1792626  1.730000
#> 2 -0.3584207  5.756667
#> 3 -0.4557666  9.783333
#> 4 -0.4523905 13.810000
#> 5 -0.3700880 17.836667
#> 6 -0.2471346 21.863333
\end{lstlisting}

\begin{lstlisting}[language=R]
pd.regr.der.ind = generatePartialDependenceData(fit.regr, bh.task, "lstat", derivative = TRUE,
  individual = TRUE)
head(pd.regr.der.ind$data)
#>         medv     lstat idx
#> 1 -0.1931323  1.730000   1
#> 2 -0.2656911  5.756667   1
#> 3 -0.2571006  9.783333   1
#> 4 -0.2033080 13.810000   1
#> 5 -0.1511472 17.836667   1
#> 6 -0.1193129 21.863333   1
\end{lstlisting}

\begin{lstlisting}[language=R]
pd.classif.der = generatePartialDependenceData(fit.classif, iris.task, "Petal.Width", derivative = TRUE)
head(pd.classif.der$data)
#>    Class Probability Petal.Width
#> 1 setosa  -0.1479385   0.1000000
#> 2 setosa  -0.2422728   0.3666667
#> 3 setosa  -0.2189893   0.6333333
#> 4 setosa  -0.2162803   0.9000000
#> 5 setosa  -0.2768042   1.1666667
#> 6 setosa  -0.2394176   1.4333333
\end{lstlisting}

\begin{lstlisting}[language=R]
pd.classif.der.ind = generatePartialDependenceData(fit.classif, iris.task, "Petal.Width", derivative = TRUE,
  individual = TRUE)
head(pd.classif.der.ind$data)
#>    Class Probability Petal.Width      idx
#> 1 setosa  0.02479474         0.1 1.setosa
#> 2 setosa  0.01710561         0.1 2.setosa
#> 3 setosa  0.01646252         0.1 3.setosa
#> 4 setosa  0.01530718         0.1 4.setosa
#> 5 setosa  0.02608577         0.1 5.setosa
#> 6 setosa  0.03925531         0.1 6.setosa
\end{lstlisting}

\subsubsection{Functional ANOVA}\label{functional-anova}

\href{http://dl.acm.org/citation.cfm?id=1014122}{Hooker (2004)} proposed
the decomposition of a learned function \(\hat{f}\) as a sum of lower
dimensional functions
\[f(\mathbf{x}) = g_0 + \sum_{i = 1}^p g_{i}(X_i) + \sum_{i \neq j} g_{ij}(x_{ij}) + \ldots\]
where \(p\) is the number of features.
\href{http://www.rdocumentation.org/packages/mlr/functions/generateFunctionalANOVAData.html}{generateFunctionalANOVAData}
estimates the individual \(g\) functions using partial dependence. When
functions depend only on one feature, they are equivalent to partial
dependence, but a \(g\) function which depends on more than one feature
is the ``effect'' of only those features: lower dimensional ``effects''
are removed.

\[\hat{g}_u(X) = \frac{1}{N} \sum_{i = 1}^N \left( \hat{f}(X) - \sum_{v \subset u} g_v(X) \right)\]

Here \(u\) is a subset of \({1, \ldots, p}\). When \(|v| = 1\) \(g_v\)
can be directly computed by computing the bivariate partial dependence
of \(\hat{f}\) on \(X_u\) and then subtracting off the univariate
partial dependences of the features contained in \(v\).

Although this decomposition is generalizable to classification it is
currently only available for regression tasks.

\begin{lstlisting}[language=R]
lrn.regr = makeLearner("regr.ksvm")
fit.regr = train(lrn.regr, bh.task)

fa = generateFunctionalANOVAData(fit.regr, bh.task, "lstat", depth = 1, fun = median)
fa
#> FunctionalANOVAData
#> Task: BostonHousing-example
#> Features: lstat
#> Target: medv
#> 
#> 
#>   effect     medv     lstat
#> 1  lstat 24.89524  1.730000
#> 2  lstat 23.76025  5.756667
#> 3  lstat 22.35052  9.783333
#> 4  lstat 20.68435 13.810000
#> 5  lstat 19.60062 17.836667
#> 6  lstat 19.01178 21.863333
#> ... (10 rows, 3 cols)

pd.regr = generatePartialDependenceData(fit.regr, bh.task, "lstat", fun = median)
pd.regr
#> PartialDependenceData
#> Task: BostonHousing-example
#> Features: lstat
#> Target: medv
#> Derivative: FALSE
#> Interaction: FALSE
#> Individual: FALSE
#>       medv     lstat
#> 1 24.89524  1.730000
#> 2 23.76025  5.756667
#> 3 22.35052  9.783333
#> 4 20.68435 13.810000
#> 5 19.60062 17.836667
#> 6 19.01178 21.863333
#> ... (10 rows, 2 cols)
\end{lstlisting}

The \lstinline!depth! argument is similar to the \lstinline!interaction!
argument in
\href{http://www.rdocumentation.org/packages/mlr/functions/generatePartialDependenceData.html}{generatePartialDependenceData}
but instead of specifying whether all of joint ``effect'' of all the
\lstinline!features! is computed, it determines whether ``effects'' of
all subsets of the features given the specified \lstinline!depth! are
computed. So, for example, with \(p\) features and depth 1, the
univariate partial dependence is returned. If, instead,
\lstinline!depth = 2!, then all possible bivariate functional ANOVA
effects are returned. This is done by computing the univariate partial
dependence for each feature and subtracting it from the bivariate
partial dependence for each possible pair.

\begin{lstlisting}[language=R]
fa.bv = generateFunctionalANOVAData(fit.regr, bh.task, c("crim", "lstat", "age"),
  depth = 2)
fa.bv
#> FunctionalANOVAData
#> Task: BostonHousing-example
#> Features: crim, lstat, age
#> Target: medv
#> 
#> 
#>       effect      medv      crim lstat age
#> 1 crim:lstat -22.68734  0.006320  1.73  NA
#> 2 crim:lstat -23.22114  9.891862  1.73  NA
#> 3 crim:lstat -24.77479 19.777404  1.73  NA
#> 4 crim:lstat -26.41395 29.662947  1.73  NA
#> 5 crim:lstat -27.56524 39.548489  1.73  NA
#> 6 crim:lstat -28.27952 49.434031  1.73  NA
#> ... (300 rows, 5 cols)

names(table(fa.bv$data$effect)) ## interaction effects estimated
#> [1] "crim:age"   "crim:lstat" "lstat:age"
\end{lstlisting}

\subsubsection{Plotting partial
dependences}\label{plotting-partial-dependences}

Results from
\href{http://www.rdocumentation.org/packages/mlr/functions/generatePartialDependenceData.html}{generatePartialDependenceData}
and
\href{http://www.rdocumentation.org/packages/mlr/functions/generateFunctionalANOVAData.html}{generateFunctionalANOVAData}
can be visualized with
\href{http://www.rdocumentation.org/packages/mlr/functions/plotPartialDependence.html}{plotPartialDependence}
and
\href{http://www.rdocumentation.org/packages/mlr/functions/plotPartialDependenceGGVIS.html}{plotPartialDependenceGGVIS}.

With one feature and a regression task the output is a line plot, with a
point for each point in the corresponding feature's grid.

\begin{lstlisting}[language=R]
plotPartialDependence(pd.regr)
\end{lstlisting}

\includegraphics{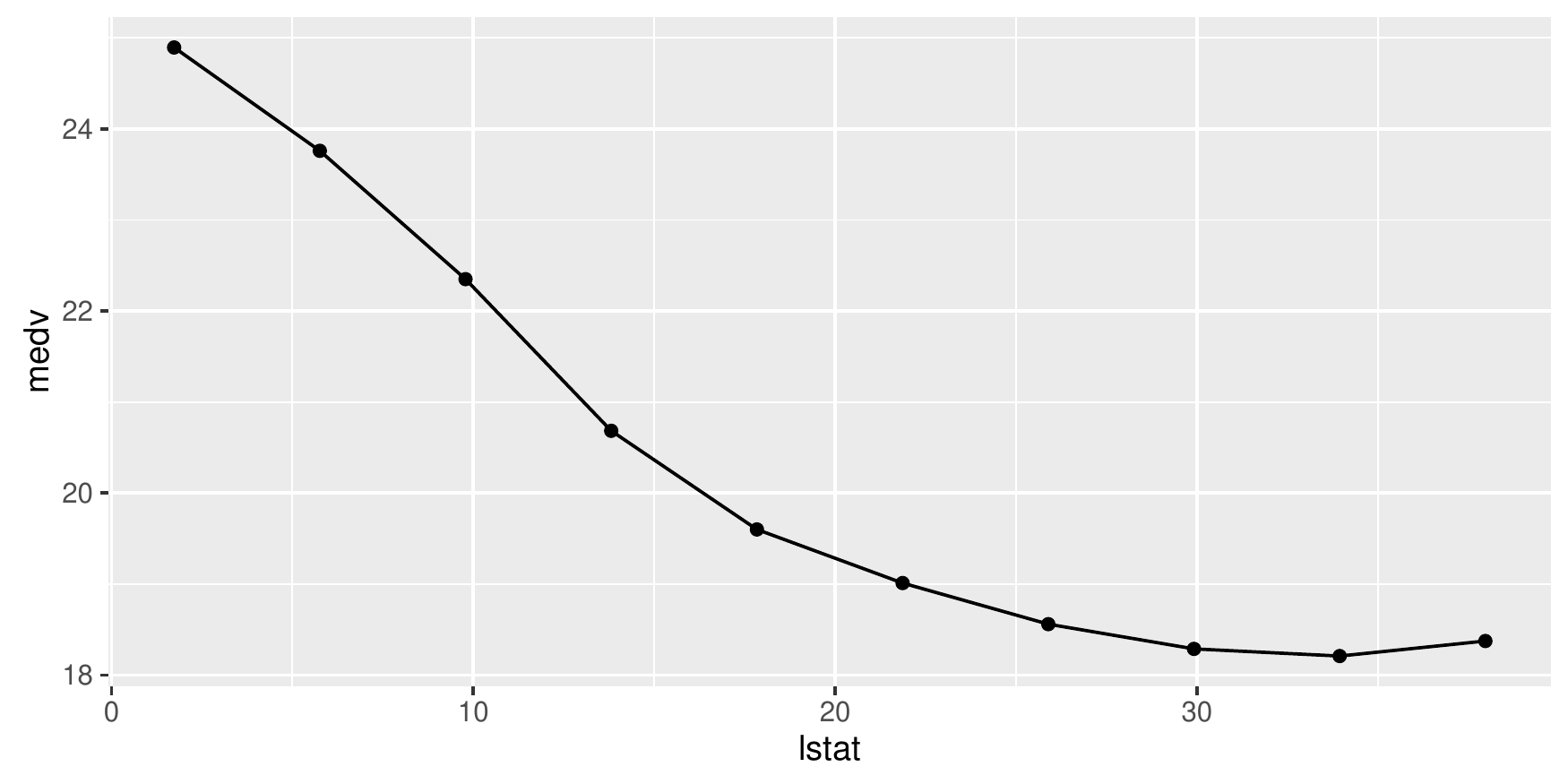}

With a classification task, a line is drawn for each class, which gives
the estimated partial probability of that class for a particular point
in the feature grid.

\begin{lstlisting}[language=R]
plotPartialDependence(pd.classif)
\end{lstlisting}

\includegraphics{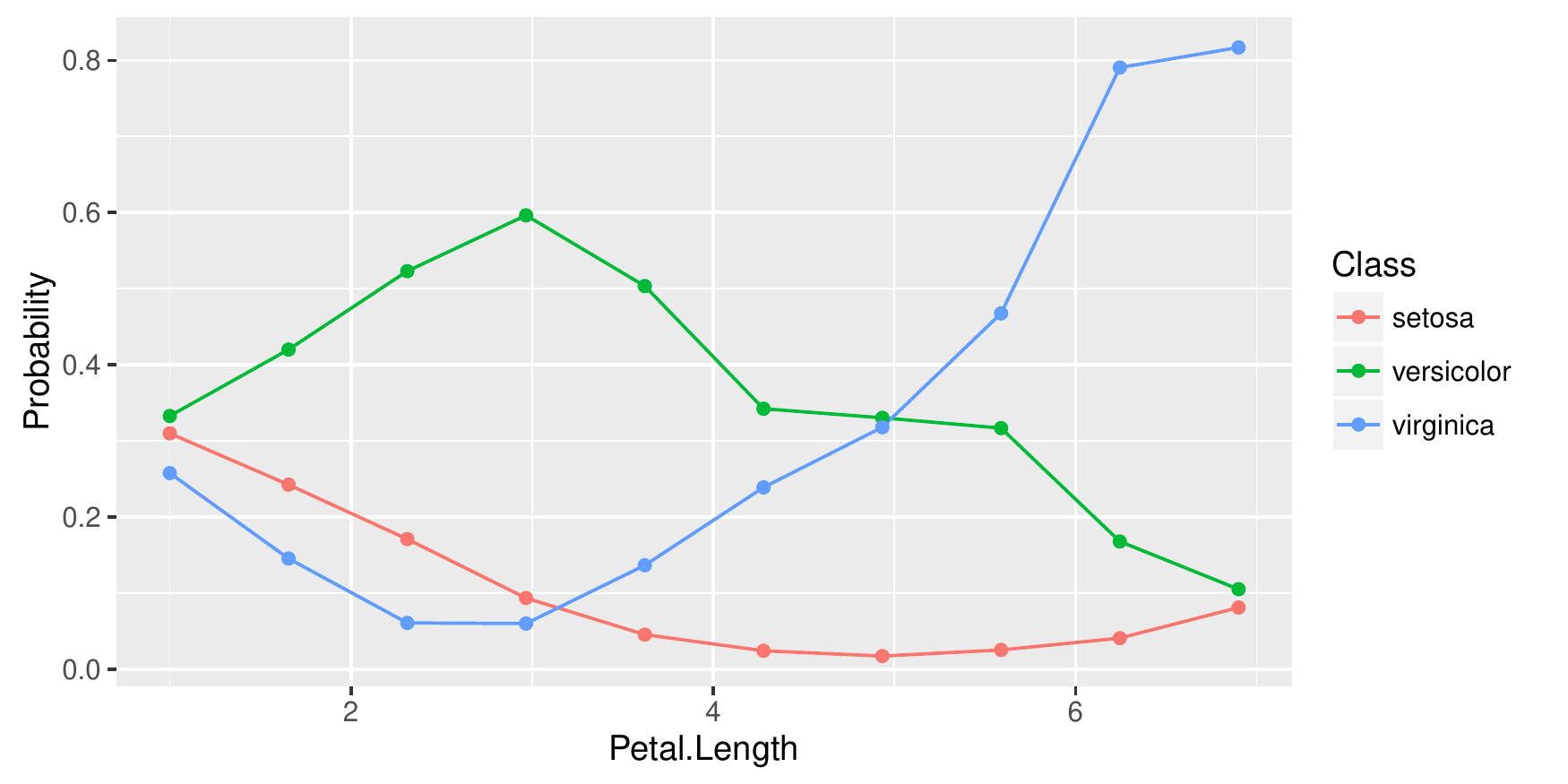}

For regression tasks, when the \lstinline!fun! argument of
\href{http://www.rdocumentation.org/packages/mlr/functions/generatePartialDependenceData.html}{generatePartialDependenceData}
is used, the bounds will automatically be displayed using a gray ribbon.

\begin{lstlisting}[language=R]
plotPartialDependence(pd.ci)
\end{lstlisting}

\includegraphics{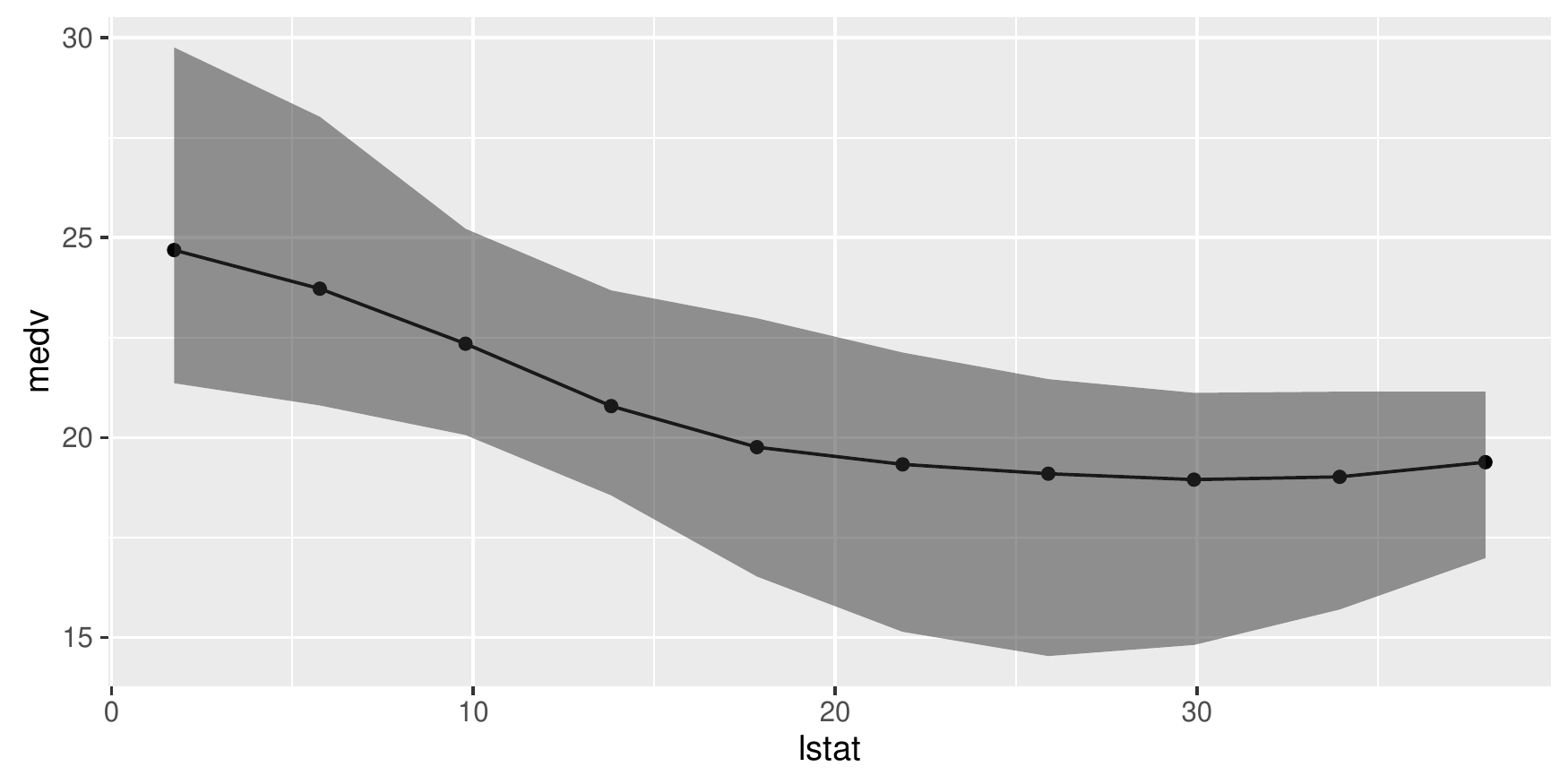}

The same goes for plots of partial dependences where the learner has
\lstinline!predict.type = "se"!.

\begin{lstlisting}[language=R]
plotPartialDependence(pd.se)
\end{lstlisting}

\includegraphics{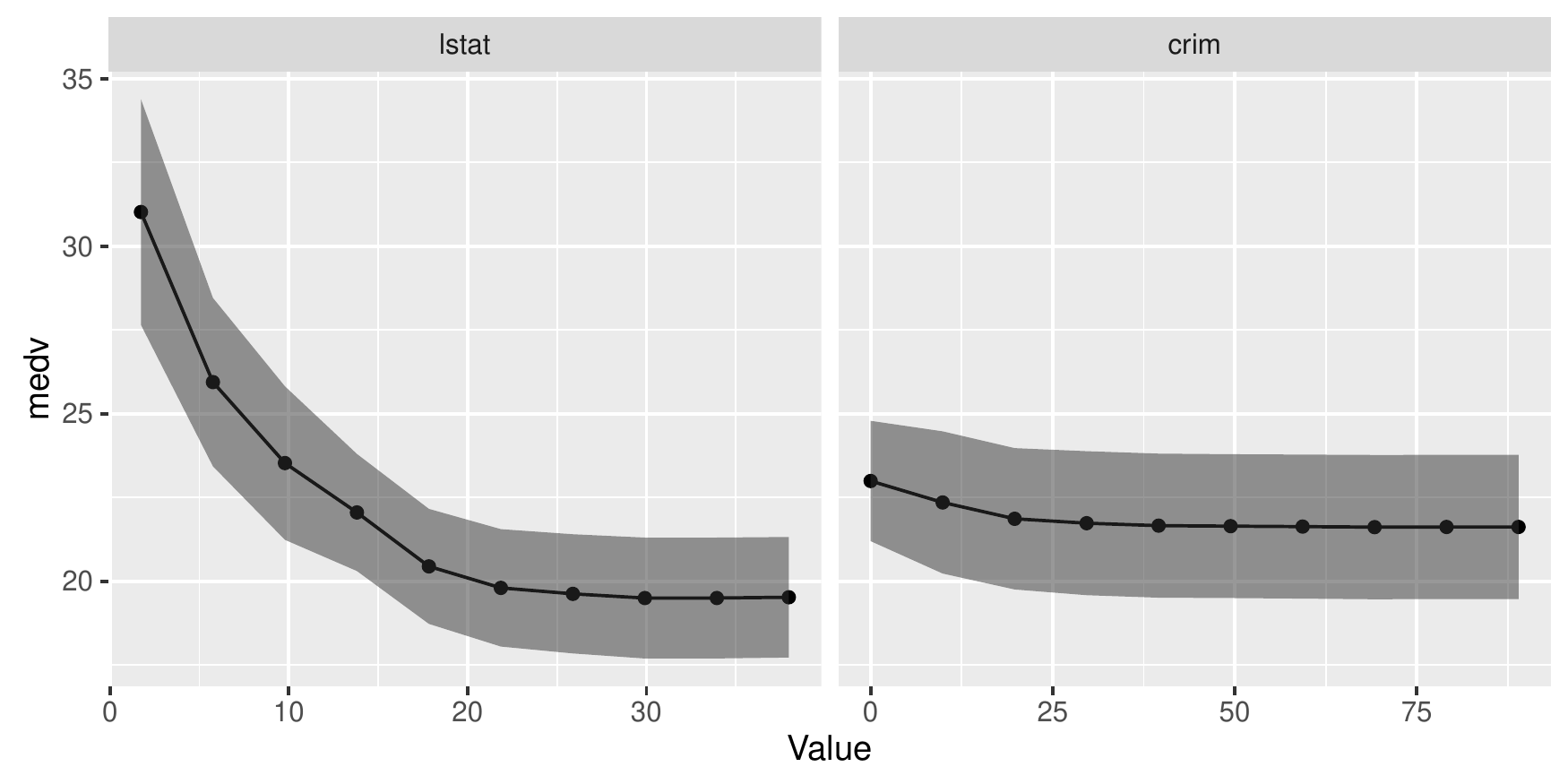}

When multiple features are passed to
\href{http://www.rdocumentation.org/packages/mlr/functions/generatePartialDependenceData.html}{generatePartialDependenceData}
but \lstinline!interaction = FALSE!, facetting is used to display each
estimated bivariate relationship.

\begin{lstlisting}[language=R]
plotPartialDependence(pd.lst)
\end{lstlisting}

\includegraphics{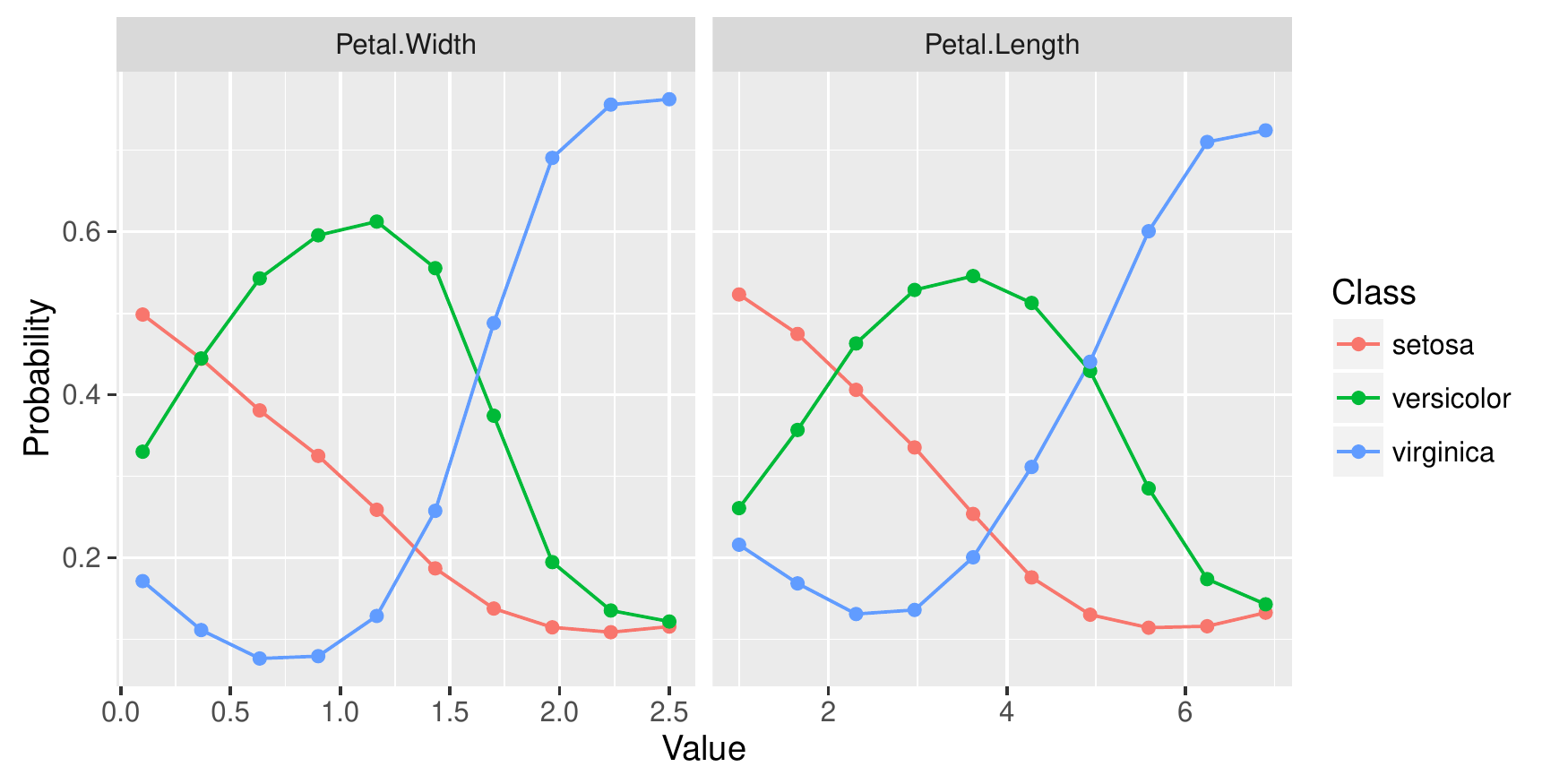}

When \lstinline!interaction = TRUE! in the call to
\href{http://www.rdocumentation.org/packages/mlr/functions/generatePartialDependenceData.html}{generatePartialDependenceData},
one variable must be chosen to be used for facetting, and a subplot for
each value in the chosen feature's grid is created, wherein the other
feature's partial dependences within the facetting feature's value are
shown. Note that this type of plot is limited to two features.

\begin{lstlisting}[language=R]
plotPartialDependence(pd.int, facet = "Petal.Length")
\end{lstlisting}

\includegraphics{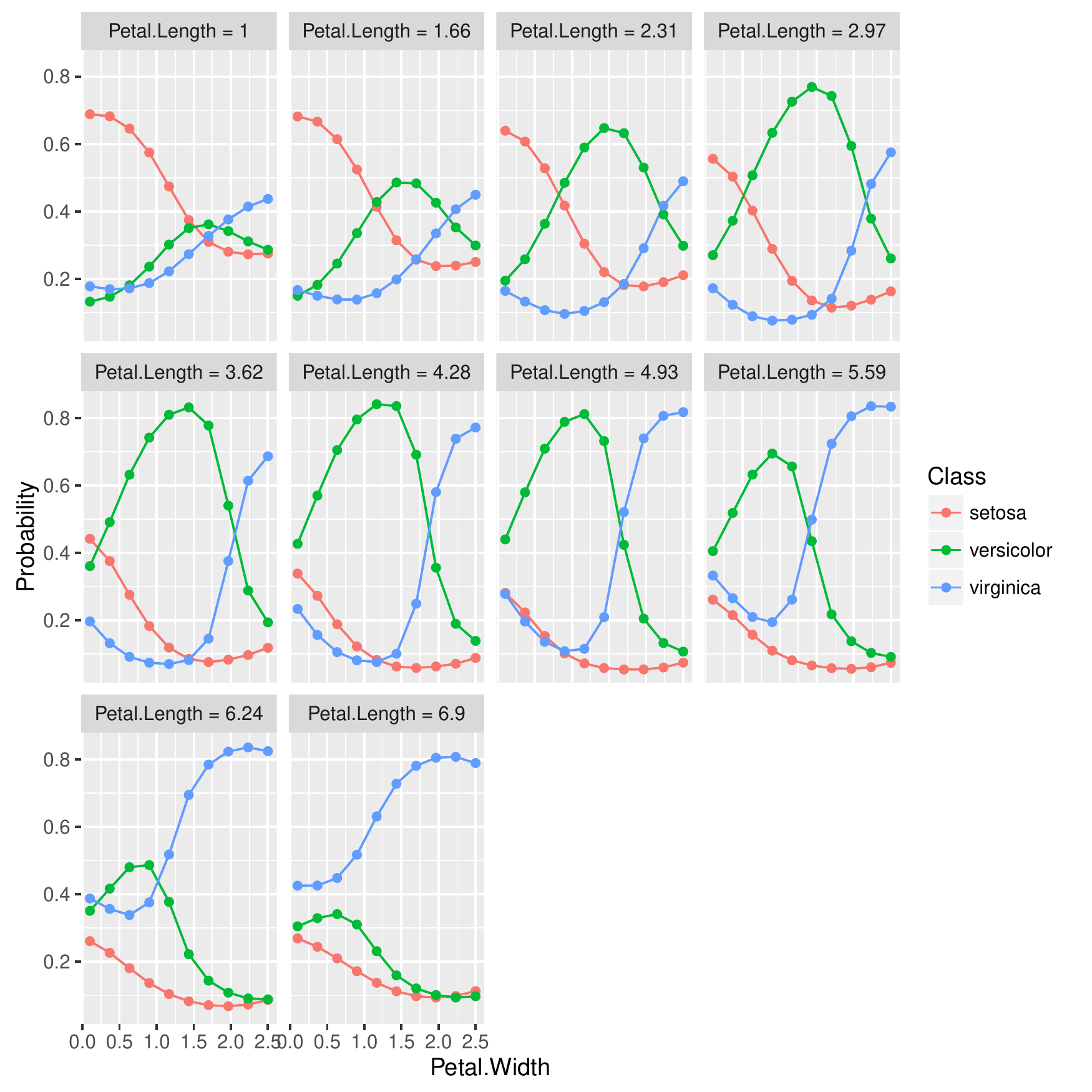}

\href{http://www.rdocumentation.org/packages/mlr/functions/plotPartialDependenceGGVIS.html}{plotPartialDependenceGGVIS}
can be used similarly, however, since
\href{http://www.rdocumentation.org/packages/ggvis/}{ggvis} currently
lacks subplotting/facetting capabilities, the argument
\lstinline!interact! maps one feature to an interactive sidebar where
the user can select a value of one feature.

\begin{lstlisting}[language=R]
plotPartialDependenceGGVIS(pd.int, interact = "Petal.Length")
\end{lstlisting}

When \lstinline!individual = TRUE! each individual conditional
expectation curve is plotted.

\begin{lstlisting}[language=R]
plotPartialDependence(pd.ind.regr)
\end{lstlisting}

\includegraphics{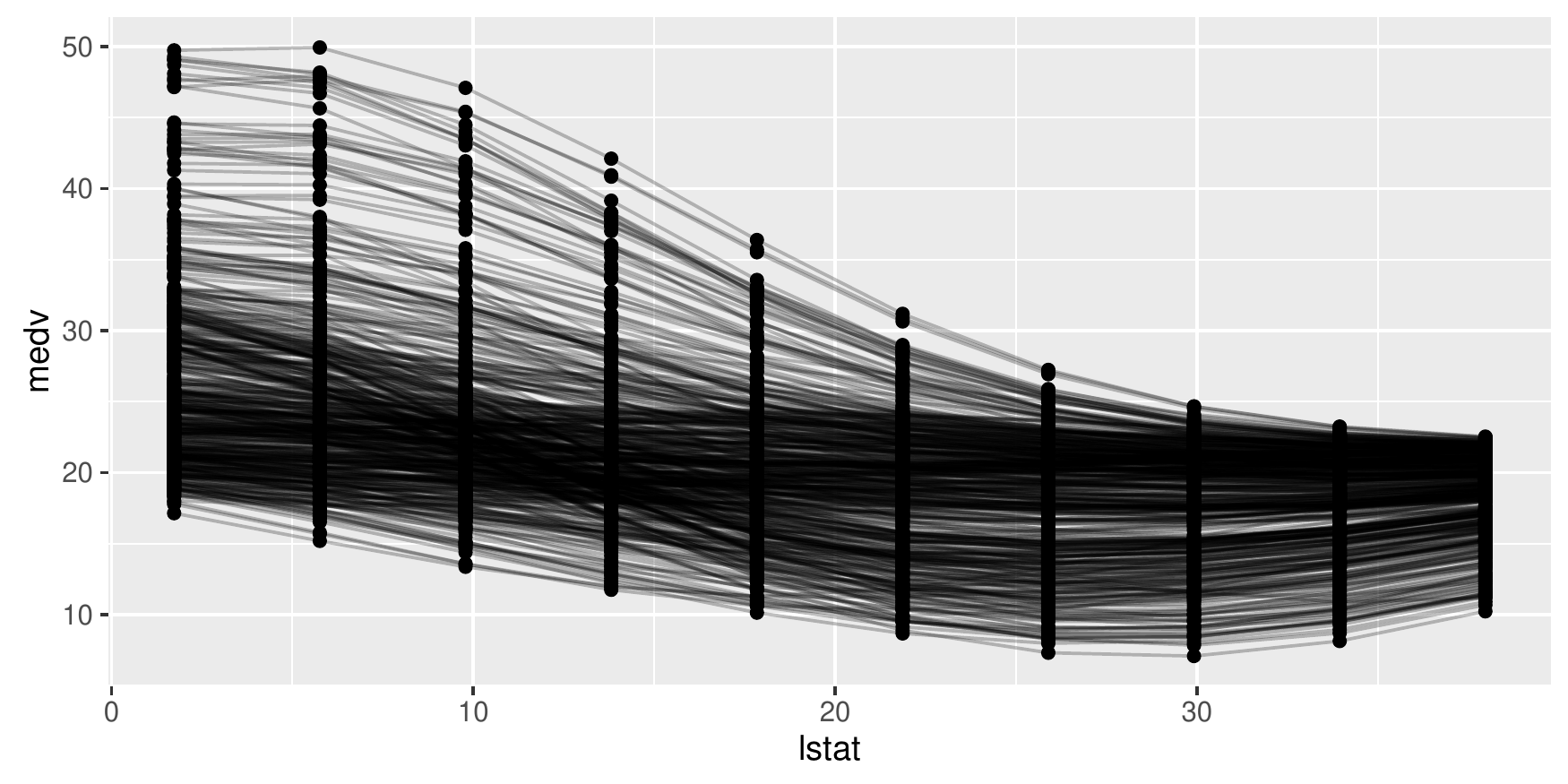}

When the individual curves are centered by subtracting the individual
conditional expectations estimated at a particular value of \(X_s\) this
results in a fixed intercept which aids in visualizing variation in
predictions made by \(\hat{f}^{(i)}_{X_s}\).

\begin{lstlisting}[language=R]
plotPartialDependence(pd.ind.classif)
\end{lstlisting}

\includegraphics{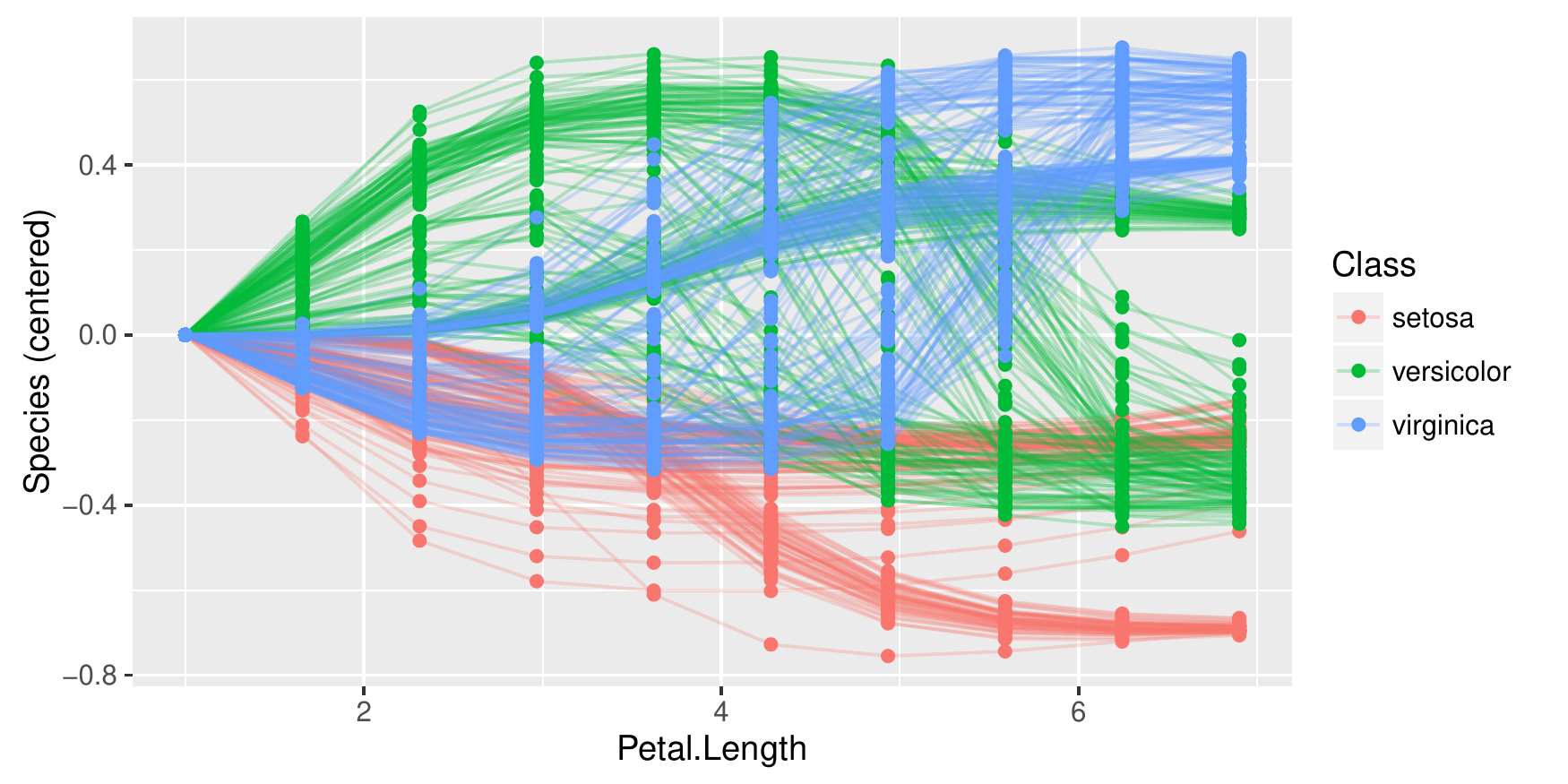}

Plotting partial derivative functions works the same as partial
dependence. Below are estimates of the derivative of the mean aggregated
partial dependence function, and the individual partial dependence
functions for a regression and a classification task respectively.

\begin{lstlisting}[language=R]
plotPartialDependence(pd.regr.der)
\end{lstlisting}

\includegraphics{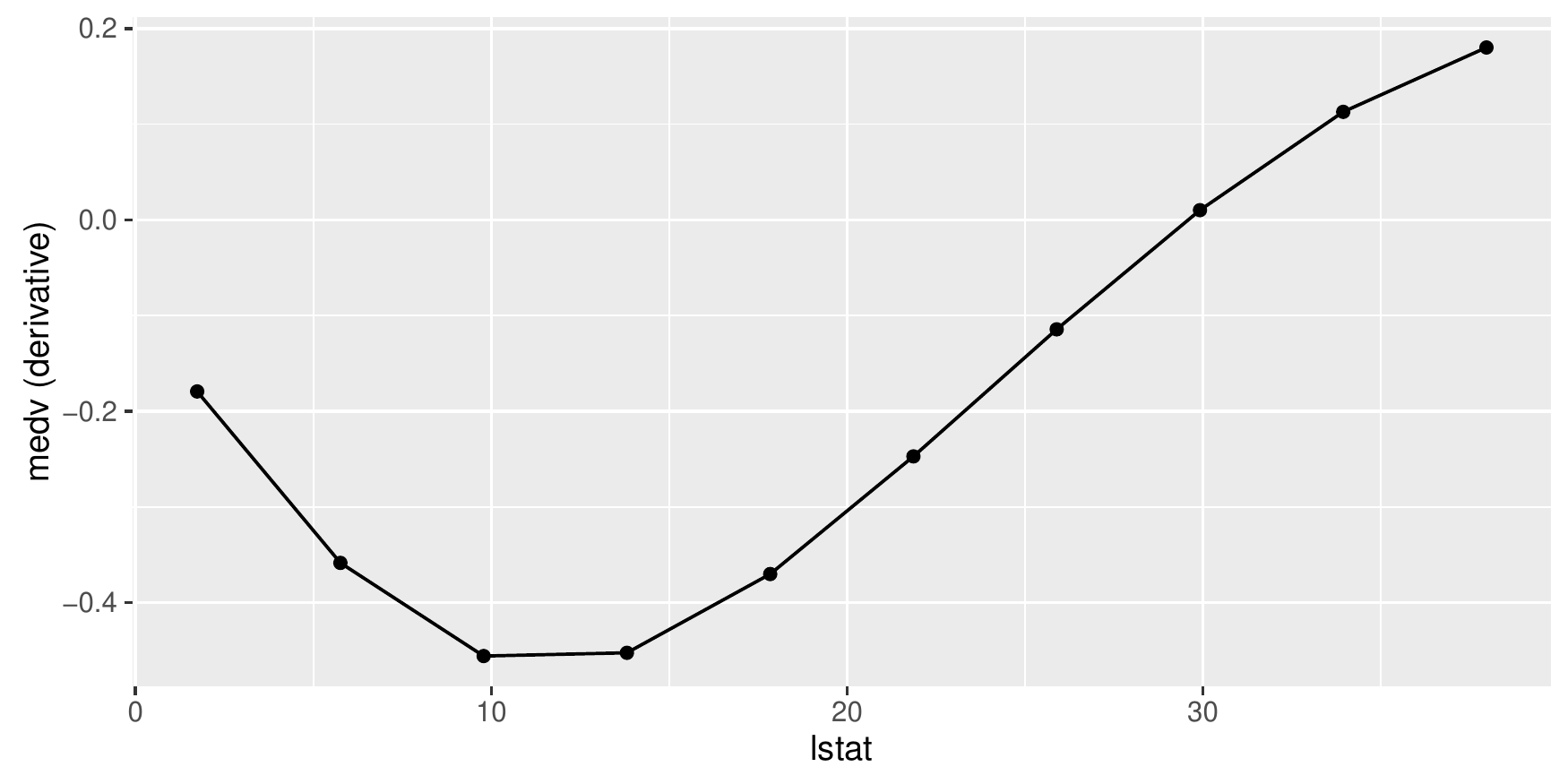}

This suggests that \(\hat{f}\) is not additive in \lstinline!lstat!
except in the neighborhood of \(25\).

\begin{lstlisting}[language=R]
plotPartialDependence(pd.regr.der.ind)
\end{lstlisting}

\includegraphics{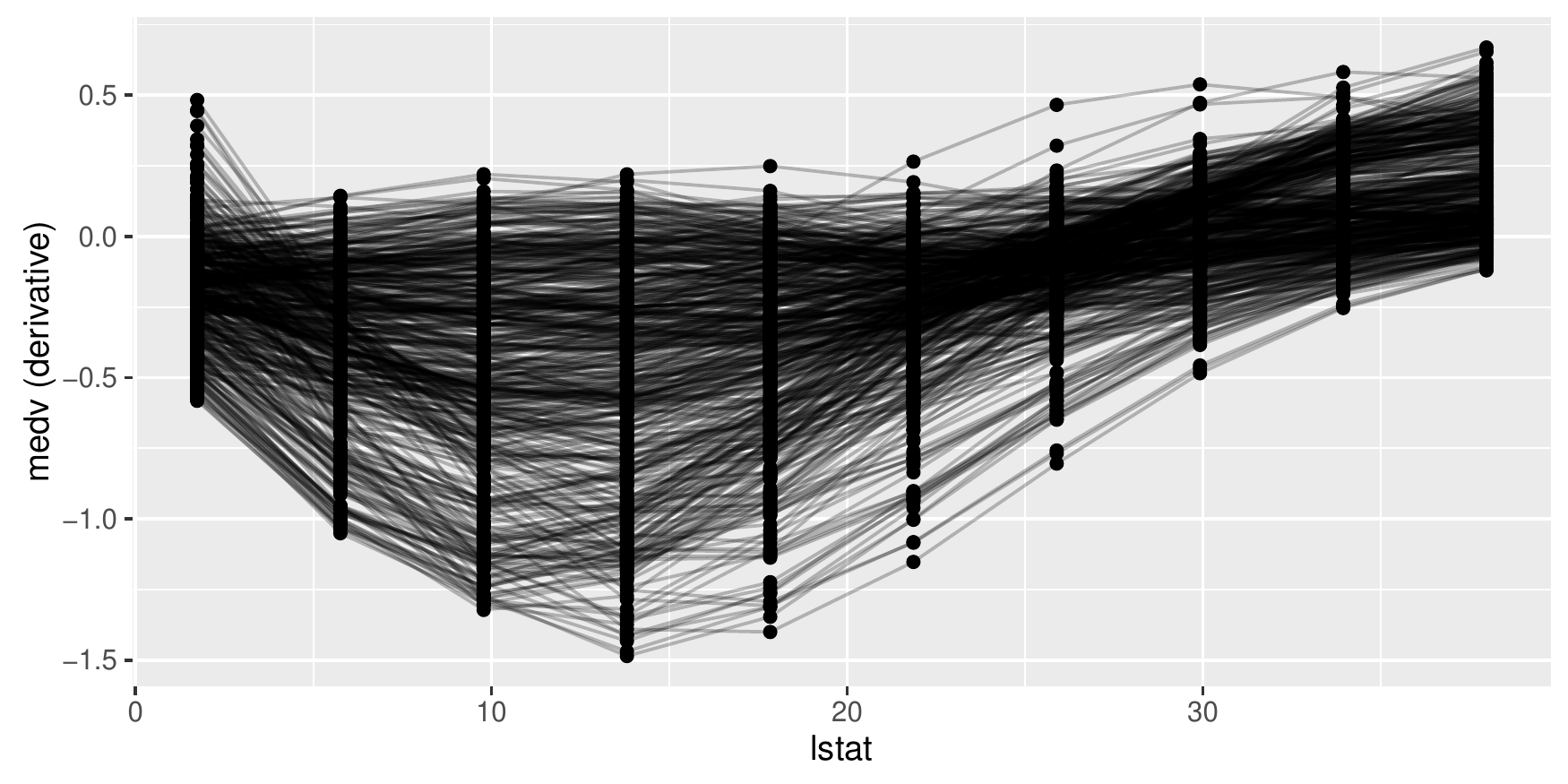}

This suggests that \lstinline!Petal.Width! interacts with some other
feature in the neighborhood of \((1.5, 2)\) for classes ``virginica''
and ``versicolor''.

\begin{lstlisting}[language=R]
plotPartialDependence(pd.classif.der.ind)
\end{lstlisting}

\includegraphics{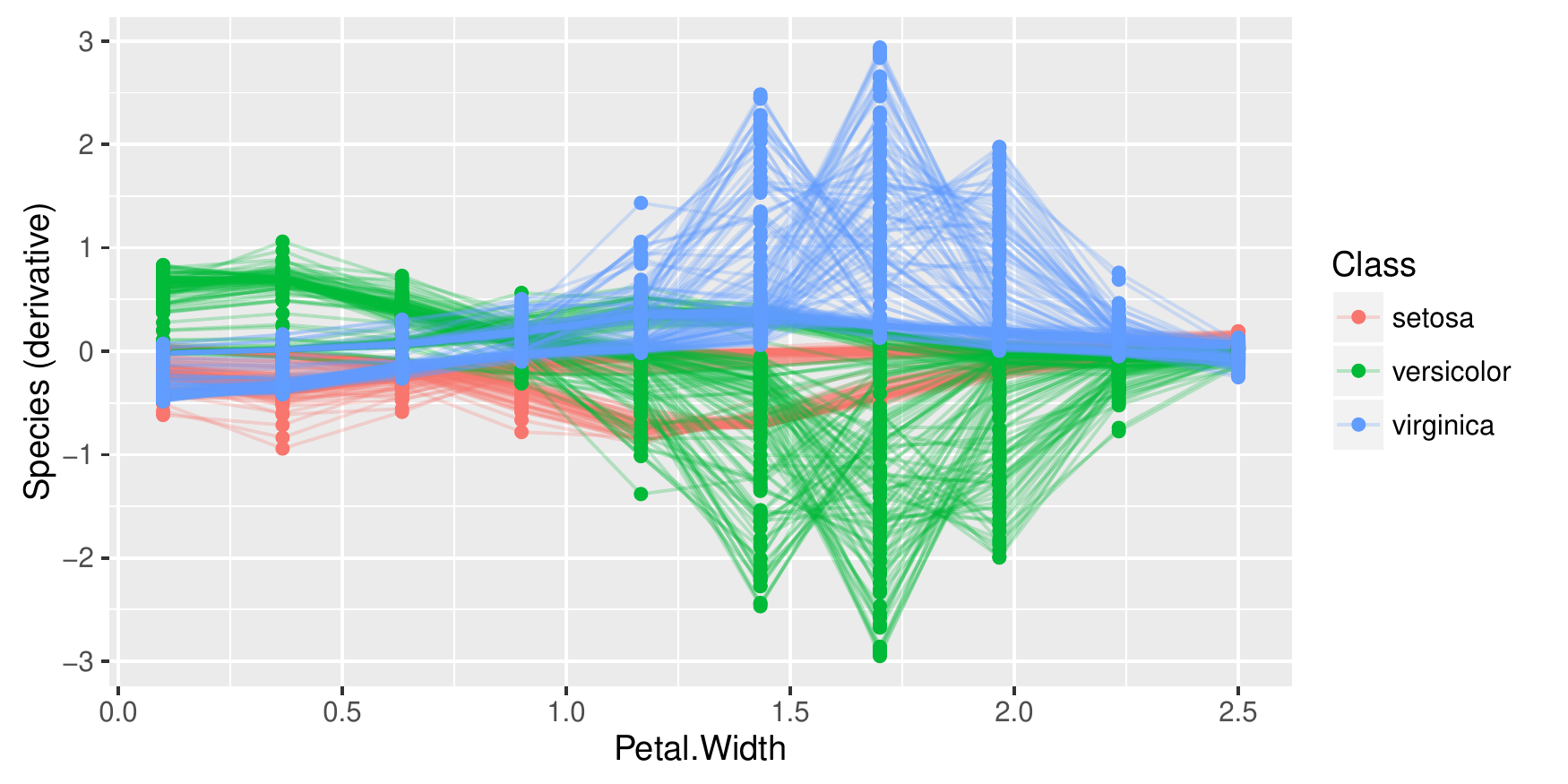}

Output from
\href{http://www.rdocumentation.org/packages/mlr/functions/generateFunctionalANOVAData.html}{generateFunctionalANOVAData}
can also be plotted using
\href{http://www.rdocumentation.org/packages/mlr/functions/plotPartialDependence.html}{plotPartialDependence}.

\begin{lstlisting}[language=R]
fa = generateFunctionalANOVAData(fit.regr, bh.task, c("crim", "lstat"), depth = 1)
plotPartialDependence(fa)
\end{lstlisting}

\includegraphics{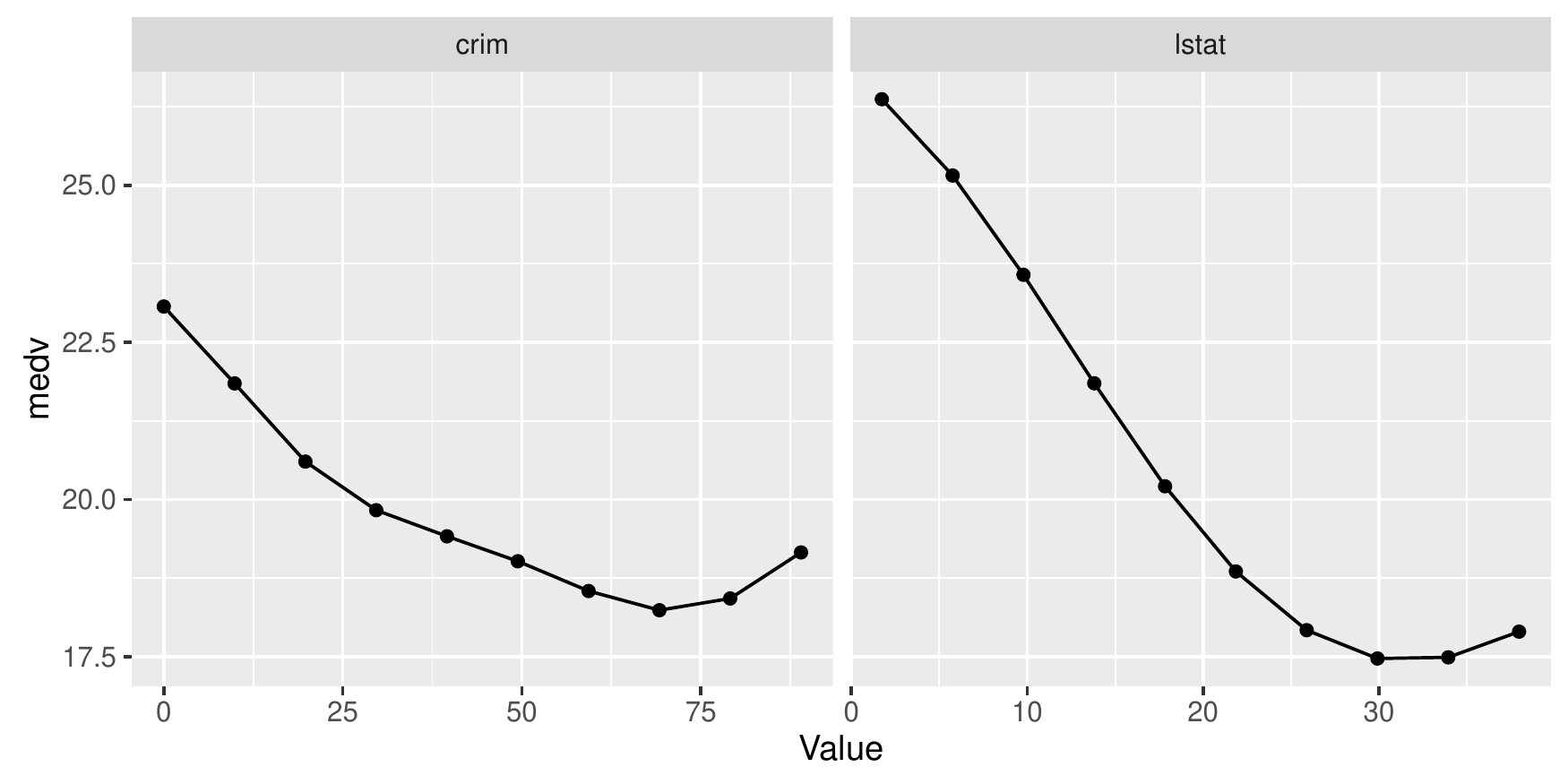}

Interactions can often be more easily visualized by using functional
ANOVA.

\begin{lstlisting}[language=R]
fa.bv = generateFunctionalANOVAData(fit.regr, bh.task, c("crim", "lstat"), depth = 2)
plotPartialDependence(fa.bv, "tile")
\end{lstlisting}

\includegraphics{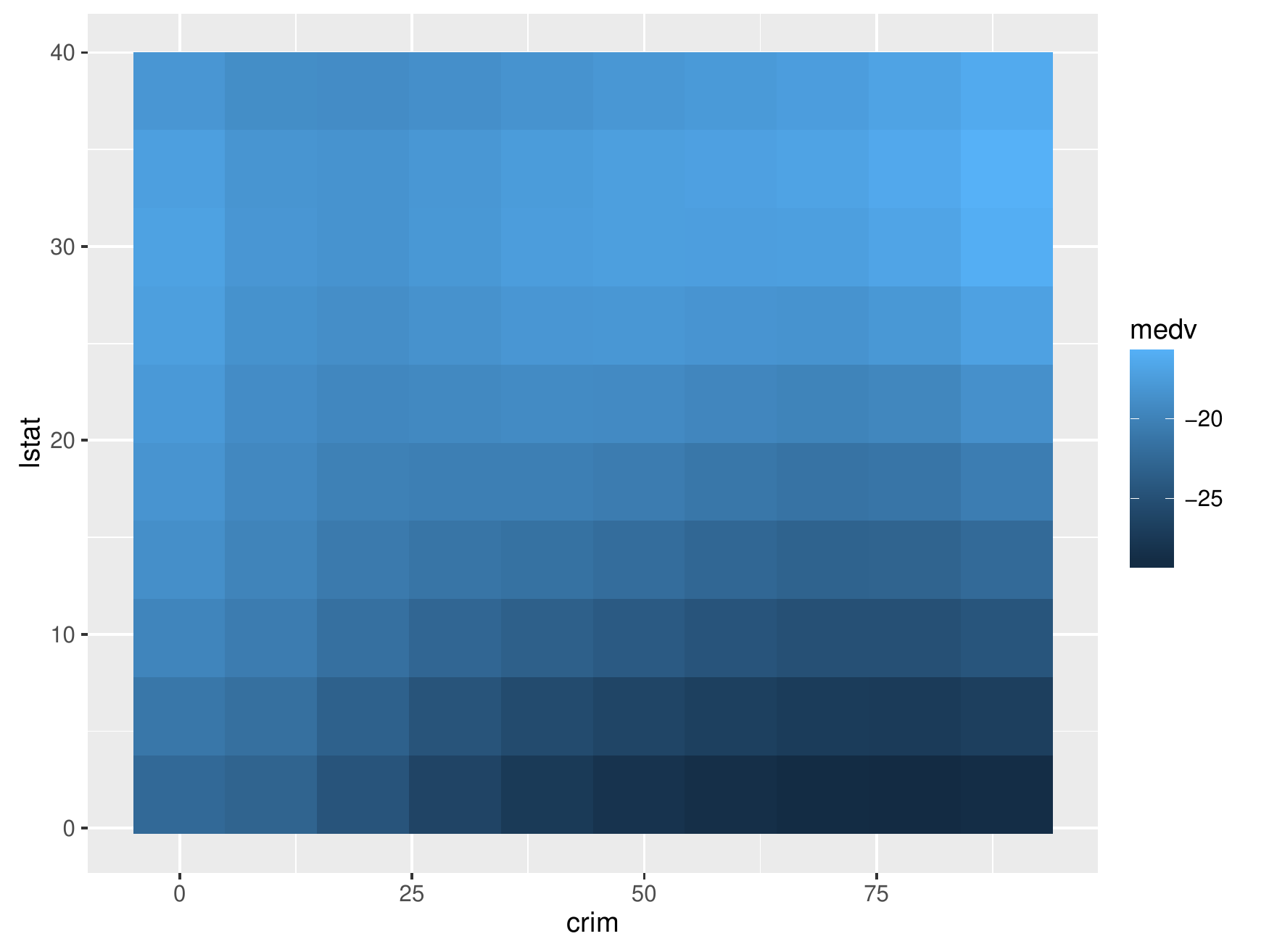}

\hypertarget{classifier-calibration}{\subsection{Classifier
Calibration}\label{classifier-calibration}}

A classifier is ``calibrated'' when the predicted probability of a class
matches the expected frequency of that class.
\href{http://www.rdocumentation.org/packages/mlr/}{mlr} can visualize
this by plotting estimated class probabilities (which are discretized)
against the observed frequency of said class in the data using
\href{http://www.rdocumentation.org/packages/mlr/functions/generateCalibrationData.html}{generateCalibrationData}
and
\href{http://www.rdocumentation.org/packages/mlr/functions/plotCalibration.html}{plotCalibration}.

\href{http://www.rdocumentation.org/packages/mlr/functions/generateCalibrationData.html}{generateCalibrationData}
takes as input
\href{http://www.rdocumentation.org/packages/mlr/functions/Prediction.html}{Prediction},
\href{http://www.rdocumentation.org/packages/mlr/functions/ResampleResult.html}{ResampleResult},
\href{http://www.rdocumentation.org/packages/mlr/functions/BenchmarkResult.html}{BenchmarkResult},
or a named list of
\href{http://www.rdocumentation.org/packages/mlr/functions/Prediction.html}{Prediction}
or
\href{http://www.rdocumentation.org/packages/mlr/functions/ResampleResult.html}{ResampleResult}
objects on a classification (multiclass or binary) task with learner(s)
that are capable of outputting probabiliites (i.e., learners must be
constructed with \lstinline!predict.type = TRUE!). The result is an
object of class
\href{http://www.rdocumentation.org/packages/mlr/functions/generateCalibrationData.html}{CalibrationData}
which has elements \lstinline!proportion!, \lstinline!data!, and
\lstinline!task!. \lstinline!proportion! gives the proportion of
observations labelled with a given class for each predicted probability
bin (e.g., for observations which are predicted to have class ``A'' with
probability \((0, 0.1]\), what is the proportion of said observations
which have class ``A''?).

\begin{lstlisting}[language=R]
lrn = makeLearner("classif.rpart", predict.type = "prob")
mod = train(lrn, task = sonar.task)
pred = predict(mod, task = sonar.task)
cal = generateCalibrationData(pred)
cal$proportion
#>      Learner       bin Class Proportion
#> 1 prediction (0.1,0.2]     M  0.1060606
#> 2 prediction (0.7,0.8]     M  0.7333333
#> 3 prediction   [0,0.1]     M  0.0000000
#> 4 prediction   (0.9,1]     M  0.9333333
#> 5 prediction (0.2,0.3]     M  0.2727273
#> 6 prediction (0.4,0.5]     M  0.4615385
#> 7 prediction (0.8,0.9]     M  0.0000000
#> 8 prediction (0.5,0.6]     M  0.0000000
\end{lstlisting}

The manner in which the predicted probabilities are discretized is
controlled by two arguments: \lstinline!breaks! and \lstinline!groups!.
By default \lstinline!breaks = "Sturges"! which uses the Sturges
algorithm in
\href{http://www.rdocumentation.org/packages/graphics/functions/hist.html}{hist}.
This argument can specify other algorithms available in
\href{http://www.rdocumentation.org/packages/graphics/functions/hist.html}{hist},
it can be a numeric vector specifying breakpoints for
\href{http://www.rdocumentation.org/packages/base/functions/cut.html}{cut},
or a single integer specifying the number of bins to create (which are
evenly spaced). Alternatively, \lstinline!groups! can be set to a
positive integer value (by default \lstinline!groups = NULL!) in which
case
\href{http://www.rdocumentation.org/packages/Hmisc/functions/cut2.html}{cut2}
is used to create bins with an approximately equal number of
observations in each bin.

\begin{lstlisting}[language=R]
cal = generateCalibrationData(pred, groups = 3)
cal$proportion
#>      Learner           bin Class Proportion
#> 1 prediction [0.000,0.267)     M 0.08860759
#> 2 prediction [0.267,0.925)     M 0.51282051
#> 3 prediction [0.925,1.000]     M 0.93333333
\end{lstlisting}

\href{http://www.rdocumentation.org/packages/mlr/functions/generateCalibrationData.html}{CalibrationData}
objects can be plotted using
\href{http://www.rdocumentation.org/packages/mlr/functions/plotCalibration.html}{plotCalibration}.
\href{http://www.rdocumentation.org/packages/mlr/functions/plotCalibration.html}{plotCalibration}
by default plots a reference line which shows perfect calibration and a
``rag'' plot, which is a rug plot on the top and bottom of the graph,
where the top pertains to ``positive'' cases, where the predicted class
matches the observed class, and the bottom pertains to ``negative''
cases, where the predicted class does not match the observed class.
Perfect classifier performance would result in all the positive cases
clustering in the top right (i.e., the correct classes are predicted
with high probability) and the negative cases clustering in the bottom
left.

\begin{lstlisting}[language=R]
plotCalibration(cal)
\end{lstlisting}

\includegraphics{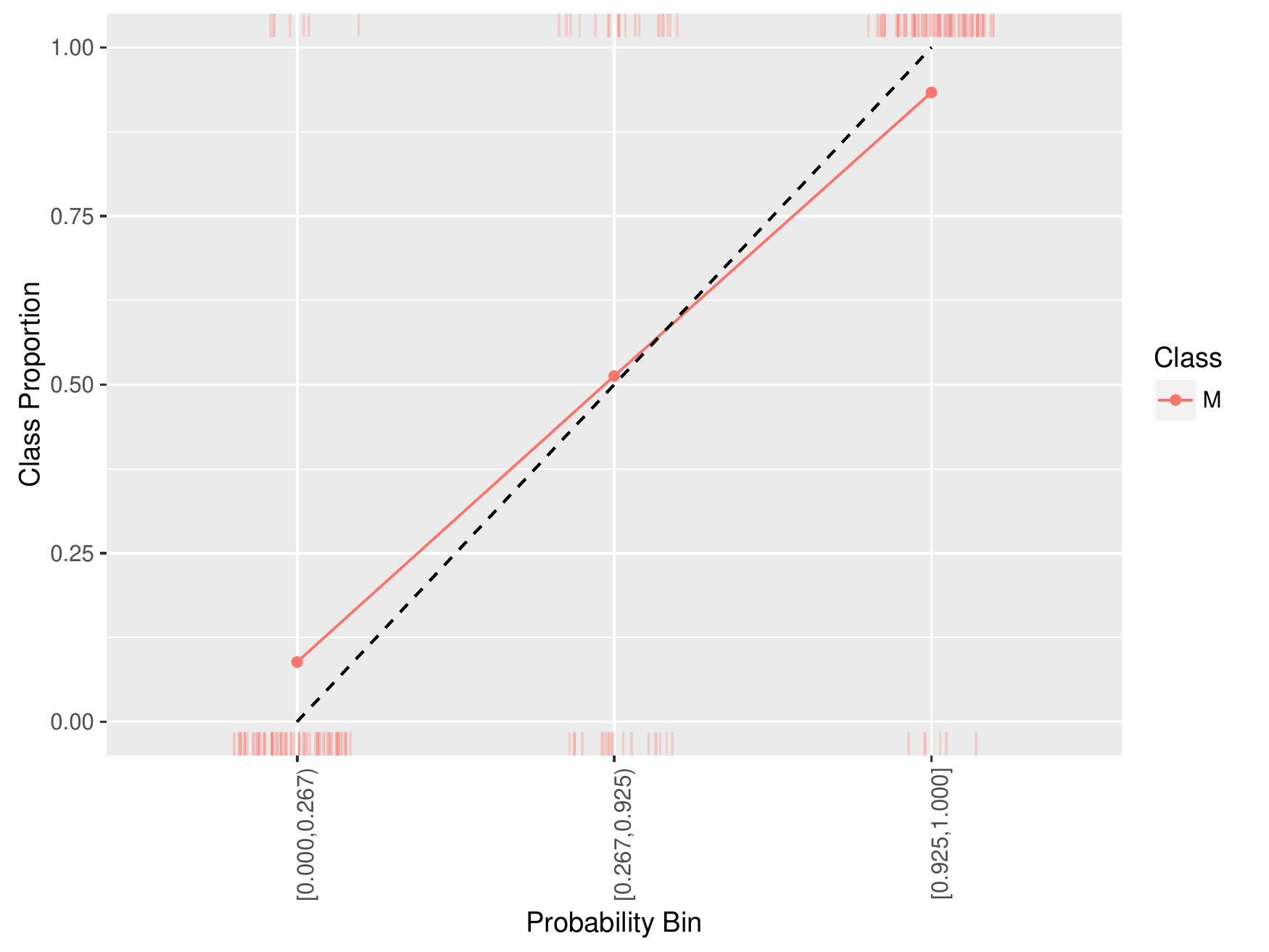}

Because of the discretization of the probabilities, sometimes it is
advantageous to smooth the calibration plot. Though
\lstinline!smooth = FALSE! by default, setting this option to
\lstinline!TRUE! replaces the estimated proportions with a loess
smoother.

\begin{lstlisting}[language=R]
cal = generateCalibrationData(pred)
plotCalibration(cal, smooth = TRUE)
\end{lstlisting}

\includegraphics{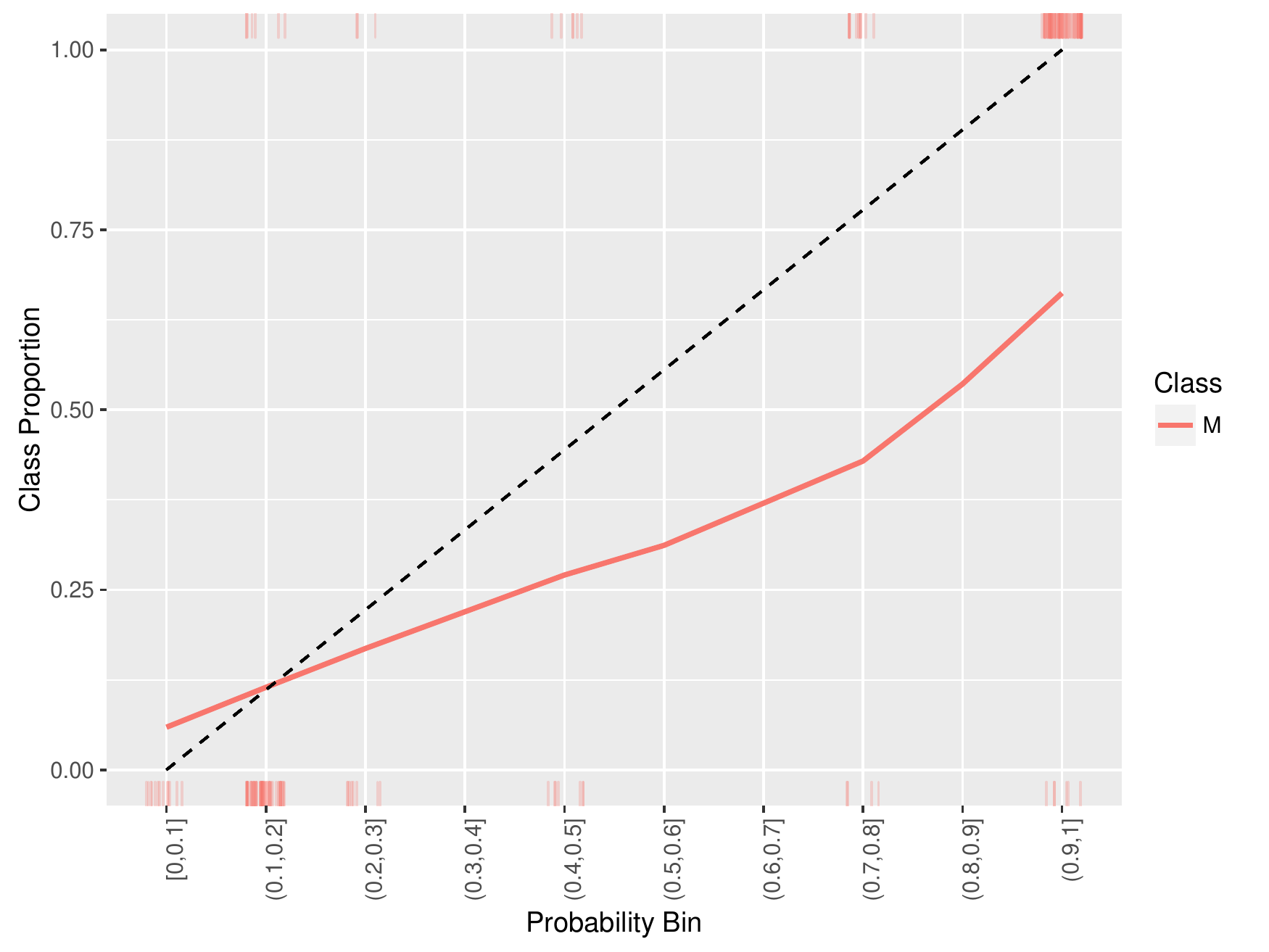}

All of the above functionality works with multi-class classification as
well.

\begin{lstlisting}[language=R]
lrns = list(
  makeLearner("classif.randomForest", predict.type = "prob"),
  makeLearner("classif.nnet", predict.type = "prob", trace = FALSE)
)
mod = lapply(lrns, train, task = iris.task)
pred = lapply(mod, predict, task = iris.task)
names(pred) = c("randomForest", "nnet")
cal = generateCalibrationData(pred, breaks = c(0, .3, .6, 1))
plotCalibration(cal)
\end{lstlisting}

\includegraphics{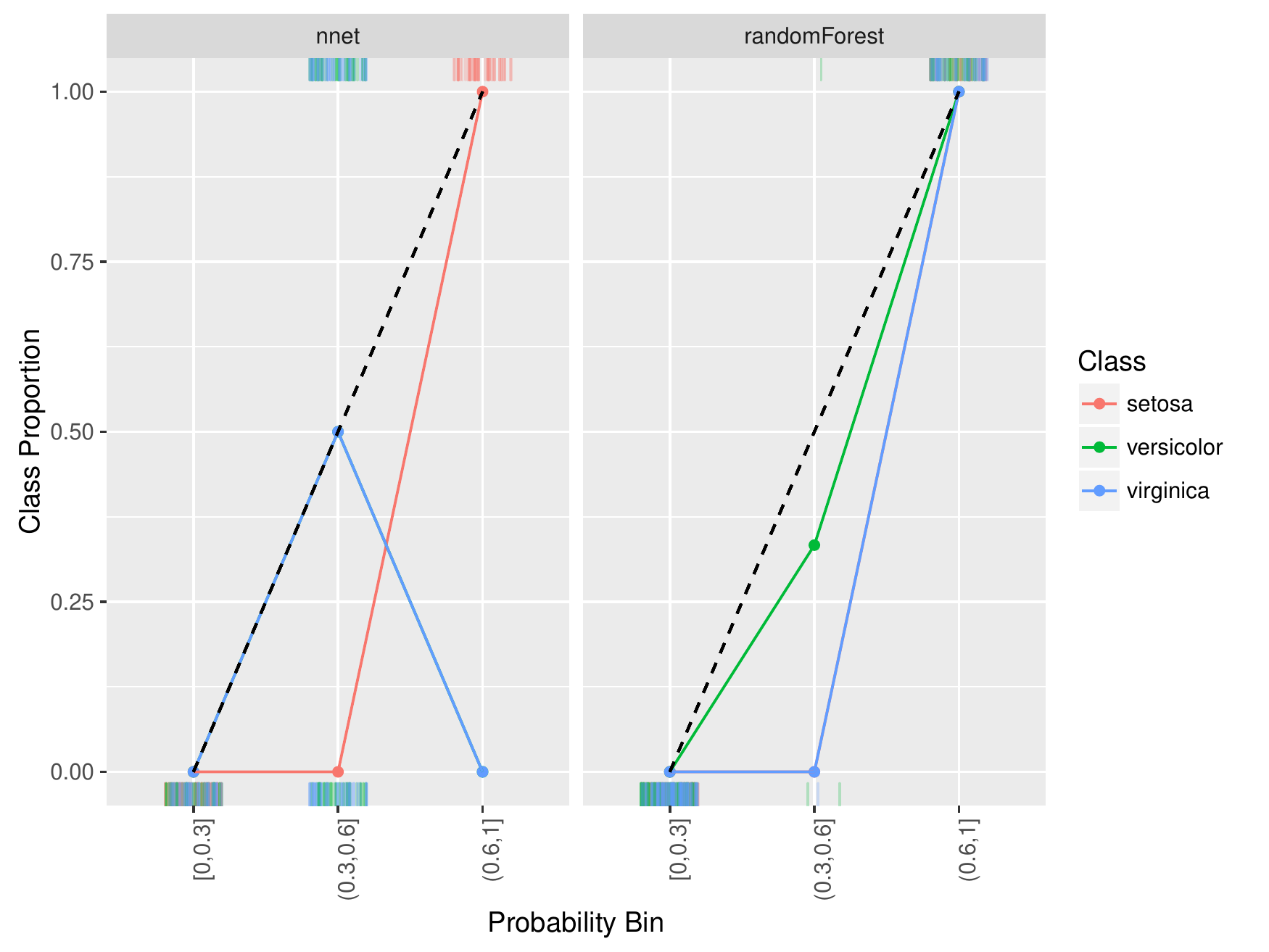}

\hypertarget{evaluating-hyperparameter-tuning}{\subsection{Evaluating
Hyperparameter Tuning}\label{evaluating-hyperparameter-tuning}}

As mentioned on the \protect\hyperlink{tuning-hyperparameters}{Tuning}
tutorial page, tuning a machine learning algorithm typically involves:

\begin{itemize}
\tightlist
\item
  the hyperparameter search space:
\end{itemize}

\begin{lstlisting}[language=R]
### ex: create a search space for the C hyperparameter from 0.01 to 0.1
ps = makeParamSet(
  makeNumericParam("C", lower = 0.01, upper = 0.1)
)
\end{lstlisting}

\begin{itemize}
\tightlist
\item
  the optimization algorithm (aka tuning method):
\end{itemize}

\begin{lstlisting}[language=R]
### ex: random search with 100 iterations
ctrl = makeTuneControlRandom(maxit = 100L)
\end{lstlisting}

\begin{itemize}
\tightlist
\item
  an evaluation method, i.e., a resampling strategy and a performance
  measure:
\end{itemize}

\begin{lstlisting}[language=R]
### ex: 2-fold CV
rdesc = makeResampleDesc("CV", iters = 2L)
\end{lstlisting}

After tuning, you may want to evaluate the tuning process in order to
answer questions such as:

\begin{itemize}
\tightlist
\item
  How does varying the value of a hyperparameter change the performance
  of the machine learning algorithm?
\item
  What's the relative importance of each hyperparameter?
\item
  How did the optimization algorithm (prematurely) converge?
\end{itemize}

\href{http://www.rdocumentation.org/packages/mlr/}{mlr} provides methods
to generate and plot the data in order to evaluate the effect of
hyperparameter tuning.

\subsubsection{Generating hyperparameter tuning
data}\label{generating-hyperparameter-tuning-data}

\href{http://www.rdocumentation.org/packages/mlr/}{mlr} separates the
generation of the data from the plotting of the data in case the user
wishes to use the data in a custom way downstream.

The
\href{http://www.rdocumentation.org/packages/mlr/functions/generateHyperParsEffectData.html}{generateHyperParsEffectData}
method takes the tuning result along with 2 additional arguments:
\lstinline!trafo! and \lstinline!include.diagnostics!. The
\lstinline!trafo! argument will convert the hyperparameter data to be on
the transformed scale in case a transformation was used when creating
the parameter (as in the case below). The
\lstinline!include.diagnostics! argument will tell
\href{http://www.rdocumentation.org/packages/mlr/}{mlr} whether to
include the eol and any error messages from the learner.

Below we perform random search on the \lstinline!C! parameter for SVM on
the famous
\href{http://www.rdocumentation.org/packages/mlbench/functions/PimaIndiansDiabetes.html}{Pima
Indians} dataset. We generate the hyperparameter effect data so that the
\lstinline!C! parameter is on the transformed scale and we do not
include diagnostic data:

\begin{lstlisting}[language=R]
ps = makeParamSet(
  makeNumericParam("C", lower = -5, upper = 5, trafo = function(x) 2^x)
)
ctrl = makeTuneControlRandom(maxit = 100L)
rdesc = makeResampleDesc("CV", iters = 2L)
res = tuneParams("classif.ksvm", task = pid.task, control = ctrl,
  measures = list(acc, mmce), resampling = rdesc, par.set = ps, show.info = FALSE)
generateHyperParsEffectData(res, trafo = T, include.diagnostics = FALSE)
#> HyperParsEffectData:
#> Hyperparameters: C
#> Measures: acc.test.mean,mmce.test.mean
#> Optimizer: TuneControlRandom
#> Nested CV Used: FALSE
#> Snapshot of data:
#>            C acc.test.mean mmce.test.mean iteration exec.time
#> 1  0.3770897     0.7695312      0.2304688         1     0.055
#> 2  3.4829323     0.7526042      0.2473958         2     0.053
#> 3  2.2050176     0.7630208      0.2369792         3     0.056
#> 4 24.9285221     0.7070312      0.2929688         4     0.060
#> 5  0.2092395     0.7539062      0.2460938         5     0.055
#> 6  0.1495099     0.7395833      0.2604167         6     0.055
\end{lstlisting}

As a reminder from the \protect\hyperlink{resampling}{resampling}
tutorial, if we wanted to generate data on the training set as well as
the validation set, we only need to make a few minor changes:

\begin{lstlisting}[language=R]
ps = makeParamSet(
  makeNumericParam("C", lower = -5, upper = 5, trafo = function(x) 2^x)
)
ctrl = makeTuneControlRandom(maxit = 100L)
rdesc = makeResampleDesc("CV", iters = 2L, predict = "both")
res = tuneParams("classif.ksvm", task = pid.task, control = ctrl,
  measures = list(acc, setAggregation(acc, train.mean), mmce, setAggregation(mmce,
    train.mean)), resampling = rdesc, par.set = ps, show.info = FALSE)
generateHyperParsEffectData(res, trafo = T, include.diagnostics = FALSE)
#> HyperParsEffectData:
#> Hyperparameters: C
#> Measures: acc.test.mean,acc.train.mean,mmce.test.mean,mmce.train.mean
#> Optimizer: TuneControlRandom
#> Nested CV Used: FALSE
#> Snapshot of data:
#>            C acc.test.mean acc.train.mean mmce.test.mean mmce.train.mean
#> 1 0.03518875     0.6510417      0.6510417      0.3489583       0.3489583
#> 2 0.17104229     0.7356771      0.7721354      0.2643229       0.2278646
#> 3 4.35326556     0.7304688      0.8828125      0.2695312       0.1171875
#> 4 0.33644238     0.7486979      0.8138021      0.2513021       0.1861979
#> 5 1.28168692     0.7500000      0.8476562      0.2500000       0.1523438
#> 6 7.36607693     0.7239583      0.8932292      0.2760417       0.1067708
#>   iteration exec.time
#> 1         1     0.074
#> 2         2     0.072
#> 3         3     0.071
#> 4         4     0.073
#> 5         5     0.072
#> 6         6     0.072
\end{lstlisting}

In the example below, we perform grid search on the \lstinline!C!
parameter for SVM on the Pima Indians dataset using nested cross
validation. We generate the hyperparameter effect data so that the
\lstinline!C! parameter is on the untransformed scale and we do not
include diagnostic data. As you can see below, nested cross validation
is supported without any extra work by the user, allowing the user to
obtain an unbiased estimator for the performance.

\begin{lstlisting}[language=R]
ps = makeParamSet(
  makeNumericParam("C", lower = -5, upper = 5, trafo = function(x) 2^x)
)
ctrl = makeTuneControlGrid()
rdesc = makeResampleDesc("CV", iters = 2L)
lrn = makeTuneWrapper("classif.ksvm", control = ctrl,
  measures = list(acc, mmce), resampling = rdesc, par.set = ps, show.info = FALSE)
res = resample(lrn, task = pid.task, resampling = cv2, extract = getTuneResult, show.info = FALSE)
generateHyperParsEffectData(res)
#> HyperParsEffectData:
#> Hyperparameters: C
#> Measures: acc.test.mean,mmce.test.mean
#> Optimizer: TuneControlGrid
#> Nested CV Used: TRUE
#> Snapshot of data:
#>            C acc.test.mean mmce.test.mean iteration exec.time
#> 1 -5.0000000     0.6640625      0.3359375         1     0.041
#> 2 -3.8888889     0.6640625      0.3359375         2     0.039
#> 3 -2.7777778     0.6822917      0.3177083         3     0.040
#> 4 -1.6666667     0.7473958      0.2526042         4     0.040
#> 5 -0.5555556     0.7708333      0.2291667         5     0.041
#> 6  0.5555556     0.7682292      0.2317708         6     0.041
#>   nested_cv_run
#> 1             1
#> 2             1
#> 3             1
#> 4             1
#> 5             1
#> 6             1
\end{lstlisting}

After generating the hyperparameter effect data, the next step is to
visualize it. \href{http://www.rdocumentation.org/packages/mlr/}{mlr}
has several methods built-in to visualize the data, meant to support the
needs of the researcher and the engineer in industry. The next few
sections will walk through the visualization support for several
use-cases.

\subsubsection{Visualizing the effect of a single
hyperparameter}\label{visualizing-the-effect-of-a-single-hyperparameter}

In a situation when the user is tuning a single hyperparameter for a
learner, the user may wish to plot the performance of the learner
against the values of the hyperparameter.

In the example below, we tune the number of clusters against the
silhouette score on the Pima dataset. We specify the x-axis with the
\lstinline!x! argument and the y-axis with the \lstinline!y! argument.
If the \lstinline!plot.type! argument is not specified,
\href{http://www.rdocumentation.org/packages/mlr/}{mlr} will attempt to
plot a scatterplot by default. Since
\href{http://www.rdocumentation.org/packages/mlr/functions/plotHyperParsEffect.html}{plotHyperParsEffect}
returns a
\href{http://www.rdocumentation.org/packages/ggplot2/functions/ggplot.html}{ggplot}
object, we can easily customize it to our liking!

\begin{lstlisting}[language=R]
ps = makeParamSet(
  makeDiscreteParam("centers", values = 3:10)
)
ctrl = makeTuneControlGrid()
rdesc = makeResampleDesc("Holdout")
res = tuneParams("cluster.kmeans", task = mtcars.task, control = ctrl,
  measures = silhouette, resampling = rdesc, par.set = ps, show.info = FALSE)
#> 
#> This is package 'modeest' written by P. PONCET.
#> For a complete list of functions, use 'library(help = "modeest")' or 'help.start()'.
data = generateHyperParsEffectData(res)
plt = plotHyperParsEffect(data, x = "centers", y = "silhouette.test.mean")
### add our own touches to the plot
plt + geom_point(colour = "red") +
  ggtitle("Evaluating Number of Cluster Centers on mtcars") +
  scale_x_continuous(breaks = 3:10) +
  theme_bw()
\end{lstlisting}

\includegraphics{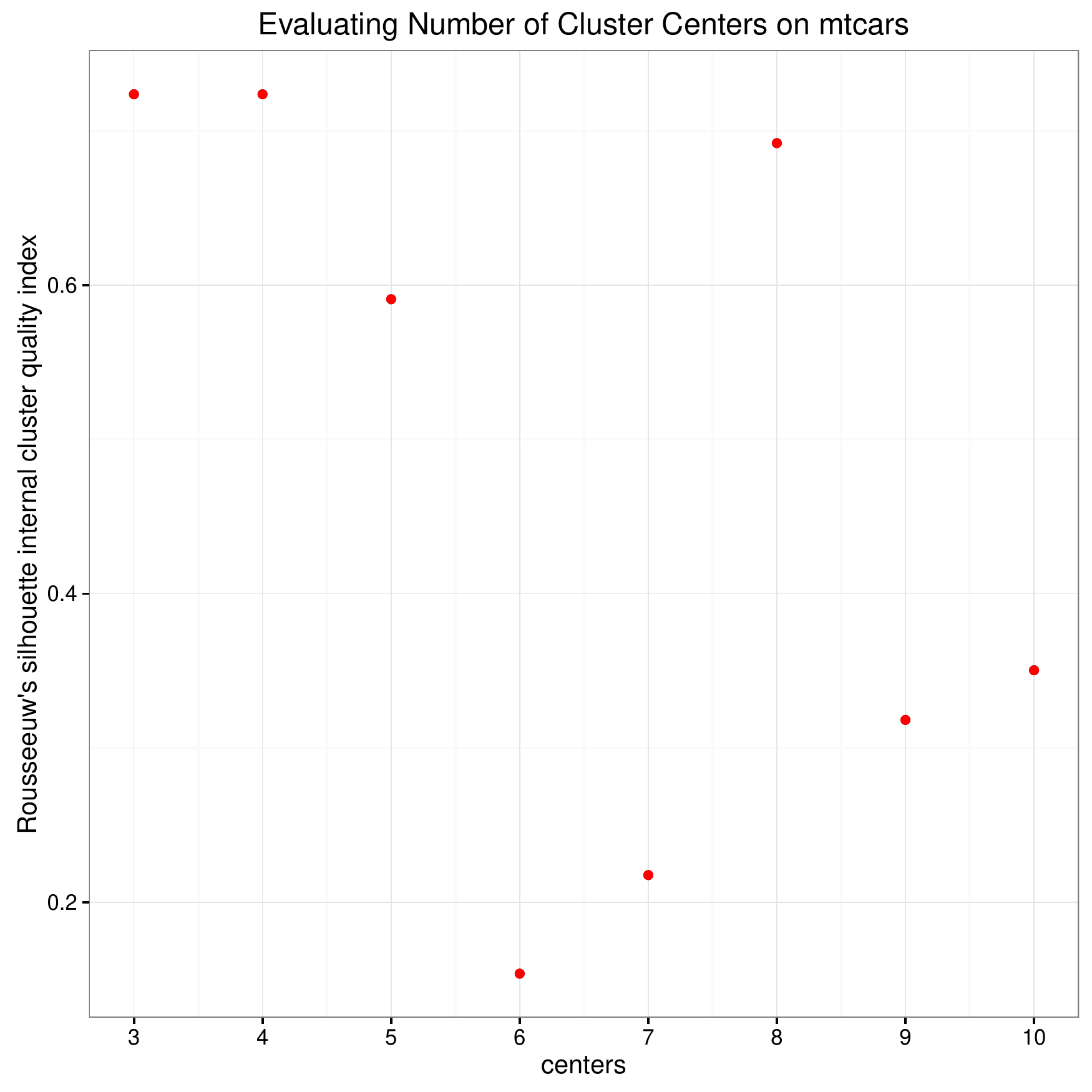}

In the example below, we tune SVM with the \lstinline!C! hyperparameter
on the Pima dataset. We will use simulated annealing optimizer, so we
are interested in seeing if the optimization algorithm actually improves
with iterations. By default,
\href{http://www.rdocumentation.org/packages/mlr/}{mlr} only plots
improvements to the global optimum.

\begin{lstlisting}[language=R]
ps = makeParamSet(
  makeNumericParam("C", lower = -5, upper = 5, trafo = function(x) 2^x)
)
ctrl = makeTuneControlGenSA(budget = 100L)
rdesc = makeResampleDesc("Holdout")
res = tuneParams("classif.ksvm", task = pid.task, control = ctrl,
  resampling = rdesc, par.set = ps, show.info = FALSE)
data = generateHyperParsEffectData(res)
plt = plotHyperParsEffect(data, x = "iteration", y = "mmce.test.mean",
  plot.type = "line")
plt + ggtitle("Analyzing convergence of simulated annealing") +
  theme_minimal()
\end{lstlisting}

\includegraphics{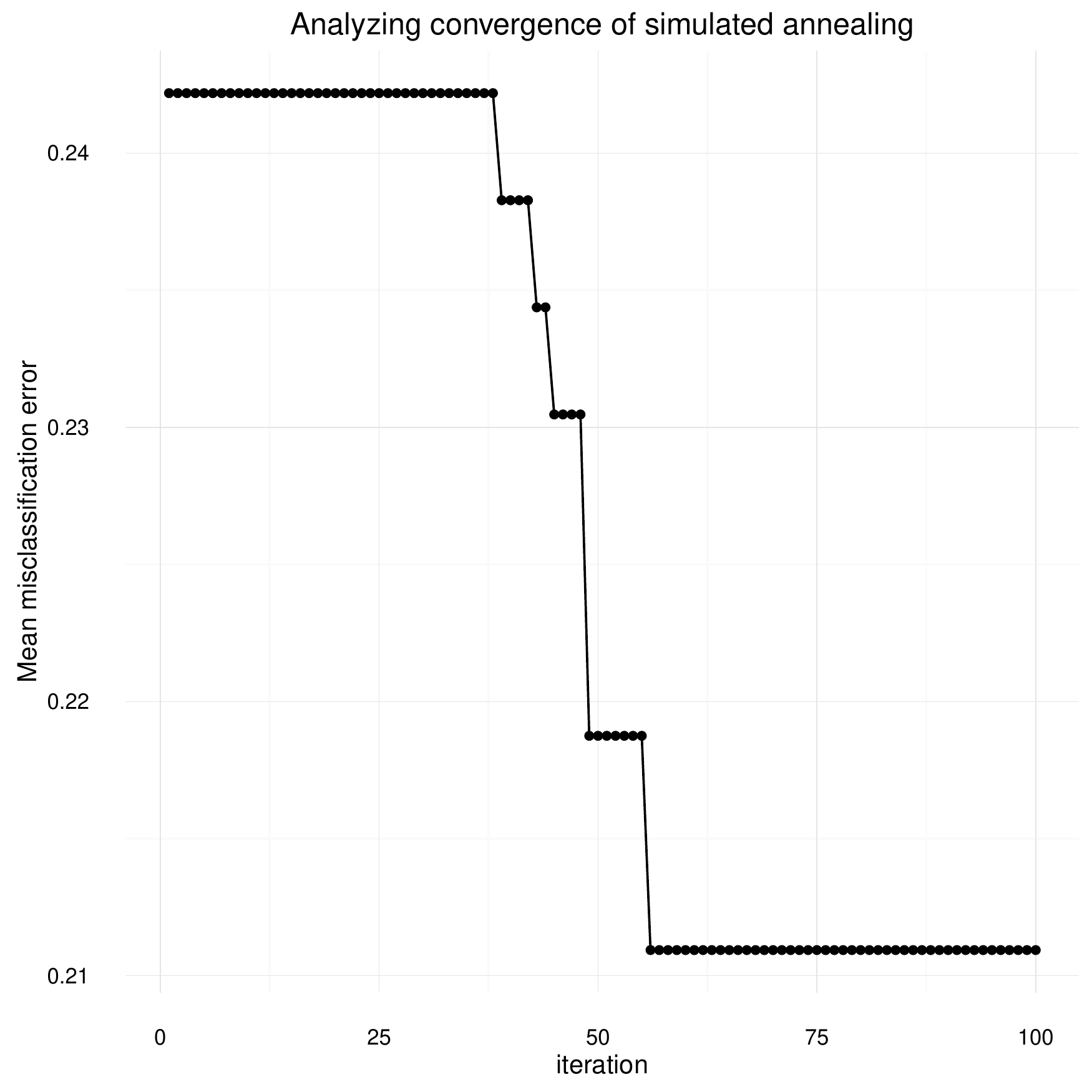}

In the case of a learner crash,
\href{http://www.rdocumentation.org/packages/mlr/}{mlr} will impute the
crash with the worst value graphically and indicate the point. In the
example below, we give the \lstinline!C! parameter negative values,
which will result in a learner crash for SVM.

\begin{lstlisting}[language=R]
ps = makeParamSet(
  makeDiscreteParam("C", values = c(-1, -0.5, 0.5, 1, 1.5))
)
ctrl = makeTuneControlGrid()
rdesc = makeResampleDesc("CV", iters = 2L)
res = tuneParams("classif.ksvm", task = pid.task, control = ctrl,
  measures = list(acc, mmce), resampling = rdesc, par.set = ps, show.info = FALSE)
data = generateHyperParsEffectData(res)
plt = plotHyperParsEffect(data, x = "C", y = "acc.test.mean")
plt + ggtitle("SVM learner crashes with negative C") +
  theme_bw()
\end{lstlisting}

\includegraphics{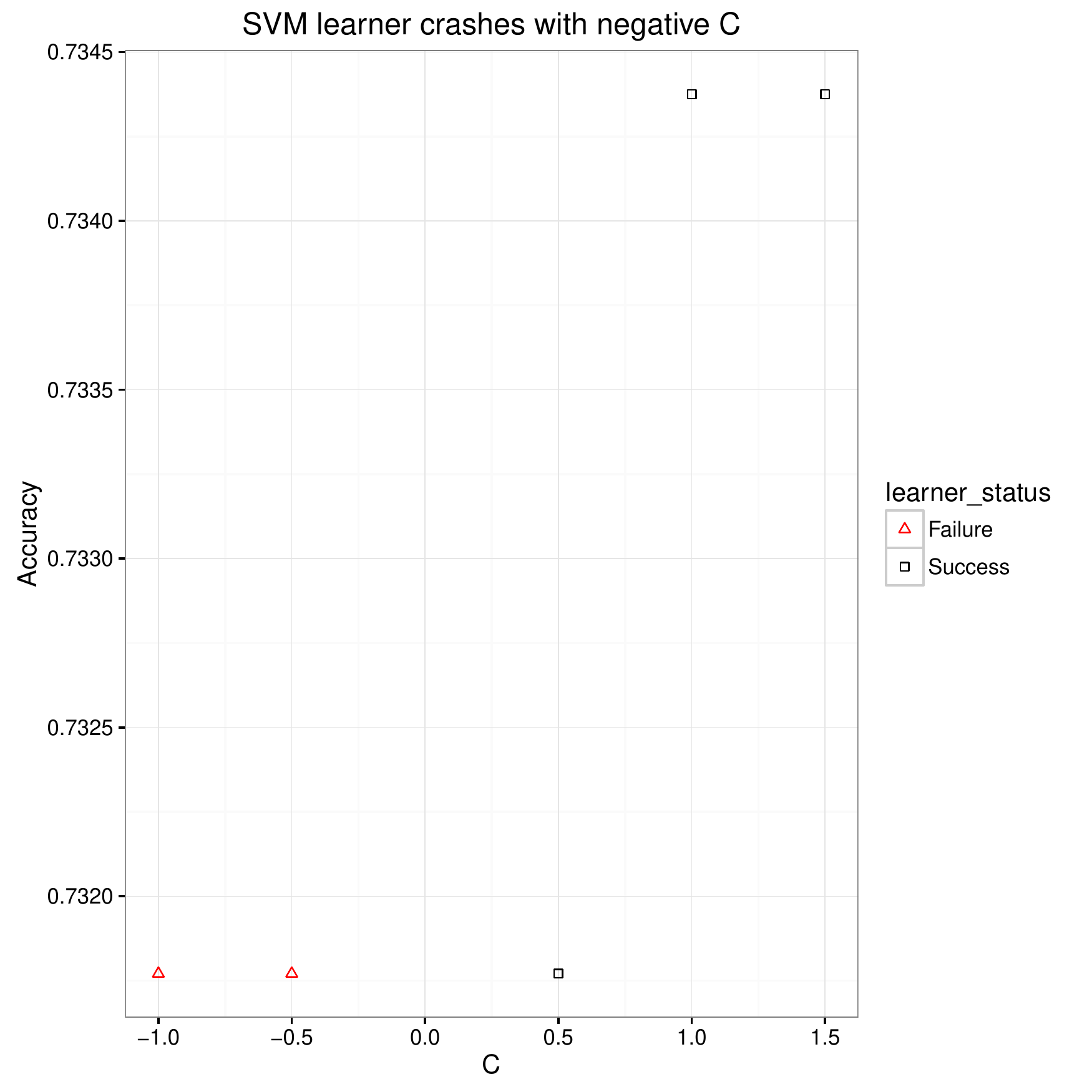}

The example below uses \protect\hyperlink{nested-resampling}{nested
cross validation} with an outer loop of 2 runs.
\href{http://www.rdocumentation.org/packages/mlr/}{mlr} indicates each
run within the visualization.

\begin{lstlisting}[language=R]
ps = makeParamSet(
  makeNumericParam("C", lower = -5, upper = 5, trafo = function(x) 2^x)
)
ctrl = makeTuneControlGrid()
rdesc = makeResampleDesc("Holdout")
lrn = makeTuneWrapper("classif.ksvm", control = ctrl,
  measures = list(acc, mmce), resampling = rdesc, par.set = ps, show.info = FALSE)
res = resample(lrn, task = pid.task, resampling = cv2, extract = getTuneResult, show.info = FALSE)
data = generateHyperParsEffectData(res)
plotHyperParsEffect(data, x = "C", y = "acc.test.mean", plot.type = "line")
\end{lstlisting}

\includegraphics{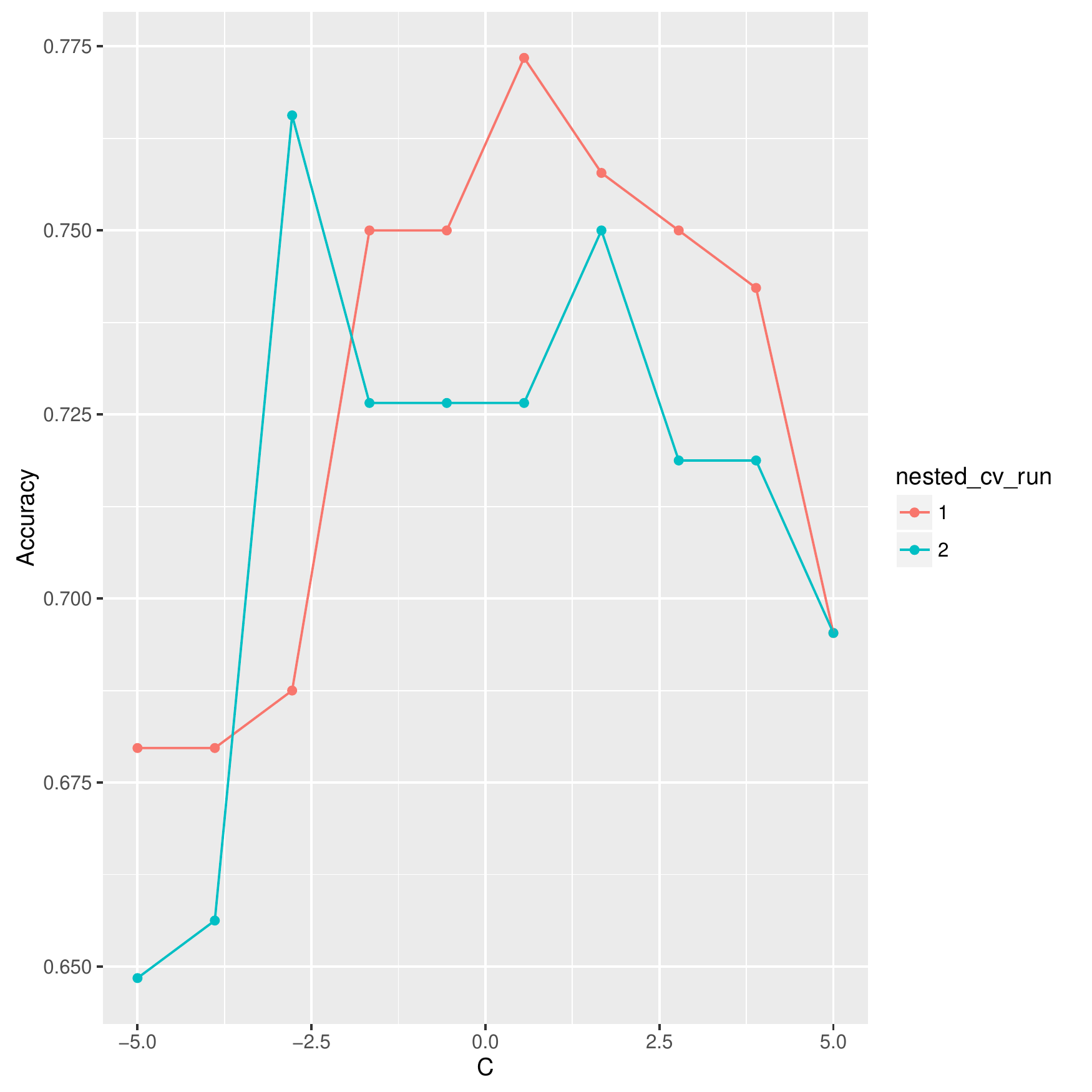}

\subsubsection{Visualizing the effect of 2
hyperparameters}\label{visualizing-the-effect-of-2-hyperparameters}

In the case of tuning 2 hyperparameters simultaneously,
\href{http://www.rdocumentation.org/packages/mlr/}{mlr} provides the
ability to plot a heatmap and contour plot in addition to a scatterplot
or line.

In the example below, we tune the \lstinline!C! and \lstinline!sigma!
parameters for SVM on the Pima dataset. We use interpolation to produce
a regular grid for plotting the heatmap. The \lstinline!interpolation!
argument accepts any regression learner from
\href{http://www.rdocumentation.org/packages/mlr/}{mlr} to perform the
interpolation. The \lstinline!z! argument will be used to fill the
heatmap or color lines, depending on the \lstinline!plot.type! used.

\begin{lstlisting}[language=R]
ps = makeParamSet(
  makeNumericParam("C", lower = -5, upper = 5, trafo = function(x) 2^x),
  makeNumericParam("sigma", lower = -5, upper = 5, trafo = function(x) 2^x))
ctrl = makeTuneControlRandom(maxit = 100L)
rdesc = makeResampleDesc("Holdout")
learn = makeLearner("classif.ksvm", par.vals = list(kernel = "rbfdot"))
res = tuneParams(learn, task = pid.task, control = ctrl, measures = acc,
  resampling = rdesc, par.set = ps, show.info = FALSE)
data = generateHyperParsEffectData(res)
plt = plotHyperParsEffect(data, x = "C", y = "sigma", z = "acc.test.mean",
  plot.type = "heatmap", interpolate = "regr.earth")
min_plt = min(data$data$acc.test.mean, na.rm = TRUE)
max_plt = max(data$data$acc.test.mean, na.rm = TRUE)
med_plt = mean(c(min_plt, max_plt))
plt + scale_fill_gradient2(breaks = seq(min_plt, max_plt, length.out = 5),
  low = "blue", mid = "white", high = "red", midpoint = med_plt)
\end{lstlisting}

\includegraphics{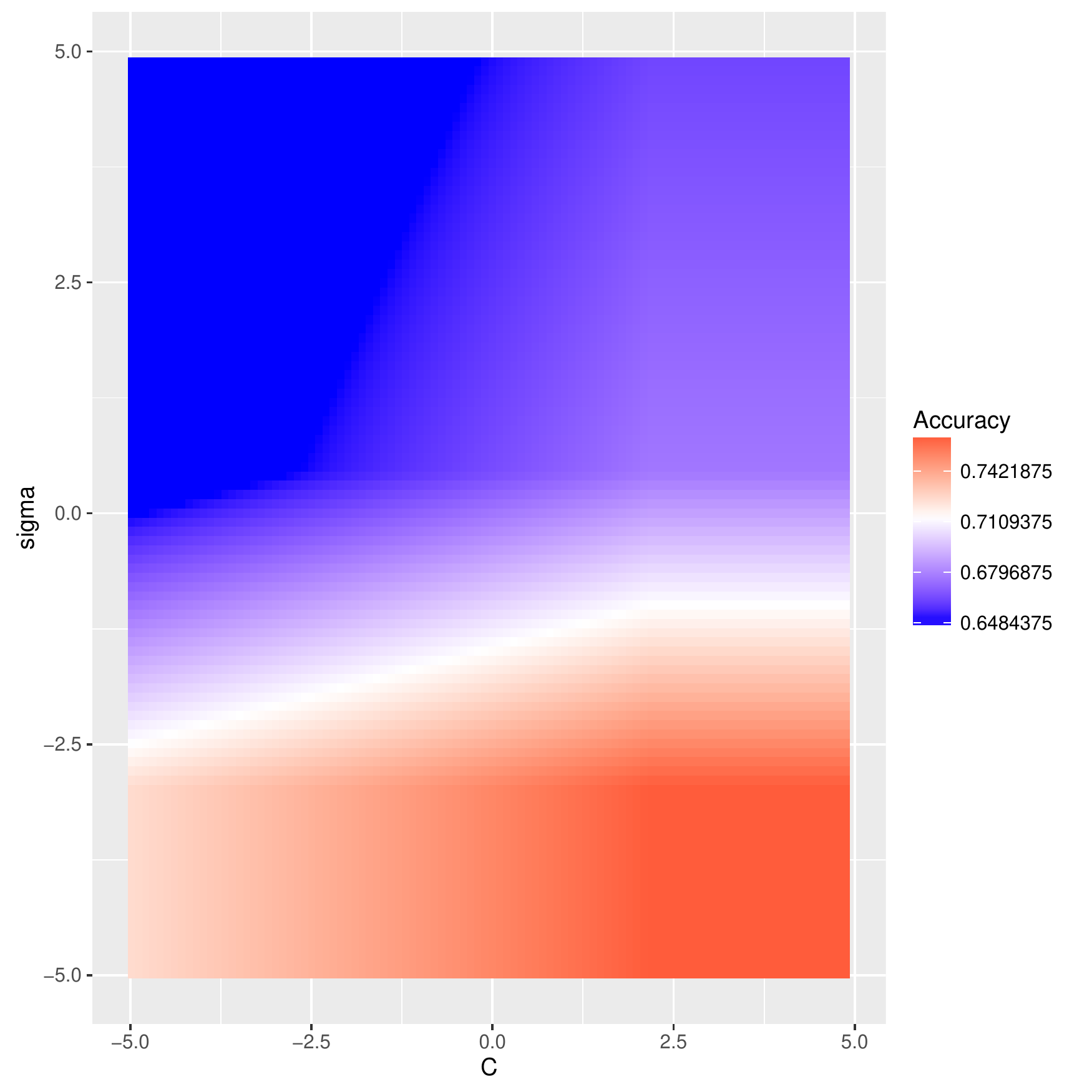}

We can use the \lstinline!show.experiments! argument in order to
visualize which points were specifically passed to the learner in the
original experiment and which points were interpolated by
\href{http://www.rdocumentation.org/packages/mlr/}{mlr}:

\begin{lstlisting}[language=R]
plt = plotHyperParsEffect(data, x = "C", y = "sigma", z = "acc.test.mean",
  plot.type = "heatmap", interpolate = "regr.earth", show.experiments = TRUE)
plt + scale_fill_gradient2(breaks = seq(min_plt, max_plt, length.out = 5),
  low = "blue", mid = "white", high = "red", midpoint = med_plt)
\end{lstlisting}

\includegraphics{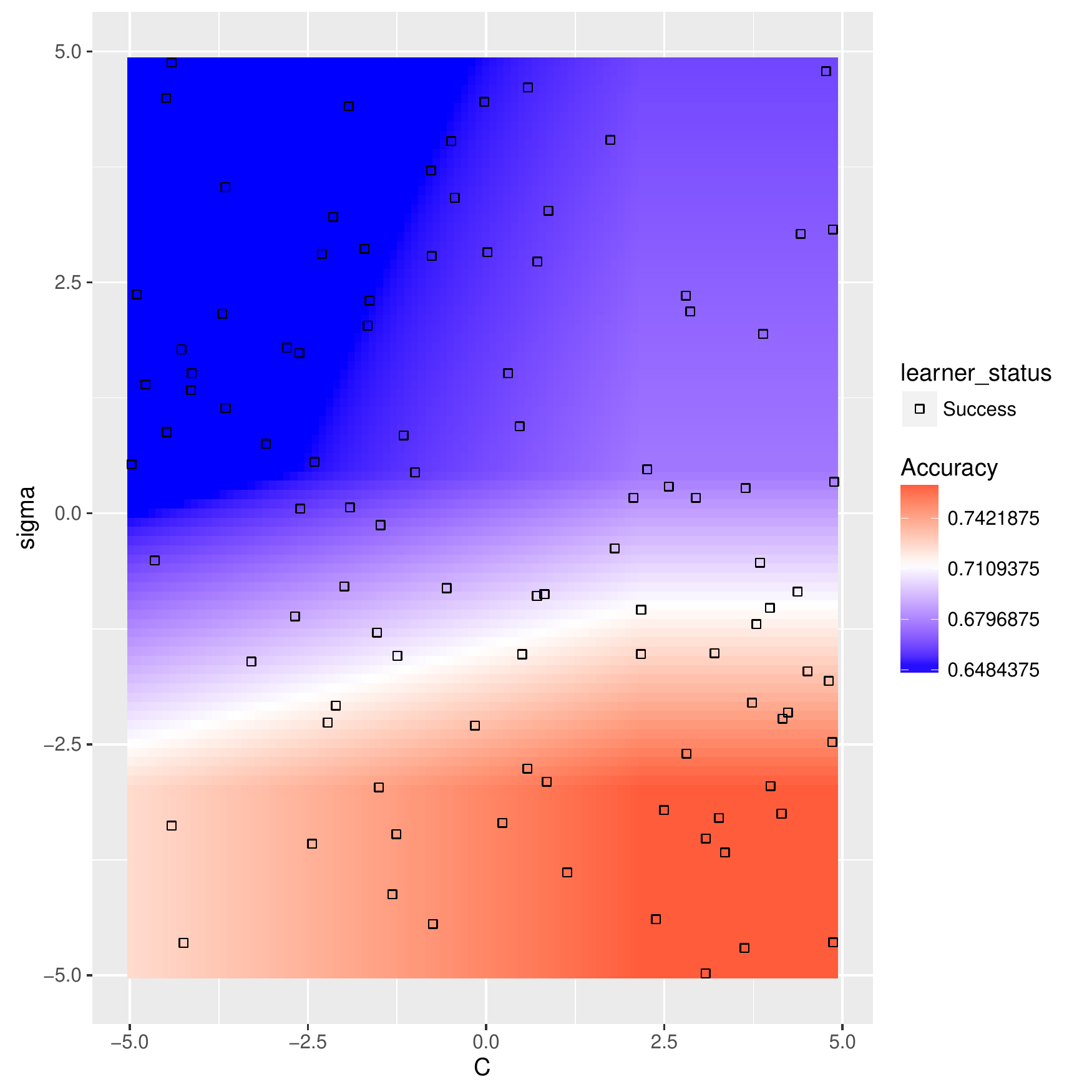}

We can also visualize how long the optimizer takes to reach an optima
for the same example:

\begin{lstlisting}[language=R]
plotHyperParsEffect(data, x = "iteration", y = "acc.test.mean",
  plot.type = "line")
\end{lstlisting}

\includegraphics{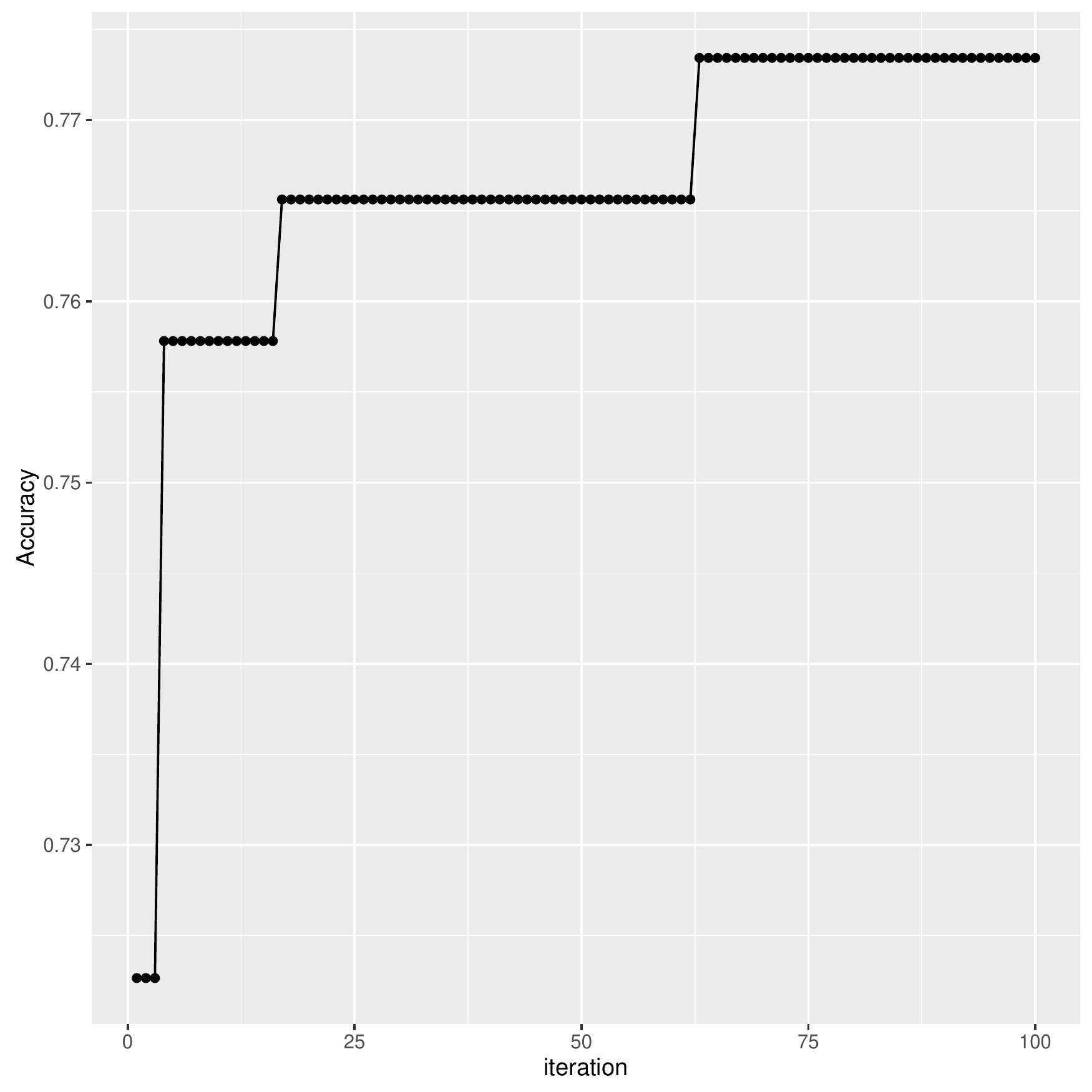}

In the case where we are tuning 2 hyperparameters and we have a learner
crash, \href{http://www.rdocumentation.org/packages/mlr/}{mlr} will
indicate the respective points and impute them with the worst value. In
the example below, we tune \lstinline!C! and \lstinline!sigma!, forcing
\lstinline!C! to be negative for some instances which will crash SVM. We
perform interpolation to get a regular grid in order to plot a heatmap.
We can see that the interpolation creates axis parallel lines resulting
from the learner crashes.

\begin{lstlisting}[language=R]
ps = makeParamSet(
  makeDiscreteParam("C", values = c(-1, 0.5, 1.5, 1, 0.2, 0.3, 0.4, 5)),
  makeDiscreteParam("sigma", values = c(-1, 0.5, 1.5, 1, 0.2, 0.3, 0.4, 5)))
ctrl = makeTuneControlGrid()
rdesc = makeResampleDesc("Holdout")
learn = makeLearner("classif.ksvm", par.vals = list(kernel = "rbfdot"))
res = tuneParams(learn, task = pid.task, control = ctrl, measures = acc,
  resampling = rdesc, par.set = ps, show.info = FALSE)
data = generateHyperParsEffectData(res)
plotHyperParsEffect(data, x = "C", y = "sigma", z = "acc.test.mean",
  plot.type = "heatmap", interpolate = "regr.earth")
\end{lstlisting}

\includegraphics{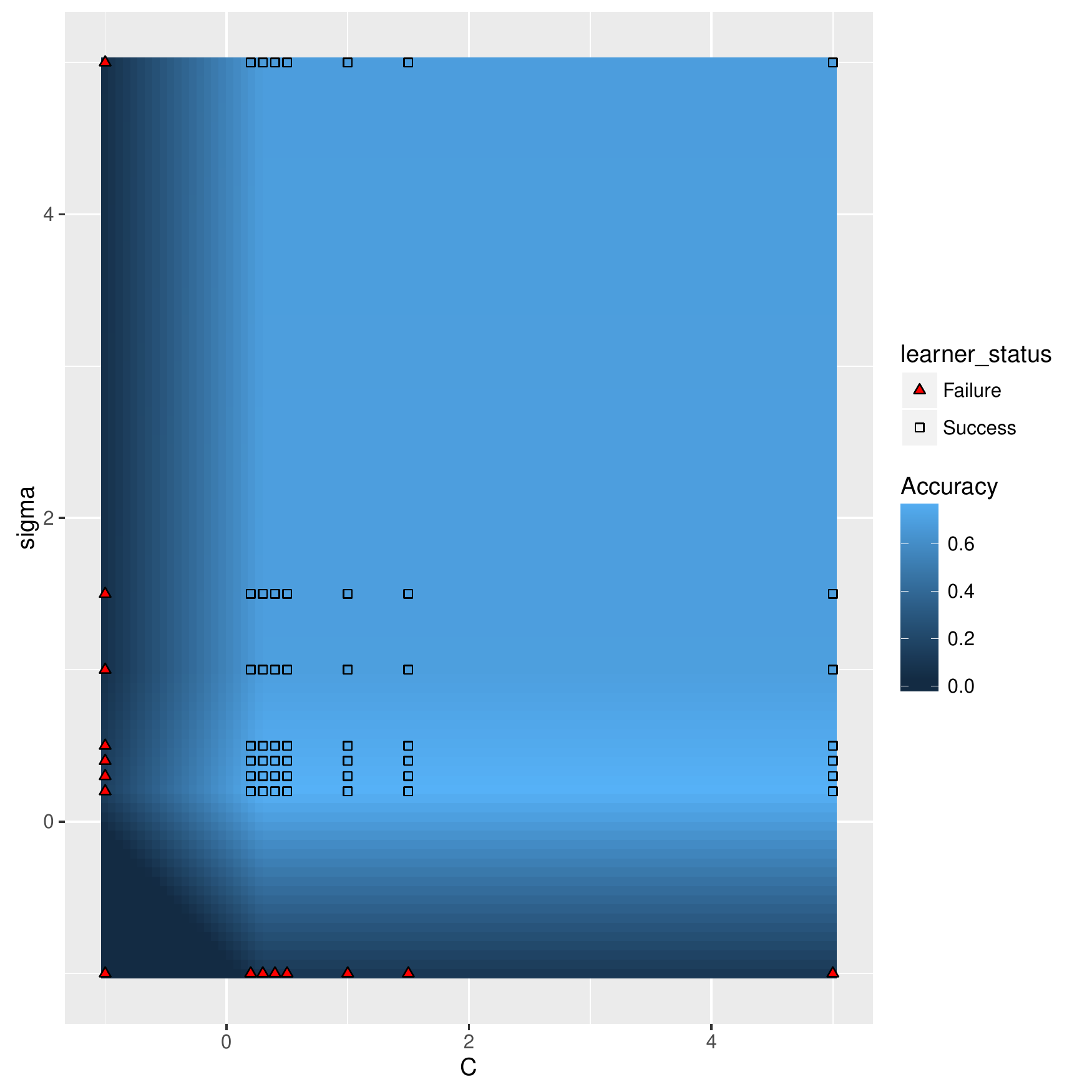}

A slightly more complicated example is using nested cross validation
while simultaneously tuning 2 hyperparameters. In order to plot a
heatmap in this case,
\href{http://www.rdocumentation.org/packages/mlr/}{mlr} will aggregate
each of the nested runs by a user-specified function. The default
function is \lstinline!mean!. As expected, we can still take advantage
of interpolation.

\begin{lstlisting}[language=R]
ps = makeParamSet(
  makeNumericParam("C", lower = -5, upper = 5, trafo = function(x) 2^x),
  makeNumericParam("sigma", lower = -5, upper = 5, trafo = function(x) 2^x))
ctrl = makeTuneControlRandom(maxit = 100)
rdesc = makeResampleDesc("Holdout")
learn = makeLearner("classif.ksvm", par.vals = list(kernel = "rbfdot"))
lrn = makeTuneWrapper(learn, control = ctrl, measures = list(acc, mmce),
  resampling = rdesc, par.set = ps, show.info = FALSE)
res = resample(lrn, task = pid.task, resampling = cv2, extract = getTuneResult, show.info = FALSE)
data = generateHyperParsEffectData(res)
plt = plotHyperParsEffect(data, x = "C", y = "sigma", z = "acc.test.mean",
  plot.type = "heatmap", interpolate = "regr.earth", show.experiments = TRUE,
  nested.agg = mean)
min_plt = min(plt$data$acc.test.mean, na.rm = TRUE)
max_plt = max(plt$data$acc.test.mean, na.rm = TRUE)
med_plt = mean(c(min_plt, max_plt))
plt + scale_fill_gradient2(breaks = seq(min_plt, max_plt, length.out = 5),
  low = "red", mid = "white", high = "blue", midpoint = med_plt)
\end{lstlisting}

\includegraphics{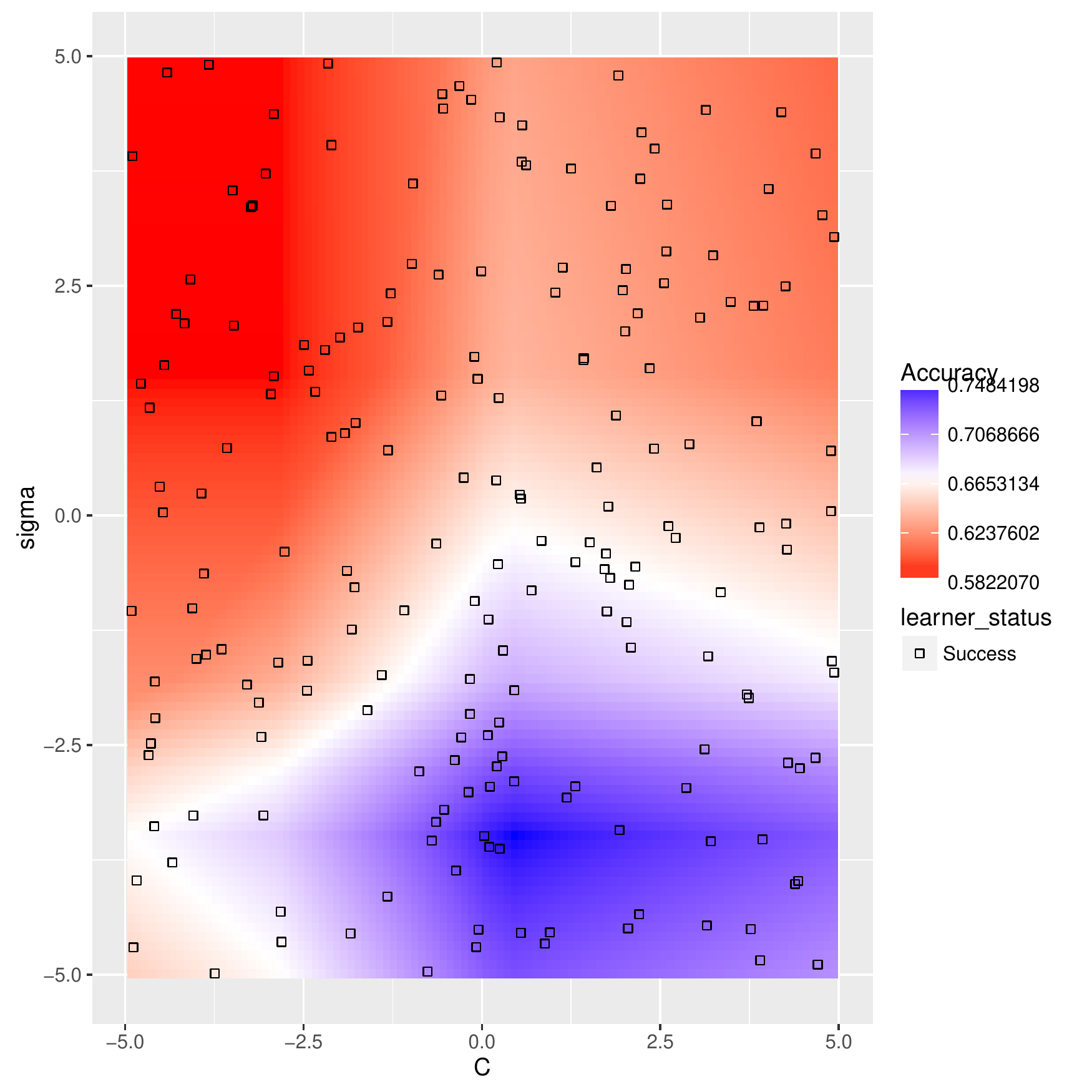}

\section{Extend}\label{extend}

\hypertarget{integrating-another-learner}{\subsection{Integrating
Another Learner}\label{integrating-another-learner}}

In order to integrate a learning algorithm into
\href{http://www.rdocumentation.org/packages/mlr/}{mlr} some interface
code has to be written. Three functions are mandatory for each learner.

\begin{itemize}
\tightlist
\item
  First, define a new learner class with a name, description,
  capabilities, parameters, and a few other things. (An object of this
  class can then be generated by
  \href{http://www.rdocumentation.org/packages/mlr/functions/makeLearner.html}{makeLearner}.)
\item
  Second, you need to provide a function that calls the learner function
  and builds the model given data (which makes it possible to invoke
  training by calling
  \href{http://www.rdocumentation.org/packages/mlr/}{mlr}'s
  \href{http://www.rdocumentation.org/packages/mlr/functions/train.html}{train}
  function).
\item
  Finally, a prediction function that returns predicted values given new
  data is required (which enables invoking prediction by calling
  \href{http://www.rdocumentation.org/packages/mlr/}{mlr}'s
  \href{http://www.rdocumentation.org/packages/mlr/functions/predict.WrappedModel.html}{predict}
  function).
\end{itemize}

Technically, integrating a learning method means introducing a new S3
\href{http://www.rdocumentation.org/packages/base/functions/class.html}{class}
and implementing the corresponding methods for the generic functions
\href{http://www.rdocumentation.org/packages/mlr/functions/RLearner.html}{makeRLerner},
\href{http://www.rdocumentation.org/packages/mlr/functions/trainLearner.html}{trainLearner},
and
\href{http://www.rdocumentation.org/packages/mlr/functions/predictLearner.html}{predictLearner}.
Therefore we start with a quick overview of the involved
\href{http://www.rdocumentation.org/packages/base/functions/class.html}{classes}
and constructor functions.

\subsubsection{Classes, constructors, and naming
schemes}\label{classes-constructors-and-naming-schemes}

As you already know
\href{http://www.rdocumentation.org/packages/mlr/functions/makeLearner.html}{makeLearner}
generates an object of class
\href{http://www.rdocumentation.org/packages/mlr/functions/makeLearner.html}{Learner}.

\begin{lstlisting}[language=R]
class(makeLearner(cl = "classif.lda"))
#> [1] "classif.lda"     "RLearnerClassif" "RLearner"        "Learner"

class(makeLearner(cl = "regr.lm"))
#> [1] "regr.lm"      "RLearnerRegr" "RLearner"     "Learner"

class(makeLearner(cl = "surv.coxph"))
#> [1] "surv.coxph"   "RLearnerSurv" "RLearner"     "Learner"

class(makeLearner(cl = "cluster.kmeans"))
#> [1] "cluster.kmeans"  "RLearnerCluster" "RLearner"        "Learner"

class(makeLearner(cl = "multilabel.rFerns"))
#> [1] "multilabel.rFerns"  "RLearnerMultilabel" "RLearner"          
#> [4] "Learner"
\end{lstlisting}

The first element of each
\href{http://www.rdocumentation.org/packages/base/functions/class.html}{class}
attribute vector is the name of the learner class passed to the
\lstinline!cl! argument of
\href{http://www.rdocumentation.org/packages/mlr/functions/makeLearner.html}{makeLearner}.
Obviously, this adheres to the naming conventions

\begin{itemize}
\tightlist
\item
  \lstinline!"classif.<R_method_name>"! for classification,
\item
  \lstinline!"multilabel.<R_method_name>"! for multilabel
  classification,
\item
  \lstinline!"regr.<R_method_name>"! for regression,
\item
  \lstinline!"surv.<R_method_name>"! for survival analysis, and
\item
  \lstinline!"cluster.<R_method_name>"! for clustering.
\end{itemize}

Additionally, there exist intermediate classes that reflect the type of
learning problem, i.e., all classification learners inherit from
\href{http://www.rdocumentation.org/packages/mlr/functions/RLearner.html}{RLearnerClassif},
all regression learners from
\href{http://www.rdocumentation.org/packages/mlr/functions/RLearner.html}{RLearnerRegr}
and so on. Their superclasses are
\href{http://www.rdocumentation.org/packages/mlr/functions/RLearner.html}{RLearner}
and finally
\href{http://www.rdocumentation.org/packages/mlr/functions/makeLearner.html}{Learner}.
For all these (sub)classes there exist constructor functions
\href{http://www.rdocumentation.org/packages/mlr/functions/RLearner.html}{makeRLearner},
\href{http://www.rdocumentation.org/packages/mlr/functions/RLearner.html}{makeRLearnerClassif},
\href{http://www.rdocumentation.org/packages/mlr/functions/RLearner.html}{makeRLearneRegr}
etc. that are called internally by
\href{http://www.rdocumentation.org/packages/mlr/functions/makeLearner.html}{makeLearner}.

A short side remark: As you might have noticed there does not exist a
special learner class for
\protect\hyperlink{cost-sensitive-classification}{cost-sensitive
classification (costsens)} with example-specific costs. This type of
learning task is currently exclusively handled through
\protect\hyperlink{wrapper}{wrappers} like
\href{http://www.rdocumentation.org/packages/mlr/functions/makeCostSensWeightedPairsWrapper.html}{makeCostSensWeightedPairsWrapper}.

In the following we show how to integrate learners for the five types of
learning tasks mentioned above. Defining a completely new type of
learner that has special properties and does not fit into one of the
existing schemes is of course possible, but much more advanced and not
covered here.

We use a classification example to explain some general principles (so
even if you are interested in integrating a learner for another type of
learning task you might want to read the following section). Examples
for other types of learning tasks are shown later on.

\subsubsection{Classification}\label{classification-1}

We show how the
\href{http://www.rdocumentation.org/packages/MASS/functions/lda.html}{Linear
Discriminant Analysis} from package
\href{http://www.rdocumentation.org/packages/MASS/}{MASS} has been
integrated into the classification learner \lstinline!classif.lda! in
\href{http://www.rdocumentation.org/packages/mlr/}{mlr} as an example.

\paragraph{Definition of the learner}\label{definition-of-the-learner}

The minimal information required to define a learner is the
\href{http://www.rdocumentation.org/packages/mlr/}{mlr} name of the
learner, its package, the parameter set, and the set of properties of
your learner. In addition, you may provide a human-readable name, a
short name and a note with information relevant to users of the learner.

First, name your learner. According to the naming conventions above the
name starts with \lstinline!classif.! and we choose
\lstinline!classif.lda!.

Second, we need to define the parameters of the learner. These are any
options that can be set when running it to change how it learns, how
input is interpreted, how and what output is generated, and so on.
\href{http://www.rdocumentation.org/packages/mlr/}{mlr} provides a
number of functions to define parameters, a complete list can be found
in the documentation of
\href{http://www.rdocumentation.org/packages/ParamHelpers/functions/LearnerParam.html}{LearnerParam}
of the
\href{http://www.rdocumentation.org/packages/ParamHelpers/}{ParamHelpers}
package.

In our example, we have discrete and numeric parameters, so we use
\href{http://www.rdocumentation.org/packages/ParamHelpers/functions/LearnerParam.html}{makeDiscreteLearnerParam}
and
\href{http://www.rdocumentation.org/packages/ParamHelpers/functions/LearnerParam.html}{makeNumericLearnerParam}
to incorporate the complete description of the parameters. We include
all possible values for discrete parameters and lower and upper bounds
for numeric parameters. Strictly speaking it is not necessary to provide
bounds for all parameters and if this information is not available they
can be estimated, but providing accurate and specific information here
makes it possible to tune the learner much better (see the section on
\protect\hyperlink{tuning-hyperparameters}{tuning}).

Next, we add information on the properties of the learner (see also the
section on \protect\hyperlink{learners}{learners}). Which types of
features are supported (numerics, factors)? Are case weights supported?
Are class weights supported? Can the method deal with missing values in
the features and deal with NA's in a meaningful way (not
\lstinline!na.omit!)? Are one-class, two-class, multi-class problems
supported? Can the learner predict posterior probabilities?

If the learner supports class weights the name of the relevant learner
parameter can be specified via argument \lstinline!class.weights.param!.

Below is the complete code for the definition of the LDA learner. It has
one discrete parameter, \lstinline!method!, and two continuous ones,
\lstinline!nu! and \lstinline!tol!. It supports classification problems
with two or more classes and can deal with numeric and factor
explanatory variables. It can predict posterior probabilities.

\begin{lstlisting}[language=R]
makeRLearner.classif.lda = function() {
  makeRLearnerClassif(
    cl = "classif.lda",
    package = "MASS",
    par.set = makeParamSet(
      makeDiscreteLearnerParam(id = "method", default = "moment", values = c("moment", "mle", "mve", "t")),
      makeNumericLearnerParam(id = "nu", lower = 2, requires = quote(method == "t")),
      makeNumericLearnerParam(id = "tol", default = 1e-4, lower = 0),
      makeDiscreteLearnerParam(id = "predict.method", values = c("plug-in", "predictive", "debiased"),
        default = "plug-in", when = "predict"),
      makeLogicalLearnerParam(id = "CV", default = FALSE, tunable = FALSE)
    ),
    properties = c("twoclass", "multiclass", "numerics", "factors", "prob"),
    name = "Linear Discriminant Analysis",
    short.name = "lda",
    note = "Learner param 'predict.method' maps to 'method' in predict.lda."
  )
}
\end{lstlisting}

\paragraph{Creating the training function of the
learner}\label{creating-the-training-function-of-the-learner}

Once the learner has been defined, we need to tell
\href{http://www.rdocumentation.org/packages/mlr/}{mlr} how to call it
to train a model. The name of the function has to start with
\lstinline!trainLearner.!, followed by the
\href{http://www.rdocumentation.org/packages/mlr/}{mlr} name of the
learner as defined above (\lstinline!classif.lda! here). The prototype
of the function looks as follows.

\begin{lstlisting}[language=R]
function(.learner, .task, .subset, .weights = NULL, ...) { }
\end{lstlisting}

This function must fit a model on the data of the task \lstinline!.task!
with regard to the subset defined in the integer vector
\lstinline!.subset! and the parameters passed in the \lstinline!...!
arguments. Usually, the data should be extracted from the task using
\href{http://www.rdocumentation.org/packages/mlr/functions/getTaskData.html}{getTaskData}.
This will take care of any subsetting as well. It must return the fitted
model. \href{http://www.rdocumentation.org/packages/mlr/}{mlr} assumes
no special data type for the return value -- it will be passed to the
predict function we are going to define below, so any special code the
learner may need can be encapsulated there.

For our example, the definition of the function looks like this. In
addition to the data of the task, we also need the formula that
describes what to predict. We use the function
\href{http://www.rdocumentation.org/packages/mlr/functions/getTaskFormula.html}{getTaskFormula}
to extract this from the task.

\begin{lstlisting}[language=R]
trainLearner.classif.lda = function(.learner, .task, .subset, .weights = NULL, ...) {
  f = getTaskFormula(.task)
  MASS::lda(f, data = getTaskData(.task, .subset), ...)
}
\end{lstlisting}

\paragraph{Creating the prediction
method}\label{creating-the-prediction-method}

Finally, the prediction function needs to be defined. The name of this
function starts with \lstinline!predictLearner.!, followed again by the
\href{http://www.rdocumentation.org/packages/mlr/}{mlr} name of the
learner. The prototype of the function is as follows.

\begin{lstlisting}[language=R]
function(.learner, .model, .newdata, ...) { }
\end{lstlisting}

It must predict for the new observations in the \lstinline!data.frame!
\lstinline!.newdata! with the wrapped model \lstinline!.model!, which is
returned from the training function. The actual model the learner built
is stored in the \lstinline!$learner.model! member and can be accessed
simply through \lstinline!.model$learner.model!.

For classification, you have to return a factor of predicted classes if
\lstinline!.learner$predict.type! is \lstinline!"response"!, or a matrix
of predicted probabilities if \lstinline!.learner$predict.type! is
\lstinline!"prob"! and this type of prediction is supported by the
learner. In the latter case the matrix must have the same number of
columns as there are classes in the task and the columns have to be
named by the class names.

The definition for LDA looks like this. It is pretty much just a
straight pass-through of the arguments to the
\href{http://www.rdocumentation.org/packages/base/functions/predict.html}{predict}
function and some extraction of prediction data depending on the type of
prediction requested.

\begin{lstlisting}[language=R]
predictLearner.classif.lda = function(.learner, .model, .newdata, predict.method = "plug-in", ...) {
  p = predict(.model$learner.model, newdata = .newdata, method = predict.method, ...)
  if (.learner$predict.type == "response") 
    return(p$class) else return(p$posterior)
}
\end{lstlisting}

\subsubsection{Regression}\label{regression-1}

The main difference for regression is that the type of predictions are
different (numeric instead of labels or probabilities) and that not all
of the properties are relevant. In particular, whether one-, two-, or
multi-class problems and posterior probabilities are supported is not
applicable.

Apart from this, everything explained above applies. Below is the
definition for the
\href{http://www.rdocumentation.org/packages/earth/functions/earth.html}{earth}
learner from the
\href{http://www.rdocumentation.org/packages/earth/}{earth} package.

\begin{lstlisting}[language=R]
makeRLearner.regr.earth = function() {
  makeRLearnerRegr(
    cl = "regr.earth",
    package = "earth",
    par.set = makeParamSet(
      makeLogicalLearnerParam(id = "keepxy", default = FALSE, tunable = FALSE),
      makeNumericLearnerParam(id = "trace", default = 0, upper = 10, tunable = FALSE),
      makeIntegerLearnerParam(id = "degree", default = 1L, lower = 1L),
      makeNumericLearnerParam(id = "penalty"),
      makeIntegerLearnerParam(id = "nk", lower = 0L),
      makeNumericLearnerParam(id = "thres", default = 0.001),
      makeIntegerLearnerParam(id = "minspan", default = 0L),
      makeIntegerLearnerParam(id = "endspan", default = 0L),
      makeNumericLearnerParam(id = "newvar.penalty", default = 0),
      makeIntegerLearnerParam(id = "fast.k", default = 20L, lower = 0L),
      makeNumericLearnerParam(id = "fast.beta", default = 1),
      makeDiscreteLearnerParam(id = "pmethod", default = "backward",
        values = c("backward", "none", "exhaustive", "forward", "seqrep", "cv")),
      makeIntegerLearnerParam(id = "nprune")
    ),
    properties = c("numerics", "factors"),
    name = "Multivariate Adaptive Regression Splines",
    short.name = "earth",
    note = ""
  )
}
\end{lstlisting}

\begin{lstlisting}[language=R]
trainLearner.regr.earth = function(.learner, .task, .subset, .weights = NULL, ...) {
  f = getTaskFormula(.task)
  earth::earth(f, data = getTaskData(.task, .subset), ...)
}
\end{lstlisting}

\begin{lstlisting}[language=R]
predictLearner.regr.earth = function(.learner, .model, .newdata, ...) {
  predict(.model$learner.model, newdata = .newdata)[, 1L]
}
\end{lstlisting}

Again most of the data is passed straight through to/from the
train/predict functions of the learner.

\subsubsection{Survival analysis}\label{survival-analysis-1}

For survival analysis, you have to return so-called linear predictors in
order to compute the default measure for this task type, the
\protect\hyperlink{implemented-performance-measures}{cindex} (for
\lstinline!.learner$predict.type! == \lstinline!"response"!). For
\lstinline!.learner$predict.type! == \lstinline!"prob"!, there is no
substantially meaningful measure (yet). You may either ignore this case
or return something like predicted survival curves (cf.~example below).

There are three properties that are specific to survival learners:
``rcens'', ``lcens'' and ``icens'', defining the type(s) of censoring a
learner can handle -- right, left and/or interval censored.

Let's have a look at how the
\href{http://www.rdocumentation.org/packages/survival/functions/coxph.html}{Cox
Proportional Hazard Model} from package
\href{http://www.rdocumentation.org/packages/survival/}{survival} has
been integrated into the survival learner \lstinline!surv.coxph! in
\href{http://www.rdocumentation.org/packages/mlr/}{mlr} as an example:

\begin{lstlisting}[language=R]
makeRLearner.surv.coxph = function() {
  makeRLearnerSurv(
    cl = "surv.coxph",
    package = "survival",
    par.set = makeParamSet(
      makeDiscreteLearnerParam(id = "ties", default = "efron", values = c("efron", "breslow", "exact")),
      makeLogicalLearnerParam(id = "singular.ok", default = TRUE),
      makeNumericLearnerParam(id = "eps", default = 1e-09, lower = 0),
      makeNumericLearnerParam(id = "toler.chol", default = .Machine$double.eps^0.75, lower = 0),
      makeIntegerLearnerParam(id = "iter.max", default = 20L, lower = 1L),
      makeNumericLearnerParam(id = "toler.inf", default = sqrt(.Machine$double.eps^0.75), lower = 0),
      makeIntegerLearnerParam(id = "outer.max", default = 10L, lower = 1L),
      makeLogicalLearnerParam(id = "model", default = FALSE, tunable = FALSE),
      makeLogicalLearnerParam(id = "x", default = FALSE, tunable = FALSE),
      makeLogicalLearnerParam(id = "y", default = TRUE, tunable = FALSE)
    ),
    properties = c("missings", "numerics", "factors", "weights", "prob", "rcens"),
    name = "Cox Proportional Hazard Model",
    short.name = "coxph",
    note = ""
  )
}
\end{lstlisting}

\begin{lstlisting}[language=R]
trainLearner.surv.coxph = function(.learner, .task, .subset, .weights = NULL, ...) {
  f = getTaskFormula(.task)
  data = getTaskData(.task, subset = .subset)
  if (is.null(.weights)) {
    mod = survival::coxph(formula = f, data = data, ...)
  } else {
    mod = survival::coxph(formula = f, data = data, weights = .weights, ...)
  }
  mod
}
\end{lstlisting}

\begin{lstlisting}[language=R]
predictLearner.surv.coxph = function(.learner, .model, .newdata, ...) {
  if (.learner$predict.type == "response") {
    predict(.model$learner.model, newdata = .newdata, type = "lp", ...)
  }
}
\end{lstlisting}

\subsubsection{Clustering}\label{clustering}

For clustering, you have to return a numeric vector with the IDs of the
clusters that the respective datum has been assigned to. The numbering
should start at 1.

Below is the definition for the
\href{http://www.rdocumentation.org/packages/RWeka/functions/FarthestFirst.html}{FarthestFirst}
learner from the
\href{http://www.rdocumentation.org/packages/RWeka/}{RWeka} package.
Weka starts the IDs of the clusters at 0, so we add 1 to the predicted
clusters. RWeka has a different way of setting learner parameters; we
use the special \lstinline!Weka_control! function to do this.

\begin{lstlisting}[language=R]
makeRLearner.cluster.FarthestFirst = function() {
  makeRLearnerCluster(
    cl = "cluster.FarthestFirst",
    package = "RWeka",
    par.set = makeParamSet(
      makeIntegerLearnerParam(id = "N", default = 2L, lower = 1L),
      makeIntegerLearnerParam(id = "S", default = 1L, lower = 1L),
      makeLogicalLearnerParam(id = "output-debug-info", default = FALSE, tunable = FALSE)
    ),
    properties = c("numerics"),
    name = "FarthestFirst Clustering Algorithm",
    short.name = "farthestfirst"
  )
}
\end{lstlisting}

\begin{lstlisting}[language=R]
trainLearner.cluster.FarthestFirst = function(.learner, .task, .subset, .weights = NULL, ...) {
  ctrl = RWeka::Weka_control(...)
  RWeka::FarthestFirst(getTaskData(.task, .subset), control = ctrl)
}
\end{lstlisting}

\begin{lstlisting}[language=R]
predictLearner.cluster.FarthestFirst = function(.learner, .model, .newdata, ...) {
  as.integer(predict(.model$learner.model, .newdata, ...)) + 1L
}
\end{lstlisting}

\subsubsection{Multilabel
classification}\label{multilabel-classification-2}

As stated in the
\protect\hyperlink{multilabel-classification}{multilabel} section,
multilabel classification methods can be divided into problem
transformation methods and algorithm adaptation methods.

At this moment the only problem transformation method implemented in
\href{http://www.rdocumentation.org/packages/mlr/}{mlr} is the
\href{http://www.rdocumentation.org/packages/mlr/functions/makeMultilabelBinaryRelevanceWrapper.html}{binary
relevance method}. Integrating more of these methods requires good
knowledge of the architecture of the
\href{http://www.rdocumentation.org/packages/mlr/}{mlr} package.

The integration of an algorithm adaptation multilabel classification
learner is easier and works very similar to the normal
multiclass-classification. In contrast to the multiclass case, not all
of the learner properties are relevant. In particular, whether one-,
two-, or multi-class problems are supported is not applicable.
Furthermore the prediction function output must be a matrix with each
prediction of a label in one column and the names of the labels as
column names. If \lstinline!.learner$predict.type! is
\lstinline!"response"! the predictions must be logical. If
\lstinline!.learner$predict.type! is \lstinline!"prob"! and this type of
prediction is supported by the learner, the matrix must consist of
predicted probabilities.

Below is the definition of the
\href{http://www.rdocumentation.org/packages/rFerns/functions/rFerns.html}{rFerns}
learner from the
\href{http://www.rdocumentation.org/packages/rFerns/}{rFerns} package,
which does not support probability predictions.

\begin{lstlisting}[language=R]
makeRLearner.multilabel.rFerns = function() {
  makeRLearnerMultilabel(
    cl = "multilabel.rFerns",
    package = "rFerns",
    par.set = makeParamSet(
      makeIntegerLearnerParam(id = "depth", default = 5L),
      makeIntegerLearnerParam(id = "ferns", default = 1000L)
    ),
    properties = c("numerics", "factors", "ordered"),
    name = "Random ferns",
    short.name = "rFerns",
    note = ""
  )
}
\end{lstlisting}

\begin{lstlisting}[language=R]
trainLearner.multilabel.rFerns = function(.learner, .task, .subset, .weights = NULL, ...) {
  d = getTaskData(.task, .subset, target.extra = TRUE)
  rFerns::rFerns(x = d$data, y = as.matrix(d$target), ...)
}
\end{lstlisting}

\begin{lstlisting}[language=R]
predictLearner.multilabel.rFerns = function(.learner, .model, .newdata, ...) {
  as.matrix(predict(.model$learner.model, .newdata, ...))
}
\end{lstlisting}

\subsubsection{Creating a new feature importance
method}\label{creating-a-new-feature-importance-method}

Some learners, for example decision trees and random forests, can
calculate feature importance values, which can be extracted from a
\href{http://www.rdocumentation.org/packages/mlr/functions/makeWrappedModel.html}{fitted
model} using function
\href{http://www.rdocumentation.org/packages/mlr/functions/getFeatureImportance.html}{getFeatureImportance}.

If your newly integrated learner supports this you need to

\begin{itemize}
\tightlist
\item
  add \lstinline!"featimp"! to the learner properties and
\item
  implement a new S3 method for function
  \href{http://www.rdocumentation.org/packages/mlr/functions/getFeatureImportanceLearner.html}{getFeatureImportanceLearner}
  (which later is called internally by
  \href{http://www.rdocumentation.org/packages/mlr/functions/getFeatureImportance.html}{getFeatureImportance}).
\end{itemize}

in order to make this work.

This method takes the
\href{http://www.rdocumentation.org/packages/mlr/functions/Learner.html}{Learner}
\lstinline!.learner!, the
\href{http://www.rdocumentation.org/packages/mlr/functions/makeWrappedModel.html}{WrappedModel}
\lstinline!.model! and potential further arguments and calculates or
extracts the feature importance. It must return a named vector of
importance values.

Below are two simple examples. In case of \lstinline!"classif.rpart"!
the feature importance values can be easily extracted from the fitted
model.

\begin{lstlisting}[language=R]
getFeatureImportanceLearner.classif.rpart = function(.learner, .model, ...) {
  mod = getLearnerModel(.model)
  mod$variable.importance
}
\end{lstlisting}

For the
\href{http://www.rdocumentation.org/packages/randomForestSRC/functions/rfsrc.html}{random
forest} from package
\href{http://www.rdocumentation.org/packages/randomForestSRC/}{randomForestSRC}
function
\href{http://www.rdocumentation.org/packages/randomForestSRC/functions/vimp.html}{vimp}
is called.

\begin{lstlisting}[language=R]
getFeatureImportanceLearner.classif.randomForestSRC = function(.learner, .model, ...) {
  mod = getLearnerModel(.model)
  randomForestSRC::vimp(mod, ...)$importance[, "all"]
}
\end{lstlisting}

\subsubsection{Registering your learner}\label{registering-your-learner}

If your interface code to a new learning algorithm exists only locally,
i.e., it is not (yet) merged into
\href{http://www.rdocumentation.org/packages/mlr/}{mlr} or does not live
in an extra package with a proper namespace you might want to register
the new S3 methods to make sure that these are found by, e.g.,
\href{http://www.rdocumentation.org/packages/mlr/functions/listLearners.html}{listLearners}.
You can do this as follows:

\begin{lstlisting}[language=R]
registerS3method("makeRLearner", "<awesome_new_learner_class>", makeRLearner.<awesome_new_learner_class>)
registerS3method("trainLearner", "<awesome_new_learner_class>", trainLearner.<awesome_new_learner_class>)
registerS3method("predictLearner", "<awesome_new_learner_class>", predictLearner.<awesome_new_learner_class>)

### And if you also have written a method to extract the feature importance
registerS3method("getFeatureImportanceLearner", "<awesome_new_learner_class>",
  getFeatureImportanceLearner.<awesome_new_learner_class>)
\end{lstlisting}

\hypertarget{integrating-another-measure}{\subsection{Integrating
Another Measure}\label{integrating-another-measure}}

In some cases, you might want to evaluate a
\href{http://www.rdocumentation.org/packages/mlr/functions/Prediction.html}{Prediction}
or
\href{http://www.rdocumentation.org/packages/mlr/functions/ResamplePrediction.html}{ResamplePrediction}
with a
\href{http://www.rdocumentation.org/packages/mlr/functions/makeMeasure.html}{Measure}
which is not yet implemented in
\href{http://www.rdocumentation.org/packages/mlr/}{mlr}. This could be
either a performance measure which is not listed in the
\protect\hyperlink{implemented-performance-measures}{Appendix} or a
measure that uses a misclassification cost matrix.

\subsubsection{Performance measures and aggregation
schemes}\label{performance-measures-and-aggregation-schemes}

Performance measures in
\href{http://www.rdocumentation.org/packages/mlr/}{mlr} are objects of
class
\href{http://www.rdocumentation.org/packages/mlr/functions/makeMeasure.html}{Measure}.
For example the
\protect\hyperlink{implemented-performance-measures}{mse} (mean squared
error) looks as follows.

\begin{lstlisting}[language=R]
str(mse)
#> List of 10
#>  $ id        : chr "mse"
#>  $ minimize  : logi TRUE
#>  $ properties: chr [1:3] "regr" "req.pred" "req.truth"
#>  $ fun       :function (task, model, pred, feats, extra.args)  
#>  $ extra.args: list()
#>  $ best      : num 0
#>  $ worst     : num Inf
#>  $ name      : chr "Mean of squared errors"
#>  $ note      : chr ""
#>  $ aggr      :List of 4
#>   ..$ id        : chr "test.mean"
#>   ..$ name      : chr "Test mean"
#>   ..$ fun       :function (task, perf.test, perf.train, measure, group, pred)  
#>   ..$ properties: chr "req.test"
#>   ..- attr(*, "class")= chr "Aggregation"
#>  - attr(*, "class")= chr "Measure"

mse$fun
#> function (task, model, pred, feats, extra.args) 
#> {
#>     measureMSE(pred$data$truth, pred$data$response)
#> }
#> <bytecode: 0xbfd1b48>
#> <environment: namespace:mlr>

measureMSE
#> function (truth, response) 
#> {
#>     mean((response - truth)^2)
#> }
#> <bytecode: 0xbc6a8f8>
#> <environment: namespace:mlr>
\end{lstlisting}

See the
\href{http://www.rdocumentation.org/packages/mlr/functions/makeMeasure.html}{Measure}
documentation page for a detailed description of the object slots.

At the core is slot \lstinline!$fun! which contains the function that
calculates the performance value. The actual work is done by function
\href{http://www.rdocumentation.org/packages/mlr/functions/measures.html}{measureMSE}.
Similar functions, generally adhering to the naming scheme
\lstinline!measure! followed by the capitalized measure ID, exist for
most performance measures. See the
\href{http://www.rdocumentation.org/packages/mlr/functions/measures.html}{measures}
help page for a complete list.

Just as
\href{http://www.rdocumentation.org/packages/mlr/functions/Task.html}{Task}
and
\href{http://www.rdocumentation.org/packages/mlr/functions/makeLearner.html}{Learner}
objects each
\href{http://www.rdocumentation.org/packages/mlr/functions/makeMeasure.html}{Measure}
has an identifier \lstinline!$id! which is for example used to annotate
results and plots. For plots there is also the option to use the longer
measure \lstinline!$name! instead. See the tutorial page on
\protect\hyperlink{visualization}{Visualization} for more information.

Moreover, a
\href{http://www.rdocumentation.org/packages/mlr/functions/makeMeasure.html}{Measure}
includes a number of \lstinline!$properties! that indicate for which
types of learning problems it is suitable and what information is
required to calculate it. Obviously, most measures need the
\href{http://www.rdocumentation.org/packages/mlr/functions/Prediction.html}{Prediction}
object (\lstinline!"req.pred"!) and, for supervised problems, the true
values of the target variable(s) (\lstinline!"req.truth"!).

For \protect\hyperlink{tuning-hyperparameters}{tuning} or
\protect\hyperlink{feature-selection}{feature selection} each
\href{http://www.rdocumentation.org/packages/mlr/functions/makeMeasure.html}{Measure}
knows its extreme values \lstinline!$best! and \lstinline!$worst! and if
it wants to be minimized or maximized (\lstinline!$minimize!).

For resampling slot \lstinline!$aggr! specifies how the overall
performance across all resampling iterations is calculated. Typically,
this is just a matter of aggregating the performance values obtained on
the test sets \lstinline!perf.test! or the training sets
\lstinline!perf.train! by a simple function. The by far most common
scheme is
\href{http://www.rdocumentation.org/packages/mlr/functions/aggregations.html}{test.mean},
i.e., the unweighted mean of the performances on the test sets.

\begin{lstlisting}[language=R]
str(test.mean)
#> List of 4
#>  $ id        : chr "test.mean"
#>  $ name      : chr "Test mean"
#>  $ fun       :function (task, perf.test, perf.train, measure, group, pred)  
#>  $ properties: chr "req.test"
#>  - attr(*, "class")= chr "Aggregation"

test.mean$fun
#> function (task, perf.test, perf.train, measure, group, pred) 
#> mean(perf.test)
#> <bytecode: 0xa8f7860>
#> <environment: namespace:mlr>
\end{lstlisting}

All aggregation schemes are objects of class
\href{http://www.rdocumentation.org/packages/mlr/functions/Aggregation.html}{Aggregation}
with the function in slot \lstinline!$fun! doing the actual work. The
\lstinline!$properties! member indicates if predictions (or performance
values) on the training or test data sets are required to calculate the
aggregation.

You can change the aggregation scheme of a
\href{http://www.rdocumentation.org/packages/mlr/functions/makeMeasure.html}{Measure}
via function
\href{http://www.rdocumentation.org/packages/mlr/functions/setAggregation.html}{setAggregation}.
See the tutorial page on \protect\hyperlink{resampling}{resampling} for
some examples and the
\href{http://www.rdocumentation.org/packages/mlr/functions/aggregations.html}{aggregations}
help page for all available aggregation schemes.

You can construct your own
\href{http://www.rdocumentation.org/packages/mlr/functions/makeMeasure.html}{Measure}
and
\href{http://www.rdocumentation.org/packages/mlr/functions/Aggregation.html}{Aggregation}
objects via functions
\href{http://www.rdocumentation.org/packages/mlr/functions/makeMeasure.html}{makeMeasure},
\href{http://www.rdocumentation.org/packages/mlr/functions/makeCostMeasure.html}{makeCostMeasure},
\href{http://www.rdocumentation.org/packages/mlr/functions/makeCustomResampledMeasure.html}{makeCustomResampledMeasure}
and
\href{http://www.rdocumentation.org/packages/mlr/functions/makeAggregation.html}{makeAggregation}.
Some examples are shown in the following.

\subsubsection{Constructing a performance
measure}\label{constructing-a-performance-measure}

Function
\href{http://www.rdocumentation.org/packages/mlr/functions/makeMeasure.html}{makeMeasure}
provides a simple way to construct your own performance measure.

Below this is exemplified by re-implementing the mean misclassification
error (\protect\hyperlink{implemented-performance-measures}{mmce}). We
first write a function that computes the measure on the basis of the
true and predicted class labels. Note that this function must have
certain formal arguments listed in the documentation of
\href{http://www.rdocumentation.org/packages/mlr/functions/makeMeasure.html}{makeMeasure}.
Then the
\href{http://www.rdocumentation.org/packages/mlr/functions/makeMeasure.html}{Measure}
object is created and we work with it as usual with the
\href{http://www.rdocumentation.org/packages/mlr/functions/performance.html}{performance}
function.

See the \textbf{R} documentation of
\href{http://www.rdocumentation.org/packages/mlr/functions/makeMeasure.html}{makeMeasure}
for more details on the various parameters.

\begin{lstlisting}[language=R]
### Define a function that calculates the misclassification rate
my.mmce.fun = function(task, model, pred, feats, extra.args) {
  tb = table(getPredictionResponse(pred), getPredictionTruth(pred))
  1 - sum(diag(tb)) / sum(tb)
}

### Generate the Measure object
my.mmce = makeMeasure(
  id = "my.mmce", name = "My Mean Misclassification Error",
  properties = c("classif", "classif.multi", "req.pred", "req.truth"),
  minimize = TRUE, best = 0, worst = 1,
  fun = my.mmce.fun
)

### Train a learner and make predictions
mod = train("classif.lda", iris.task)
pred = predict(mod, task = iris.task)

### Calculate the performance using the new measure
performance(pred, measures = my.mmce)
#> my.mmce 
#>    0.02

### Apparently the result coincides with the mlr implementation
performance(pred, measures = mmce)
#> mmce 
#> 0.02
\end{lstlisting}

\subsubsection{Constructing a measure for ordinary misclassification
costs}\label{constructing-a-measure-for-ordinary-misclassification-costs}

For in depth explanations and details see the tutorial page on
\protect\hyperlink{cost-sensitive-classification}{cost-sensitive
classification}.

To create a measure that involves ordinary, i.e., class-dependent
misclassification costs you can use function
\href{http://www.rdocumentation.org/packages/mlr/functions/makeCostMeasure.html}{makeCostMeasure}.
You first need to define the cost matrix. The rows indicate true and the
columns predicted classes and the rows and columns have to be named by
the class labels. The cost matrix can then be wrapped in a
\href{http://www.rdocumentation.org/packages/mlr/functions/makeMeasure.html}{Measure}
object and predictions can be evaluated as usual with the
\href{http://www.rdocumentation.org/packages/mlr/functions/performance.html}{performance}
function.

See the \textbf{R} documentation of function
\href{http://www.rdocumentation.org/packages/mlr/functions/makeCostMeasure.html}{makeCostMeasure}
for details on the various parameters.

\begin{lstlisting}[language=R]
### Create the cost matrix
costs = matrix(c(0, 2, 2, 3, 0, 2, 1, 1, 0), ncol = 3)
rownames(costs) = colnames(costs) = getTaskClassLevels(iris.task)

### Encapsulate the cost matrix in a Measure object
my.costs = makeCostMeasure(
  id = "my.costs", name = "My Costs",
  costs = costs,
  minimize = TRUE, best = 0, worst = 3
)

### Train a learner and make a prediction
mod = train("classif.lda", iris.task)
pred = predict(mod, newdata = iris)

### Calculate the average costs
performance(pred, measures = my.costs)
#>   my.costs 
#> 0.02666667
\end{lstlisting}

\subsubsection{Creating an aggregation
scheme}\label{creating-an-aggregation-scheme}

It is possible to create your own aggregation scheme using function
\href{http://www.rdocumentation.org/packages/mlr/functions/makeAggregation.html}{makeAggregation}.
You need to specify an identifier \lstinline!id!, the
\lstinline!properties!, and write a function that does the actual
aggregation. Optionally, you can \lstinline!name! your aggregation
scheme.

Possible settings for \lstinline!properties! are \lstinline!"req.test"!
and \lstinline!"req.train"! if predictions on either the training or
test sets are required, and the vector
\lstinline!c("req.train", "req.test")! if both are needed.

The aggregation function must have a certain signature detailed in the
documentation of
\href{http://www.rdocumentation.org/packages/mlr/functions/makeAggregation.html}{makeAggregation}.
Usually, you will only need the performance values on the test sets
\lstinline!perf.test! or the training sets \lstinline!perf.train!. In
rare cases, e.g., the
\href{http://www.rdocumentation.org/packages/mlr/functions/Prediction.html}{Prediction}
object \lstinline!pred! or information stored in the
\href{http://www.rdocumentation.org/packages/mlr/functions/Task.html}{Task}
object might be required to obtain the aggregated performance. For an
example have a look at the
\href{https://github.com/mlr-org/mlr/blob/master/R/aggregations.R\#L211}{definition}
of function
\href{http://www.rdocumentation.org/packages/mlr/functions/aggregations.html}{test.join}.

\paragraph{Example: Evaluating the range of
measures}\label{example-evaluating-the-range-of-measures}

Let's say you are interested in the range of the performance values
obtained on individual test sets.

\begin{lstlisting}[language=R]
my.range.aggr = makeAggregation(id = "test.range", name = "Test Range",
  properties = "req.test",
  fun = function (task, perf.test, perf.train, measure, group, pred)
    diff(range(perf.test))
)
\end{lstlisting}

\lstinline!perf.train! and \lstinline!perf.test! are both numerical
vectors containing the performances on the train and test data sets. In
most cases (unless you are using bootstrap as resampling strategy or
have set \lstinline!predict = "both"! in
\href{http://www.rdocumentation.org/packages/mlr/functions/makeResampleDesc.html}{makeResampleDesc})
the \lstinline!perf.train! vector is empty.

Now we can run a feature selection based on the first measure in the
provided
\href{http://www.rdocumentation.org/packages/base/functions/list.html}{list}
and see how the other measures turn out.

\begin{lstlisting}[language=R]
### mmce with default aggregation scheme test.mean
ms1 = mmce

### mmce with new aggregation scheme my.range.aggr
ms2 = setAggregation(ms1, my.range.aggr)

### Minimum and maximum of the mmce over test sets
ms1min = setAggregation(ms1, test.min)
ms1max = setAggregation(ms1, test.max)

### Feature selection
rdesc = makeResampleDesc("CV", iters = 3)
res = selectFeatures("classif.rpart", iris.task, rdesc, measures = list(ms1, ms2, ms1min, ms1max),
  control = makeFeatSelControlExhaustive(), show.info = FALSE)

### Optimization path, i.e., performances for the 16 possible feature subsets
perf.data = as.data.frame(res$opt.path)
head(perf.data[1:8])
#>   Sepal.Length Sepal.Width Petal.Length Petal.Width mmce.test.mean
#> 1            0           0            0           0     0.70666667
#> 2            1           0            0           0     0.31333333
#> 3            0           1            0           0     0.50000000
#> 4            0           0            1           0     0.09333333
#> 5            0           0            0           1     0.04666667
#> 6            1           1            0           0     0.28666667
#>   mmce.test.range mmce.test.min mmce.test.max
#> 1            0.16          0.60          0.76
#> 2            0.02          0.30          0.32
#> 3            0.22          0.36          0.58
#> 4            0.10          0.04          0.14
#> 5            0.08          0.02          0.10
#> 6            0.08          0.24          0.32

pd = position_jitter(width = 0.005, height = 0)
p = ggplot(aes(x = mmce.test.range, y = mmce.test.mean, ymax = mmce.test.max, ymin = mmce.test.min,
  color = as.factor(Sepal.Width), pch = as.factor(Petal.Width)), data = perf.data) +
  geom_pointrange(position = pd) +
  coord_flip()
print(p)
\end{lstlisting}

\includegraphics{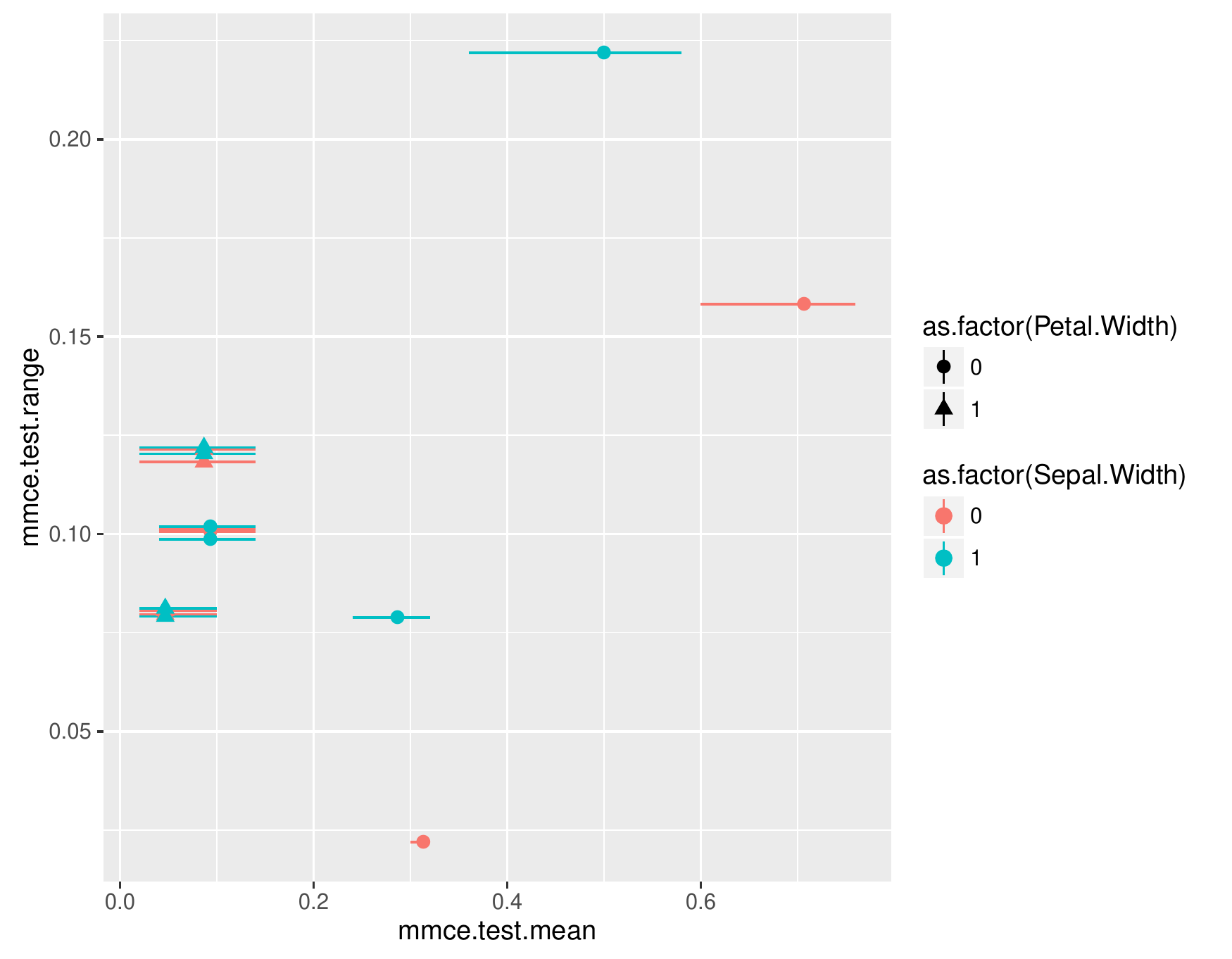}

The plot shows the range versus the mean misclassification error. The
value on the y-axis thus corresponds to the length of the error bars.
(Note that the points and error bars are jittered in y-direction.)

\hypertarget{creating-an-imputation-method}{\subsection{Creating an
Imputation Method}\label{creating-an-imputation-method}}

Function
\href{http://www.rdocumentation.org/packages/mlr/functions/makeImputeMethod.html}{makeImputeMethod}
permits to create your own imputation method. For this purpose you need
to specify a \emph{learn} function that extracts the necessary
information and an \emph{impute} function that does the actual
imputation. The \emph{learn} and \emph{impute} functions both have at
least the following formal arguments:

\begin{itemize}
\tightlist
\item
  \lstinline!data! is a
  \href{http://www.rdocumentation.org/packages/base/functions/data.frame.html}{data.frame}
  with missing values in some features.
\item
  \lstinline!col! indicates the feature to be imputed.
\item
  \lstinline!target! indicates the target variable(s) in a supervised
  learning task.
\end{itemize}

\subsubsection{Example: Imputation using the
mean}\label{example-imputation-using-the-mean}

Let's have a look at function
\href{http://www.rdocumentation.org/packages/mlr/functions/imputations.html}{imputeMean}.

\begin{lstlisting}[language=R]
imputeMean = function() {
  makeImputeMethod(learn = function(data, target, col) mean(data[[col]], na.rm = TRUE), 
    impute = simpleImpute)
}
\end{lstlisting}

\href{http://www.rdocumentation.org/packages/mlr/functions/imputations.html}{imputeMean}
calls the unexported
\href{http://www.rdocumentation.org/packages/mlr/}{mlr} function
\lstinline!simpleImpute! which is defined as follows.

\begin{lstlisting}[language=R]
simpleImpute = function(data, target, col, const) {
  if (is.na(const)) 
    stopf("Error imputing column '%s'. Maybe all input data was missing?", col)
  x = data[[col]]
  if (is.factor(x) && const %nin% levels(x)) {
    levels(x) = c(levels(x), as.character(const))
  }
  replace(x, is.na(x), const)
}
\end{lstlisting}

The \emph{learn} function calculates the mean of the non-missing
observations in column \lstinline!col!. The mean is passed via argument
\lstinline!const! to the \emph{impute} function that replaces all
missing values in feature \lstinline!col!.

\subsubsection{Writing your own imputation
method}\label{writing-your-own-imputation-method}

Now let's write a new imputation method: A frequently used simple
technique for longitudinal data is \emph{last observation carried
forward} (LOCF). Missing values are replaced by the most recent observed
value.

In the \textbf{R} code below the \emph{learn} function determines the
last observed value previous to each \lstinline!NA! (\lstinline!values!)
as well as the corresponding number of consecutive \lstinline!NA's!
(\lstinline!times!). The \emph{impute} function generates a vector by
replicating the entries in \lstinline!values! according to
\lstinline!times! and replaces the \lstinline!NA's! in feature
\lstinline!col!.

\begin{lstlisting}[language=R]
imputeLOCF = function() {
  makeImputeMethod(
    learn = function(data, target, col) {
      x = data[[col]]
      ind = is.na(x)
      dind = diff(ind)
      lastValue = which(dind == 1)  # position of the last observed value previous to NA
      lastNA = which(dind == -1)    # position of the last of potentially several consecutive NA's
      values = x[lastValue]         # last observed value previous to NA
      times = lastNA - lastValue    # number of consecutive NA's
      return(list(values = values, times = times))
    },
    impute = function(data, target, col, values, times) {
      x = data[[col]]
      replace(x, is.na(x), rep(values, times))
    }
  )
}
\end{lstlisting}

Note that this function is just for demonstration and is lacking some
checks for real-world usage (for example `What should happen if the
first value in \lstinline!x! is already missing?'). Below it is used to
impute the missing values in features \lstinline!Ozone! and
\lstinline!Solar.R! in the
\href{http://www.rdocumentation.org/packages/datasets/functions/airquality.html}{airquality}
data set.

\begin{lstlisting}[language=R]
data(airquality)
imp = impute(airquality, cols = list(Ozone = imputeLOCF(), Solar.R = imputeLOCF()),
  dummy.cols = c("Ozone", "Solar.R"))
head(imp$data, 10)
#>    Ozone Solar.R Wind Temp Month Day Ozone.dummy Solar.R.dummy
#> 1     41     190  7.4   67     5   1       FALSE         FALSE
#> 2     36     118  8.0   72     5   2       FALSE         FALSE
#> 3     12     149 12.6   74     5   3       FALSE         FALSE
#> 4     18     313 11.5   62     5   4       FALSE         FALSE
#> 5     18     313 14.3   56     5   5        TRUE          TRUE
#> 6     28     313 14.9   66     5   6       FALSE          TRUE
#> 7     23     299  8.6   65     5   7       FALSE         FALSE
#> 8     19      99 13.8   59     5   8       FALSE         FALSE
#> 9      8      19 20.1   61     5   9       FALSE         FALSE
#> 10     8     194  8.6   69     5  10        TRUE         FALSE
\end{lstlisting}

\subsection{Integrating Another Filter
Method}\label{integrating-another-filter-method}

A lot of feature filter methods are already integrated in
\href{http://www.rdocumentation.org/packages/mlr/}{mlr} and a complete
list is given in the
\protect\hyperlink{integrated-filter-methods}{Appendix} or can be
obtained using
\href{http://www.rdocumentation.org/packages/mlr/functions/listFilterMethods.html}{listFilterMethods}.
You can easily add another filter, be it a brand new one or a method
which is already implemented in another package, via function
\href{http://www.rdocumentation.org/packages/mlr/functions/makeFilter.html}{makeFilter}.

\subsubsection{Filter objects}\label{filter-objects}

In \href{http://www.rdocumentation.org/packages/mlr/}{mlr} all filter
methods are objects of class
\href{http://www.rdocumentation.org/packages/mlr/functions/makeFilter.html}{Filter}
and are registered in an environment called \lstinline!.FilterRegister!
(where
\href{http://www.rdocumentation.org/packages/mlr/functions/listFilterMethods.html}{listFilterMethods}
looks them up to compile the list of available methods). To get to know
their structure let's have a closer look at the
\lstinline!"rank.correlation"! filter which interfaces function
\href{http://www.rdocumentation.org/packages/FSelector/functions/correlation.html}{rank.correlation}
in package
\href{http://www.rdocumentation.org/packages/FSelector/}{FSelector}.

\begin{lstlisting}[language=R]
filters = as.list(mlr:::.FilterRegister)
filters$rank.correlation
#> Filter: 'rank.correlation'
#> Packages: 'FSelector'
#> Supported tasks: regr
#> Supported features: numerics

str(filters$rank.correlation)
#> List of 6
#>  $ name              : chr "rank.correlation"
#>  $ desc              : chr "Spearman's correlation between feature and target"
#>  $ pkg               : chr "FSelector"
#>  $ supported.tasks   : chr "regr"
#>  $ supported.features: chr "numerics"
#>  $ fun               :function (task, nselect, ...)  
#>  - attr(*, "class")= chr "Filter"

filters$rank.correlation$fun
#> function (task, nselect, ...) 
#> {
#>     y = FSelector::rank.correlation(getTaskFormula(task), data = getTaskData(task))
#>     setNames(y[["attr_importance"]], getTaskFeatureNames(task))
#> }
#> <bytecode: 0xc73ee50>
#> <environment: namespace:mlr>
\end{lstlisting}

The core element is \lstinline!$fun! which calculates the feature
importance. For the \lstinline!"rank.correlation"! filter it just
extracts the data and formula from the \lstinline!task! and passes them
on to the
\href{http://www.rdocumentation.org/packages/FSelector/functions/correlation.html}{rank.correlation}
function.

Additionally, each
\href{http://www.rdocumentation.org/packages/mlr/functions/makeFilter.html}{Filter}
object has a \lstinline!$name!, which should be short and is for example
used to annotate graphics (cp.
\href{http://www.rdocumentation.org/packages/mlr/functions/plotFilterValues.html}{plotFilterValues}),
and a slightly more detailed description in slot \lstinline!$desc!. If
the filter method is implemented by another package its name is given in
the \lstinline!$pkg! member. Moreover, the supported task types and
feature types are listed.

\subsubsection{Writing a new filter
method}\label{writing-a-new-filter-method}

You can integrate your own filter method using
\href{http://www.rdocumentation.org/packages/mlr/functions/makeFilter.html}{makeFilter}.
This function generates a
\href{http://www.rdocumentation.org/packages/mlr/functions/makeFilter.html}{Filter}
object and also registers it in the \lstinline!.FilterRegister!
environment.

The arguments of
\href{http://www.rdocumentation.org/packages/mlr/functions/makeFilter.html}{makeFilter}
correspond to the slot names of the
\href{http://www.rdocumentation.org/packages/mlr/functions/makeFilter.html}{Filter}
object above. Currently, feature filtering is only supported for
supervised learning tasks and possible values for
\lstinline!supported.tasks! are \lstinline!"regr"!,
\lstinline!"classif"! and \lstinline!"surv"!.
\lstinline!supported.features! can be \lstinline!"numerics"!,
\lstinline!"factors"! and \lstinline!"ordered"!.

\lstinline!fun! must be a
\href{http://www.rdocumentation.org/packages/base/functions/function.html}{function}
with at least the following formal arguments:

\begin{itemize}
\tightlist
\item
  \lstinline!task! is a
  \href{http://www.rdocumentation.org/packages/mlr/}{mlr} learning
  \href{http://www.rdocumentation.org/packages/mlr/functions/Task.html}{Task}.
\item
  \lstinline!nselect! corresponds to the argument of
  \href{http://www.rdocumentation.org/packages/mlr/functions/generateFilterValuesData.html}{generateFilterValuesData}
  of the same name and specifies the number of features for which to
  calculate importance scores. Some filter methods have the option to
  stop after a certain number of top-ranked features have been found, in
  order to save time and ressources when the number of features is high.
  The majority of filter methods integrated in
  \href{http://www.rdocumentation.org/packages/mlr/}{mlr} doesn't
  support this and thus \lstinline!nselect! is ignored in most cases. An
  exception is the minimum redundancy maximum relevance filter from
  package \href{http://www.rdocumentation.org/packages/mRMRe/}{mRMRe}.
\item
  \lstinline!...! for additional arguments.
\end{itemize}

\lstinline!fun! must return a named vector of feature importance values.
By convention the most important features receive the highest scores.

If \lstinline!nselect! is actively used \lstinline!fun! can either
return a vector of \lstinline!nselect! scores or a vector as long as the
numbers of features in the task that contains \lstinline!NAs! for all
features whose scores weren't calculated.

For writing \lstinline!fun! many of the getter functions for
\href{http://www.rdocumentation.org/packages/mlr/functions/Task.html}{Task}s
come in handy, particularly
\href{http://www.rdocumentation.org/packages/mlr/functions/getTaskData.html}{getTaskData},
\href{http://www.rdocumentation.org/packages/mlr/functions/getTaskFormula.html}{getTaskFormula}
and
\href{http://www.rdocumentation.org/packages/mlr/functions/getTaskFeatureNames.html}{getTaskFeatureNames}.
It's worth having a closer look at
\href{http://www.rdocumentation.org/packages/mlr/functions/getTaskData.html}{getTaskData}
which provides many options for formatting the data and recoding the
target variable.

As a short demonstration we write a totally meaningless filter that
determines the importance of features according to alphabetical order,
i.e., giving highest scores to features with names that come first
(\lstinline!decreasing = TRUE!) or last (\lstinline!decreasing = FALSE!)
in the alphabet.

\begin{lstlisting}[language=R]
makeFilter(
  name = "nonsense.filter",
  desc = "Calculates scores according to alphabetical order of features",
  pkg = "",
  supported.tasks = c("classif", "regr", "surv"),
  supported.features = c("numerics", "factors", "ordered"),
  fun = function(task, nselect, decreasing = TRUE, ...) {
    feats = getTaskFeatureNames(task)
    imp = order(feats, decreasing = decreasing)
    names(imp) = feats
    imp
  }
)
#> Filter: 'nonsense.filter'
#> Packages: ''
#> Supported tasks: classif,regr,surv
#> Supported features: numerics,factors,ordered
\end{lstlisting}

The \lstinline!nonsense.filter! is now registered in
\href{http://www.rdocumentation.org/packages/mlr/}{mlr} and shown by
\href{http://www.rdocumentation.org/packages/mlr/functions/listFilterMethods.html}{listFilterMethods}.

\begin{lstlisting}[language=R]
listFilterMethods()$id
#>  [1] anova.test                 carscore                  
#>  [3] cforest.importance         chi.squared               
#>  [5] gain.ratio                 information.gain          
#>  [7] kruskal.test               linear.correlation        
#>  [9] mrmr                       nonsense.filter           
#> [11] oneR                       permutation.importance    
#> [13] randomForest.importance    randomForestSRC.rfsrc     
#> [15] randomForestSRC.var.select rank.correlation          
#> [17] relief                     rf.importance             
#> [19] rf.min.depth               symmetrical.uncertainty   
#> [21] univariate                 univariate.model.score    
#> [23] variance                  
#> 23 Levels: anova.test carscore cforest.importance ... variance
\end{lstlisting}

You can use it like any other filter method already integrated in
\href{http://www.rdocumentation.org/packages/mlr/}{mlr} (i.e., via the
\lstinline!method! argument of
\href{http://www.rdocumentation.org/packages/mlr/functions/generateFilterValuesData.html}{generateFilterValuesData}
or the \lstinline!fw.method! argument of
\href{http://www.rdocumentation.org/packages/mlr/functions/makeFilterWrapper.html}{makeFilterWrapper};
see also the page on \protect\hyperlink{feature-selection}{feature
selection}).

\begin{lstlisting}[language=R]
d = generateFilterValuesData(iris.task, method = c("nonsense.filter", "anova.test"))
d
#> FilterValues:
#> Task: iris-example
#>           name    type nonsense.filter anova.test
#> 1 Sepal.Length numeric               2  119.26450
#> 2  Sepal.Width numeric               1   49.16004
#> 3 Petal.Length numeric               4 1180.16118
#> 4  Petal.Width numeric               3  960.00715

plotFilterValues(d)
\end{lstlisting}

\includegraphics{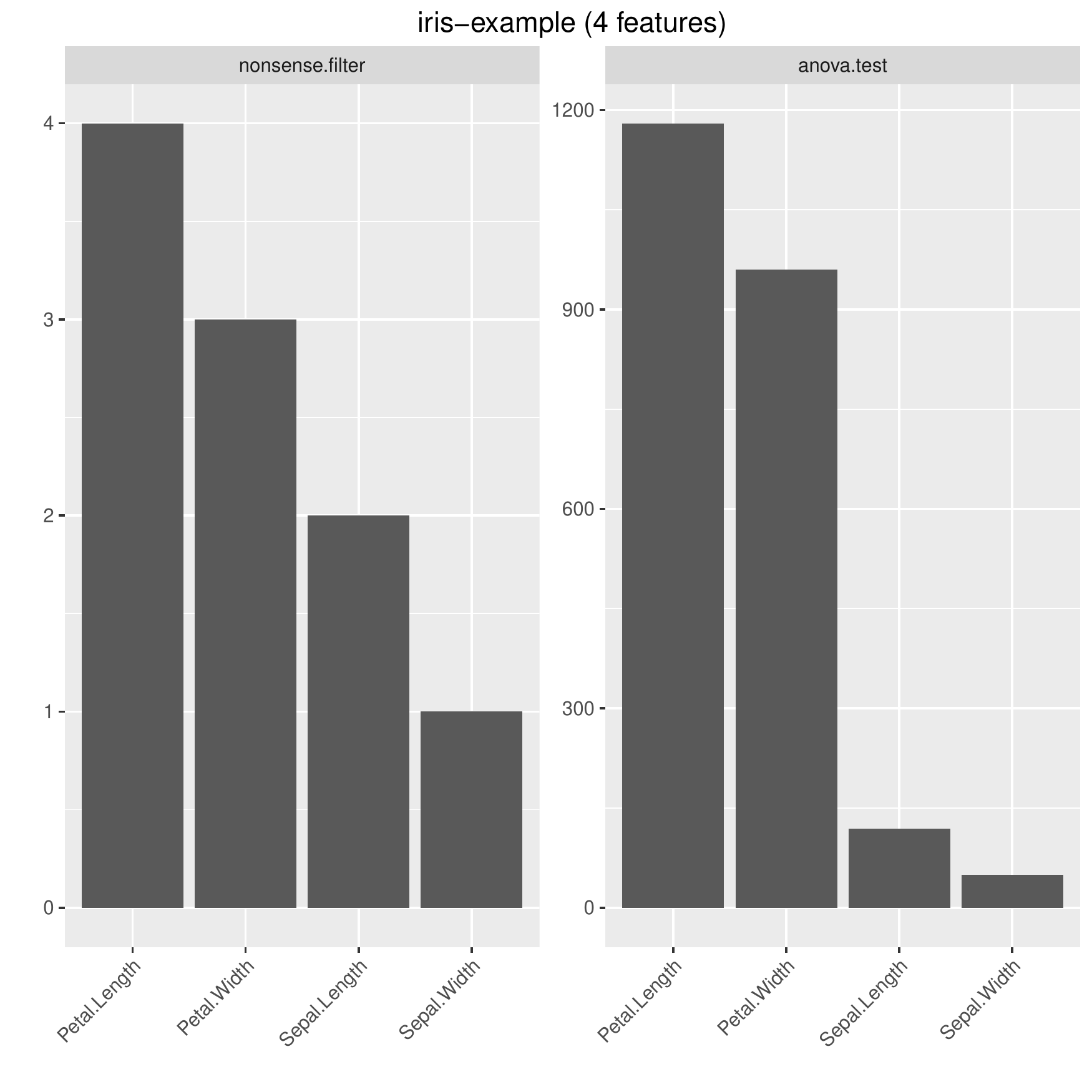}

\begin{lstlisting}[language=R]
iris.task.filtered = filterFeatures(iris.task, method = "nonsense.filter", abs = 2)
iris.task.filtered
#> Supervised task: iris-example
#> Type: classif
#> Target: Species
#> Observations: 150
#> Features:
#> numerics  factors  ordered 
#>        2        0        0 
#> Missings: FALSE
#> Has weights: FALSE
#> Has blocking: FALSE
#> Classes: 3
#>     setosa versicolor  virginica 
#>         50         50         50 
#> Positive class: NA

getTaskFeatureNames(iris.task.filtered)
#> [1] "Petal.Length" "Petal.Width"
\end{lstlisting}

You might also want to have a look at the
\href{https://github.com/mlr-org/mlr/blob/master/R/Filter.R\#L95}{source
code} of the filter methods already integrated in
\href{http://www.rdocumentation.org/packages/mlr/}{mlr} for some more
complex and meaningful examples.

\end{document}